\documentclass{article}

\usepackage{arxiv}

\usepackage[utf8]{inputenc} 
\usepackage[T1]{fontenc}    
\usepackage{hyperref}       
\usepackage{url}            
\usepackage{booktabs}       
\usepackage{amsfonts}       
\usepackage{nicefrac}       
\usepackage{microtype}      
\usepackage{fancyhdr}       
\usepackage{graphicx}       
\usepackage{amsmath}
\graphicspath{{media/}}     
\usepackage{natbib}
\usepackage[table,xcdraw]{xcolor}
\usepackage{dsfont}
\usepackage{subfigure}
\usepackage{multicol}
\usepackage{mdframed}
\usepackage{amsthm}
\usepackage{hyperref}

\usepackage{caption}
\usepackage{subcaption}
\usepackage{tcolorbox}
\AtBeginEnvironment{tcolorbox}{\small}
\usepackage{listings}
\tcbuselibrary{breakable}
\usepackage{pdfpages}
\usepackage{float}
\usepackage{siunitx}
\usepackage{array}
\usepackage{changepage}

\definecolor{instructioncolor}{RGB}{255, 69, 0}  
\definecolor{querycolor}{RGB}{34, 139, 34}       
\definecolor{subquestioncolor}{RGB}{0, 0, 255}   
\definecolor{sourcescolor}{RGB}{255, 165, 0}     
\definecolor{responsecolor}{RGB}{75, 0, 130}     

\newcounter{egbox}[section]
\renewcommand{\theegbox}{\thesection.\arabic{egbox}} 

\newtcolorbox{egbox}{
    colback=white, 
    colframe=black,
    boxrule=0.5mm,
    arc=5mm,
}

\newcounter{promptbox}[section]
\renewcommand{\thepromptbox}{\thesection.\arabic{promptbox}} 

\newmdenv[
    backgroundcolor=white,
    linecolor=black,
    linewidth=1pt,
    roundcorner=10pt,
    font=\ttfamily,  
]{promptbox}

\title{The Extractive-Abstractive Spectrum: Uncovering Verifiability Trade-offs in LLM Generations}

\usepackage{authblk}
\author{
  \textbf{Theodora Worledge, Tatsunori Hashimoto, Carlos Guestrin}\thanks{Stanford Institute for Human-Centered Artificial Intelligence} \ \thanks{Chan Zuckerberg Biohub}
}
\affil{Computer Science Department\\Stanford University\\\texttt{worledge@stanford.edu}}

\begin{document}
\maketitle

\begin{abstract}
Across all fields of academic study, well-respected experts cite their sources when sharing information. While large language models (LLMs) excel at synthesizing information, they do not provide reliable citation to sources, making it difficult to trace and verify the origins of the information they present. In contrast, search engines make sources readily accessible to users and place the burden of synthesizing information on the user. Through a survey, we find that users prefer search engines over LLMs for high-stakes queries, where concerns regarding information provenance outweigh the perceived utility of LLM responses. To rigorously examine the interplay between the verifiability and utility of information-sharing tools, we introduce the \textit{extractive-abstractive spectrum}, in which search engines and LLMs are extreme endpoints encapsulating multiple unexplored intermediate operating points. Search engines are \textit{extractive} because they respond to queries with snippets of sources with links (citations) to the original webpages. LLMs are \textit{abstractive} because they address queries with answers that synthesize and logically transform relevant information from training and in-context sources without reliable citation. We define five operating points that span the extractive-abstractive spectrum and conduct human evaluations on seven systems across four diverse query distributions that reflect real-world QA settings: web search, language simplification, multi-step reasoning, and medical advice. As outputs become more abstractive, we find that perceived utility improves by as much as 200\%, while the proportion of properly cited sentences decreases by as much as 50\% and users take up to 3 times as long to verify cited information. Our findings recommend distinct operating points for different domain-specific LLM systems. Additionally, our failure analysis informs strategies for citation identification following generation, information retrieval tasks specific to citations, and LLM systems that leverage multiple operating points. By studying the extractive-abstractive spectrum, we can create high-utility LLM systems that empower users to easily verify provided information.
\end{abstract}

\section{Introduction}
\label{sec:Introduction}
Large language models (LLMs) are increasingly capable of answering questions fluently and effectively \citep{Ouyang2022TrainingLM, openai2024gpt4technicalreport}. Users should be able to develop trust in LLMs to provide reliable answers, even on unfamiliar subjects. However, current LLMs lack accuracy in long-tail knowledge settings \citep{Sun2023HeadtoTailHK}, align with unreliable sources\footnote{\href{https://www.bbc.com/news/articles/cd11gzejgz4o}{BBC news article: "Glue pizza and eat rocks: Google AI search errors go viral"}} \citep{anagnostidis2024susceptible}, and disseminate biased perspectives and misinformation \citep{urman2023silence}. In light of such shortcomings, it is critical to equip users in verifying LLM-generated claims, especially as LLMs are applied more widely across domains and users become more dependent on them. 

\begin{figure*}
    \centering
    \includegraphics[width=\textwidth]{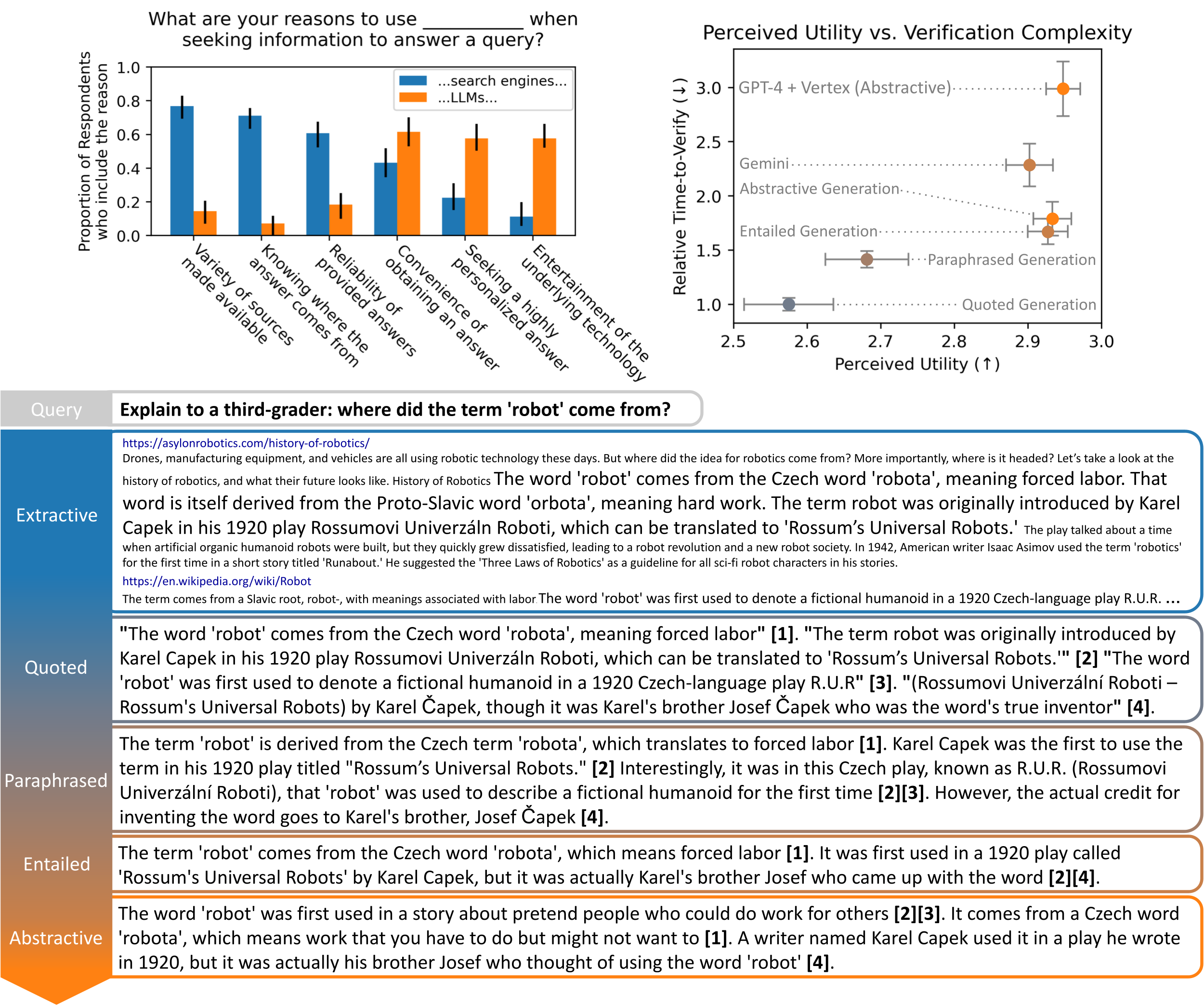}
    \caption{(Top left) Survey results from search engine and LLM users regarding the reasons they prefer search engines versus LLMs; search engines are desired for their ability to provide sources directly whereas LLMs are appreciated for their ability to synthesize information. The full survey questions are included in the appendix (\ref{subsec:human_preference_survey}). (Top right) Human evaluations show that perceived utility increases as the relative time taken to evaluate whether sentences are properly supported by citation increases across the extractive-abstractive spectrum. (Bottom) Examples of the five operating points from the reference implementations spanning the extractive-abstractive spectrum. More abstractive generations are more concise and better suit the reading level as requested in the query than more extractive generations, which more closely reflect the original source context, making them easier to verify.}
    \label{fig:EQUIP_EA}
\end{figure*}

To meet the need for verifiability, numerous generative search engines, including Google Gemini, Perplexity, and Microsoft Copilot, have combined LLMs with search engines to provide citations. However, these systems struggle to provide high quality citation; \cite{liu2023evaluating} show that only 74.5\% of citations are accurate and only 51.5\% of generated sentences requiring citation are actually cited to sources correctly, across several deployed systems. Consequently, search engines remain an appealing option for users prioritizing verifiability. Search engine outputs have high verifiability because users obtain information from the sources themselves, but they sacrifice the fluency, concision, and synthesis of LLM responses; search engine users must open webpages, identify relevant information, and synthesize information across multiple sources. Our user survey reflects this contrast between search engines and LLMs; the former are preferred for reliable answers from a variety of original sources, while the latter are instead preferred for convenient and personalized answers (Figure \ref{fig:EQUIP_EA}, top left).

In this work, we interrogate whether we can achieve verifiability while retaining the benefits of LLM responses by studying the landscape of verifiable systems between search engines and LLMs.  We view search engines and LLMs as two extremes on the \textit{extractive-abstractive spectrum}. Search engines are extractive with respect to known sources in that they answer queries with snippets and urls of the sources themselves. Accordingly, citations are inherent to search engine outputs. In contrast, LLMs are abstractive in that they synthesize content across disparate sources, often without providing citation. Our work identifies five operating points that span the extractive-abstractive spectrum and make distinct guarantees to users (Figure \ref{fig:EQUIP_EA}). For instance, the \textit{quoted operating point} takes a step from extraction toward abstraction by extracting direct quotations from multiple sources and incorporating them inline (with citations) into a single response. Another operating point is the \textit{entailed generation}, where each sentence in the response is textually entailed by cited quotes from sources. Despite intense scrutiny over improving citation systems \citep{nakano2021webgpt, gao2022rarr, menick2022teaching, thoppilan2022lamda, slobodkin2024attributefirstgeneratelocallyattributable}, existing work predominantly seeks to achieve a single operating point---the entailed generation. At the same time, quotation and paraphrase have received relatively little attention in cited generation systems, despite having long been quintessential communication modes in journalism and media \citep{gibson1999student}.

We perform careful human evaluations of seven different systems across the extractive-abstractive spectrum over four query distributions.\footnote{Code and human evaluation data available at \href{https://github.com/guestrin-lab/extractive-abstractive-spectrum}{https://github.com/guestrin-lab/extractive-abstractive-spectrum}.} An ideal system provides relevant and fluent answers that are \textit{verifiable}, i.e., have exact and thorough citations that users can verify with minimal effort. To this end, we obtain human evaluations of the quality of answers (perceived utility and fluency), citation quality (precision and coverage), and the complexity of verification (time-to-verify). Our results show trade-offs in fluency, perceived utility, and verifiability---we observe as much as a 200\% increase in perceived utility and 50\% decrease in the proportion of properly cited sentences from extractive to abstractive cited generation systems, in addition to users taking up to three times as long to verify cited information. Deployed systems are subject to these trade-offs as well. Despite having high utility, Google Gemini only properly cites 15.0\% of generated sentences.\footnote{Google Gemini generations are the cited responses obtained using the "Double-check response" feature on the Gemini web app available at \href{https://gemini.google.com/app}{https://gemini.google.com/app}.}

The four query distributions we evaluate represent real world question-answering (QA) tasks. These tasks include simplifying the style in which information is presented relative to the original sources \citep{jeblick2024chatgpt, 7776380}, multihop reasoning where conclusions must be drawn from premises across multiple sources \citep{ho2020constructing}, answering Google Search queries submitted by users \citep{kwiatkowski2019natural}, and providing high-stakes medical advice \citep{zhu2020question}. 

By characterizing the trade-offs of generations across the extractive-abstractive spectrum, we provide concrete recommendations for different application settings. Furthermore, we perform an extensive failure analysis that highlights five perceived utility failure modes, six citation quality failure modes, and three abstention modes. Our analysis highlights opportunities for future work in identifying citations after generation, information retrieval tasks specific to citations, and LLM systems that leverage multiple operating points. 

Our main contributions are as follows:
\begin{enumerate}
\item We run a human preference survey to understand the goals of evaluating different cited generation systems.

\item We define the extractive-abstractive spectrum with five notable operating points that make distinct guarantees to users. 

\item After using human evaluation to characterize the trade-offs between generation quality and verifiability across seven system implementations spanning the extractive-abstractive spectrum, including Google Gemini, we recommend different operating points for different application settings.

\item We provide a thorough failure analysis and discuss future work to improve verifiable LLM systems across the extractive-abstractive spectrum.

\end{enumerate}
\section{Human Preference Survey: LLMs versus Search Engines}
\label{sec:motivation}
Over the last 30 years, search engines have prevailed as the main tool to seek information. With LLM systems as new contenders for meeting user information needs, it is timely to understand what we gain and what we lose between the search engine and LLM system user experiences. 

\begin{figure}[ht]
    \centering
    \resizebox{.43\textwidth}{!}{
    \begin{subfigure}
        \centering
        \includegraphics[width=.25\textwidth]{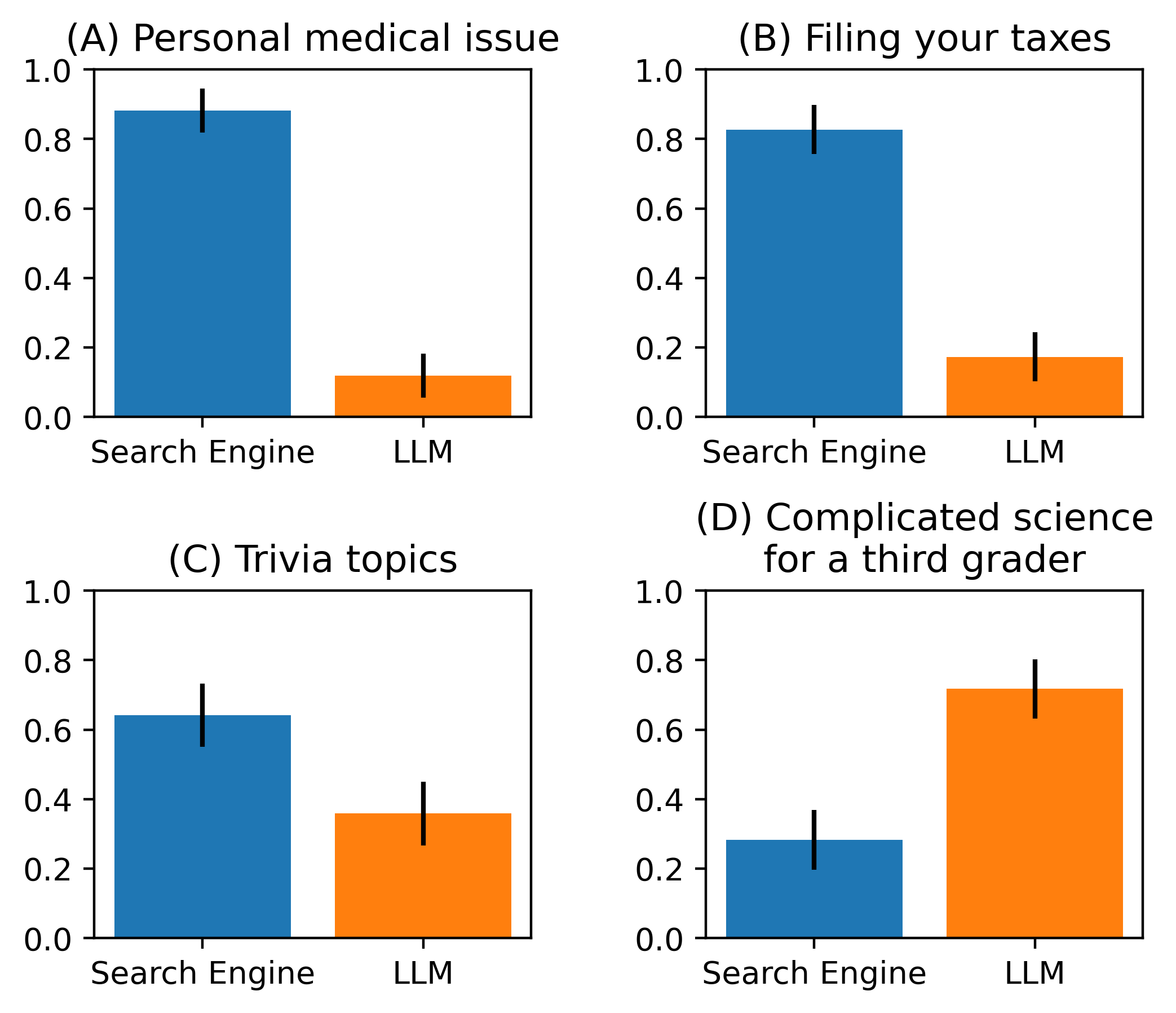}
        \label{subfig:prolific_pp}
    \end{subfigure}\hfill
    }
\caption{We survey 200 individuals evenly stratified over the Gen Z, Millenial, Gen X, and Boomer generations and report results for those who have used both search engines and LLMs. For answering a query regarding A, B, C, or D, different proportions of users prefer to use search engines or LLMs. The original survey questions are included in the appendix (\ref{subsec:human_preference_survey}). The error bars represent 95\% confidence intervals.}
    \label{fig:prolific_pp}
\end{figure}

In a survey of 200 individuals, we find that users prefer search engines to LLMs without citations, like ChatGPT and Claude, for answering high-stakes questions, while they prefer LLMs for research-intensive tasks with lower stakes (Figure \ref{fig:prolific_pp}). Not only do users have platform preferences, but their preferences are informed by specific reasons that reflect a lack of trust in LLMs. Users prefer search engines for the variety of available sources, understanding of information provenance, and reliability of provided information, in contrast to preferring LLMs for convenience and personalized answers (Figure \ref{fig:EQUIP_EA}, top left). Our evaluations of citation quality and perceived utility are motivated by the lack of user confidence in LLMs to provide supporting sources and in search engines to provide personalized answers.

While user preferences may change as LLMs become increasingly commonplace, these results indicate that users perceive different information-seeking tasks as being best served by entirely different systems. Accordingly, there is an exciting opportunity to design systems preferred by users across a broad range of queries (e.g. all queries in Figure \ref{fig:prolific_pp}). To map out the extractive-abstractive spectrum toward this end, our work here takes the user's perspective to propose distinct operating points that bridge the gap between search engines and LLMs. By evaluating systems across this extractive-abstractive spectrum, we inform the design of future systems that achieve verifiability and user desiderata across a broad range of information-seeking queries.

\section{Cited Generation Formalization}
\label{sec:task_definition}
We consider the task of cited generation for queries in the open-ended QA setting. Given a user-provided natural language query, $x$, the systems we study produce a natural language generation $y$ that contains inline citations to sources. The generation $y$ is partitioned into $n$ attributable units, $U = \{u_1,...,u_n\}$, which our work defines as passages of text, often sentences. Attributable unit $u_i$ is cited to $m_i$ quotes from sources, $Q_i = \{q_{i,1},...,q_{i,m_i}\}$, where $q_{i,j}$ is a direct quote from a known source. Accordingly, any claim contained within $q_{i,j}$ is interpreted in the context of its original source text. We refer to the set of source texts from which quotes across all $Q_i$ originate as $S$. Lastly, $Q_i$ does not contain any cited quotes if $u_i$ does not require citation.\footnote{We use Google's Vertex API to identify filler sentences that do not require citation, such as "Great question!" or "Happy to explain further!", and find such sentences are relatively rare in the systems we evaluate. Determining what to cite can be a nuanced task that depends on common knowledge and user needs in an application area. Our work expects citations for all sentences containing claims of a factual nature.} For example:

\begin{table}[H]
\begin{tabular}{m{.3cm}m{15cm}}
\centering
$x$: & What is the diameter of Jupiter's largest moon? \\
\\
$y$: & $\underbrace{\text{Great question!}}_{u_1} \underbrace{\text{Jupiter's largest moon, Ganymede, has a diameter of 5,268 km}}_{u_2}$ [1][2]. \\
\\
$Q_1$: & $\emptyset$ \\
\\
$Q_2$: \newline \newline & [1] "Jupiter’s moon Ganymede is the largest moon in our solar system, bigger than the planet Mercury \newline and \& dwarf planet Pluto." \newline [2] "Ganymede has a diameter of 3270 miles (5,268 km) and is larger than the planet Mercury." \\
\\
$S$: \newline & [1] Text of \href{https://science.nasa.gov/jupiter/moons/ganymede/facts/}{https://science.nasa.gov/jupiter/moons/ganymede/facts/} \newline [2] Text of \href{https://coolcosmos.ipac.caltech.edu/ask/201-What-is-the-largest-moon-in-the-Solar-System}{https://coolcosmos.ipac.caltech.edu/ask/201-What-is-the-largest-moon-in-the-Solar-System} \\
\end{tabular}
\end{table}

The cited source quotes $Q_i$, interpreted in their original source context, should support information in $u_i$ following the intuitive test from \cite{rashkin2022measuringattributionnaturallanguage}: "According to $Q_i$, $u_i$." Because citations are ultimately a tool for users, we emphasize that citations should also have low verification complexity, i.e. be easy for users to check. A sentence is \textit{verifiable} if users can determine that cited quotes from known sources support the sentence with ease. For this reason, we study citation systems where $q_{i,j}$ is markedly shorter than the source that contains it, in contrast to systems that have higher verification complexity due to citing longer passages or even entire sources. We evaluate Google Gemini because Perplexity, SearchGPT, and YouChat cite entire sources, whereas Gemini orients users towards specific sentences within cited sources. We define metrics to evaluate citation quality and verification complexity in Section \ref{subsec:metrics}. 

Ideally, cited sources are trustworthy and up-to-date. In this work, we assume that cited sources retrieved from Google meet these criteria within reason. Because evaluating source reliability may require domain expertise and careful examination of multiple sources in relation to one another, we leave a more thoughtful approach for this challenging task to future work. 
\section{The Extractive-Abstractive Spectrum}
\label{sec:spectrum}

A response to a query is extractive with respect to cited source quotes if the response is composed of the source quotes themselves. For example, search engines respond to queries with snippets of webpages. On the other hand, a response is abstractive with respect to cited source quotes if the cited information is reworded, pruned, supplemented, and logically transformed, e.g. the abstractive generation in Figure \ref{fig:EQUIP_EA} (bottom). We define notable operating points (OPs) over the extractive–abstractive spectrum and describe our implementation of each operating point, as well as additional baselines.

\subsection{Operating Point Definitions}
\label{sec:op_definitions}

Early works on automatic summarization frame abstraction as a means to achieving coherence \citep{10.5555/315149.315355, 10.1016/S0004-3702(02)00222-9}. Other works discern the abstraction of generated summaries with respect to source text using n-gram based metrics \citep{ladhak-etal-2022-faithful, dreyer-etal-2023-evaluating, weller-etal-2024-according}. However, neither coherence nor n-gram based metrics capture how generated text semantically relates to source text. In contrast, \cite{10.1145/312624.312666} study how humans use the semantic operations of paraphrase and information removal to write abstractive summaries of a source text. In the same spirit, we define operating points across the extractive-abstractive spectrum that correspond to distinct user experiences regarding the semantic relation between generations and cited source quotes. Our definitions for the quoted, paraphrased, and entailed generations draw from work on meaning relations \citep{gold-etal-2019-annotating} applied to the setting of citations by \cite{10516637}. Figure \ref{fig:EQUIP_EA} (bottom) provides an example of each OP.

\textbf{Extractive Output:} An extractive output is a set of quotes from the sources themselves. In the context of extractive outputs, we refer to these quotes as snippets, as in Google Search. Citations are inherent to this operating point. Formally, $\forall \ u_i, \ |Q_i|=1, \ u_i = q_{i,1}$, where $u_i$ may be multiple sentences.

\textbf{Quoted Generation:} A quoted generation achieves grounding by incorporating quotes of claims from known sources inline with unquoted words used sparingly to facilitate fluency. The quotes are taken \textit{word-for-word} from the sources, are not misrepresented from the original context, and are clearly demarcated from non-quoted text in the generation. Formally, $\forall \ u_i, \ Q_i \Leftrightarrow u_i \text{ and every } q_{i,j} \text{ is a quoted substring of } u_i$.

\textbf{Paraphrased Generation:} A paraphrased generation achieves grounding by rephrasing and rewording information from cited source quotes. Each sentence in the paraphrased generation conveys \textit{the same information} as its cited source quotes; the generation does not remove any information, does not contract logical steps of reasoning, and does not add new claims, relative to the cited quotes. Formally, $\forall \ u_i, \ Q_i \Leftrightarrow u_i$. 

\textbf{Entailed Generation:} An entailed generation achieves grounding by rephrasing, rewording, removing, and logically transforming information relative to cited source quotes. Each sentence in the entailed generation conveys \textit{no more information} than its cited source quotes; the generation may remove extraneous information and contract logical steps of reasoning, but does not add new claims, relative to the cited quotes. Formally, $\forall \ u_i, \ Q_i \Rightarrow u_i$.

\textbf{Abstractive Generation:} Each sentence of an abstractive generation may contain claims that are not cited ($u_{i, \text{uncited claims}}$), in addition to claims that are cited ($u_{i, \text{cited claims}}$). Accordingly, an abstractive sentence is not necessarily grounded and may convey \textit{more information} than the cited source quotes. Formally, if $u_{i, \text{cited claims}} \neq \emptyset, \text{ then } Q_i \Rightarrow u_{i, \text{cited claims}}.$

We note that the quoted OP is distinct from more abstractive OPs for two reasons. First, quotation marks delineate cited content from uncited words in the generation. Second, because quotes in the generation exactly match text from a source, users only need to ensure the quote is not taken out of context from the source when verifying a cited quote. In contrast, when verifying citations from more abstractive OPs, users have the additional burden of ensuring that the cited sentence is faithfully re-written from the source.

While other OPs of interest may lie on the extractive-abstractive spectrum, we highlight the OPs defined above because they either correspond to deployed systems or bridge the gap between familiar user experiences. For instance, conventional search engines provide extractive responses. To the best of our knowledge, prior work has neither defined nor implemented quoted and paraphrased cited generations in the context of cited question-answering LLM systems. We refer readers interested in systems from prior work and their relation to our operating points to Sections \ref{subsec:extractive_quoted_systems} and \ref{subsec:entailed_systems}. Most prior work on cited LLM generations seek to implement the entailed operating point. LLM generations without citations are a commonplace example of the abstractive generation, where there are no cited claims ($u_{i, \text{cited claims}} = \emptyset$). The internals of deployed generative search engines are not public, but we speculate they fall around the entailed and abstractive operating points.

\subsection{Reference Operating Point Implementations}
\label{sec:op_implementations}

We design reference OP implementations to study trade-offs across the extractive-abstractive spectrum. In order to make each OP comparable to the others, we ground responses with the same set of source quotes for each query.\footnote{While our implementation sheds light on favorable design choices and yields high citation quality, we hope to see even stronger and more efficient implementations of individual OPs in future work without design choices specific to studying the extractive-abstractive spectrum.} We describe the reference implementation for each OP and provide further details in the appendix (\ref{subsec:more_implementation_details}).

The \textbf{extractive output} is a list of source snippets. Each snippet is about 1000 characters long and contains a quote used in the reference quoted generation. These snippets are either retrieved or gold-standard snippets (when they are provided by a dataset). To retrieve snippets given a query, we scrape text from the top 50 Google Search results for the query, chunk the text into snippets about 1000 characters long, and select the ten most relevant snippets to the query using Dense Passage
Retrieval \citep{karpukhin-etal-2020-dense}.

The \textbf{quoted generation} is generated 
from GPT-4 with few-shot examples, the query, and relevant source snippets provided in-context. We instruct the model to respond to the query using word-for-word quotes from the provided sources. The prompt with few-shot examples is provided in the appendix (\ref{Prompts for the Quoted Output}). Citations are identified by matching quotes in the generation with their word-for-word counterparts in the snippets provided in-context, using string matching. We generate each reference quoted generation 10 times and keep the generation with the fewest unquoted words. Across our evaluated queries, 82\% of words in quoted generations are quoted and on average, quotes are 14.5 words long. We denote this implementation as $\mathcal{M}_\text{quoted}(x, \ S)$. 

Each of the \textbf{paraphrased}, \textbf{entailed}, and \textbf{abstractive generations} is a revision of the quoted generation. These generations are obtained by prompting GPT-4 with few-shot examples, the query, and the corresponding quoted generation (not the source snippets). The paraphrased instruction requests a rewrite, the entailed instruction requests a more concise generation, and the abstractive instruction requests the model to improve the quoted generation by adding additional information. The prompts for each OP are provided in the appendix (\ref{Prompts for the Paraphrased Output}, \ref{Prompts for the Entailed Output}, \ref{Prompts for the Abstractive Output}). For each sentence in the generation, citations are obtained via few-shot prompting and a citation instruction that identifies supporting quotes from those used in the quoted generation. Each supporting quote is cited within its original snippet. The prompts for citation identification are available in the appendix (\ref{Prompts for Citation Generation}). We denote the paraphrased implementation as $\mathcal{M}_{\text{paraphrased}}(x, \ \mathcal{M}_\text{quoted}(x, \ S))$, with the same function signature applying to $\mathcal{M}_{\text{entailed}}$ and $\mathcal{M}_{\text{abstractive}}$.

The citations provided by all reference generations are both corroborative \textit{and} contributive attributions; they provide users with sources that both support claims in the generation and contributed to the creation of the generation \citep{10516637}.

\subsection{Additional Evaluated Systems}
\label{sec:baselines}
In addition to the reference implementations of the OPs, we study two more methods of cited generation. One method combines publicly available APIs for generation and citation, while the other is a deployed system. Further details about these implementations are included in the appendix (\ref{gpt-4 output details}, \ref{gemini output details}).

The \textbf{GPT-4 + Vertex generation} is another instantiation of the abstractive OP. The response is generated by conditioning GPT-4 on the query, without retrieved sources. Using Google's Vertex API\footnote{\href{https://cloud.google.com/generative-ai-app-builder/docs/check-grounding}{https://cloud.google.com/generative-ai-app-builder/docs/check-grounding}}, we identify citations for individual generated sentences from snippets retrieved using Google search and a semantic similarity model\footnote{\href{https://huggingface.co/sentence-transformers/all-MiniLM-L6-v2}{https://huggingface.co/sentence-transformers/all-MiniLM-L6-v2}}, in addition to any gold-standard source snippets provided by the query dataset. Because cited sources are only retrieved and identified after response generation, we refer to these citations as \textit{post-hoc} citations. 

We also evaluate \textbf{Google Gemini} as a deployed system baseline. Google has specified that Gemini can use retrieved sources to answer queries but that cited sources are not necessarily those used to generate the response.\footnote{\href{https://cloud.google.com/blog/products/ai-machine-learning/rag-and-grounding-on-vertex-ai}{https://cloud.google.com/blog/products/ai-machine-learning/rag-and-grounding-on-vertex-ai}}\footnote{\href{https://support.google.com/gemini/answer/14143489}{https://support.google.com/gemini/answer/14143489}} In other words, some Gemini citations are post-hoc citations. We scrape Gemini responses and citations to queries using the "Double-check response" feature. The responses were generated and scraped throughout August and September 2024.
\section{Evaluation of the Extractive-Abstractive Spectrum} 
\label{sec:evaluation}

We conduct human evaluations of the reference, GPT-4 + Vertex, and Gemini generations. To capture the quality and verifiability of generations from each operating point, we record five metrics over four query distributions. 

\subsection{Metrics}
\label{subsec:metrics}
An ideal cited generation addresses the query fluently, effectively, and verifiably. For the first two properties, we remove citation markers and evaluate fluency and perceived utility for the entire generation. To measure verifiability, we record the precision of each citation, the citation coverage of each sentence, and the time to verify each sentence. We define each metric below.

\textbf{Fluency: }Annotators are asked to count misprints, incoherent sentences, and abrupt transitions between sentences against fluency (Fig. \ref{fig:annotation_interface_utility_fluency}). The fluency of a response to a query is evaluated on a three-point scale, 3 being fluent and 1 not being fluent. 

\textbf{Perceived Utility: }
Annotators are asked to consider the extent to which they believe a generation addresses the query, is concise, and is written in an appropriate style (Fig. \ref{fig:annotation_interface_utility_fluency}). We note that perceived utility does not necessarily measure the correctness of generated text because annotators may not have the domain knowledge to fact-check generated information. The utility of a generation for a query is evaluated on a three-point scale, 3 being high utility and 1 being low utility. 

\textbf{Citation Precision: }A citation accompanying text is precise if the cited source supports at least one claim made in the text \citep{liu2023evaluating}. An imprecise citation is one that is either irrelevant to the text, or, in the worst case, disagrees with the text. For each citation in a generation, annotators are given the binary task of determining whether the cited source supports at least one claim in the generated sentence containing the citation. We leave the task of identifying claims within a sentence to the discretion of the human annotators and their experience as information-seeking users. Annotators are trained to use the intuitive test from \cite{rashkin2022measuringattributionnaturallanguage} where a cited source, $s$, supports a claim, $c$, if the annotator believes it is truthful to say, "According to $s$, $c$."

\textbf{Citation Coverage: }A sentence is covered if the cited sources accompanying the sentence support all claims made in the sentence. We consider citation coverage as a quantity that best captures a user's verifiability concerns; even with perfect citation precision, it is possible for generated claims to be unsupported. For each sentence in a generation, annotators are given the binary task of determining whether all claims in the sentence are supported by at least one citation accompanying the sentence. Annotators are trained to use the intuitive test from \cite{rashkin2022measuringattributionnaturallanguage} to determine whether the cited sources together support the generated sentence. 

\textbf{Time to Verify (T2V): }T2V is the wall clock time it takes a human to determine whether a generated sentence is covered by its accompanying citations. T2V captures the complexity a user faces when verifying a sentence with cited sources. To account for differences in reading speeds between annotators throughout the study, we report \textit{relative T2V}, which is the average normalized T2V; each T2V value is normalized by its annotator's average T2V over sentences from quoted generations. We measure T2V over the task of determining the coverage of a single sentence, rather than the precision of a citation, because determining coverage more holistically represents the user experience of verifying select sentences. We discuss and present results on other variants of relative T2V, including absolute T2V and the time to verify entire responses in the appendix (\ref{subsec:closer_look_at_t2v}).

While the \textbf{fluency} and \textbf{perceived utility} ratings are given for the entire response, our results for \textbf{citation precision}, \textbf{coverage}, and \textbf{time-to-verify} only consider sentences that require citation. We use the Vertex API to identify filler sentences, such as "That's a good question", that do not require citation and find there are few filler sentences across all generations. Results for citation precision, coverage, and T2V over all sentences are nearly identical to those only over sentences that require citation and are included in the appendix (\ref{subsec:results_on_all_sentences}). 

\subsection{Query Distributions}

We obtain human evaluations for each of seven cited generations over 480 open-ended question-answering (QA) queries, drawn from four query distributions that reflect different information-seeking scenarios. The reference implementations of the operating points require sources to create the cited generations. For each query distribution, we either retrieve sources from Google or use gold-standard sources. Gold-standard sources are sources provided by the query dataset that are guaranteed to answer the query. Our snippet retrieval pipeline is detailed in Section \ref{sec:op_implementations}.

\textbf{Natural Questions (NQ)} \citep{kwiatkowski2019natural}: The NQ query distribution is composed of real user queries to Google Search. We retrieve the sources for each NQ query.

\textbf{Explain to a Third Grader (Eta3G):} The Eta3G query distribution is a variant of the NQ query distribution \citep{kwiatkowski2019natural} where NQ queries are prepended with the prefix, "Explain to a third-grader: ". Queries from this distribution not only seek relevant information in response to the query, but also require that the information is presented in a style suitable for a third-grade reading level and maturity. We retrieve the sources for each Eta3G query.

\textbf{Wikipedia and Wikidata Multi-Hop (2WikiMH)} \citep{ho2020constructing}: Answering queries from 2WikiMH requires combining premises from multiple sources to draw logical conclusions not necessarily present in any individual source. The queries themselves are about the relation between people, places, dates, and other proper nouns, e.g. "Who is the paternal grandfather of Prince Alexander Of Imereti (1760–1780)?" We use the gold-standard source snippets provided by 2WikiMH for each query.

\textbf{Multiple Answer Spans Healthcare (MASH)} \citep{zhu2020question}: The queries from the MASH distribution are from the perspective of consumers seeking medical advice, a high risk setting where unreliable information has consequences on human health. We use the gold-standard source snippets from WebMD articles provided by MASH for each query.

\subsection{Evaluation Procedure}
\begin{figure}[ht]
    \centering
    \includegraphics[width=\textwidth]{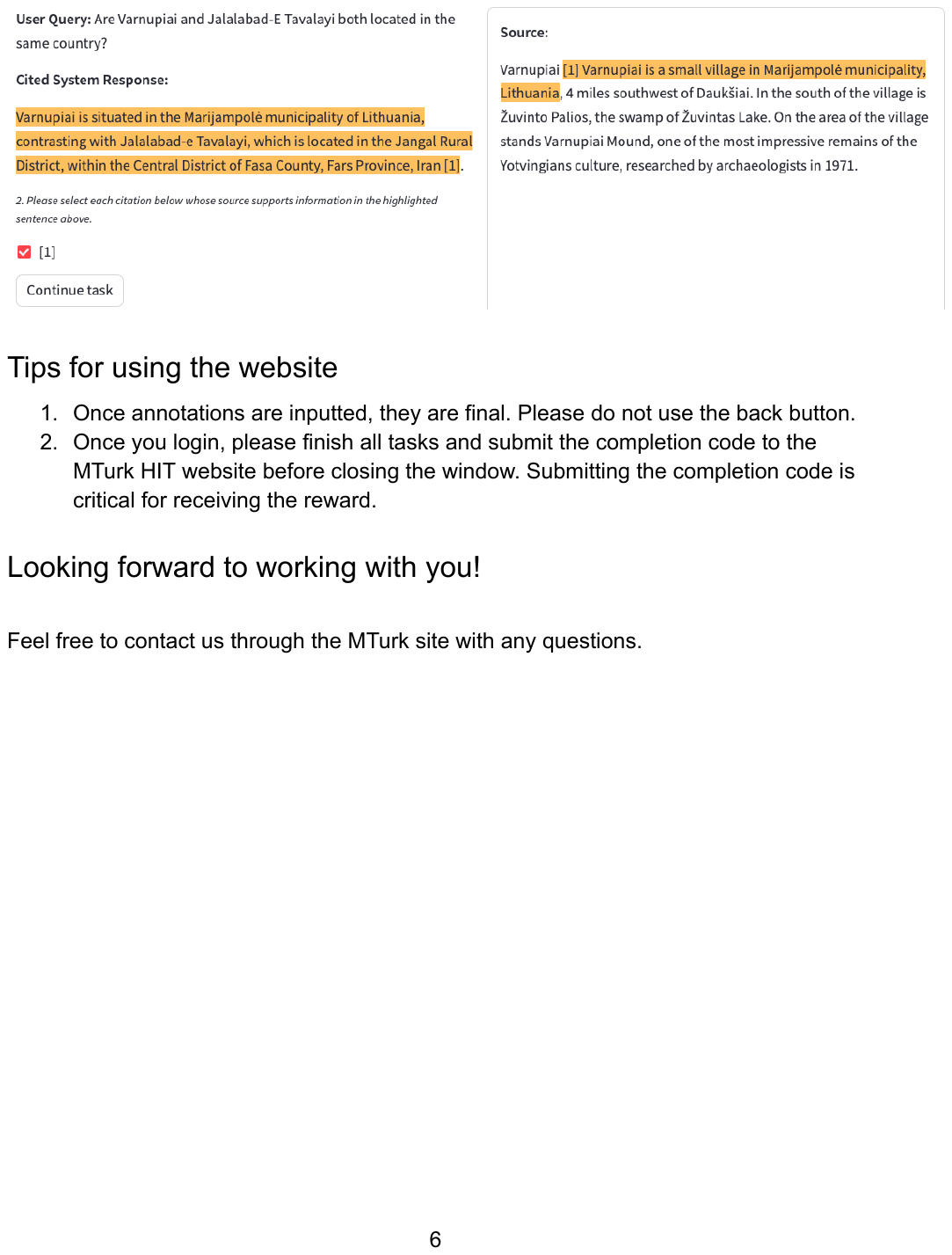}
    \caption{Annotation guidelines provided to annotators throughout the study.}
    \label{fig:Annotation_Task_Guidelines6}
\end{figure}

Amazon Mechanical Turk (MTurk) annotators evaluated cited generations for fluency, utility, citation precision, citation coverage, and T2V. MTurk workers residing in the United States with HIT success rates over 96\% were invited to read the annotation guidelines (Fig. \ref{fig:Annotation_Task_Guidelines6}) and take a screener HIT to qualify for the study. We provided annotators with high performance on the screener with personalized feedback and only admitted those who acknowledged the correspondence. Thirty-one annotators were ultimately admitted to the study. Throughout the study, we maintained an active presence through email and a dedicated channel on the Turker Nation Slack workspace.  

We assigned each annotator to evaluate an equal number of generations from each reference operating point and an equal number of GPT-4 + Vertex and Gemini generations. In order to avoid any effects of familiarization with the queries and sources on T2V, no annotator evaluated citations for the same query in more than one OP. We provide further details about the evaluation procedure, annotator compensation, annotation guidelines, and a pilot study performed before the MTurk evaluations in Appendix \ref{subsec:human_evaluation_details}. 
\section{Results and Analysis}
\label{sec:results}

\begin{figure*}[ht]
\raggedright
\resizebox{\textwidth}{!}{
    \begin{subfigure}
        \centering
        \includegraphics{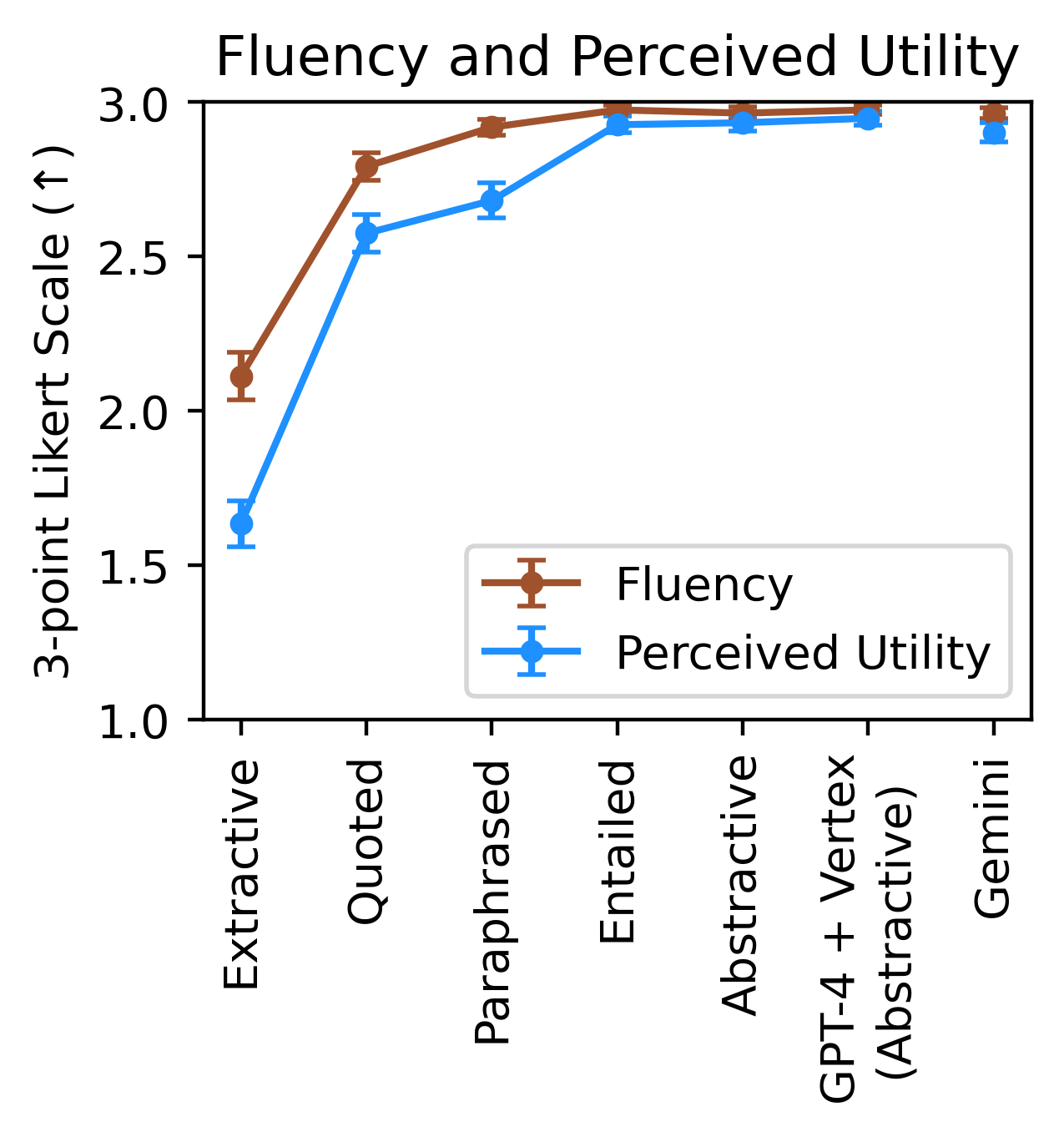}
        \label{fig:utility_fluency_aggregate}
    \end{subfigure}
    \begin{subfigure}
        \centering
        \includegraphics{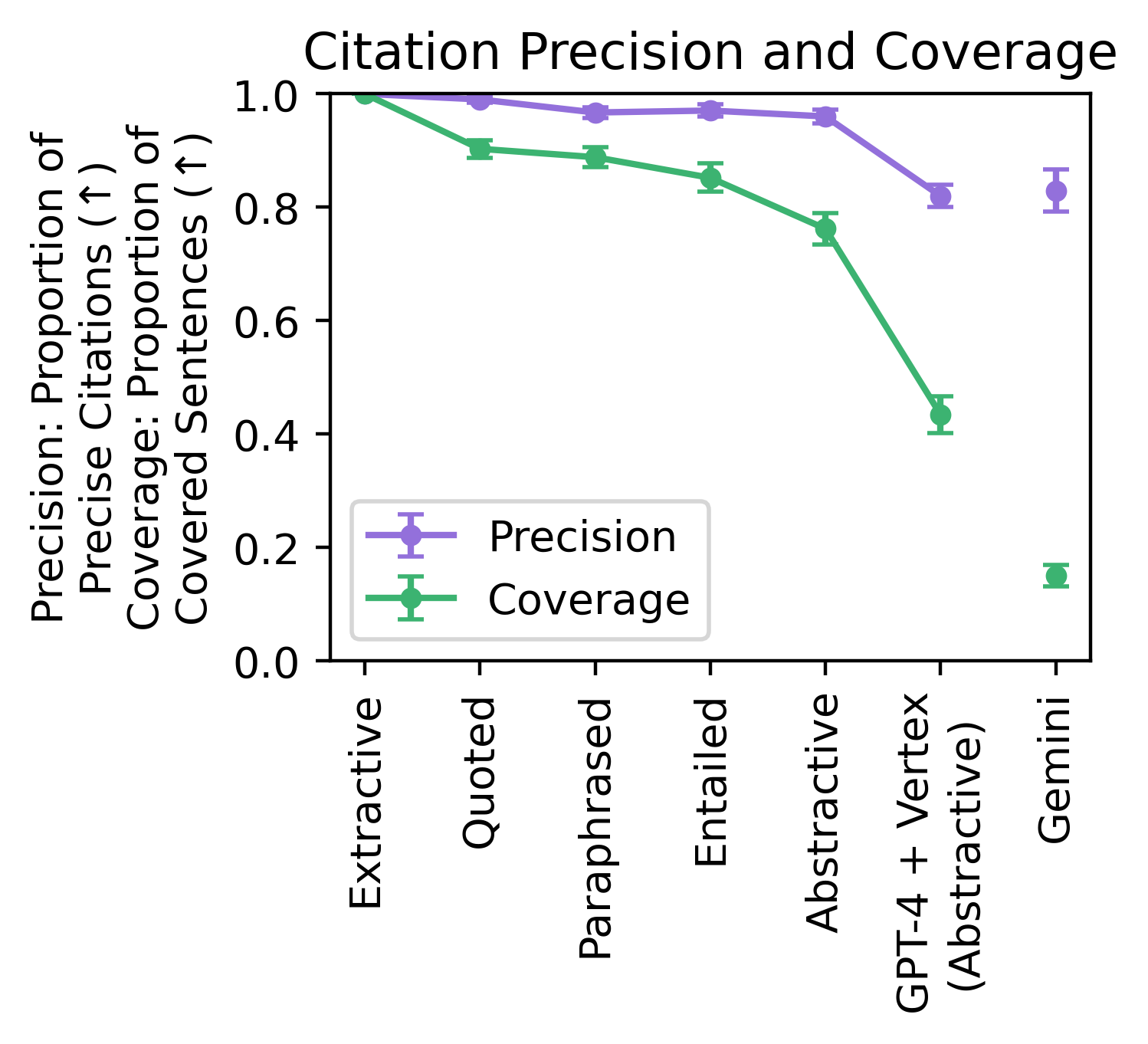}
        \label{fig:precision_coverage_aggregate}
    \end{subfigure}
    \begin{subfigure}
        \centering
        \includegraphics{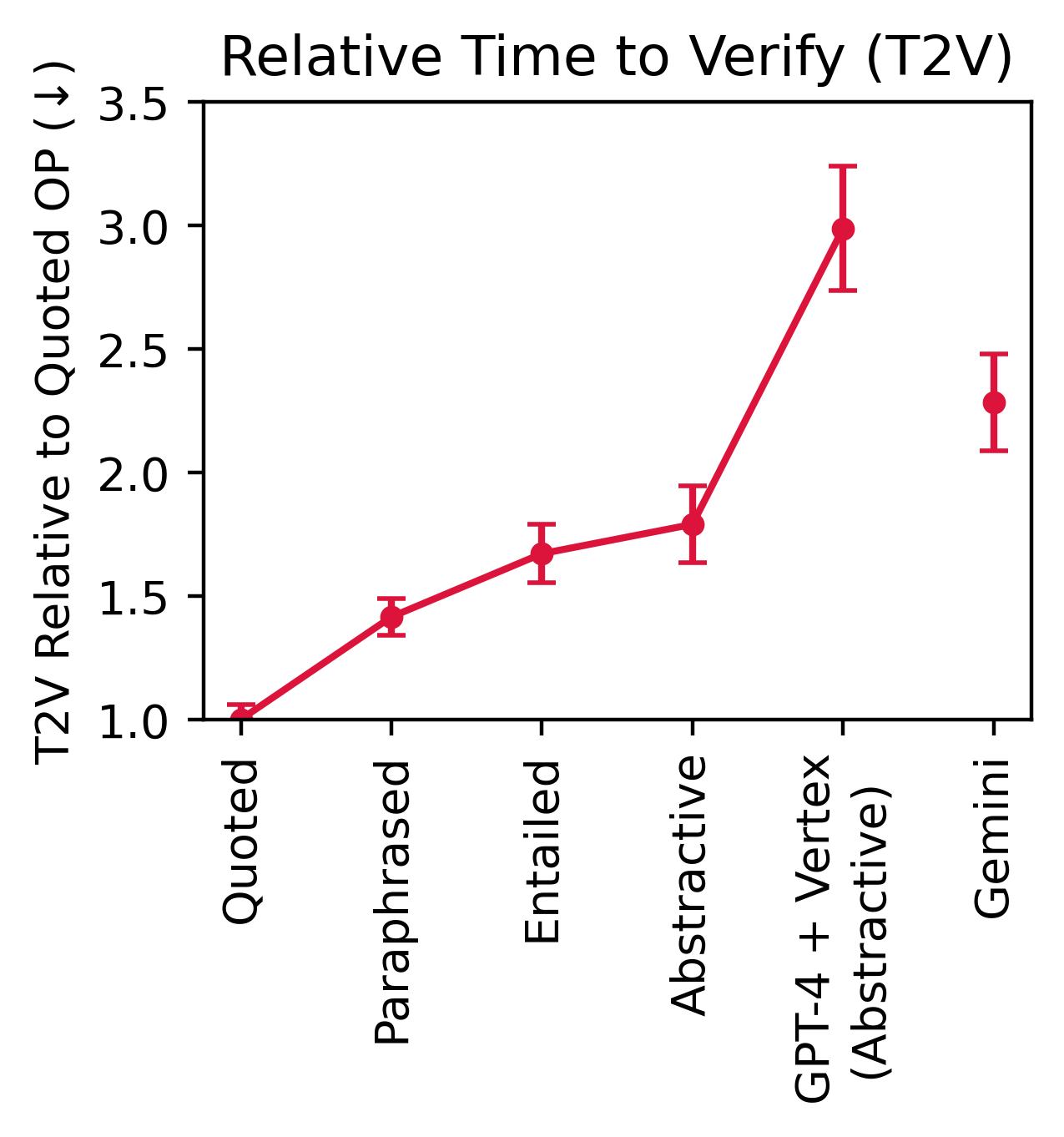}
        \label{fig:t2v_aggregate}
    \end{subfigure}
    }
\caption{Human evaluation results averaged over the four query distributions. Fluency and perceived utility increase with abstraction, while citation precision and coverage decrease. Annotators take longer to evaluate coverage as generations become more abstractive. The error bars represent 95\% confidence intervals.}
\label{fig:results_in_aggregate}
\end{figure*}

\subsection{Summary of results across the extractive-abstractive spectrum}
\label{sec:results_in_aggregate}

While fluency and perceived utility increase as generations become more abstractive relative to cited sources, verifiability degrades. In particular, perceived utility increases by over 200\% and citation coverage decreases by over 50\% between the extractive and GPT-4 + Vertex abstractive generations (Figure \ref{fig:results_in_aggregate}, left and middle). Simultaneously, humans take three times longer to verify cited sentences from GPT-4 + Vertex generations relative to those from quoted generations (Figure \ref{fig:results_in_aggregate}, right). The deployed Gemini system is Pareto-dominated by the entailed generations; while Gemini generations demonstrate similar fluency and perceived utility compared to the entailed generations, they exhibit over 80\% lower coverage (Figure \ref{fig:results_in_aggregate}, left and middle). Humans also take significantly longer to verify cited sentences from Gemini than from the entailed generations (Figure \ref{fig:results_in_aggregate}, right).

\subsubsection{Perceived Generation Quality}

Generations that are more abstractive with respect to sources are more fluent and have higher perceived utility than generations that are more extractive (Figure \ref{fig:results_in_aggregate} left). Fluency is lowest for the extractive OP, which provides sources in a list without smooth transitions. The quoted OP significantly improves fluency by incorporating source quotes inline within a single response, but the flexibility of the paraphrased, entailed, and abstractive OPs leads to more natural and fluent generations. 

Similarly, the extractive OP exhibits the lowest perceived utility; source snippets often contain irrelevant information and may not present information in a suitable style for the query. The quoted OP significantly improves perceived utility by excising irrelevant information from the snippets. While the paraphrased OP further improves perceived utility, a more significant increase is achieved by the entailed OP. This increase is likely because both paraphrased and entailed generations customize generated language to the query, but only the entailed OP generations can remove information and draw entailed conclusions from the cited source quotes. 

Gemini generations are comparable to the more abstractive OPs in fluency and perceived utility. Examples, categories, and frequencies of perceived utility failures are presented in Section \ref{subsec:uf_failure_quoted_outputs}.

\subsubsection{Citation Precision Results}
Citation precision is relatively similar (95\%-100\%) across the reference OP generations, but is significantly lower for the GPT-4 + Vertex and Gemini generations (Figure \ref{fig:results_in_aggregate}, middle). This result indicates that the paradigm employed by the reference implementation of first choosing source quotes to cite and then generating a response using them, or more succinctly, "attribute first, then generate" \citep{slobodkin2024attributefirstgeneratelocallyattributable}, mitigates the trade-off between abstraction and precision observed by \cite{liu2023evaluating} among deployed implementations of the entailed OP. In contrast, methods like the GPT-4 + Vertex pipeline and Gemini that provide post-hoc citations are more susceptible to precision errors. We discuss examples, categories, and frequencies of precision errors by implementation in Section \ref{subsec:citation_precision_failures}.

\subsubsection{Citation Coverage Results}
As generations become more abstractive, citation coverage trends downward in Figure \ref{fig:results_in_aggregate} (middle). As expected, coverage drops significantly for the reference abstractive generations---by definition, the abstractive OP contains claims not entailed by cited sources. Coverage is even lower for GPT-4 + Vertex and Gemini generations, at 43.4\% and 15.0\% respectively. The low coverage of Gemini generations indicate that achieving high coverage in deployed settings may present additional challenges, such as cost and latency. Regardless, the finding that Gemini and abstractive generations have both high perceived utility and low coverage leads to the concern that convincing generations may lead users to believe insufficiently supported information.

\begin{table}[ht]
\centering
\begin{tabular}{lc}
\toprule
\textbf{Type of Generation} & \textbf{Number of Citations per Cited Sentence} \\ 
\midrule
Quoted                      & 1.19 \\ 
Paraphrased                 & 1.33 \\ 
Entailed                    & 1.90 \\ 
Abstractive                 & 1.75 \\ 
GPT-4 + Vertex (Abstractive) & 3.14 \\ 
Gemini                      & 1.00 \\ 
\bottomrule
\end{tabular}
\caption{Average number of citations per cited sentence. Cited sentences from more abstractive generations tend to have more citations than those from more extractive generations. Gemini consistently provides at most one citation per sentence.}
\label{fig:num_citations_by_op}
\end{table}

\subsubsection{Relative T2V Results}
As generations become increasingly abstractive with respect to cited source quotes, humans take longer to verify sentences with accompanying citations (Figure \ref{fig:results_in_aggregate}, right). Relative to quoted generations, we find that some abstractive generations take up to three times as long to verify. The quoted OP significantly expedites human verification of cited text in comparison to all other OPs; simply rewording quoted generations to be paraphrased generations increases T2V by over 40\%. The positive relation between T2V and generation abstraction reflects the cognitive load placed on users verifying citations by reworded and logically transformed cited source quotes. 

We also examine T2V in relation to the number of citations in the sentence. The more citations in a sentence, the more citations may need to be evaluated in order to determine the coverage of the sentence. In Table \ref{fig:num_citations_by_op}, we observe that citations per sentence often increase with generation abstraction as information is increasingly combined across multiple sources. The high relative T2V of the GPT-4 + Vertex generations may be attributed to the higher number of citations per sentence. However, verification complexity is not solely driven by citation count because Gemini generations have fewer citations on average per sentence yet still higher relative T2V than the reference generations. We present further analysis of T2V, including T2V presented as absolute values, in the appendix (\ref{subsec:closer_look_at_t2v}).

\subsection{Results by query distribution across the extractive-abstractive spectrum}
\label{sec:qd_results}
\begin{figure*}[ht]
    \raggedright
    \resizebox{\textwidth}{!}{
    \begin{subfigure}
        \centering
        \includegraphics{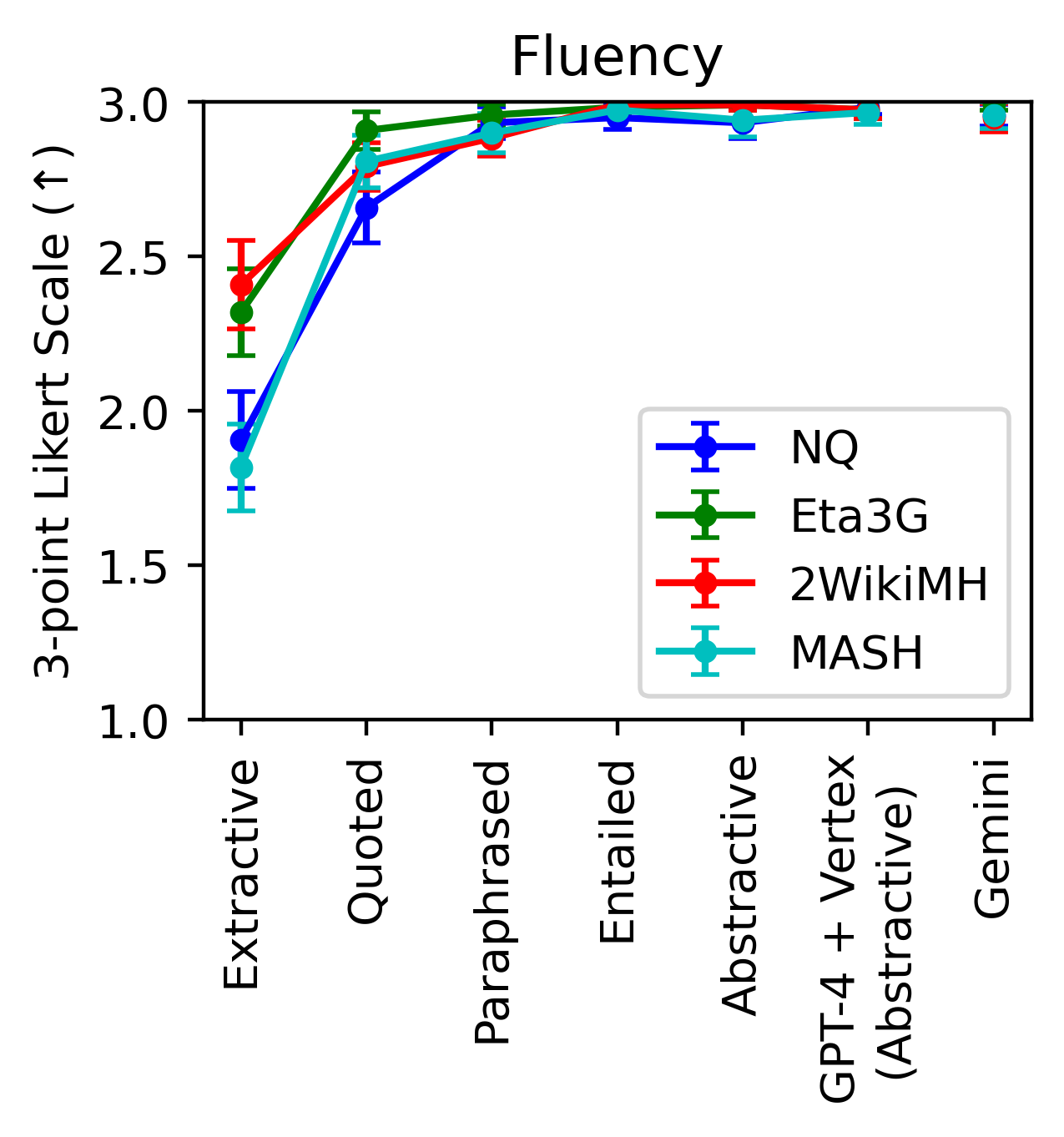}
        \label{fig:fluency_by_qd}
    \end{subfigure}
    \begin{subfigure}
        \centering
        \includegraphics{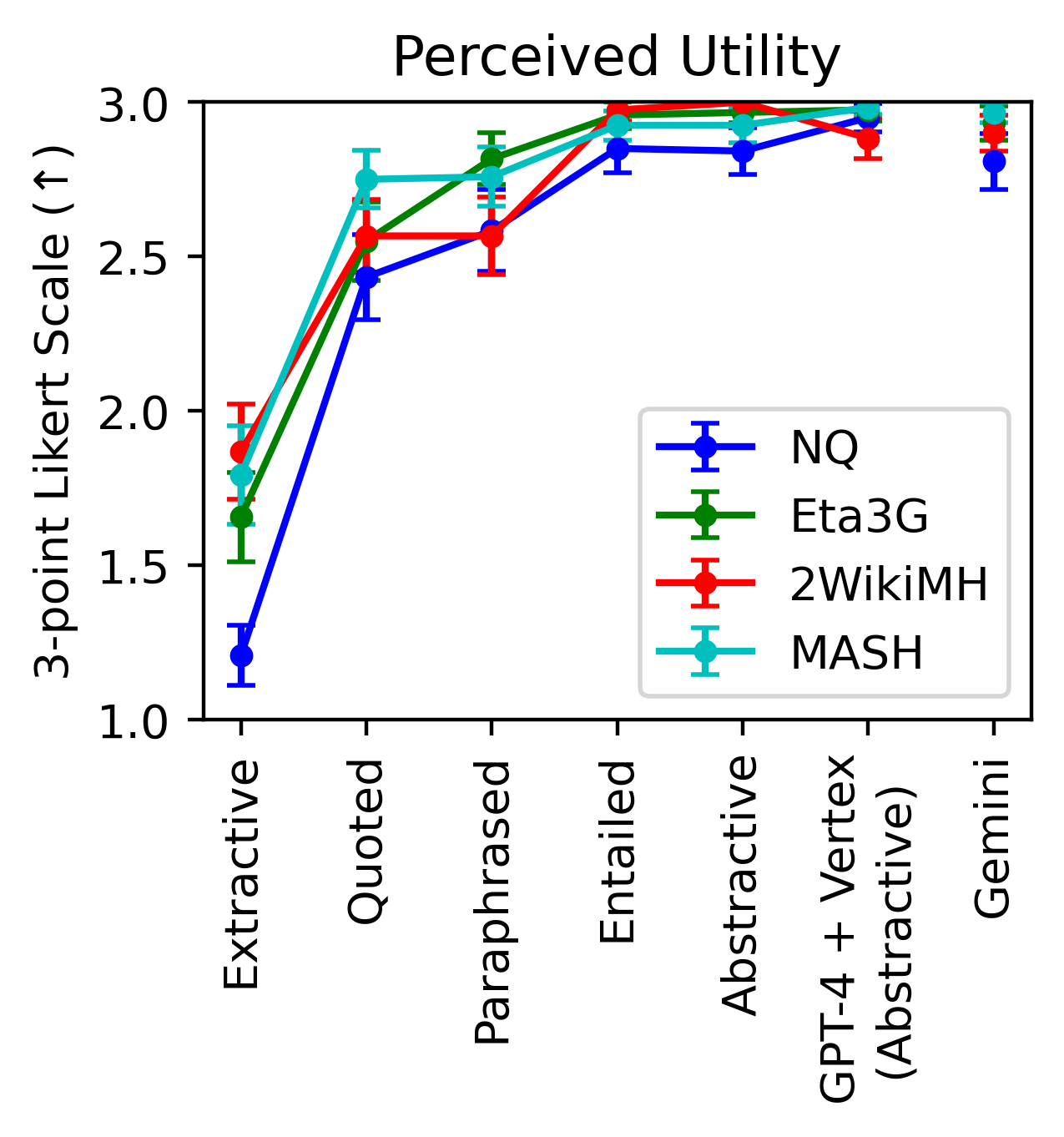}
        \label{fig:utility_by_qd}
    \end{subfigure}
    \begin{subfigure}
        \centering
        \includegraphics{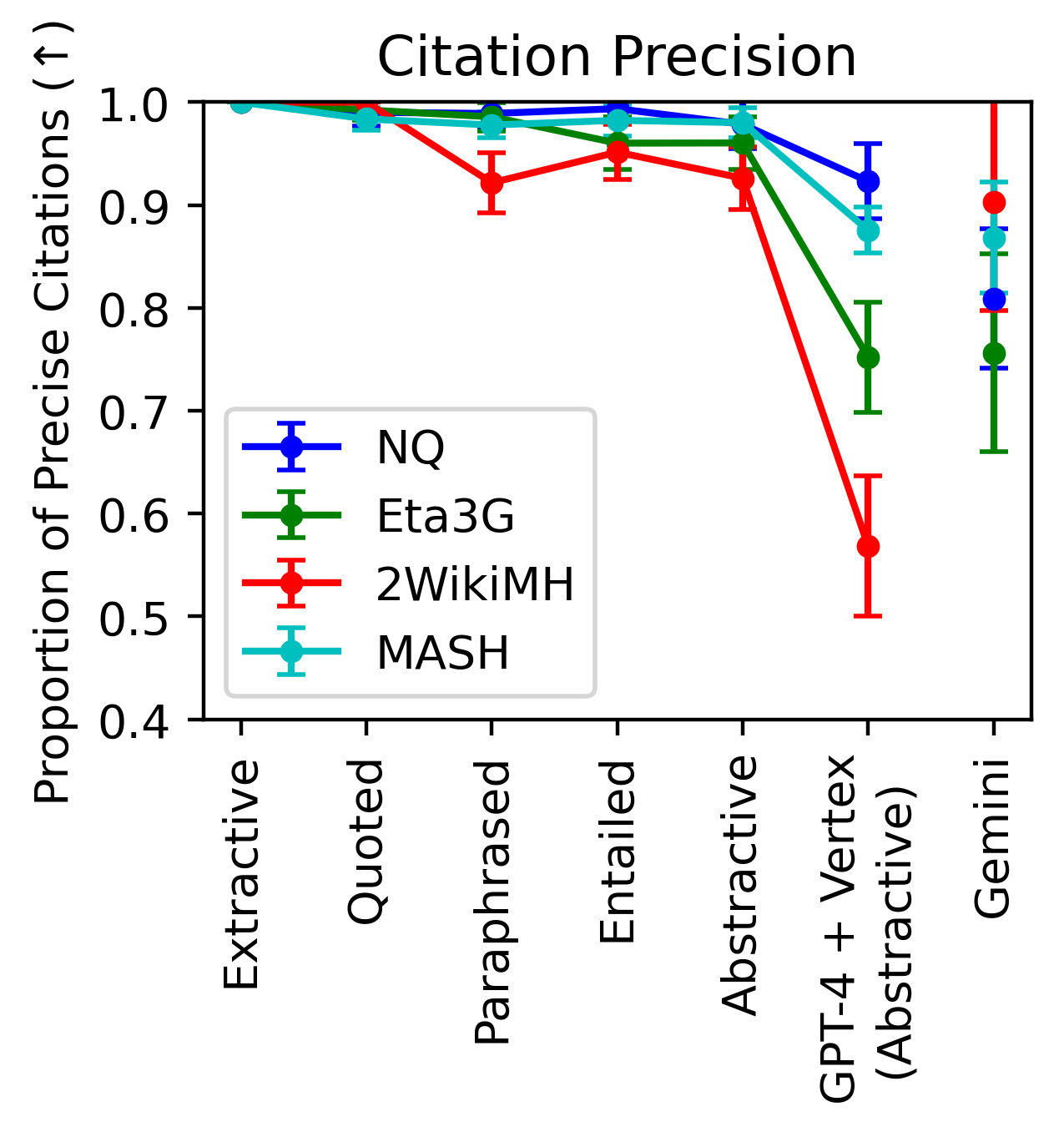}
        \label{fig:precision_by_qd}
    \end{subfigure}
    }
    \resizebox{\textwidth}{!}{
    \begin{subfigure}
        \centering
        \includegraphics{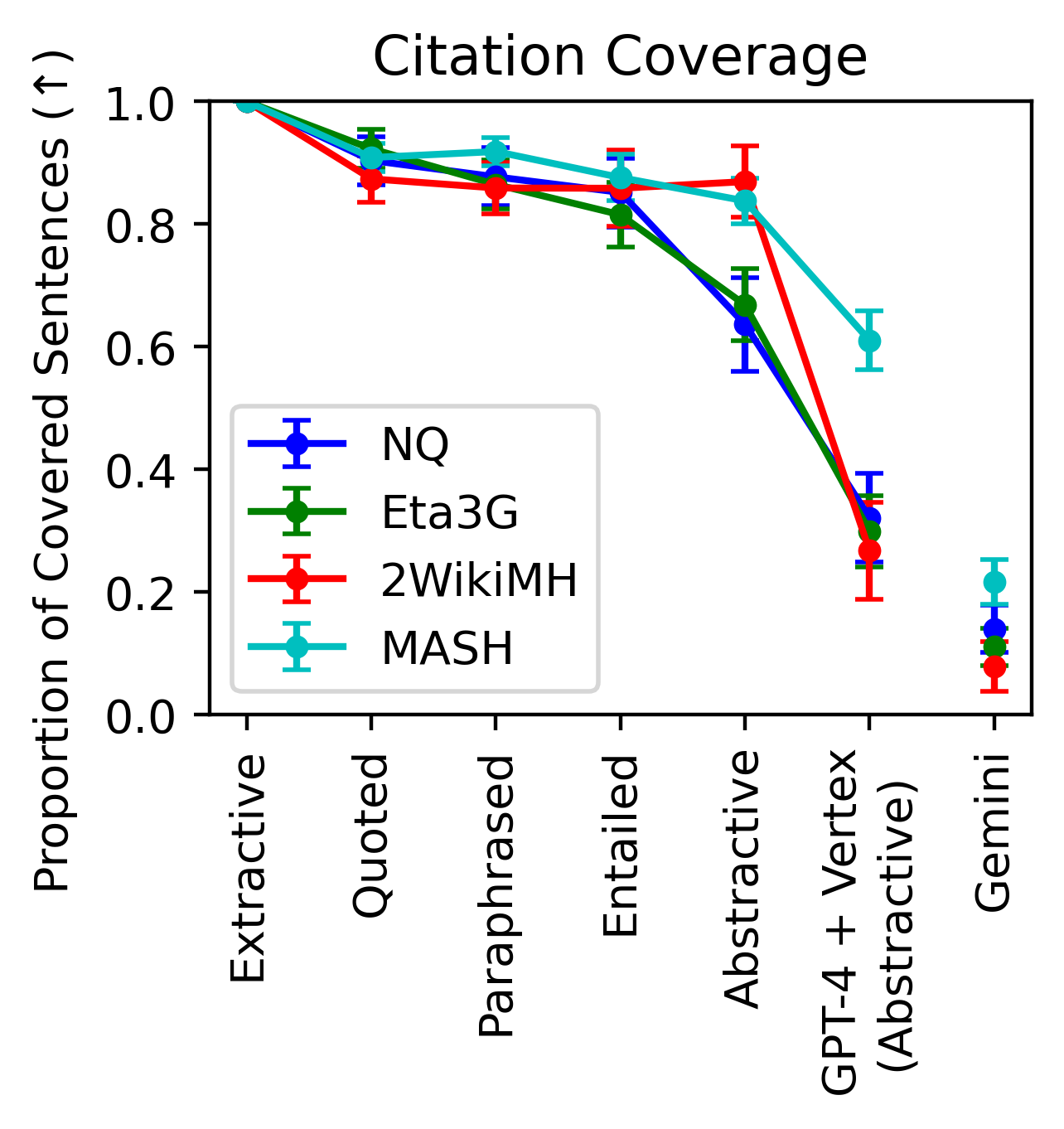}
        \label{fig:coverage_by_qd}
    \end{subfigure}
    \begin{subfigure}
        \centering
        \includegraphics{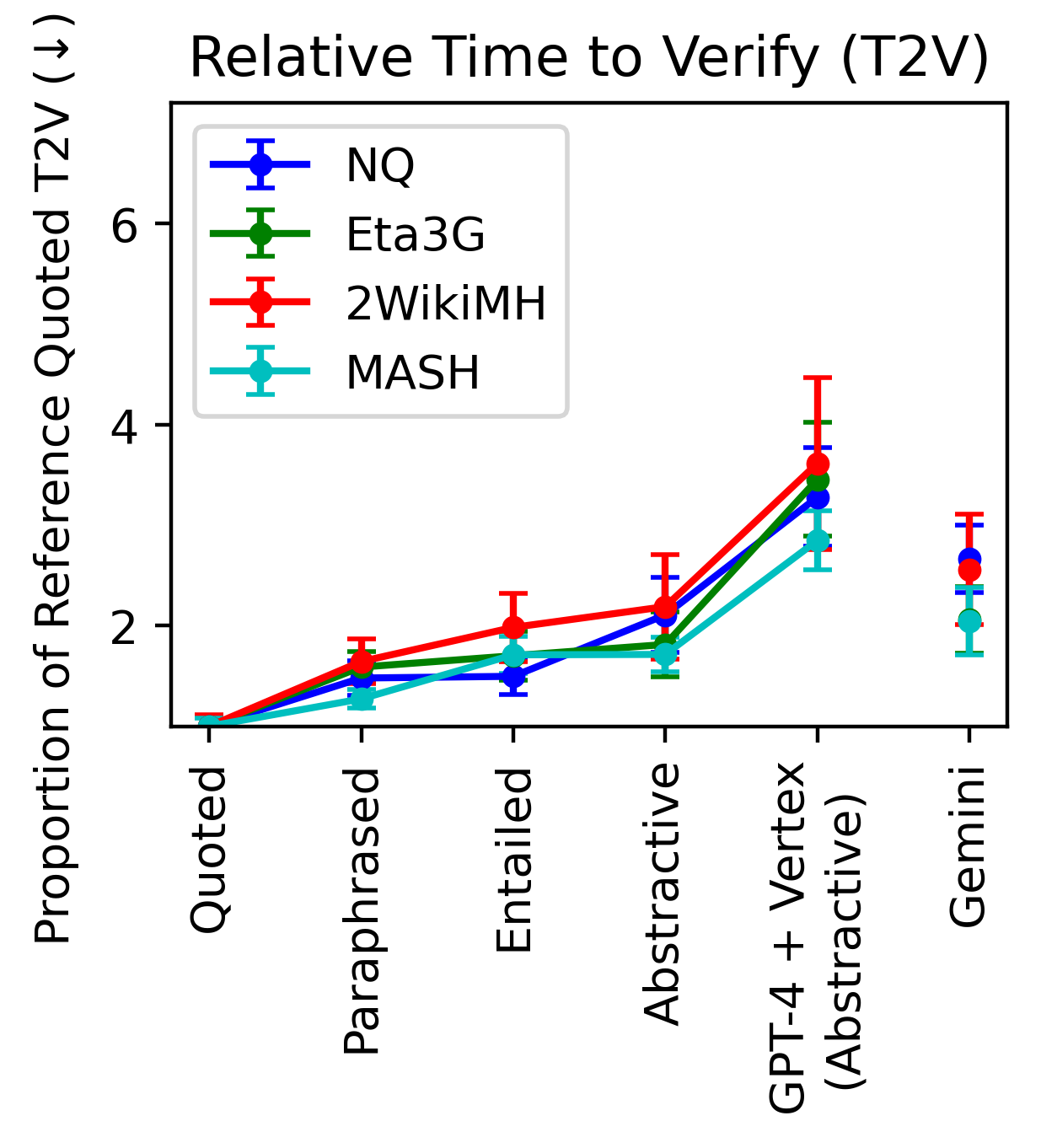}
        \label{fig:t2v_by_qd}
    \end{subfigure}
    \begin{subfigure}
        \centering
        \includegraphics[width=.48\textwidth]{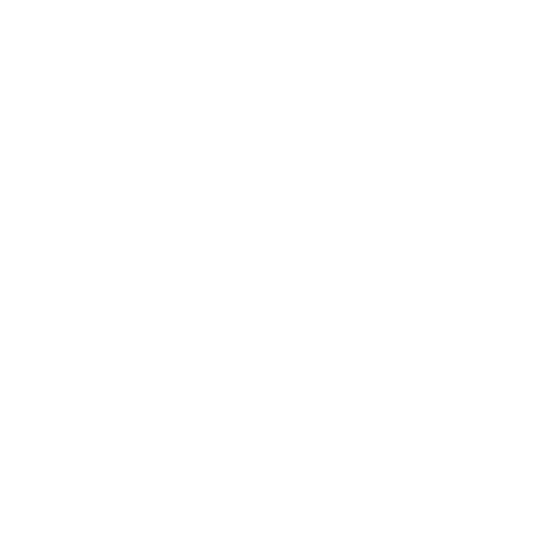}
        \label{fig:blank}
    \end{subfigure}
    }
\caption{Human evaluation results by query distribution. As generations become more abstractive, fluency increases similarly for all query distributions, while perceived utility increases at different OPs for different query distributions. Citation precision and coverage decrease while relative T2V increases across each query distributions as generations become more abstractive. The error bars represent 95\% confidence intervals.}
\label{fig:results_by_qd}
\end{figure*}

For each query distribution individually, fluency and perceived utility increase, citation precision and coverage decrease, and T2V increases with generation abstraction (Figure \ref{fig:results_by_qd}). Furthermore, different query distributions reap gains in perceived utility at different OPs. Perceived utility improves between the quoted and entailed OPs for all evaluated query distributions, but stagnates between the quoted and paraphrased OPs for 2WikiMH and MASH queries (Figure \ref{fig:results_by_qd}, top middle). We provide examples of the shortcomings of the quoted and paraphrased OPs that likely contribute to this effect.

\refstepcounter{egbox}
\begin{tcolorbox}[adjusted title=Example \theegbox: Value of abstraction for multi-hop queries, colback=white, arc=4mm, fontupper=\sffamily]
\raggedright
\noindent
\textbf{2WikiMH Query: }Do both films La Vuelta De Martín Fierro and Joy Division (2006 Film) have the directors from the same country?\vspace{.2cm}\\\textbf{Quoted Generation: }"La Vuelta de Martín Fierro is a 1974 Argentine film directed by Enrique Dawi" [1]. "Enrique Dawi( 1927- 1988) was an Argentine film director and screenplay writer" [2]. "Joy Division is a 2006 British- German- Hungarian film directed by Reg Traviss" [3]. "Reginald Stephen' Reg' Traviss( born 12 February 1977) is a British film director and writer" [4].\vspace{.2cm}\\\textbf{Paraphrased Generation: }'La Vuelta de Martín Fierro,' a film from Argentina released in 1974, was directed by Enrique Dawi [1]. Enrique Dawi (1927-1988) was recognized for his contributions to Argentine cinema as both a director and a screenwriter [1][2]. On the other hand, 'Joy Division,' a film produced in collaboration between Britain, Germany, and Hungary in 2006, was helmed by Reg Traviss [3][4]. Reginald Stephen 'Reg' Traviss, born on February 12, 1977, is known for his work in the British film industry as both a director and a screenwriter [3][4].\vspace{.2cm}\\\textbf{Entailed Generation: }No, 'La Vuelta de Martín Fierro' was directed by Argentine Enrique Dawi, while 'Joy Division' was directed by British filmmaker Reg Traviss [1][3][4].
\label{utility_failure_premises_only}
\end{tcolorbox}

Queries that require multihop reasoning, such as 2WikiMH queries, achieve significant gains in perceived utility at the entailed OP. Example \ref{utility_failure_premises_only} shows how entailed generations can draw logical conclusions entailed by the cited source quotes, unlike the quoted and paraphrased generations. This example is consistent with how the perceived utility of generations for 2WikiMH queries in Figure \ref{fig:results_by_qd} remains the same between the quoted and paraphrased OPs and significantly increases for the entailed OP. 

Perhaps due to the use of gold sources with a plethora of relevant information, MASH queries often result in lengthy quoted generations. Only the entailed and abstractive OPs, not the paraphrased OP, can make generations more concise relative to cited source quotes, as seen in Example \ref{utility_failure_verbose}. This improvement in concision is consistent with how the MASH perceived utility is steady across the quoted and paraphrased OPs and increases for the entailed OP in Figure \ref{fig:results_by_qd}. 
\refstepcounter{egbox}
\begin{tcolorbox}[adjusted title=Example \theegbox: Value of abstraction for improving concision, colback=white, arc=4mm, fontupper=\sffamily]
\raggedright
\noindent
\textbf{MASH Query: }How often should I check my blood sugar if I have diabetes?\vspace{.2cm}\\\textbf{Quoted Generation: }"When you have diabetes, you may need to check your blood sugar throughout the day" [1]. "Give yourself a blood sugar test as often as your doctor advises" [2]. "To do it, you use a gadget that pricks your finger with a tiny needle" [3]. "You'll put a drop of blood onto a test strip" [4]. "The strip goes into a handheld device that measures your blood sugar level" [5]. "Record the test results, so you can share it with your doctor" [6]. "A1c Test: This is a blood test you'll get in your doctor's office at least twice a year, or as often as he recommends" [7]. "The results show your average blood sugar control for the past 2 to 3 months" [8]. "Continuous Glucose Monitoring System: If you choose this method, your doctor will place a tiny sensor under your skin to check blood sugar levels every 5 minutes" [9].\vspace{.2cm}\\\textbf{Entailed Generation: }If you have diabetes, check your blood sugar as frequently as recommended by your doctor using a blood glucose meter, which involves pricking your finger for a blood sample [1][2][3]. Record your results for doctor consultations [6]. Additionally, undergo an A1c test at least twice a year to monitor your average blood sugar over the past few months [7][8]. Alternatively, consider a Continuous Glucose Monitoring System for real-time blood sugar tracking [9].
\label{utility_failure_verbose}
\end{tcolorbox}

In contrast, generations for queries that require style change, like the Eta3G queries, steadily increase in perceived utility across the quoted, paraphrased, and entailed OPs (Figure \ref{fig:results_by_qd}). This effect is seen in Example \ref{utility_failure_style}, where the paraphrased generation rewords the quoted generation to be more suitable to a third-grader and the entailed generation further simplifies the presented information. By studying the 2WikiMH, MASH, and Eta3G query distributions, we find that different tasks require different levels of generation abstraction with respect to cited source quotes.

\refstepcounter{egbox}
\begin{tcolorbox}[adjusted title=Example \theegbox: Value of abstraction for style change, colback=white, arc=4mm, fontupper=\sffamily]
\raggedright
\noindent
\textbf{ETa3G Query: }Explain to a third-grader: what is the hierarchy of the episcopal church\vspace{.2cm}\\\textbf{Quoted Generation: }"In contrast, the Episcopal Church puts an emphasis on shared authority" [1]. "It operates under a polity that includes both episcopal and synodical governance" [2]. "The Episcopal Church is governed by a bicameral General Convention, which meets every three years, and by an Executive Council during interim years" [3]. "The General Convention consists of the House of Bishops and the House of Deputies" [4].\vspace{.2cm}\\\textbf{Paraphrased Generation: }The Episcopal Church focuses on everyone having a say in how things are run [1]. It uses a system that combines leaders making decisions and groups of people coming together to decide things [2]. This church is led by a big meeting called the General Convention that happens every three years, and by a group called the Executive Council in the years in between [3]. At the General Convention, there are two main groups: one is made up of bishops, and the other is made up of deputies [4].\vspace{.2cm}\\\textbf{Entailed Generation: }The Episcopal Church is like a big team that makes decisions together [1]. It has leaders called bishops and other members who meet every three years to decide on important things [3]. They work together to make sure everyone agrees [1].
\label{utility_failure_style}
\end{tcolorbox}
\section{Discussion: Optimal operating point differs across tasks}
\label{sec:discussion_results}
Due to the observed trade-offs across the extractive-abstractive spectrum, there is no one-size-fits-all operating point across applications. In particular, the low verification complexity and high citation coverage of more extractive OPs are often desirable in high-stakes settings that require frequent verification, as indicated by our user preference survey (Figure \ref{fig:EQUIP_EA}, top left; Figure \ref{fig:prolific_pp}). On the other hand, more abstractive generations with higher fluency and perceived utility may be more desirable for settings where verification is less important. The extractive-abstractive trade-offs motivate a segmentation of QA settings into three categories (Table \ref{tab:discussion_summary}). Model developers should consider the priorities of their application area in the context of trade-offs across the extractive-abstractive spectrum and design their cited generation systems to achieve favorable operating points. 

\begin{table}[h]
\centering
\begin{tabular}{p{0.17\linewidth} p{0.11\linewidth} p{0.63\linewidth}}
\toprule
\textbf{QA Settings} & \textbf{Ideal OPs} & \textbf{Examples} \\
\midrule
\begin{tabular}[c]{@{}l@{}}High-stakes settings\\ with dispersed \\information \\in presentable form \\ (Section \ref{subsec:high_stakes_presentable})\end{tabular} & 
\begin{tabular}[c]{@{}l@{}}- Extractive \\ - Quoted\end{tabular} & 
\begin{tabular}[c]{@{}l@{}}- Supporting lawyers in finding and reading legal documents \\ - Assisting healthcare providers in reading patient histories \& case studies \\ - Supporting researchers in finding and digesting relevant papers\end{tabular} \\
\midrule
\begin{tabular}[c]{@{}l@{}}Low-stakes settings \\with information\\ in need of refinement \\ (Section \ref{subsec:low_stakes_refinement})\end{tabular} & 
- Abstractive & 
\begin{tabular}[c]{@{}l@{}}- General writing assistance for grammar, rephrasing, and vocabulary \\ - Creative writing (e.g., slogans, video game character dialogue) \\ - Brainstorming (e.g., social event planning, gift shopping)\end{tabular} \\
\midrule
\begin{tabular}[c]{@{}l@{}}High-stakes settings \\ with information\\ in need of refinement \\ (Section \ref{subsec:high_stakes_refinement})\end{tabular} & 
\begin{tabular}[c]{@{}l@{}}- Paraphrased \\ - Entailed\end{tabular} & 
\begin{tabular}[c]{@{}l@{}}- Writing clinical medical notes from patient visit dialogue \\ - Simplifying medical records for patient understanding \\ - Simplifying legal documents for layperson understanding\end{tabular} \\
\bottomrule
\end{tabular}
\caption{A categorization of QA settings with examples best served by different operating points on the extractive-abstractive spectrum. The first row corresponds to high-stakes settings that require high verifiability and seek pieces of information within a broader corpus written in the same or similar style. The second row refers to low-stakes settings that do not necessarily require high verifiability but instead seek pieces of information reformulated into new ideas or a different style of writing. The third and most challenging row corresponds to high-stakes settings that require high verifiability as well as logical transformations of source information or stylistic changes.}
\label{tab:discussion_summary}
\end{table}

\subsection{High-stakes settings with dispersed information in presentable form}
\label{subsec:high_stakes_presentable}
In some QA settings, queries require a search over refined corpora for information that is already written in an appropriate form, but may be dispersed over multiple sources. In this search setting, LLMs can play a critical role in processing queries to aid with retrieval and in generating responses that stitch together information across multiple sources. Due to the importance of verifiability in these high-stakes settings and the limited involvement of LLMs in generating responses, these settings are best served by more extractive operating points, such as quoted and extractive generations. 

An example of this setting is the use of LLMs to assist lawyers in writing legal documents. Errors in legal documents arising from misinformation from LLM systems lead to serious consequences, including suspension of legal licenses, fines, and reputational harm.\footnote{\href{https://www.cbsnews.com/colorado/news/colorado-lawyer-artificial-intelligence-suspension/}{https://www.cbsnews.com/colorado/news/colorado-lawyer-artificial-intelligence-suspension/}}$^{,}$\footnote{\href{https://www.cbsnews.com/news/chatgpt-judge-fines-lawyers-who-used-ai/}{https://www.cbsnews.com/news/chatgpt-judge-fines-lawyers-who-used-ai/}} Because verifiability is critical in legal applications of LLMs, extractive and quoted operating points may best suit this application, given the current limitations of LLMs. For example, Harvey\footnote{\href{https://www.harvey.ai/}{https://www.harvey.ai/}} has deployed an LLM-based instantiation of the extractive operating point for law firms. As long as lawyers themselves are formulating the legal arguments, the logical transformations achieved by more abstractive operating points are not necessary, whereas citations to relevant sections of case materials are critical. 

\subsection{Low-stakes settings with information in need of refinement}
\label{subsec:low_stakes_refinement}
Other settings require reformulating and recombining information from original sources into a different style. In cases where information errors have insignificant consequences, verifiability is a low priority. Accordingly, the abstractive operating point is ideal for these settings. 

Low-stakes LLM applications seeking creativity, such as creative writing assistance \citep{10.1145/3544548.3581225} or brainstorming light-hearted ideas for further consideration \citep{shaer2024aiaugmentedbrainwritinginvestigatinguse}, do not require verifiability. Rather, these applications intend to generate novel content not necessarily found in existing sources; the abstractive operating point is the only operating point that can achieve this goal. Citations may still be desired to determine whether generated content is taken from other sources. In particular, \cite{lee2023language} study verbatim, paraphrase, and idea plagiarism, which roughly correspond to the cases of quoted, paraphrased, and entailed cited generations, without citation.

\subsection{High-stakes settings with information in need of refinement}
\label{subsec:high_stakes_refinement}
In other QA settings, queries require reformulating information from one or more sources into a different style while providing high verifiability. These are the most challenging settings to address because verifiability trades off with generation abstraction.

Healthcare is a high-stakes domain for LLM applications; errors in generation may lead to serious consequences regarding patient health and legal action. Tasks such as transforming patient visit dialogue into medical notes \citep{biswas2024intelligent, leong2024gen} require high faithfulness to the original dialogue and substantive style change from colloquial patient descriptions of health issues to standard medical terminology. Similarly, tasks such as the simplification of patient medical records for layperson understanding \citep{jeblick2024chatgpt, 7776380} or the simplification of legal contracts for lower reading grade levels \citep{ali2024bridging} also require both high verifiability and significant changes in style. While the extractive and quoted OPs prioritize verifiability, they are not expressive enough to effectively transform the style of the generation for these applications (see Section \ref{subsubsec:utility_style_failure}). Paraphrased and entailed generations provide more of a balance between verifiability and stylistic change. However, the risk of an erroneous generation leading to severe consequences remains significant; experts are still required to verify generations for these applications, underscoring the importance of simple verification \citep{ali2024bridging, mathews2024aihealthcare, jeblick2022chatgptmakesmedicineeasy}. 

Under current extractive-abstractive spectrum trade-offs, none of the operating points achieve high verifiability and perceived utility simultaneously, as is desirable for high-stakes tasks that require frequent verification in addition to style change and logical transformation. Such applications would be best served by future implementations of the entailed operating point that further improve citation quality and new strategies for citation that reduce verification complexity. In particular, novel approaches to citation in settings simplifying terminology should be designed to be accessible and educational to the target audience. 
\section{Failure Analysis} 
\label{sec:failure_analysis}
In this section, we categorize observed failure modes of generation utility, fluency, and citation quality. Understanding these failure modes in relation to the operating points and methods used for cited output generation provides insight for future methods. We divide our failure analysis into four sections. We examine fluency and utility failures in the quoted OP (Section \ref{subsec:uf_failure_quoted_outputs}), citation precision failures (Section \ref{subsec:citation_precision_failures}), citation coverage failures (Section \ref{subsec:citation_coverage_failures}), and abstentions (Section \ref{subsec:abstentions}). For each section, we carefully review generations to determine failure categories. Then, we manually code randomly sampled generations to obtain the frequency of each failure mode. We additionally provide illustrative examples of the different failure modes.


\begin{table}[h]
\centering
\begin{tabular}{lcc}
\toprule
 & \% Improved by Paraphrased OP & \% Improved by Entailed OP \\ 
\midrule
Fluency & 91.4 & 98.5 \\ 
Perceived Utility & 65.4 & 89.6 \\ 
\bottomrule
\end{tabular}
\caption{The percentage of queries with quoted generations assigned a fluency (perceived utility) rating lower than the max with corresponding paraphrased (entailed) generations with the maximum fluency (perceived utility) rating.}
\label{tab:utility_fluency_improvements}
\end{table}

\subsection{Fluency and utility failures}
\label{subsec:uf_failure_quoted_outputs}
While more verifiable than more abstractive OPs, the quoted OP struggles with lower perceived utility and fluency. Table \ref{tab:utility_fluency_improvements} shows that nearly all fluency failures and the majority of utility failures in quoted generations are addressed by their paraphrased and entailed counterparts. This section focuses on quoted generations and highlights failures uniquely prevalent in this OP. We categorize these failures to guide improvements in future quoted OP implementations. Table \ref{tab:quoted_utility_failure_prevalences} contains the prevalence of five perceived utility failure modes discussed further below.

\begin{table}[h]
\centering
\begin{tabular}{lcccccc}
\toprule
 & \begin{tabular}[c]{@{}c@{}}Overall \\ prevalence of \\ low utility\end{tabular} 
 & Verbose 
 & \begin{tabular}[c]{@{}c@{}}Premises without\\ conclusions\end{tabular} 
 & \begin{tabular}[c]{@{}c@{}}Unsuitable \\ style\end{tabular} 
 & \begin{tabular}[c]{@{}c@{}}Failure to interpret\\ sources in-context\end{tabular} 
 & \begin{tabular}[c]{@{}c@{}}Failure in \\ information \\ retrieval\end{tabular} \\ 
\midrule
NQ      
& 40/131 
& \textbf{37.5\%}   
& 10\%                                                                      
& 2.5\%                                                        
& \textbf{25.0\%}                                                                             
& \textbf{27.5\%} \\ 

Eta3G   
& 39/141 
& \textbf{43.6\%}   
& 7.7\%                                                                      
& \textbf{71.8\%}                                                       
& 12.8\%                                                                             
& 5.1\% \\ 

2WikiMH 
& 40/153 
& \textbf{62.5\%}   
& \textbf{52.5\%}                                                                      
& 0\%                                                        
& 2.5\%                                                                             
& 2.5\% \\ 

MASH    
& 25/129 
& \textbf{80\%}   
& 8\%                                                                      
& 0\%                                                        
& 4\%                                                                             
& 8\% \\ 
\bottomrule
\end{tabular}
\caption{Prevalence of different perceived utility failure categories in the quoted generations by query distribution. We evaluate 40 low utility generations, or as many are available, from each query distribution. The overall prevalence column on the left provides the occurrence of these utility failures (any generation without the highest perceived utility rating) among sampled quoted generations. The other columns provide the rate of the specified failure category out of all utility failures in the row, with prevalent failure categories for each query distribution in bold. A utility failure may fall under more than one category. The failures in information retrieval for 2WikiMH and MASH were due to faulty gold sources.}
\label{tab:quoted_utility_failure_prevalences}
\end{table}

\subsubsection{Fluency failure: Disfluent inline quote integration}
\label{subsubsec:disfluent_quoted_outputs}
Fluency failures predominantly arise from awkward inline integrations of quotes. The choice of quote may also contribute to this failure; some quotes may be easier to incorporate inline that others. For instance, the second quote in Example \ref{fluency_failure} is not integrated fluently. This disfluency is corrected in the paraphrased output.

\refstepcounter{egbox}
\begin{tcolorbox}[adjusted title=Example \theegbox: Value of abstraction for improving fluency, colback=white, arc=4mm, fontupper=\sffamily]
\raggedright
\noindent
\textbf{ETa3G Query: }Explain to a third-grader: when does the next allegiant movie come out\vspace{.2cm}\\\textbf{Quoted Generation: }"The Divergent Series: Allegiant – Part 1" [1] was released on March 18, 2016, and there will not be a next movie as "Ascendent, but the film never saw the light of day" [2].\vspace{.2cm}\\\textbf{Paraphrased Generation: }The movie "The Divergent Series: Allegiant – Part 1" came out on March 18, 2016 [1]. There won't be another movie called "Ascendent" because it was never made [2].
\label{fluency_failure}
\end{tcolorbox}

\subsubsection{Utility failure: Verbose}
\label{subsubsec:verbose_quoted_outputs}
Some responses to queries have low utility because they are unnecessarily long. Verbose generations are a common perceived utility failure mode across all query distributions (Table \ref{tab:quoted_utility_failure_prevalences}). The quoted generation in Example \ref{utility_failure_verbose} contains relevant information that can be condensed into a more concise response, as in the entailed generation. Furthermore, Table \ref{tab:word_count_by_op} illustrates that the paraphrased generation does not reduce word count relative to the quoted generation. Only the entailed and more abstractive OPs achieve shorter generations than those of the quoted OP.


\begin{table}[ht]
\centering
\begin{tabular}{lc}
\toprule
\textbf{Type of Generation} & \textbf{Generation Word Count} \\ 
\midrule
Snippet                     & 236 \\ 
Quoted                      & 57  \\ 
Paraphrased                 & 60  \\ 
Entailed                    & 33  \\ 
Abstractive                 & 38  \\ 
GPT-4 + Vertex (Abstractive) & 34  \\ 
Gemini                      & 48  \\ 
\bottomrule
\end{tabular}
\caption{The average number of words for each generation type. The snippet, quoted, and paraphrased generations are longer than the entailed and more abstractive generations.}
\label{tab:word_count_by_op}
\end{table}

\subsubsection{Utility failure: Premises without the conclusion}
\label{subsubsec:no_conclusion_quoted_outputs}
Sometimes, a generation has low utility because it only enumerates the premises of the answer, without drawing the query's sought-after conclusion. Table \ref{tab:quoted_utility_failure_prevalences} shows that this failure mode is most common in quoted generations for the 2WikiMH query distribution, likely because multi-step reasoning often requires drawing conclusions not explicitly stated in premises from sources. 

The quoted generation in Example \ref{utility_failure_premises_only} presents premises without drawing the conclusion that addresses the query. Unlike the quoted and paraphrased generations, the entailed generation succeeds in presenting the conclusion "No" that answers the question.

\subsubsection{Utility failure: Unsuitable style}
\label{subsubsec:utility_style_failure}
Quoted generations may also have low utility due to being written in a style that does not suit the query. Table \ref{tab:quoted_utility_failure_prevalences} shows that this failure mode is most common in quoted generations for the Eta3G query distribution, which follows from relatively few internet sources being written for third-graders. 

The quoted generation in Example \ref{utility_failure_style} relies on vocabulary, such as "bicameral", "interim", and "synodical", that third-graders typically do not understand. In contrast, the paraphrased generation simplifies vocabulary and the entailed generation is written in a significantly more appropriate style than the quoted generation.

\subsubsection{Utility failure: Failure to interpret sources in-context}
\label{subsubsec:utility_failure_interpret}
Sometimes, quoted generations fail to quote parts of the in-context sources that contain the answer. This failure mode affects any OP implementation that generates responses using in-context sources, such as the reference OP implementations. Table \ref{tab:quoted_utility_failure_prevalences} shows that this failure mode is more common for NQ and Eta3G queries than for queries accompanied by gold-standard sources, suggesting that even among relevant sources, some are more interpretable than others.

In Example \ref{utility_failure_misinterpretation}, even though the retrieved sources contain the answer to the query, Elaine Stritch, the quoted generation does not provide this pertinent information. This issue is addressed in the abstractive generation, where parametric knowledge is permitted. We include the retrieved sources for this example in appendix resource \ref{utility_failure_misinterpretation_full}.

\refstepcounter{egbox}
\begin{tcolorbox}[adjusted title=Example \theegbox: Value of abstraction for filling in missing knowledge, colback=white, arc=4mm, fontupper=\sffamily]
\raggedright
\noindent
\textbf{NQ Query: }who plays jack's mom in 30 rock 2\vspace{.2cm}\\\textbf{Quoted Generation: }"From 2007 to 2012, she had a recurring role as Colleen Donaghy on the NBC sitcom 30 Rock, a role that won her a third Emmy in 2007" [1].\vspace{.2cm}\\\textbf{Abstractive Generation: }Elaine Stritch played Jack Donaghy's mother, Colleen Donaghy, on the NBC sitcom "30 Rock" from 2007 to 2012, a role that earned her an Emmy in 2007 [1].
\label{utility_failure_misinterpretation}
\end{tcolorbox}

\subsubsection{Utility failure: Failure in information retrieval}
\label{subsubsec:utility_failure_retrieval}
The generation process may also fail to produce a useful response if the sources provided in-context are not relevant enough to address the query. In these cases, the failure originates with the information retrieval system. Again, this failure mode affects any OP implementation that generates using in-context sources. As expected, Table \ref{tab:quoted_utility_failure_prevalences} shows that this failure mode is more common for NQ queries, where we retrieve sources, than for 2WikiMH and MASH, which provide gold-standard sources. Although Eta3G sources are also retrieved, this failure mode appears less common for Eta3G queries, perhaps due to lower annotator expectations of information content in generations suitable for third graders. From the user perspective, this failure case is indistinguishable from that of Example \ref{utility_failure_misinterpretation}, where more abstractive generations are sometimes able to provide an answer from parametric knowledge. We provide an actual example of this failure mode in the appendix (\ref{utility_failure_info_retrieval_full}).

\subsection{Citation precision failures} 
\label{subsec:citation_precision_failures}
We categorize different types of citation precision failures to the end of improving future cited generation systems. In this section of the failure analysis, we focus on the quoted, entailed, GPT-4 + Vertex, and Gemini generations. Table \ref{tab:precision_prevalence} contains the prevalence of each of the three precision failure modes discussed below.

\begin{table}[h]
\centering
\begin{tabular}{lcccc}
\toprule
 & \begin{tabular}[c]{@{}c@{}}Overall prevalence \\ of these failure categories\end{tabular} & \begin{tabular}[c]{@{}c@{}}Claim taken \\ out of context\end{tabular} & \begin{tabular}[c]{@{}c@{}}Irrelevant source \\ quote: Keywords \\ as proxies\end{tabular} & \begin{tabular}[c]{@{}c@{}}Relevant source \\ quote: Claim \\ too specific\end{tabular} \\ 
\midrule
Quoted       & 3/1383  & \textbf{100\%} & 0\%  & 0\%  \\ 
Entailed     & 23/998  & 4.3\%           & 34.8\% & \textbf{60.9\%} \\ 
GPT-4 + Vertex & 34/193  & 14.7\%          & \textbf{67.6\%} & 17.6\% \\ 
Gemini       & 28/244  & 3.6\%           & \textbf{75.0\%} & 21.4\% \\ 
\bottomrule
\end{tabular}
\caption{Prevalence of different precision failure categories in the quoted, entailed, GPT-4 + Vertex, and Gemini generations. The overall prevalence column on the left provides the occurrence of these precision failure categories out of sampled citations. The other columns provide the rate of the specified failure category out of all precision failures in the row, with the most common failure category for each row in bold. We evaluate 40 imprecise citation annotations, or as many are available, from each implementation. False positive imprecise citation annotations are discussed in the appendix (\ref{subsubsec:citation_subjectivity}).}
\label{tab:precision_prevalence}
\end{table}

\subsubsection{Precision failure: Claim taken out of context}
\label{subsubsec:precision_failures_out_of_context}

Taking a cited source quote out of context, with respect to the source, results in a citation precision error. Even if one part of a source appears to provide support for a claim, other parts of the source may indicate otherwise. Table \ref{tab:precision_prevalence} shows that this is the only failure type that occurs for quoted generation citations; because quoted generations are not re-worded, cited quotes are predominantly misrepresented by being taken out of context \citep{zhang-etal-2023-extractive}. Other operating points are susceptible to this failure mode as well.

The quoted generation in Example \ref{precision_failure_out_of_context} quotes an answer option for a multiple choice question. Taken out of context, the citation appears to be precise, but a source presenting the claim as a multiple choice option is not the same as a source presenting the claim as the truth.

\refstepcounter{egbox}
\begin{tcolorbox}[adjusted title=Example \theegbox: Quote taken out of context, colback=white, arc=4mm, fontupper=\sffamily]
\raggedright
\noindent
\textbf{NQ Query: }right to property according to the constitution of india is a\vspace{.2cm}\\\textbf{Quoted Generation: }"Right to property is legal right but not a Fundamental Right" \textcolor{orange}{[1]}.\vspace{.2cm}\\\textbf{Source 1: }\href{https://www.toppr.com/ask/question/the-right-to-private-property-in-india-today-is-a/}{https://www.toppr.com/ask/question/the-right-to-private-property-in-india-today-is-a/}\\Consider the following statements regarding Right to Property:  1. \textcolor{orange}{[1] Right to property is legal right but not a Fundamental Right.}  2. Article 300 A was inserted in the Constitution of India by the 44th Constitutional Amendment.  Which of the statements given above is/are correct?
\label{precision_failure_out_of_context}
\end{tcolorbox}

\subsubsection{Precision failure: Relevant keywords serving as proxies for supporting claims} 
\label{subsubsec:precision_failures_relevant_keywords}

Some citation precision errors are due to a source quote being mistakenly flagged as entailing a target sentence when it in fact does not, simply due to the source sharing similar keywords with the sentence. Table \ref{tab:precision_prevalence} shows that this type of error is common for GPT-4 + Vertex and Gemini generations, perhaps because Vertex citations and some Gemini citations are post-hoc, i.e., identified after generation occurs from sources not necessarily used in generation. This failure mode is far less frequent for the quoted and entailed OP implementations, which directly generate responses from the cited source quotes.

In Example \ref{precision_failure_isource}, citation 3 discusses the International Organization for Standardization (ISO) document language style, rather than plain English, which is the subject of the query. Citation 3 was likely mistakenly included because the words used to describe the purpose of ISO language style are similar to those used to describe the purpose of plain English.

\refstepcounter{egbox}
\begin{tcolorbox}[adjusted title=Example \theegbox: Cited source quote only relevant up to keywords, colback=white, arc=4mm, fontupper=\sffamily]
\raggedright
\noindent
\textbf{NQ Query: }explain what is meant by plain english and why is it used\vspace{.2cm}\\\textbf{GPT-4 + Vertex Generation: }Plain English is a style of writing that uses simple, straightforward language to communicate clearly [0]. It is used to ensure that information is accessible and easily understood by a wide audience [1][2]\textcolor{orange}{[3]}.\vspace{.2cm}\\\textbf{Source 1: }\href{https://www.iso.org/ISO-house-style.html}{https://www.iso.org/ISO-house-style.html}\\Punctuation helps make meaning clear, but there are changes in practice over time and different writers adopt different approaches. \textcolor{orange}{[3] The priority for ISO documents is to be easily understandable and unambiguous.} Use punctuation consistently within a document. and/or The phrase 'and/or' is often used in English to express 'either or both' of two options. The meaning can be ambiguous, especially in translation to other languages where the '/' is not a recognized punctuation mark. Avoid using 'and/or' in a document to avoid confusion and misapplication. Use the construction 'either x or y, or both' instead. and or \& ISO documents do not use the ampersand (\&) in ordinary running text within the document or in the document title, table titles or figure titles. Use the word 'and' instead. Exceptionally, the ampersand is found in certain limited cases, including within code and abbreviated terms.
\label{precision_failure_isource}
\end{tcolorbox}

\subsubsection{Precision failure: Supporting claim is highly relevant, but not specific enough}

Some citation precision errors result from identifying a source claim that is relevant to the generation, but makes a broader claim than the generation. This mismatch can make it difficult for a reader to confirm support for the generation without additional research. Table \ref{tab:precision_prevalence} shows that this type of error is common for the entailed generations, perhaps because rewording the source quote can distort the original claim to be more specific. 

In Example \ref{precision_failure_rsource}, while the generation specifies the "coxal bone of the pelvis", citation 1 discusses "the hip", a more general term.

\refstepcounter{egbox}
\begin{tcolorbox}[adjusted title=Example \theegbox: Cited source quote highly relevant yet misses a detail, colback=white, arc=4mm, fontupper=\sffamily]
\raggedright
\noindent
\textbf{NQ Query: }the joint between a coxal bone of the pelvis and the sacrum\vspace{.2cm}\\\textbf{GPT-4 + Vertex Generation: }The joint between a coxal bone of the pelvis and the sacrum is known as the sacroiliac joint \textcolor{orange}{[1]}. This joint plays a crucial role in transferring weight and movement between the upper body and legs.\vspace{.2cm}\\\textbf{Source 1: }\href{https://www.spine-health.com/conditions/spine-anatomy/sacroiliac-joint-anatomy}{https://www.spine-health.com/conditions/spine-anatomy/sacroiliac-joint-anatomy}\\\textcolor{orange}{[1] The unique sacroiliac articulation between the sacrum and the hip provides a transition between the spine and the pelvis.} While the SI joint has the features of a typical moveable joint, the movements within the joint are very limited. The sacroiliac joint has a C-shaped contour A fully developed adult sacroiliac joint is formed by two C-shaped (or inverted L-shaped) layers of bone with several irregular ridges and depressions.1Vleeming A, Schuenke MD, Masi AT, Carreiro JE, Danneels L, Willard FH. The sacroiliac joint: an overview of its anatomy, function and potential clinical implications. J Anat. 2012;221(6):537-567. doi:10.1111/j.1469-7580.2012.01564.x
\label{precision_failure_rsource}
\end{tcolorbox}

\begin{table}[ht]
\centering
\begin{tabular}{lcccc}
\toprule
\textbf{}                  & \begin{tabular}[c]{@{}c@{}}Overall prevalence \\ of coverage failures\end{tabular} & \begin{tabular}[c]{@{}c@{}}Sentences with \\ no citation, \\ i.e., $Q_i = \emptyset$\end{tabular} & \begin{tabular}[c]{@{}c@{}}Sentences that only \\ contain imprecise \\ citation, \\ i.e., $Q_i \neq \emptyset$, \\ $u_{i, \text{cited claims}} = \emptyset$\end{tabular} & \begin{tabular}[c]{@{}c@{}}Sentences partially \\ covered with \\ precise citation, \\ i.e., $u_{i, \text{cited claims}} \neq \emptyset$\end{tabular} \\ 
\midrule
Quoted      & 134/1381 & \textbf{67.9\%} & 6.0\%  & 26.1\%  \\ 
Paraphrased & 145/1288 & 29.0\%          & 12.4\% & \textbf{58.6\%}  \\ 
Entailed    & 115/776  & 20.0\%          & 13.0\% & \textbf{67.0\%}  \\ 
Abstractive & 216/907  & 42.1\%          & 6.0\%  & \textbf{51.9\%}  \\ 
GPT-4 + Vertex    & 520/920  & \textbf{67.7\%} & 11.7\% & 20.6\%  \\ 
Gemini            & 1157/1391 & \textbf{83.8\%} & 5.5\%  & 10.7\%  \\ 
\bottomrule
\end{tabular}
\caption{Prevalence of different coverage failure categories across all OP implementations and Gemini. The overall prevalence column on the left contains the occurrence of coverage failures out of all sentences. The other columns provide the percentage of coverage failures in that row that are occurrences of the specified failure category. The most common failure category for each row is in bold.}
\label{tab:coverage_prevalence}
\end{table}

\subsection{Citation coverage failures}
\label{subsec:citation_coverage_failures}
A sentence lacking citation coverage is either the result of there being no citation at all, only imprecise citation, or precise citation that only \textit{partially covers} the sentence. Table \ref{tab:coverage_prevalence} shows the prevalence of these three failure modes. 

\label{subsubsec:coverage_failures}
\subsubsection{Coverage failure: No citations}
Table \ref{tab:coverage_prevalence} shows that a complete lack of citation is the most common failure mode of the quoted generations. This means that the citation coverage failure is evident upon first glance of a sentence, leading to low verification complexity. 

The absence of citations is also the most common failure mode for GPT-4 + Vertex and Gemini. Additionally, these implementations respectively have over 6x and 8x the coverage failure rate of the quoted generations. These findings for the GPT-4 + Vertex generations underscore the challenges of identifying post-hoc citations, in contrast to the reference abstractive OP implementation, which has both a lower coverage failure rate and partial coverage in a majority of failure cases. Lastly, Gemini coverage failures arise from the absence of citations at the highest rate across all evaluated systems. This may be the result of deployment considerations such as latency and user-interface design that limit citation to select sentences. 

\subsubsection{Coverage failure: Partial coverage}
Some coverage errors result from citations that are precise, but are not sufficient to support all claims within a sentence. Table \ref{tab:coverage_prevalence} shows that in contrast to the quoted generations, the majority of paraphrased and entailed generation coverage failures result from partial coverage. This pattern follows expectations that increasingly abstractive generations synthesize information across sources, and therefore require multiple citations to properly cover a sentence. Furthermore, the partial coverage failure mode is not instantly recognizable as a coverage error; verification complexity is higher than for the case of no citations because users must verify each provided citation. 

In Example \ref{coverage_failure_partial_coverage}, the second sentence of the generation is not covered. While the claim that the song was on the film soundtrack for \textit{The Great Gatsby} is supported, the claim that the song was released as part of movie's soundtrack in 2013 is not supported by citation 2.
\refstepcounter{egbox}
\begin{tcolorbox}[adjusted title=Example \theegbox: Sentence partially covered by citation, colback=white, arc=4mm, fontupper=\sffamily]
\raggedright
\noindent
\textbf{NQ Query: }who sings will you still love me when i'm not young and beautiful\vspace{.2cm}\\\textbf{Gemini Generation: }The song "Will You Still Love Me When I'm No Longer Young and Beautiful" is sung by Lana Del Rey [1]. It was released in 2013 as part of the soundtrack for the film The Great Gatsby \textcolor{orange}{[2]}. The song explores themes of love, aging, and the fear of losing one's beauty [3].\vspace{.2cm}\\\textbf{Source 1: }\href{https://en.wikipedia.org/wiki/Young\_and\_Beautiful\_(Lana\_Del\_Rey\_song)\#:\~:text=\%22Young\%20and\%20Beautiful\%22\%20is\%20a,drama\%20film\%20The\%20Great\%20Gatsby.}{https://en.wikipedia.org/wiki/Young\_and\_Beautiful\_(Lana\_Del\_Rey\_song)\#:\~:text=\%22Young\%20a}\\nded remix) Label Interscope Polydor Water Tower Songwriter(s) Lana Del Rey Rick Nowels Producer(s) Rick Nowels, Al Shux (original version) Dan Heath (DH orchestral version) Lana Del Rey singles chronology 'Dark Paradise' (2013) 'Young and Beautiful' (2013) 'Once Upon a Dream' (2014) Cedric Gervais singles chronology 'Summertime Sadness' (remix)(2013) 'Young and Beautiful' (remix)(2013) 'Adore You' (remix)(2014) Music video'Young and Beautiful' on YouTube'Young and Beautiful' (remix) on YouTube \textcolor{orange}{[2] 'Young and Beautiful' is a song by American singer and songwriter Lana Del Rey used for the soundtrack to the drama film The Great Gatsby.} Contemporary music critics lauded the single, calling it 'haunting' and 'somber'. Lyrically, 'Young and Beautiful' follows a young lover's apprehension about whether love can last. A music video, directed by Chris Sweeney and filmed by Sophie Muller, was released on May 10, 2013. 'Young and Beautiful' reached number one in the Commonwealth of Independent Stat
\label{coverage_failure_partial_coverage}
\end{tcolorbox}

\subsubsection{Coverage failure: Only imprecise citations}

If all the citations provided to a sentence are imprecise, then the sentence is not covered. While this failure mode does not dominate the coverage failures of any OP implementation or Gemini, the non-zero values across this column indicate potential for achieving better coverage by improving precision. 

\subsection{Abstentions}
\label{subsec:abstentions}
A model "abstains" from answering a query when it declines to provide a response. By abstaining from responding to a query, a system may opt out of providing unreliable information at the expense of disrupting the user experience. Model abstentions also reflect failures of different system components; we examine abstention rates due to source retrieval and LLM generation in Tables \ref{tab:abstention_prevalence_retrieved_qds} and \ref{tab:abstention_prevalence_gold_qds}.

\begin{table}[h]
\centering
\begin{tabular}{lcccc}
\toprule
               & \begin{tabular}[c]{@{}c@{}}Overall Abstention Rate \\ on NQ and Eta3G\end{tabular} & \begin{tabular}[c]{@{}c@{}}Due to generation \\ abstention\end{tabular} & \begin{tabular}[c]{@{}c@{}}Due to response \\ retrieval failure\end{tabular} & \begin{tabular}[c]{@{}c@{}}Due to citation \\ retrieval failure\end{tabular} \\ 
\midrule
Reference OPs  & 44 / 300  & 27.3\%    & 72.7\%    & -         \\ 
GPT-4 + Vertex & 4 / 300   & 50.0\%    & -         & 50.0\%    \\ 
Gemini         & 13 / 300  & -         & -         & -         \\ 
\bottomrule
\end{tabular}
\caption{Prevalence of different abstention categories across the reference, GPT-4 + Vertex, and Gemini generations over queries accompanied by retrieved sources. The first column contains the prevalence of abstentions out of 300 randomly sampled NQ and Eta3G queries. The other columns provide the percentage of abstentions in that row that are occurrences of the specified failure.}
\label{tab:abstention_prevalence_retrieved_qds}
\end{table}

\begin{table}[h]
\centering
\begin{tabular}{lcccc}
\toprule
               & \begin{tabular}[c]{@{}c@{}}Overall Abstention Rate \\ on 2WikiMH and MASH\end{tabular} & \begin{tabular}[c]{@{}c@{}}Due to generation \\ abstention\end{tabular} & \begin{tabular}[c]{@{}c@{}}Due to response \\ retrieval failure\end{tabular} & \begin{tabular}[c]{@{}c@{}}Due to citation \\ retrieval failure\end{tabular} \\ 
\midrule
Reference OPs  & 0 / 300  & 0\%       & 0\%       & -         \\ 
GPT-4 + Vertex & 28 / 300 & 17.9\%    & -         & 82.1\%    \\ 
Gemini         & 42 / 300 & -         & -         & -         \\ 
\bottomrule
\end{tabular}
\caption{Prevalence of different abstention categories across the reference, GPT-4 + Vertex, and Gemini generations over queries accompanied by gold-standard sources. The first column contains the prevalence of abstentions out of 300 randomly sampled 2WikiMH and MASH queries. The other columns provide the percentage of abstentions in that row that are occurrences of the specified failure.}
\label{tab:abstention_prevalence_gold_qds}
\end{table}

For the reference OP generations, in which response generation relies upon retrieved sources, a \textit{generation abstention} occurs when retrieved sources address the query, but the LLM still opts out of responding. A \textit{response retrieval failure} occurs when either no sources are surfaced by the retrieval system or the retrieved sources fail to address the query and the model abstains.

For the GPT-4 + Vertex generations, in which response generation does not rely upon retrieved sources but the citation generation does, a \textit{generation abstention} occurs when the LLM opts out of responding. We consider the system to abstain due to a \textit{citation retrieval failure} when retrieval fails for the citation generation step. We report the overall Gemini abstention rate as we do not have access to its retrieval and generation steps.

The high reference OP abstention rates on the two query distributions with retrieved sources (Table \ref{tab:abstention_prevalence_retrieved_qds}) in contrast to the absence of abstention on the query distributions with gold-standard sources (Table \ref{tab:abstention_prevalence_gold_qds}) indicate the importance of a strong retriever when generating model responses grounded in retrieved sources. GPT-4 + Vertex and Gemini likely suffer from high abstention rates in Table \ref{tab:abstention_prevalence_gold_qds} because 2WikiMH responses frequently require long-tail knowledge.
\section{Discussion: Recommendations for future work}
\label{sec:failure_analysis_discussion}
Given the results of our failure analysis, we identify three key areas for future work in improving cited generation systems across the extractive-abstractive spectrum. Our discussion here is complementary to our recommendations for cited generation systems in applied settings included in Section \ref{sec:discussion_results}.

\subsection{Improving post-hoc citations}

The low citation precision and coverage of GPT-4 + Vertex and Gemini, relative to those of the reference implementations, indicate that post-hoc citations are challenging to identify. The predominant precision failure for GPT-4 + Vertex and Gemini citations is when semantic similarity between an irrelevant source and a generated sentence is mistaken for the source entailing the generated sentence (Section \ref{subsubsec:precision_failures_relevant_keywords}). This result reflects the tendency of models to rely on the spurious correlation of word overlap between source and target sentence for the natural language inference (NLI) task of determining whether the source entails the target sentence \citep{mccoy-etal-2019-right}. Post-hoc citation identification is further complicated by the practical implementation of selecting candidate sources for their semantic similarity to the generated text; we emphasize the importance of ensuring the robustness of NLI methods from the benchmark setting to downstream application settings.

While continued improvement in NLI capabilities is critical for post-hoc citation identification, we highlight that these two tasks are not identical. NLI classifies the meaning relation between a text and a single source. In contrast, citation identification is a combinatorial search problem: given a text and a set of candidate sources, the task is to identify a subset of the sources that together entail the text. For instance, a response to a query that requires multi-hop reasoning may only be entailed by a set of sources and no individual source alone. For instance, the claim “The maternal grandfather of X is Z” is only entailed by the union of sources that state “The mother of X is Y” and “The father of Y is Z”; neither of these sources alone entail the claim. The combinatorial complexity of identifying post-hoc citations makes the paradigm of grounding a generation in specific source quotes pre-designated for citation particularly appealing.

The especially low citation coverage of GPT-4 + Vertex and Gemini generations, with the majority of uncovered sentences without any citation, may reflect the failure mode of seeking sources that entail the claim individually. After all, the Vertex API provides scores indicating the extent to which individual sources support the provided text and Gemini provides at most one citation per sentence. Unfortunately, the naive approach for identifying an entailing subset of sources requires combinatorial applications of the NLI module. To address this, we hope to see future work on citation identification that first decomposes text into sub-claims and then individually supports each sub-claim with a single source. \cite{min-etal-2023-factscore} shows promising results with such decompositions for fact verification, while \cite{wanner-etal-2024-closer} demonstrates that decomposition strategy affects verification results. Further work is required to obtain effective claim decompositions, especially on generations for multi-hop queries, where important sub-claims may be implicit. Furthermore, citation identification for users requires low verification complexity. Accordingly, the subset of cited sources should not be redundant; the removal of any one source should nullify the entailment relation. Future work on these initial results from fact verification should also ensure the resulting set of cited sources is deduplicated and verifiable by humans with low T2V. 

\subsection{Task-specific information retrieval}
Throughout the extractive-abstractive spectrum, information retrieval (IR) is fundamental for different steps in cited output generation. We encourage system developers to implement distinct IR systems for three different tasks. The first task of \textbf{query-source} retrieval identifies sources likely to contain an answer, given a query. Query-source retrieval is broadly recognized as a fundamental IR task for QA systems that ground generations in relevant sources \citep{karpukhin-etal-2020-dense}. However, even if a relevant source is retrieved, we observe in Section \ref{subsubsec:utility_failure_interpret} that responses grounded in retrieved sources may focus on irrelevant sections of the source. This failure mode illuminates a second finer-grain \textbf{query-quote} retrieval task of identifying specific quotes within a retrieved source, given a query. This task is fundamental to the quoted OP and the "attribute-first-then-generate" paradigm from \cite{slobodkin2024attributefirstgeneratelocallyattributable}, which both require identifying quotes within relevant sources to answer a given query. For the quoted OP, this retrieval task has additional objectives such as identifying concise quotes that are also possible to fluently incorporate inline.

The third task of \textbf{text-entailment} retrieval identifies sources that are not only relevant to a given text, but \textit{entail} the text. Systems that identify post-hoc citations rely on this task. This task is often implemented in two stages. First, sources are retrieved for their semantic similarity to the given text. Second, an NLI method is applied to re-rank the retrieved sources. We observe in Section \ref{subsubsec:precision_failures_relevant_keywords} that the most common citation precision failure is due to the cited source having semantic similarity to the text, rather than sharing an entailment relation with the text. While shortcomings in current NLI systems contribute to this failure mode, it also motivates an IR system that seeks entailment relations, rather than semantic similarity, from the start. \cite{10.5555/1667583.1667603} has considered this IR task before, but to the best of our knowledge, modern dense retrieval methods have not yet been used to implement this task. 

\subsection{Leveraging different OPs within the same system}

Some LLM systems need to handle a variety of queries. In Section \ref{sec:discussion_results} we discuss how different queries are best served by different operating points; to best serve users across \textit{all} queries, we recommend that general-purpose systems consider routing different types of queries to different operating points. For instance, a query for health advice should be routed for a more extractive generation than a query for birthday party ideas because health advice carries higher stakes. In light of the perceived utility failures in Section \ref{subsec:uf_failure_quoted_outputs}, queries requiring style change or concise logical conclusions should be routed to the entailed or abstractive operating points. A thoughtful method for routing queries to different points across the extractive-abstractive spectrum must either leverage personalized verifiability preferences or learn a nuanced understanding of the potential downstream consequences of different queries on human decisions and actions. Alternatively, systems can directly provide users with a choice between different operating point generations for any given query. 

We also recommend that system developers consider interpolating between operating points within a generation for an individual query. For example, pertinent information in a generation may need to be more verifiable than other parts of the generation. Such pertinent information could be quoted and the rest of the generation could be written as a more abstractive OP, similar to how quotations are used in journalism to communicate important points from relevant sources. This approach requires a method that recognizes certain sentences or claims as requiring more verifiability than others; it is a generalization of the task of determining which sentences require citation, which itself is challenging (Appendix \ref{subsec:results_on_all_sentences}). By leveraging multiple operating points within the same system or generation, developers can design systems to provide useful and verifiable generations for diverse queries.

\section{Related Works} 
\label{sec:related_works}

\subsection{Entailed Generation Systems}
\label{subsec:entailed_systems}
Many works have proposed systems that achieve the entailed operating point. For example, \cite{nakano2021webgpt} use reinforcement learning from human preferences (RLHP) to fine-tune WebGPT which interacts with a web interface and generates responses with citation to web page passages. \cite{menick2022teaching} also use RLHP to fine-tune GopherCite to generate responses with citation to succinct source quotes retrieved through Google Search. \cite{thoppilan2022lamda} fine-tune LaMDA models to generate responses with citation to URLs. \cite{gao2022rarr} and \cite{chen2023purrefficientlyeditinglanguage} propose systems that identify citations to evidence snippets after a response has been generated and revised in accordance with retrieved sources, rather than generating the original response and citations from retrieved sources. \cite{shao-etal-2024-assisting} and \cite{jiang-etal-2024-unknown} generate long reports with citations to entailing source snippets and entire sources, respectively. \cite{slobodkin2024attributefirstgeneratelocallyattributable} propose the \textit{"Attribute First, then Generate"} method, in which specific source quotes are designated as citations for each sentence even before the response is generated. This paradigm is shared by our reference implementation, in which the specific quotes incorporated into the quoted generation are then used to generate the paraphrased, entailed, and abstractive cited generations. \textit{While the aforementioned systems implement the entailed operating point, our work studies different operating points spanning the extractive-abstractive spectrum.} 

\subsection{Extractive and Quoted Generation Systems}
\label{subsec:extractive_quoted_systems}
Extractive generations have long been considered for QA \citep{kolomiyets2011survey}. More recent works explore the use of quotations in generations, although selected quotes are often entities or sentence fragments and are too short to contain full claims. Notably, \cite{schuster2024semqasemiextractivemultisourcequestion} propose the task of Semi-Extractive Multi-Source Question Answering (SEMQA), in which short quotes of factual dates, entities, and statements are "glued together" with generated text.  Both \cite{weller-etal-2024-according} and \cite{zhang2024verifiabledesignaligninglanguage} propose stepping away from citations in favor of improving verifiability by quoting pre-training data, using prompting and reward-base optimization respectively. In these works quoting pre-training data, verifiability is measured with a character 25-gram metric that accepts short quotations. It is unclear where these methods fall on the extractive-abstractive spectrum. While these works require source quotes to be substrings of the generated sentence, \textit{our quoted operating point additionally specifies that claims within cited source quotes should entail the generated sentence.} Additionally, we provide a quoted generation implementation that leverages up-to-date non-parametric knowledge, but future implementations of the quoted operating point could leverage parametric content as well.

\subsection{Faithfulness in relation to abstraction}

Prior work in the area of automatic text summarization studies the relationship between summary abstraction and faithfulness \citep{ladhak-etal-2022-faithful, dreyer-etal-2023-evaluating}. Through generating faithfulness-abstractiveness trade-off curves across multiple models, \cite{ladhak-etal-2022-faithful} demonstrate that the abstraction of a summary with respect to the original text, as measured by n-gram metrics, is correlated with the unfaithfulness of the summary across models. \cite{dreyer-etal-2023-evaluating} study this same trade-off for individual models. In contrast to the extractive-abstractive operating points, papers in this line of work focus on a smooth notion of abstraction, rather than distinct levels of abstraction that correspond to interpretable differences in user experience. Furthermore, these works consider neither the task of citation identification nor the verification complexity experienced by users.

In the setting of cited LLM generations, trade-offs between cited generation desiderata have been observed in audits of deployed systems. In studying deployed citation systems that fall under the entailed operating point, \cite{liu2023evaluating} present positive correlation between citation precision and generation extractiveness relative to cited web pages, as measured by BLEU and BERTScore, and negative correlation between citation precision and perceived utility. \textit{This work explores noteworthy trade-offs associated with different levels of abstraction achieved within a single operating point, while we study trade-offs across the entire extractive-abstractive spectrum.} 

\subsection{Evaluating verifiability}
Prior works have also established guidelines for evaluating citations. \cite{rashkin2022measuringattributionnaturallanguage} define the attributed to identified sources (AIS) framework used for human evaluations of whether a statement is interpretable and supported by provided sources. Expanding from the AIS framework, \cite{bohnet2023attributedquestionansweringevaluation} propose a framework for attributed question answering and \cite{gao-etal-2023-enabling} propose an automated evaluation framework for the fluency, correctness, and citation quality of cited generations and demonstrate their correlation with human evaluations. 

Prior work has also considered the wall clock time of citation evaluations. \cite{rashkin2022measuringattributionnaturallanguage} record completion times across the tasks of rating both interpretability and citation coverage for entire responses as a notion annotation quality. \cite{schuster2024semqasemiextractivemultisourcequestion} also reports annotation duration timed across multiple tasks, including those unrelated to citation accuracy evaluation. Most similar to our work, \cite{slobodkin2024attributefirstgeneratelocallyattributable} record annotator completion times for entire responses across only the citation coverage task as a measure of citation verification complexity. \cite{rashkin2022measuringattributionnaturallanguage} observes that annotators became faster at the AIS evaluation throughout their study, perhaps due to increased familiarity with the task. \textit{In order to account for this effect in addition to variation in annotator reading speed when drawing comparisons across different cited generation types, we propose normalizing each annotator's time to evaluate citation coverage by their time on a baseline citation coverage task evaluated in parallel with every cited generation type.} We present this metric as relative T2V for individual sentences in order to capture the user experience verifying sentences with citation.
\section{Conclusion} 
\label{sec:conclusion}

We investigated cited generations across the extractive-abstractive spectrum through the lens of human user experience to inform the design of future information systems. Our work shows that while users perceive generations to be more useful as they become more abstractive, citation coverage is up to 50\% lower and humans take up to three times longer to verify cited sentences. The deployed Google Gemini system is subject to these trade-offs as well; despite having high fluency and perceived utility, only 15.0\% of sentences generated by Gemini are fully supported by citations to sources. Our trade-off results show that designing a one-size-fits-all cited LLM system requires strategically leveraging operating points across the extractive-abstractive spectrum within a single system. Through a failure analysis of generations across the spectrum, we recommend three implementation opportunities for future cited LLM systems: identifying post-hoc citations on the granularity of claims within sentences, combinations of operating points within the same system or generation, and information retrieval for specific source quotes and entailment relations. We hope our work helps model developers situate their application area and build systems that best maximize utility and verifiability. 

One limitation of our study is that our reference implementations do not meet the compute efficiency and latency criteria required in deployed settings. Because citation generation often requires additional rounds of retrieval and inference, we hypothesize that these deployment considerations could exacerbate the already low citation coverage of more abstractive generations. Another limitation stems from our assumption that retrieved sources are inherently trustworthy. Determining whether a source reflects objective truth to the best of human knowledge is a nuanced task that can require domain expertise and examination of sources in relation to one another. Additionally, our study focuses solely on whether citations to sources are accurate, and not on whether these citations effectively assist users in detecting model hallucinations; especially in systems with high T2V, users may not examine citations thoroughly enough to detect unverified content. These limitations reflect directions for future work in building low latency citation systems, identifying trustworthy sources, and understanding the downstream impact of citations on user decision-making.

Understanding the trade-off between information system utility and verifiability is timely. Not only are LLM systems becoming more commonplace across domains, but LLMs are also becoming increasingly capable. We hope that future LLM developments will push the Pareto frontier across fluency, perceived utility and verifiability. While improvements in LLM capabilities may naturally lead to more accurate and comprehensive citations, it is unclear whether generations with such citations will also be easier for users to verify. Without deliberate efforts to control verification complexity in LLM systems capable of addressing increasingly complex queries, verifying generations from such systems may naturally become burdensome for users. Yet, verifiability is critical to navigate our world where disinformation abounds, long-tail knowledge constantly evolves, and LLMs retain bias from training, even as their capabilities scale. We hope the extractive-abstractive spectrum inspires future information systems that play upon the strengths of multiple operating points in order to maximize both utility and verifiability for users.

\section{Acknowledgments} 
\label{sec:acknowledgments}
We thank the 31 annotators whose contributions to the cited generation evaluations made this work possible. We are grateful to Nelson Liu for evaluation advice and resources in early discussions, Percy Liang for evaluation infrastructure, and Marco Tulio Ribeiro for feedback on in-progress results. We also thank Irena Gao, Nicole Meister, Luke Bailey, Marina Zhang, Judy Shen, and other members of the Guestrin and Hashimoto labs for feedback on this paper. This work is funded by the Stanford Institute for Human-Centered Artificial Intelligence. TW is supported by the National Science Foundation Graduate Research Fellowship Program under Grant No. DGE-2146755. Any opinion, findings, and conclusions or recommendations expressed in this material are those of the authors(s) and do not necessarily reflect the views of the National Science Foundation.

\bibliographystyle{unsrtnat}  
\bibliography{references}  

\section{Appendix} 
\label{sec:appendix}

\subsection{Human Preference Survey}
\label{subsec:human_preference_survey}

Our survey first asks participants whether they have used both search engines and LLMs, referred to as \texttt{"chatbots (e.g. ChatGPT, Claude)"} in the survey. We only consider results over participants who have used both search engines and LLMs.

We then ask the following questions.

Question A: \texttt{If you have a question concerning how to handle a personal medical issue, which platform would you use to seek an answer?}

Question B: \texttt{If you have a question concerning topics in a friendly game of trivia, which platform would you use to seek an answer?}

Question C: \texttt{If you had a question concerning how to file your taxes, which platform would you use to seek an answer?}

Question D: \texttt{If you were trying to explain a complicated science topic to your third grade niece, which platform would you use to seek an answer?} 

Question E: \texttt{What are your reasons to use search engines (e.g. Google, Bing) instead of chatbots (e.g. ChatGPT, Claude) when seeking information to answer a query?}

Question F: \texttt{What are your reasons to use chatbots (e.g. ChatGPT, Claude) instead of search engines (e.g. Google, Bing) when seeking information to answer a query?}

For all questions, we randomize the answer choice order across individuals to eliminate bias arising from the order in which answer choices are presented. Questions A-D have the mutually exclusive answer options of:
\begin{enumerate}
  \item \texttt{A traditional search engine (e.g. Google and Bing Search)}
  \item \texttt{A hybrid search engine and chatbot (e.g. Perplexity, Gemini, and CoPilot)}
  \item \texttt{A chatbot (e.g. ChatGPT and Claude)}
\end{enumerate}
Because relatively few study participants have previously interacted with a hybrid system, we report the proportion of participants who chose \texttt{traditional search engine} or \texttt{chatbot} out of these two categories.

We chose the mutually inclusive answer options for Questions E and F by consolidating free-text input from survey respondents in a trial run. The answer options are as follows:
\begin{enumerate}
  \item \texttt{Reliability of provided answers}
  \item \texttt{Convenience of obtaining an answer}
  \item \texttt{Knowing where the answer comes from}
  \item \texttt{For the entertainment factor of interacting with the underlying technology}
  \item \texttt{The variety of sources made available}
  \item \texttt{Seeking a highly personalized answer}
\end{enumerate}

We solicit survey responses through the Prolific platform and compensate workers at \$13.71/hr. 

\subsection{Coverage and T2V results on all generated sentences}
\label{subsec:results_on_all_sentences}
\begin{figure*}[ht]
\centering
\resizebox{.7\textwidth}{!}{
    \begin{subfigure}
        \centering
        \includegraphics{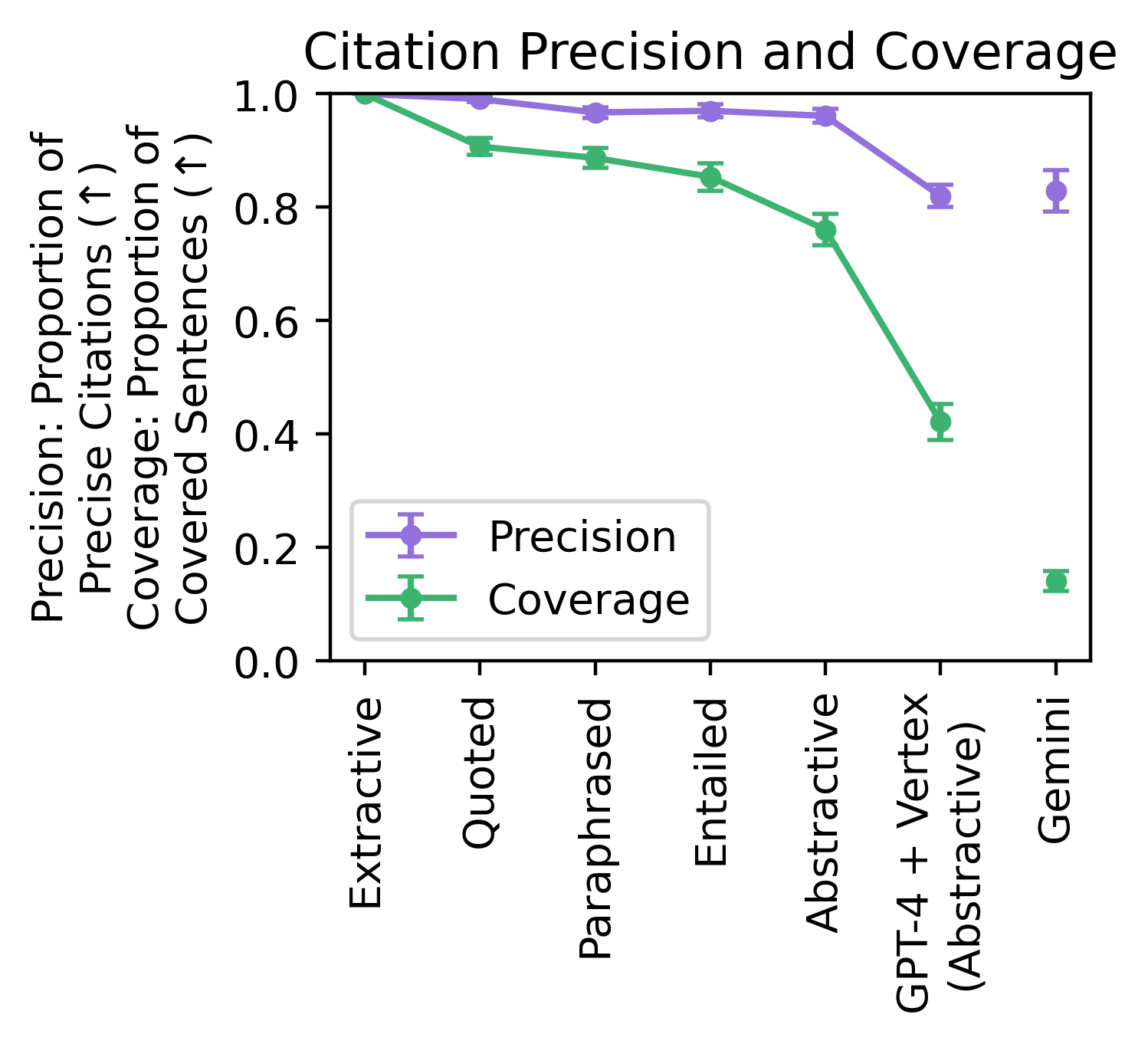}
    \end{subfigure}
    \begin{subfigure}
        \centering
        \includegraphics{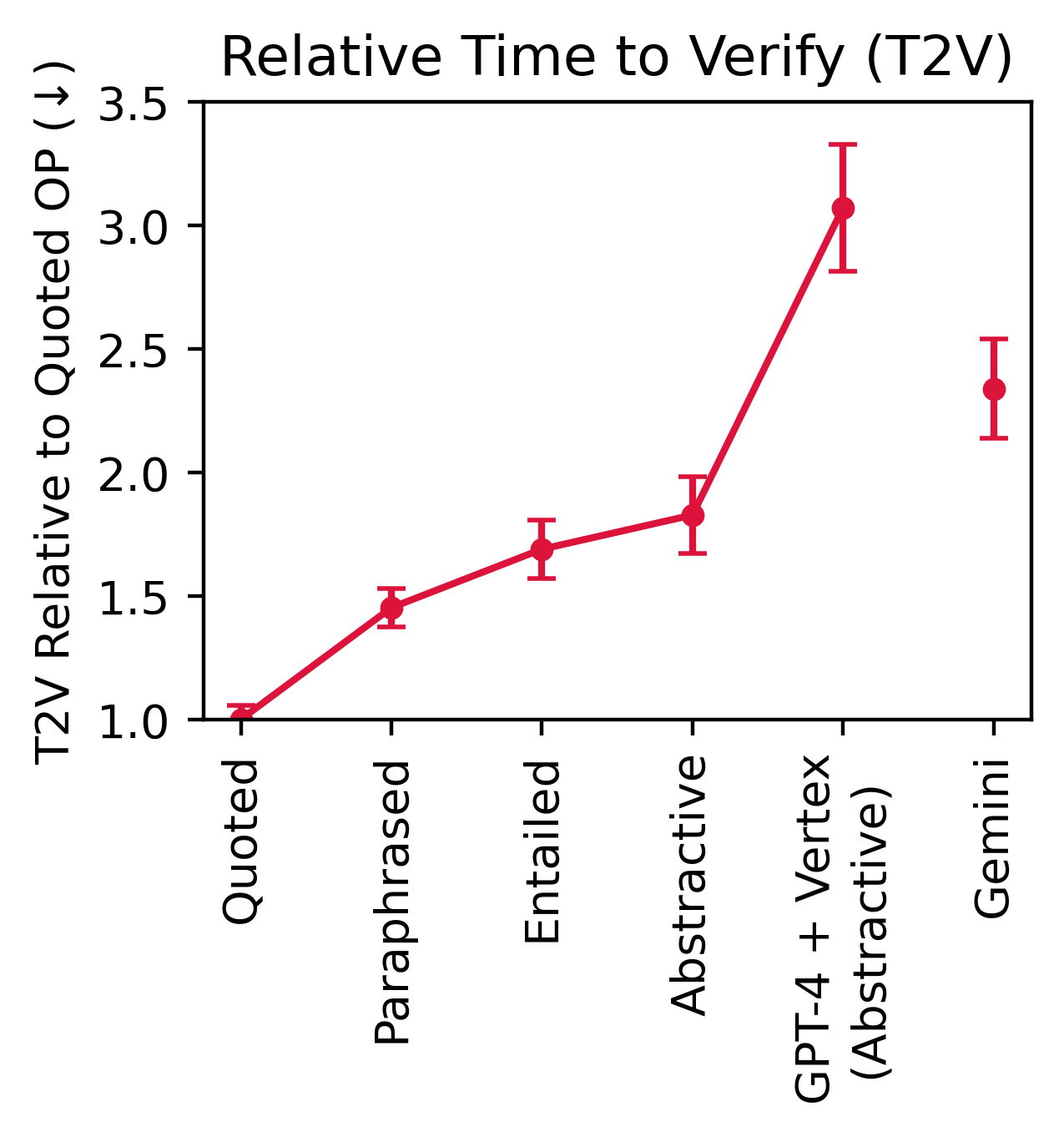}
    \end{subfigure}
    }
\caption{Human evaluation results across the four query distributions for \textit{all} sentences. The error bars represent 95\% confidence intervals.}
\label{fig:results_in_aggregate_all_sentences}
\end{figure*}

\begin{figure*}[ht]
    \raggedright
    \resizebox{\textwidth}{!}{
    \begin{subfigure}
        \centering
        \includegraphics{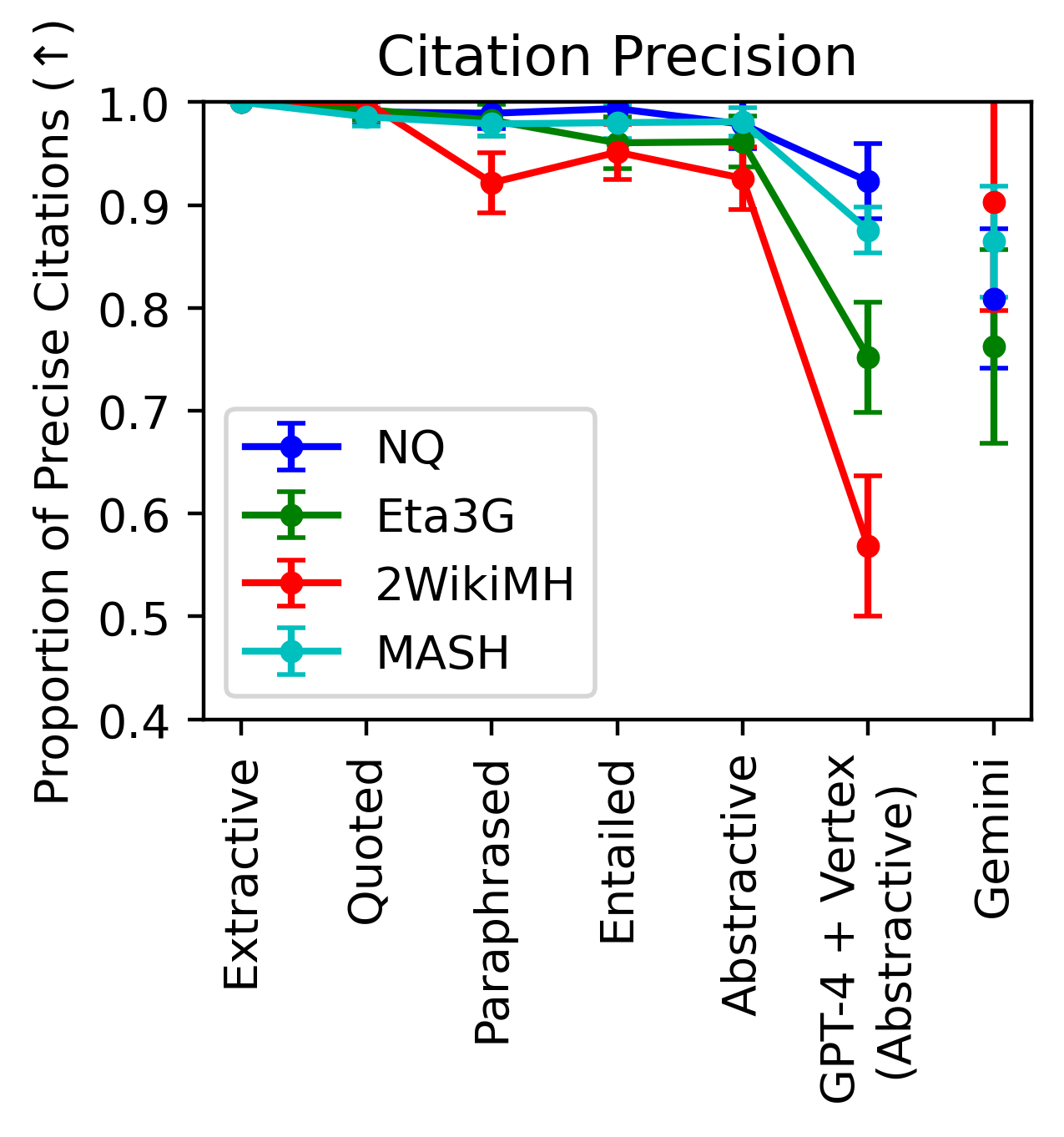}
    \end{subfigure}
    \begin{subfigure}
        \centering
        \includegraphics{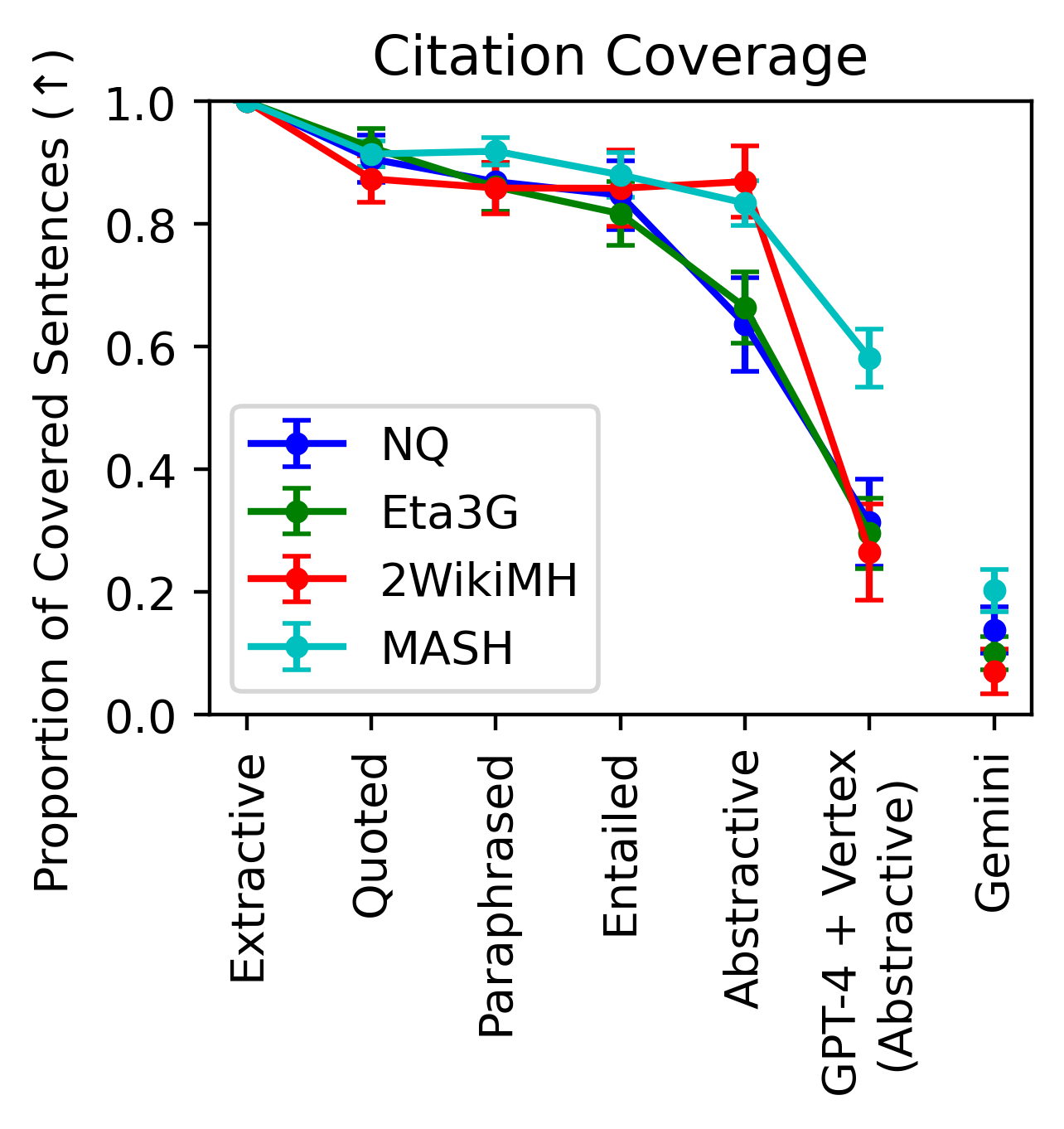}
    \end{subfigure}
    \begin{subfigure}
        \centering
        \includegraphics{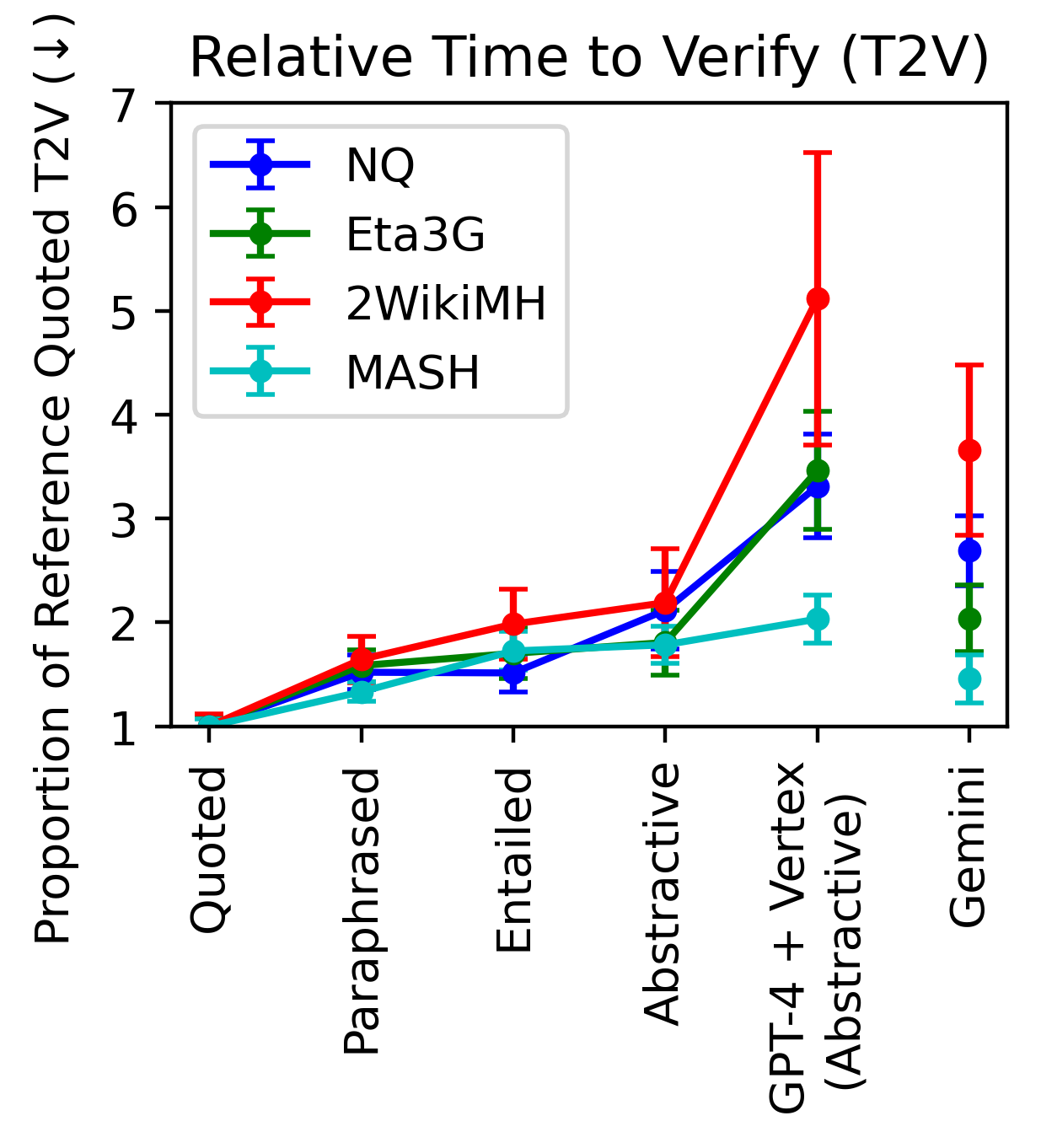}
    \end{subfigure}
    }
\caption{Human evaluation results by query distribution for \textit{all} sentences. The error bars represent 95\% confidence intervals.}

\label{fig:results_by_qd_all_sentences}
\end{figure*}

The citation precision, coverage, and T2V results in Section \ref{sec:results} are evaluated over sentences judged to require citation by the Vertex API. However, we observe that the Vertex API flags imperatives as sentences that do not require citation, even though responses to MASH queries often contain commands that provide health directives (e.g. "Do not give adult cough medicine to children under age 4."). We hope to see future work on the task of determining which sentences require citation. The results in Figures \ref{fig:results_in_aggregate_all_sentences} and \ref{fig:results_by_qd_all_sentences} show that evaluating the citation precision, coverage, and T2V over all generated sentences does not affect the trade-offs observed in Figures \ref{fig:results_in_aggregate} and \ref{fig:results_by_qd}. 

\subsection{Alternative measures of T2V}
\label{subsec:closer_look_at_t2v}

\begin{figure*}[h]
    \centering
    \resizebox{\textwidth}{!}{
    \begin{subfigure}
        \centering
        \includegraphics{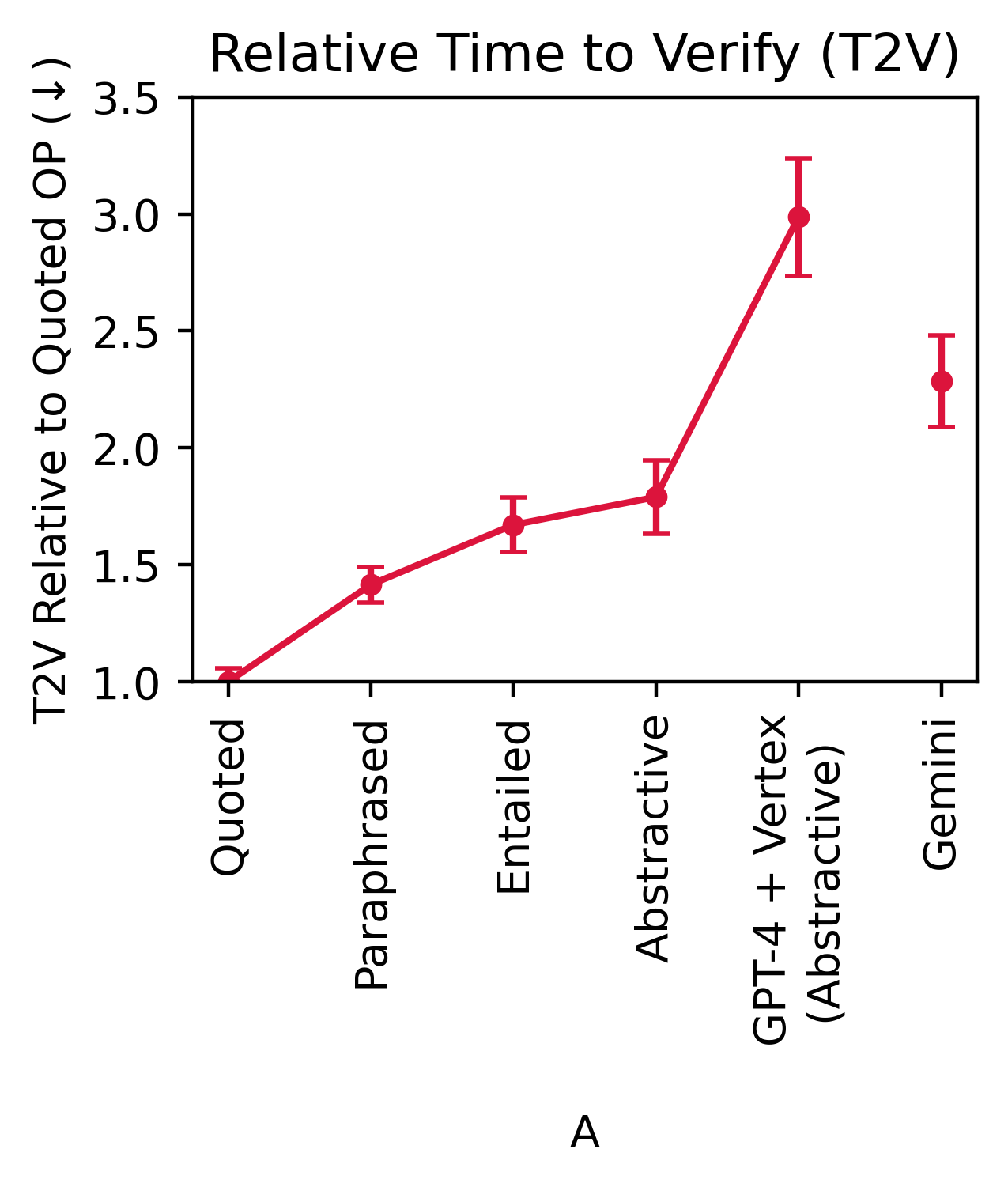}
    \end{subfigure}
    \begin{subfigure}
        \centering
        \includegraphics{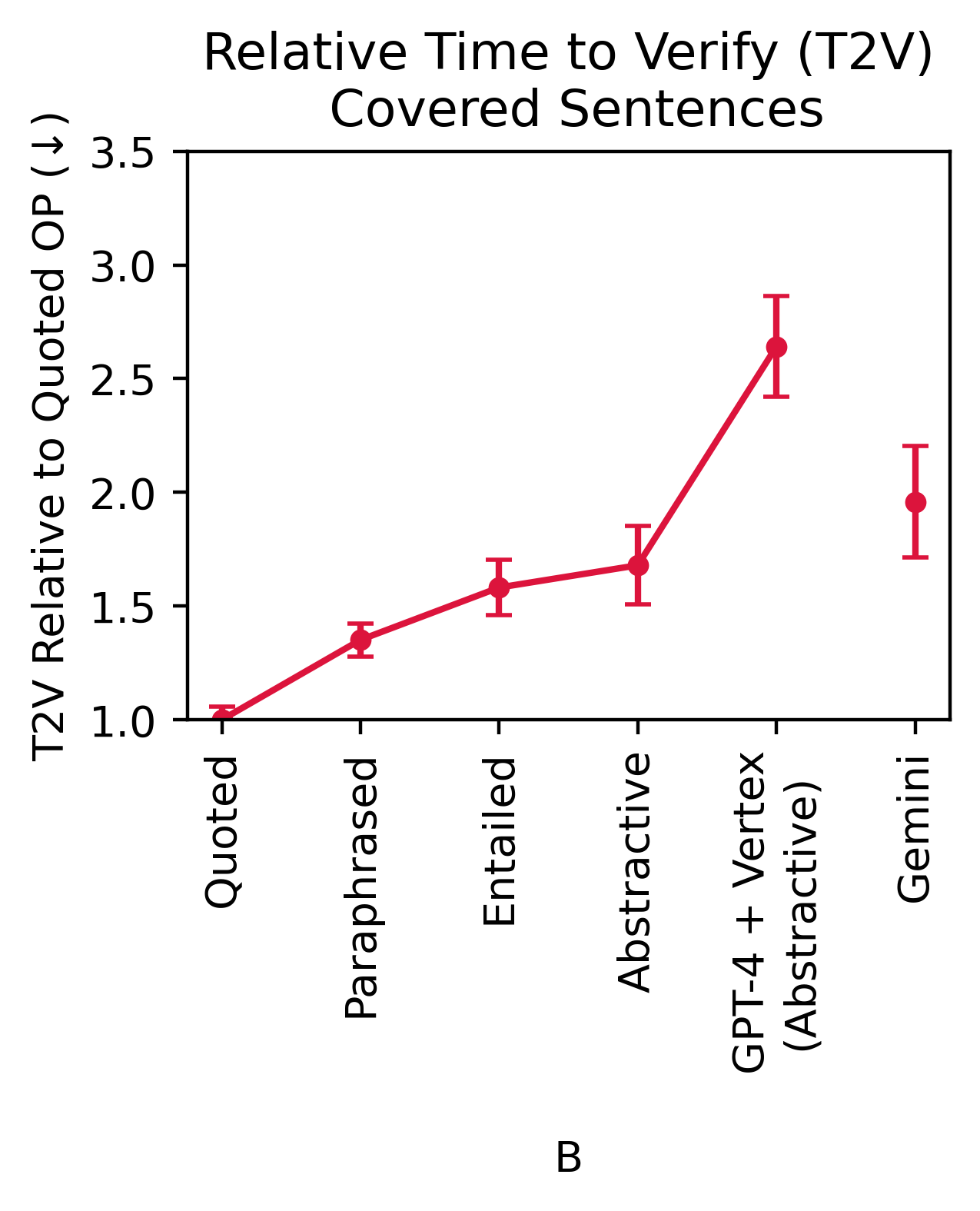}
    \end{subfigure}
    \begin{subfigure}
        \centering
        \includegraphics{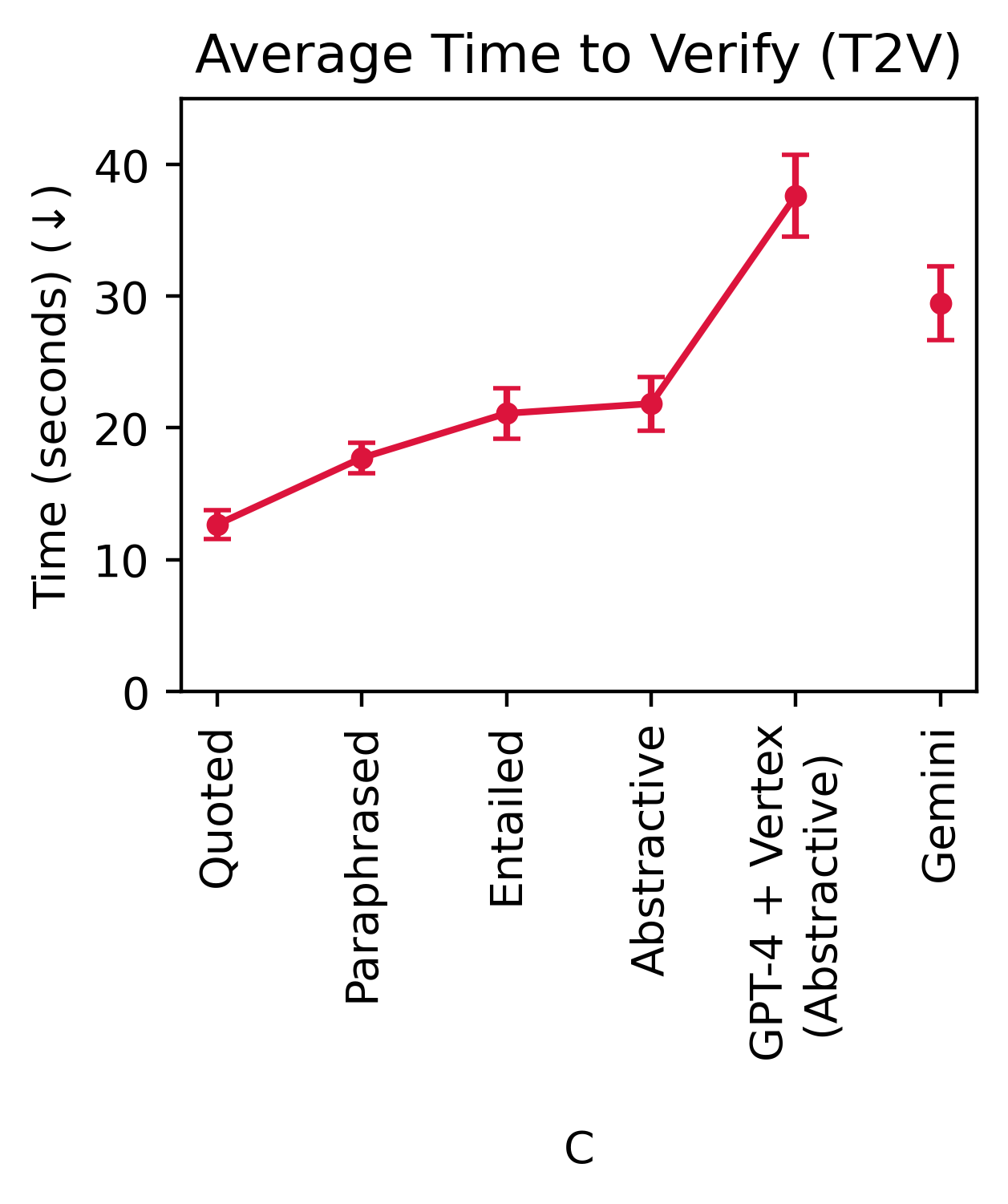}
    \end{subfigure}
    }
\caption{(A) Figure \ref{fig:results_in_aggregate} copied for convenience. Over all sentences, relative T2V increases alongside generation abstraction. (B) Over sentences with citation coverage, relative T2V increases as generation abstraction increases. (C) Over all sentences, absolute T2V increases as generation abstraction increases. Absolute T2V is the wall clock time in seconds taken to verify a cited sentence.}
\label{fig:t2v_alternatives}
\end{figure*}

We investigate whether the increase in relative T2V is driven by the increase in coverage errors as generations become more abstractive. Figure \ref{fig:t2v_alternatives}B shows that relative T2V increases as generations become more abstractive over sentences that are properly covered by citation. While users evidently take longer to evaluate sentences with coverage failures, this does not drive the differences in relative T2V across operating points observed in Figure \ref{fig:t2v_alternatives}A. We also report absolute T2V in seconds in Figure \ref{fig:t2v_alternatives}C.

\begin{figure*}[h]
\centering
    \resizebox{.7\textwidth}{!}{
    \begin{subfigure}
        \centering
        \includegraphics{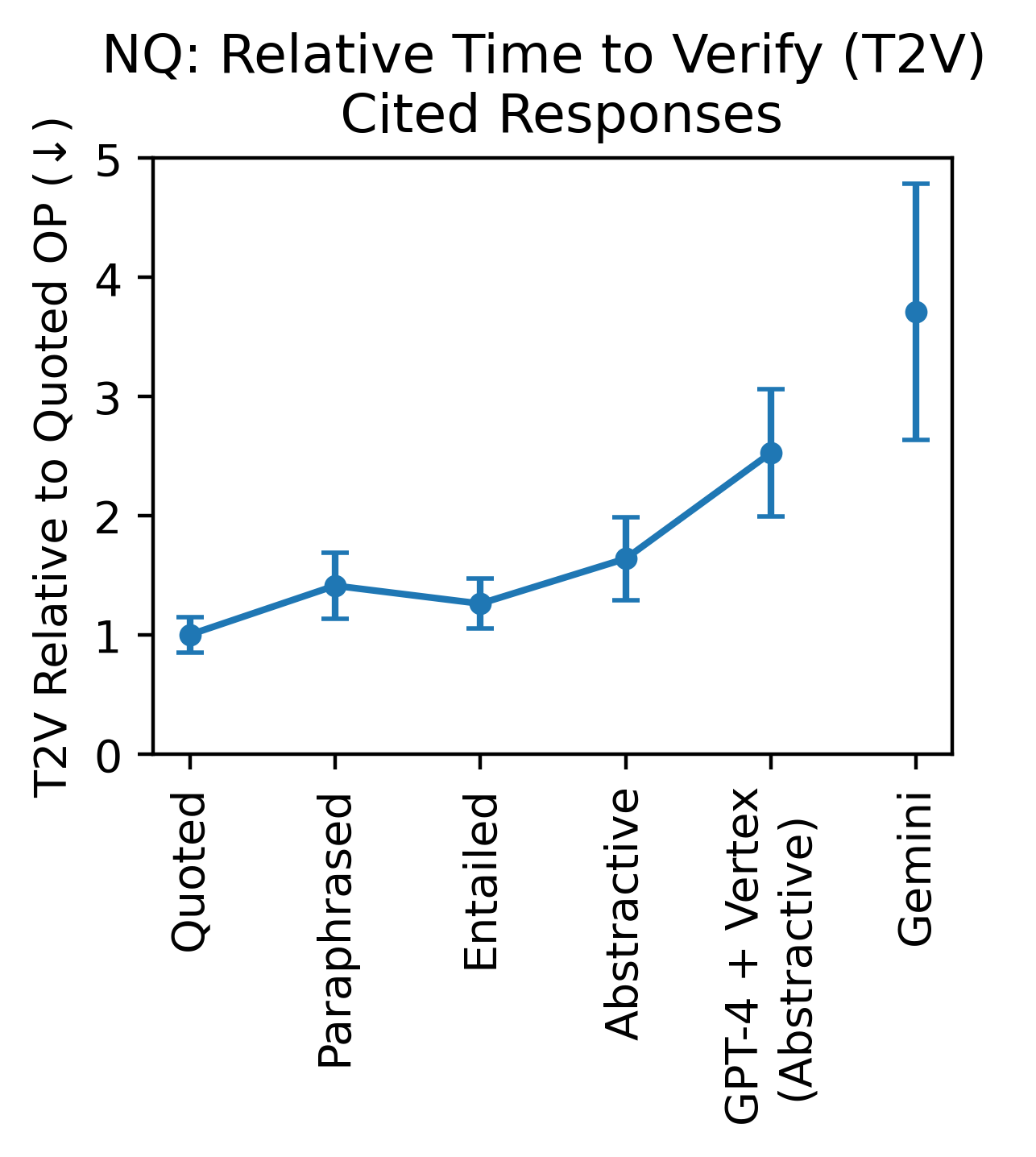}
    \end{subfigure}
    \begin{subfigure}
        \centering
        \includegraphics{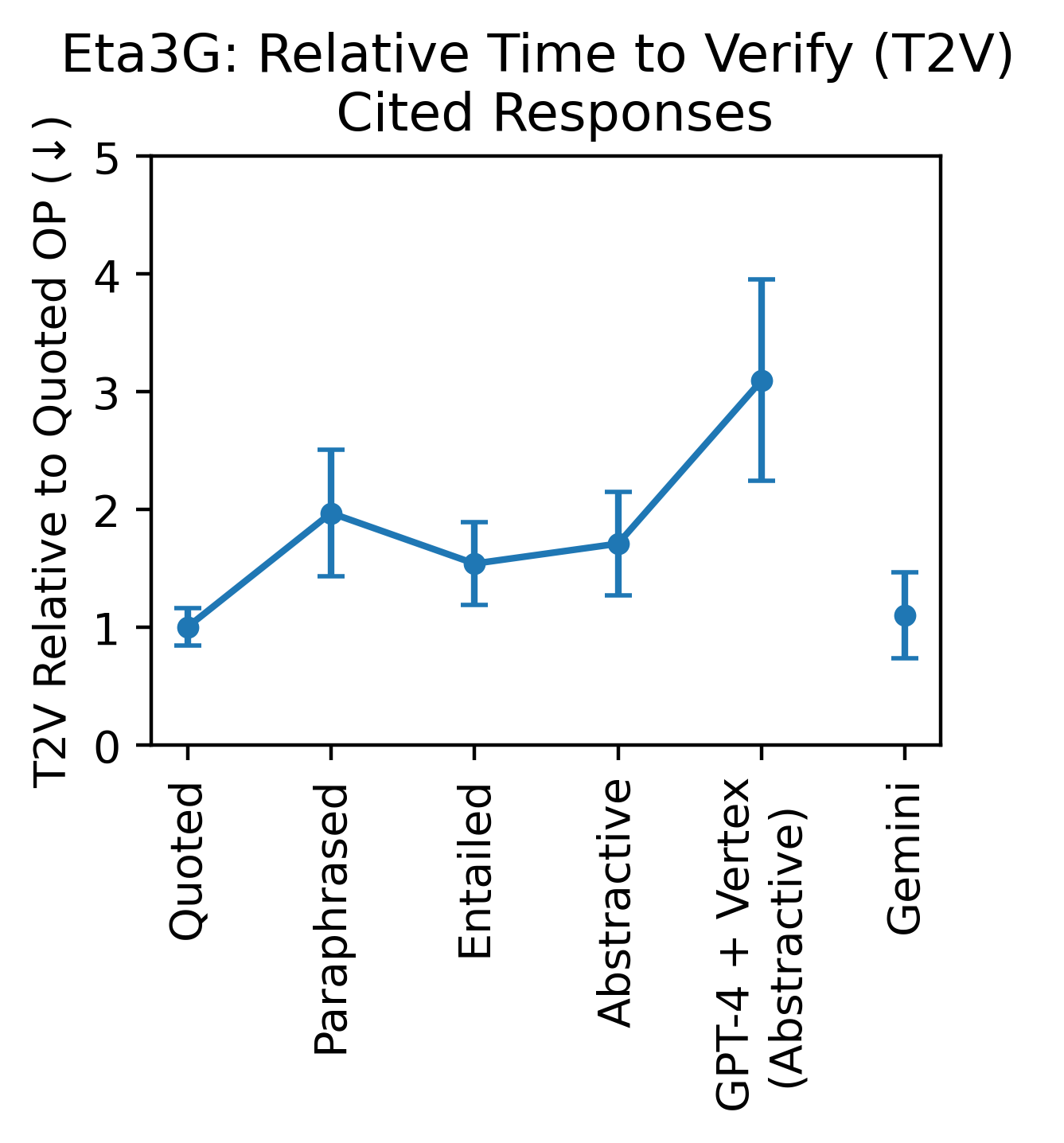}
    \end{subfigure}
    }
    \resizebox{.7\textwidth}{!}{
    \begin{subfigure}
        \centering
        \includegraphics{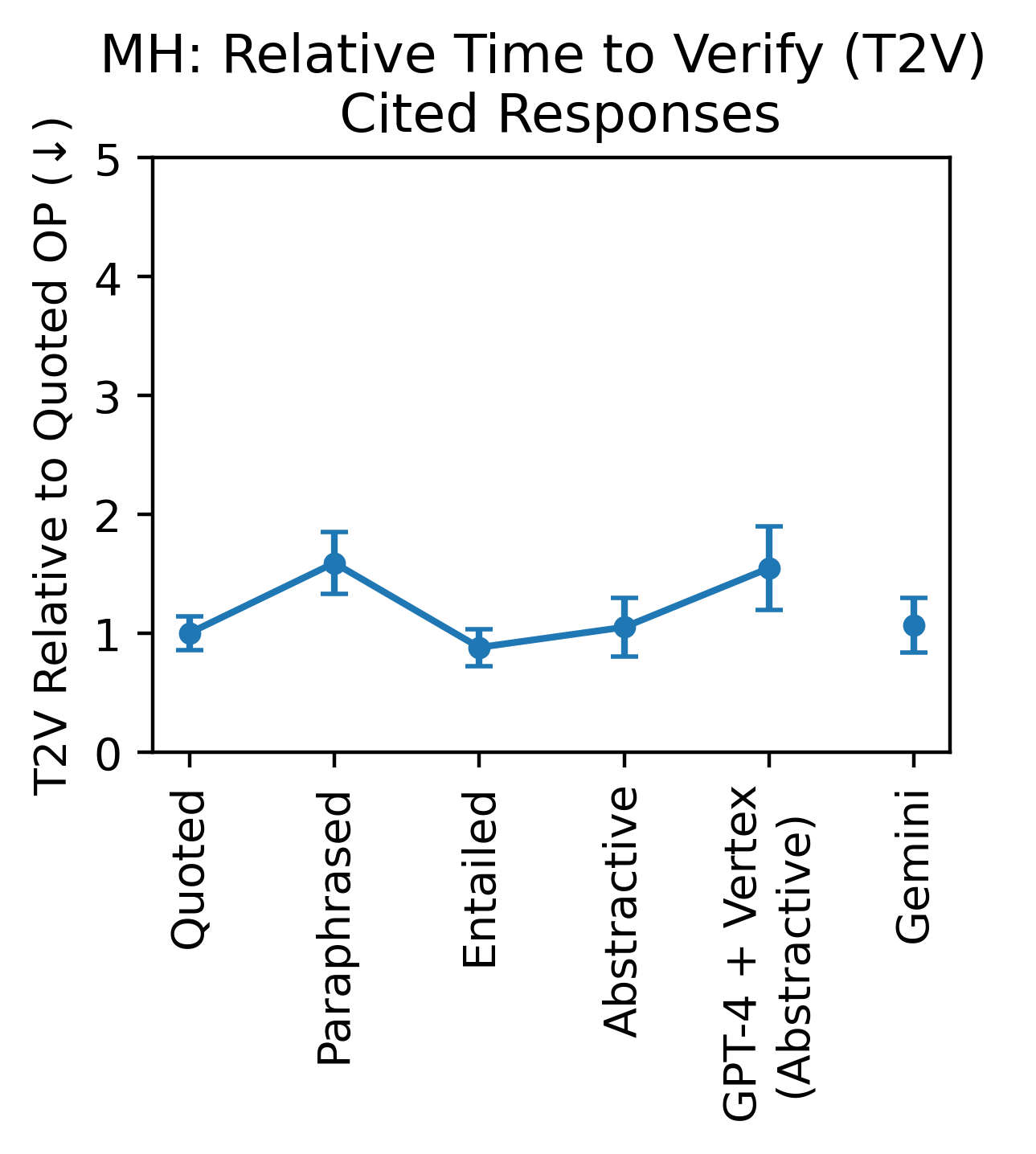}
    \end{subfigure}
    \begin{subfigure}
        \centering
        \includegraphics{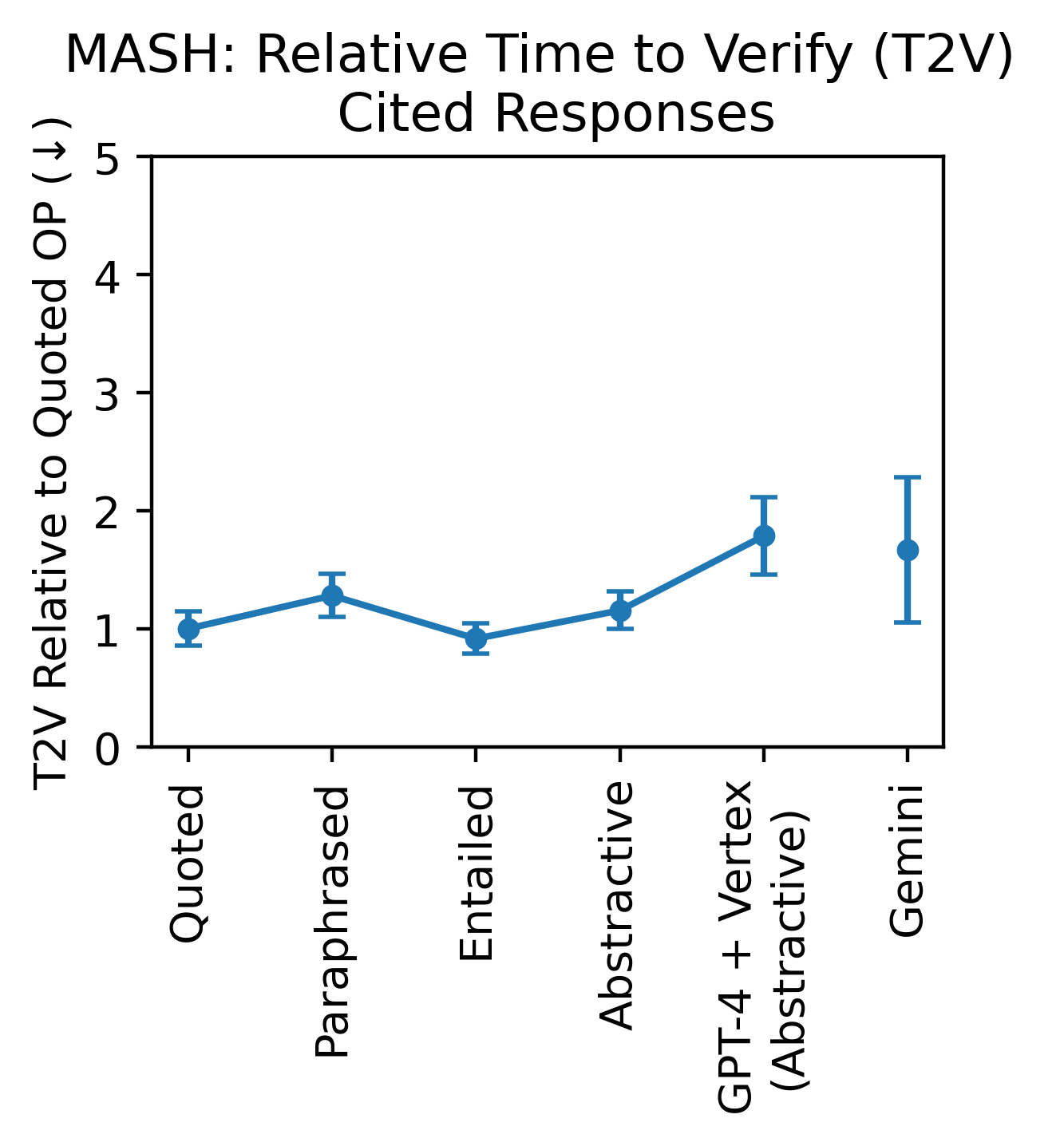}
    \end{subfigure}
    }
\caption{Time to verify entire responses by OP, relative to each annotator's time to verify entire quoted responses. While the results are noisier, T2V trends upward alongside abstraction, with the exception of the paraphrased generation. Results are aggregated over responses where each sentence is cited.}

\label{fig:response_t2v_by_qd}
\end{figure*}

Additionally, we examine the time that annotators take to verify entire responses. While our work considers T2V as the time taken to evaluate the coverage of an \textit{individual sentence} to emulate the user experience of verifying whether part of a generation is correct, verifying the \textit{entire response} is also a meaningful user experience. However, taking response-level T2V across all generations introduces a confounding effect. More abstractive generations tend to have lower coverage in part because more sentences that require citation simply have no citation at all. Because T2V is only measured for sentences with citation, responses with sentences without citation then appear to be evaluated far faster than responses where all sentences have citation. We report the relative time to verify entire responses where each sentence has at least one citation, by query distribution (Figure \ref{fig:response_t2v_by_qd}). However, we note that only evaluating responses that have associated citation potentially introduces another confounder. It is possible that generations that remain cited at more abstractive OPs are those that are the easiest to cite (and thereby verify). This may confound T2V results aggregated across properly covered generations by leading to lower relative T2Vs for more abstractive OPs.

Figure \ref{fig:response_t2v_by_qd} shows that the time taken to verify entire cited responses, relative to those of the quoted OP, generally trends upward alongside abstractiveness with the exception of the paraphrased OP. This elevated complexity of evaluating entire paraphrased responses may be explained by paraphrased generations having similarly high word count as the quoted generations, while lacking the convenience of inline quotation. In contrast, entailed and abstractive generations have substantially lower word count than quoted generations (Table \ref{tab:word_count_by_op}). Despite this, the verification times for entailed and abstractive generations are either similar or greater than those of quoted generations in Figure \ref{fig:response_t2v_by_qd}.

\subsection{Human Evaluation Details}
\label{subsec:human_evaluation_details}

\subsubsection{Pilot Evaluation Results}
\label{subsubsec:pilot}
\begin{figure*}[ht]
\raggedright
\resizebox{\textwidth}{!}{
    \begin{subfigure}
        \centering
        \includegraphics{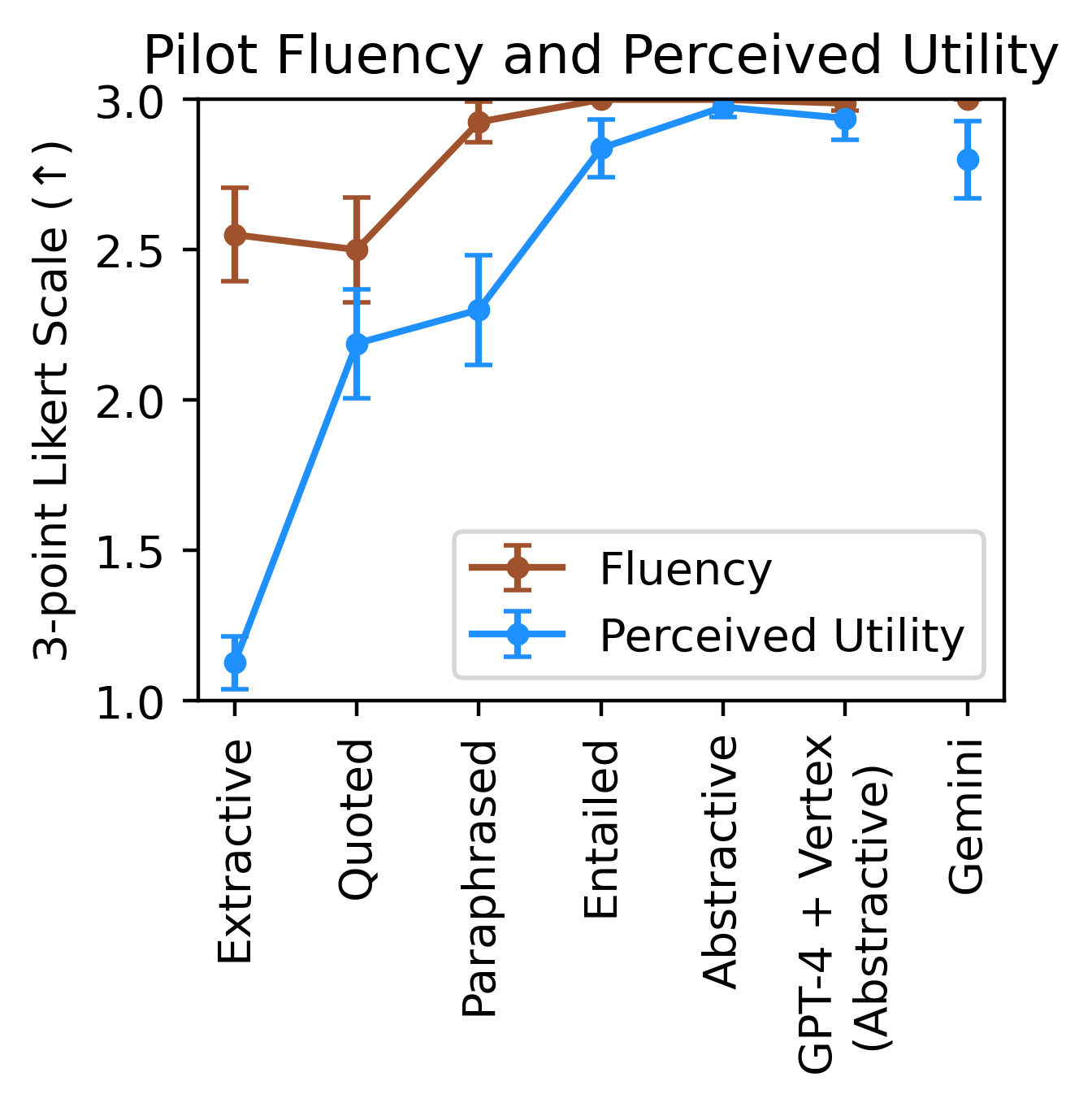}
    \end{subfigure}
    \begin{subfigure}
        \centering
        \includegraphics{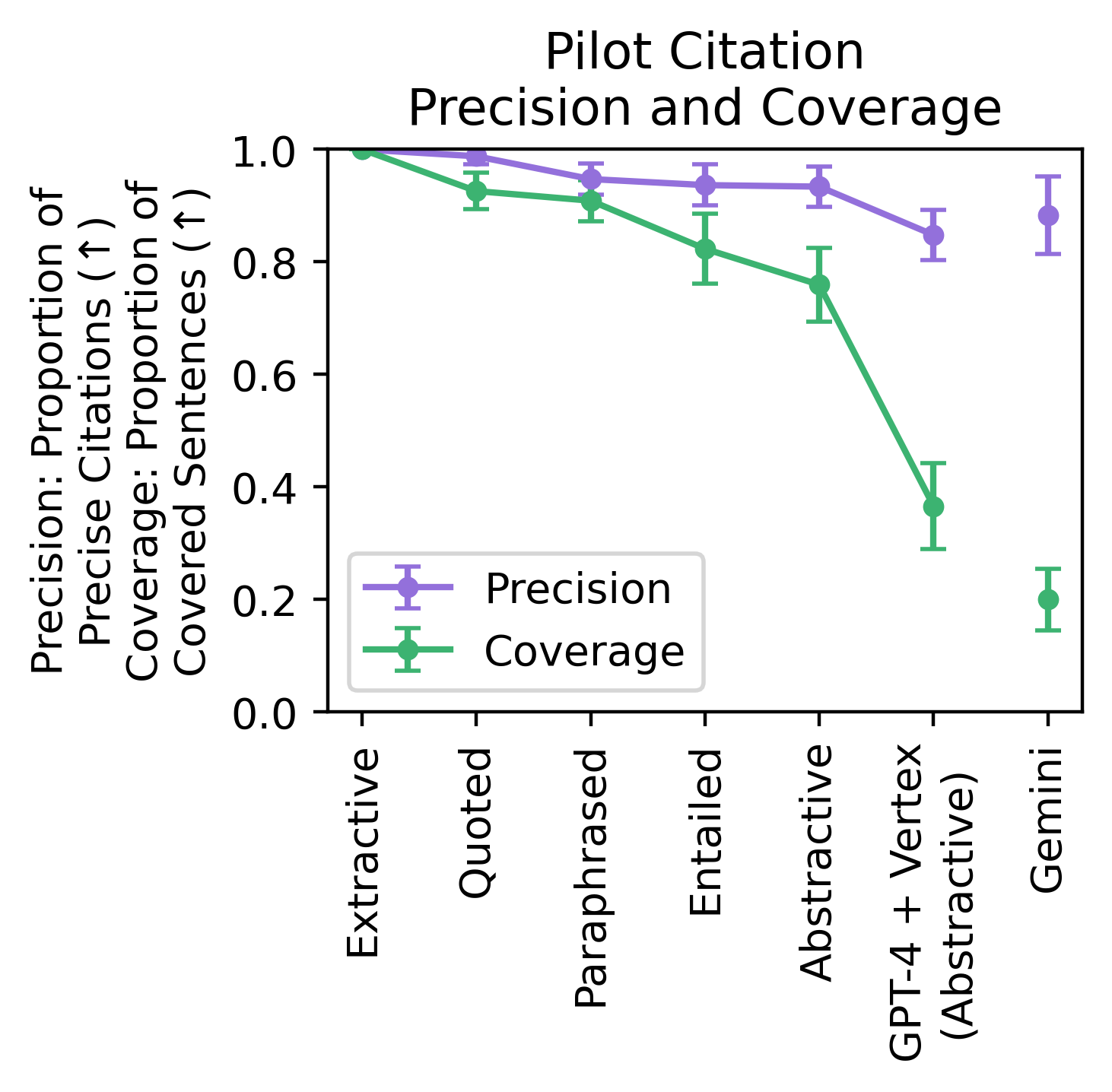}
    \end{subfigure}
    \begin{subfigure}
        \centering
        \includegraphics{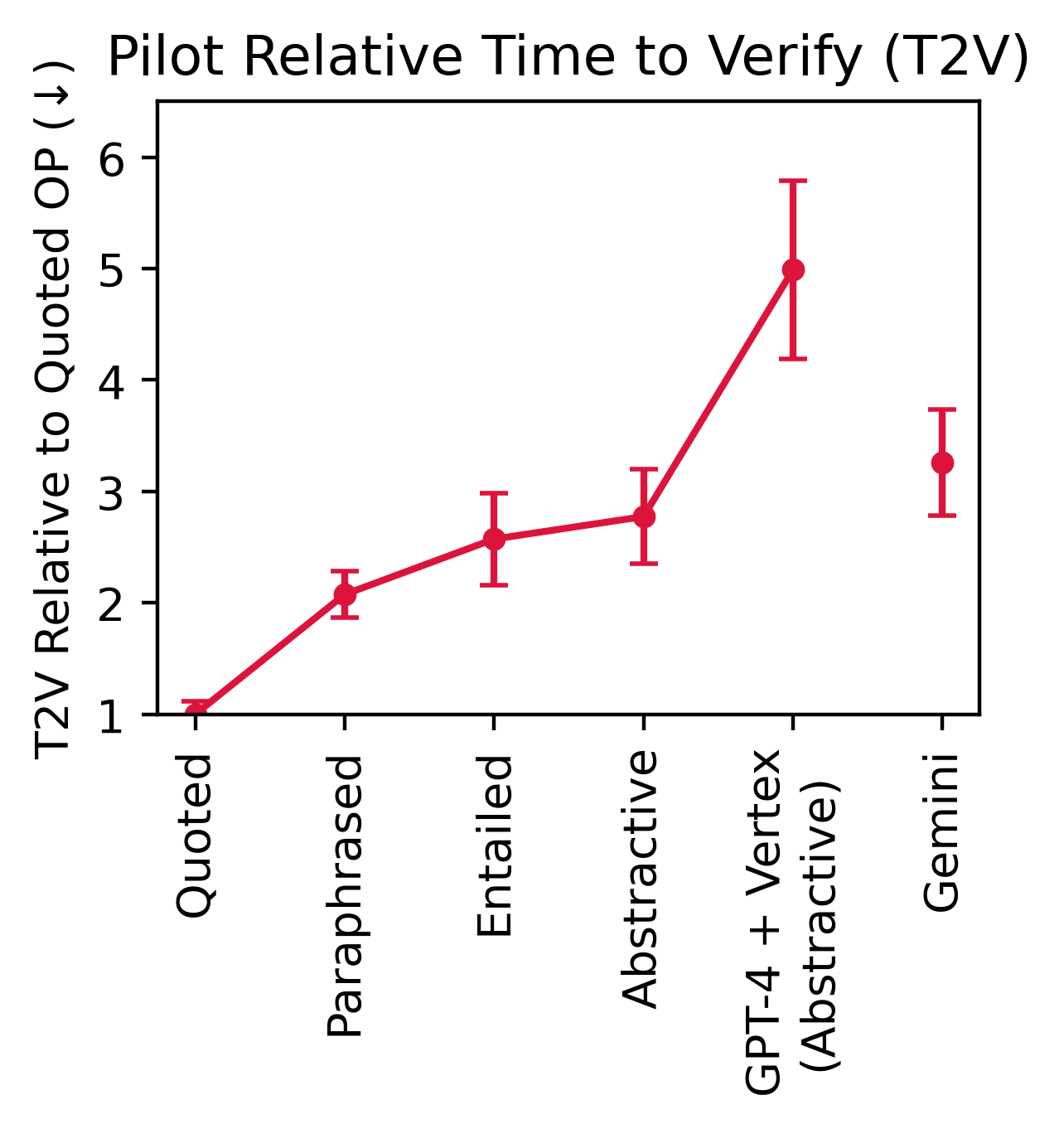}
    \end{subfigure}
    }
\caption{Pilot human evaluation results averaged over the four query distributions. Similar to the MTurk evaluations, fluency and perceived utility increase with abstraction, while citation precision and coverage decrease. Evaluating coverage takes longer as generations become more abstractive. The error bars represent 95\% confidence intervals.}
\label{fig:pilot_results_in_aggregate}
\end{figure*}

\begin{figure*}[ht]
    \raggedright
    \resizebox{\textwidth}{!}{
    \begin{subfigure}
        \centering
        \includegraphics{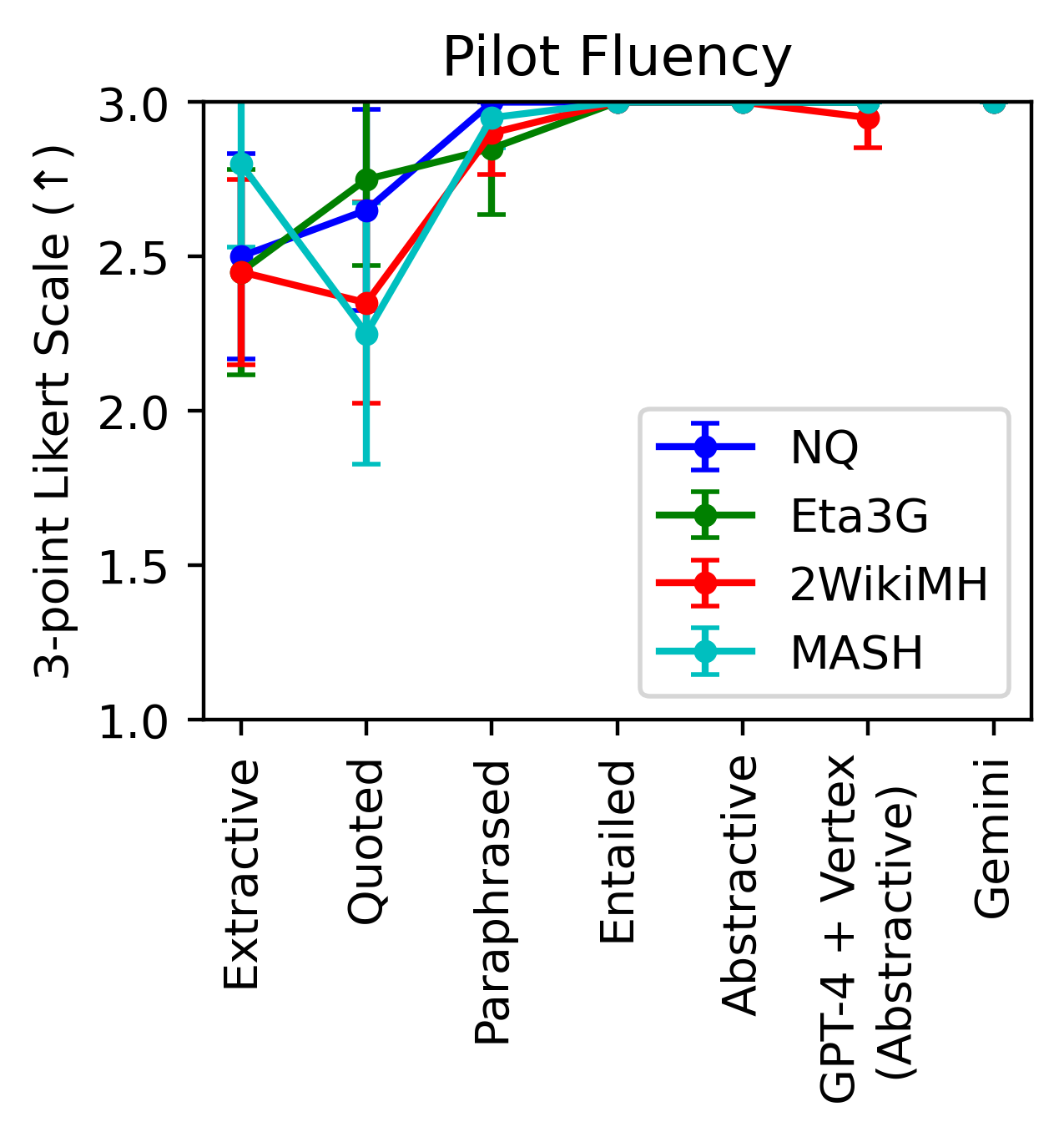}
    \end{subfigure}
    \begin{subfigure}
        \centering
        \includegraphics{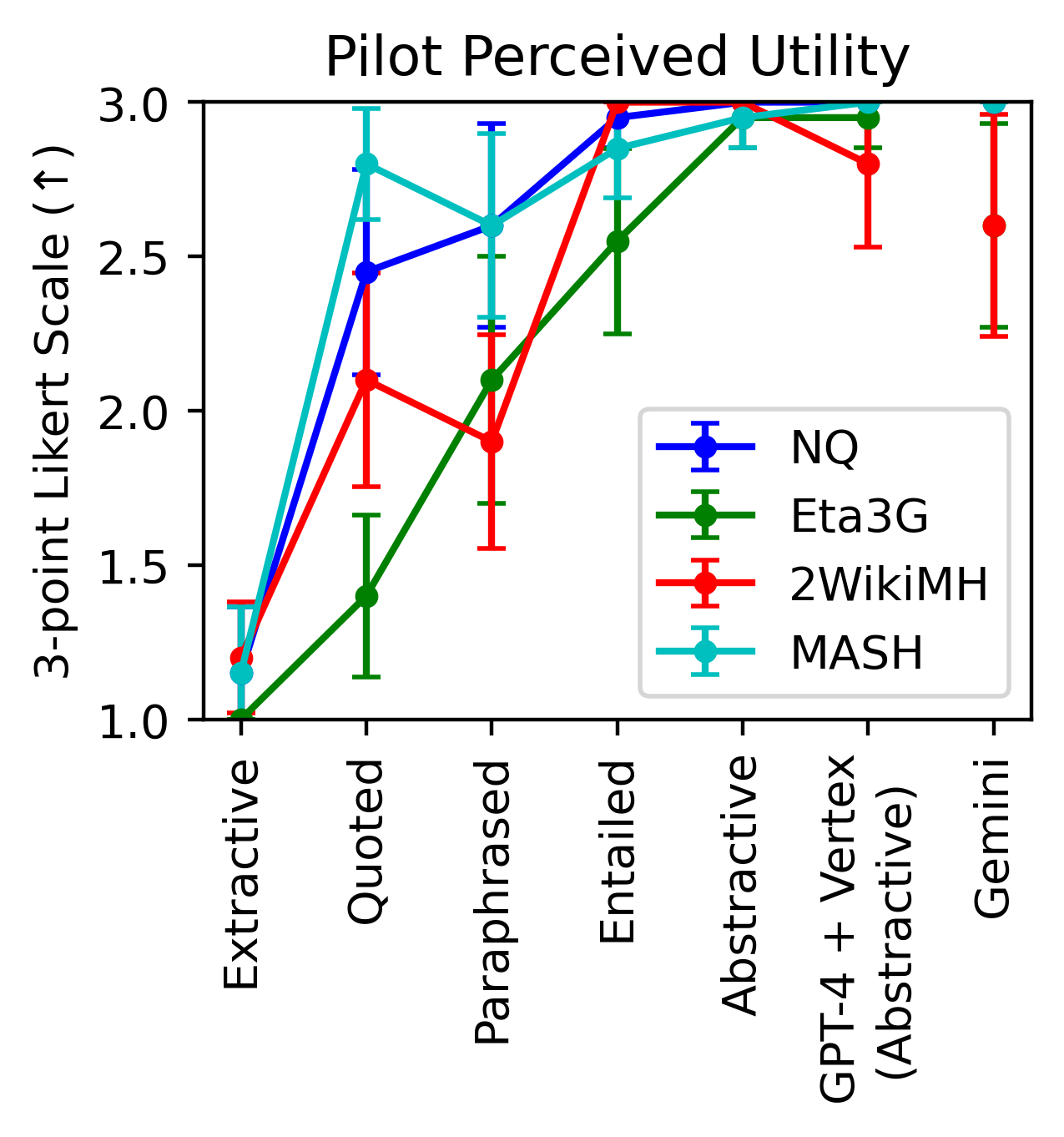}
    \end{subfigure}
    \begin{subfigure}
        \centering
        \includegraphics{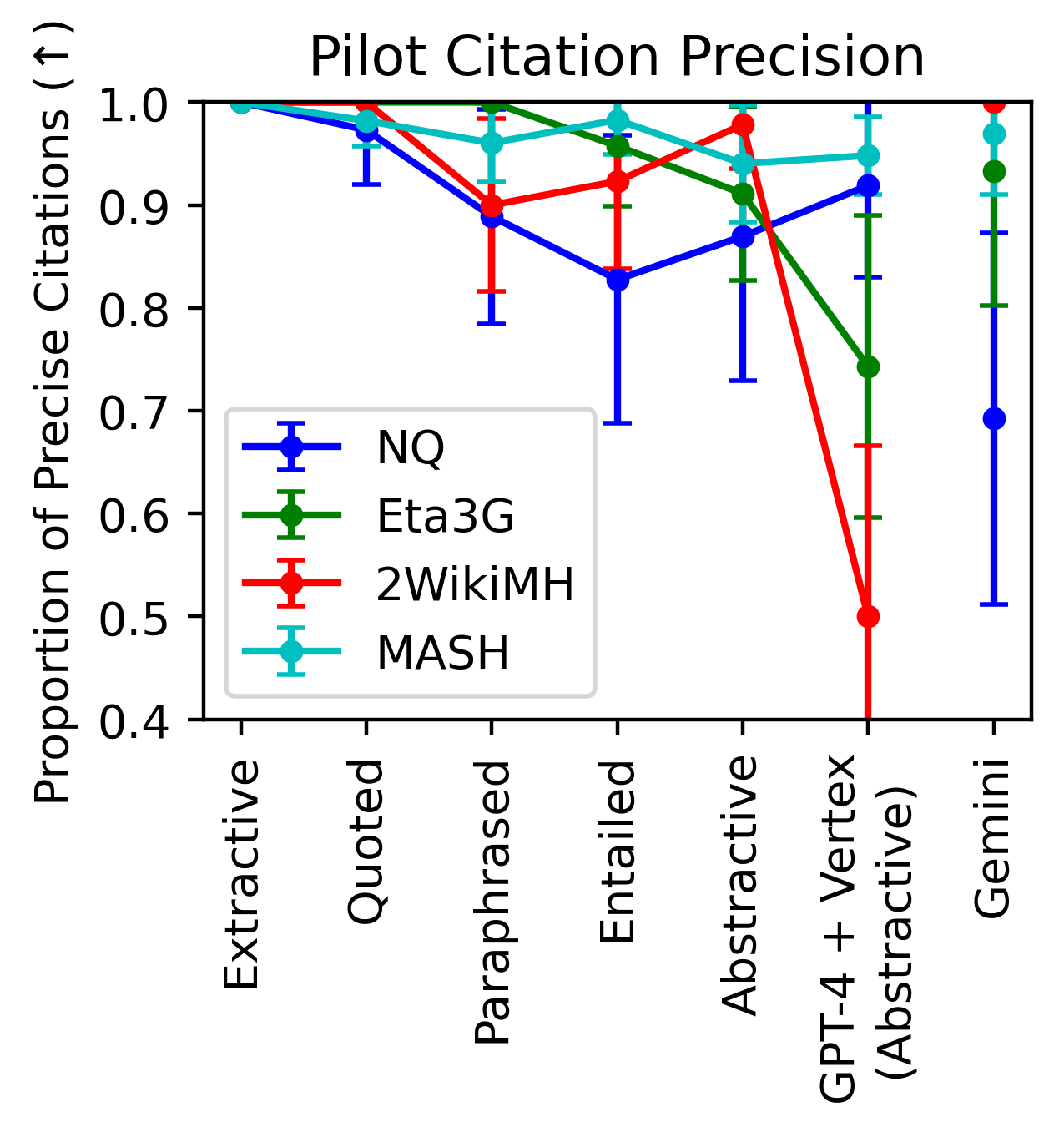}
    \end{subfigure}
    }
    \resizebox{\textwidth}{!}{
    \begin{subfigure}
        \centering
        \includegraphics{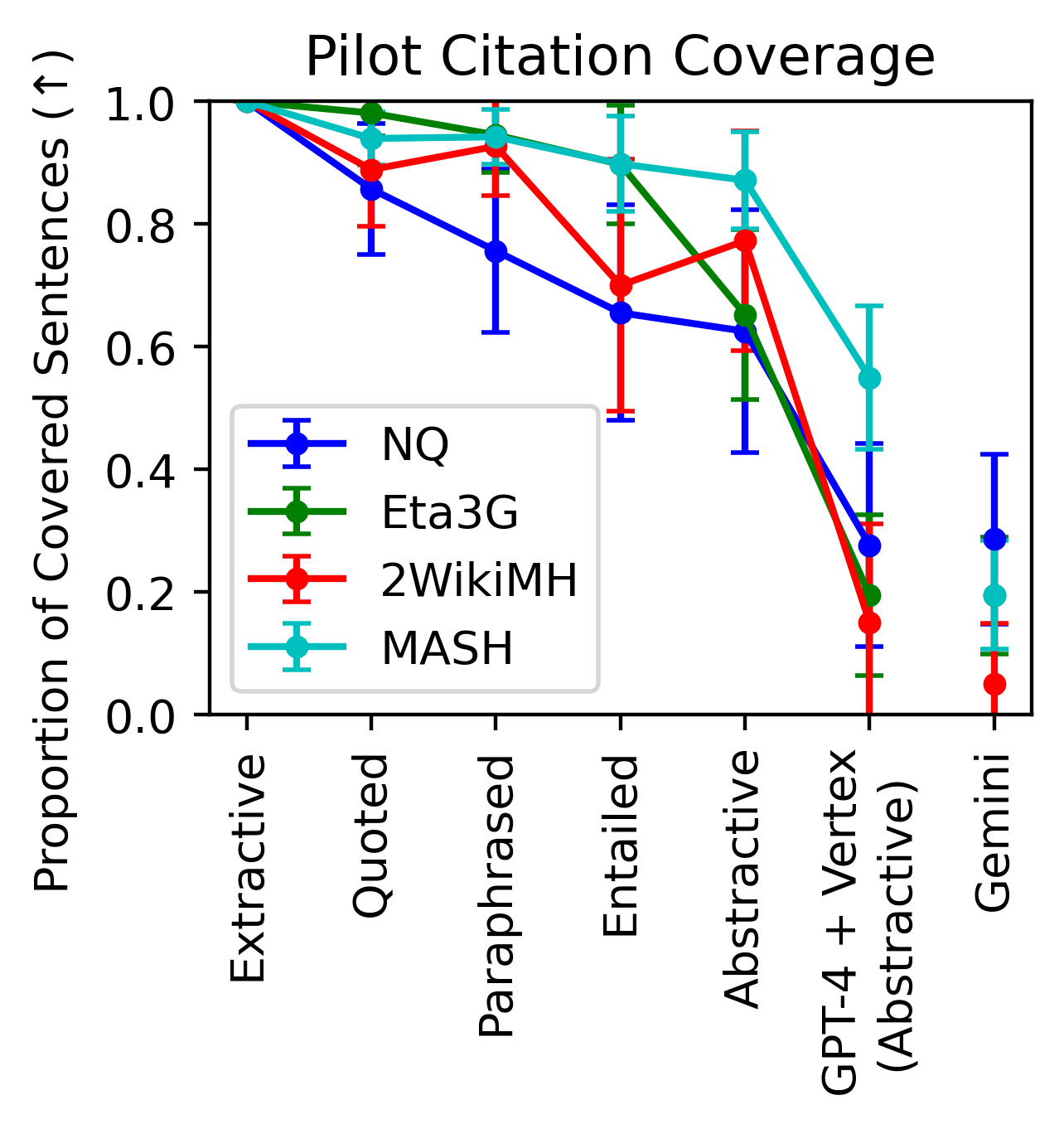}
    \end{subfigure}
    \begin{subfigure}
        \centering
        \includegraphics{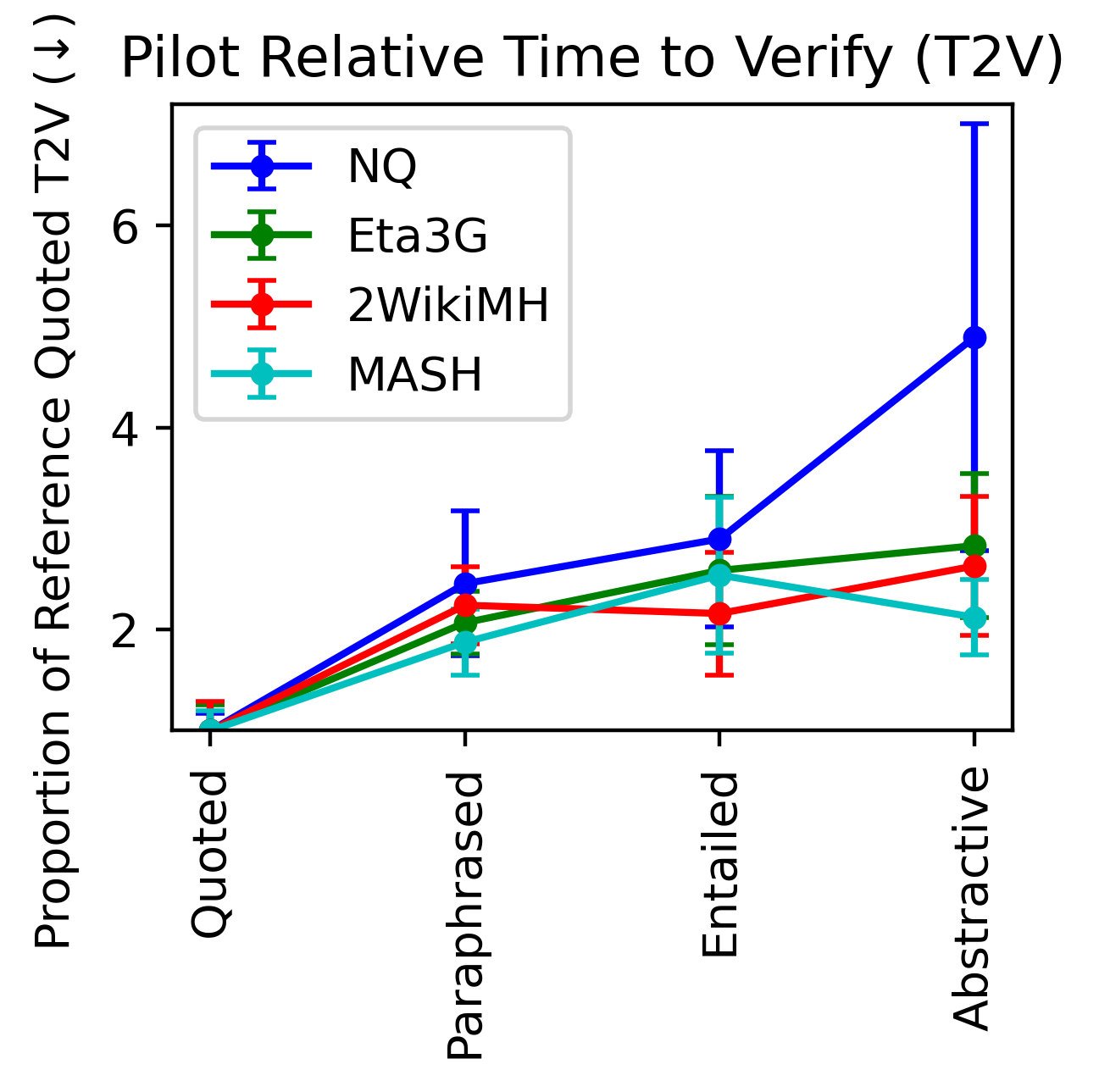}
    \end{subfigure}
    \begin{subfigure}
        \centering
        \includegraphics[width=.48\textwidth]{images/filler.png}
    \end{subfigure}
    }
\caption{Pilot human evaluation results by query distribution. The MTurk human evaluations reflect the same trends as the pilot results---as generations become more abstractive, fluency increases similarly for all query distributions, while perceived utility increases at different OPs (also stagnating between the quoted and paraphrased OPs for the MH and MASH distributions). Also, citation precision and coverage decrease while relative T2V increases across each query distributions as generations become more abstractive. The error bars represent 95\% confidence intervals.}

\label{fig:pilot_results_by_qd}
\end{figure*}

Before running any human evaluations through MTurk, we performed pilot evaluations of all seven systems across 20 queries for each of the four query distributions ourselves. Figure \ref{fig:pilot_results_in_aggregate} shows the pilot results in aggregate and Figure \ref{fig:pilot_results_by_qd} shows the pilot results by query distribution. Corroborating the trends we report from the MTurk evaluations, fluency and utility trend upward with increased abstractiveness, while citation coverage deteriorates and relative T2V increases. 

\subsubsection{Order of Experiments}
\label{subsubsec:order_of_experiments}
The results for citation coverage, precision, and T2V were obtained across two batches. First, we ran a batch for the reference OP implementations; here five query-generation pairs were assigned to an annotator for each HIT, one for each OP. Then, we ran a batch to evaluate the GPT-4 + Vertex and Gemini generations, which assigned three query-generation pairs per HIT, one for each of the GPT-4 + Vertex, Gemini, and quoted generations (which was re-evaluated in order to normalize T2V). We observed differences in the average quoted fluency and utility ratings across these two batches, implying that annotators were calibrating their evaluations of fluency and utility to the generations of each batch. In order to achieve consistent fluency and utility annotations across all OP implementations, we obtained the final fluency and utility results by re-running the fluency and utility evaluation over all OP implementations simultaneously; seven query-generation pairs were assigned per HIT. Across the two rounds of citation evaluation, we ensured no annotator saw the same query more than once in order to avoid any effects of familiarization with the queries and sources on T2V. 

\subsubsection{Annotation Quality Control}
\label{subsubsec:annotation_quality_control}
Because a month had elapsed in between the first and second batches (discussed in Section \ref{subsubsec:order_of_experiments}), we ran another screener HIT to refresh annotators on the task guidelines before admitting them to the second batch; 14 of the original annotators still active on the MTurk platform continued with the study. Again, personalized feedback was sent to and acknowledged by each one. Throughout the study, we dropped annotators who failed spot checks or quickly clicked through our annotation interface. 

\subsubsection{Annotator Compensation}
\label{subsubsec:annotator_compensation}
Annotators were paid \$16/hr. HITs with 5 utility/fluency + citation evaluation tasks were compensated at \$2.55/HIT, those with 3 utility/fluency + citation evaluation tasks at \$1.75/HIT, and those with 7 utility/fluency tasks at \$0.50/HIT. Screener HITs were compensated at \$2.55/HIT with a \$4 bonus if accepted into the study.

\clearpage
\begin{figure}[ht]
    \centering
    \includegraphics[width=\textwidth]{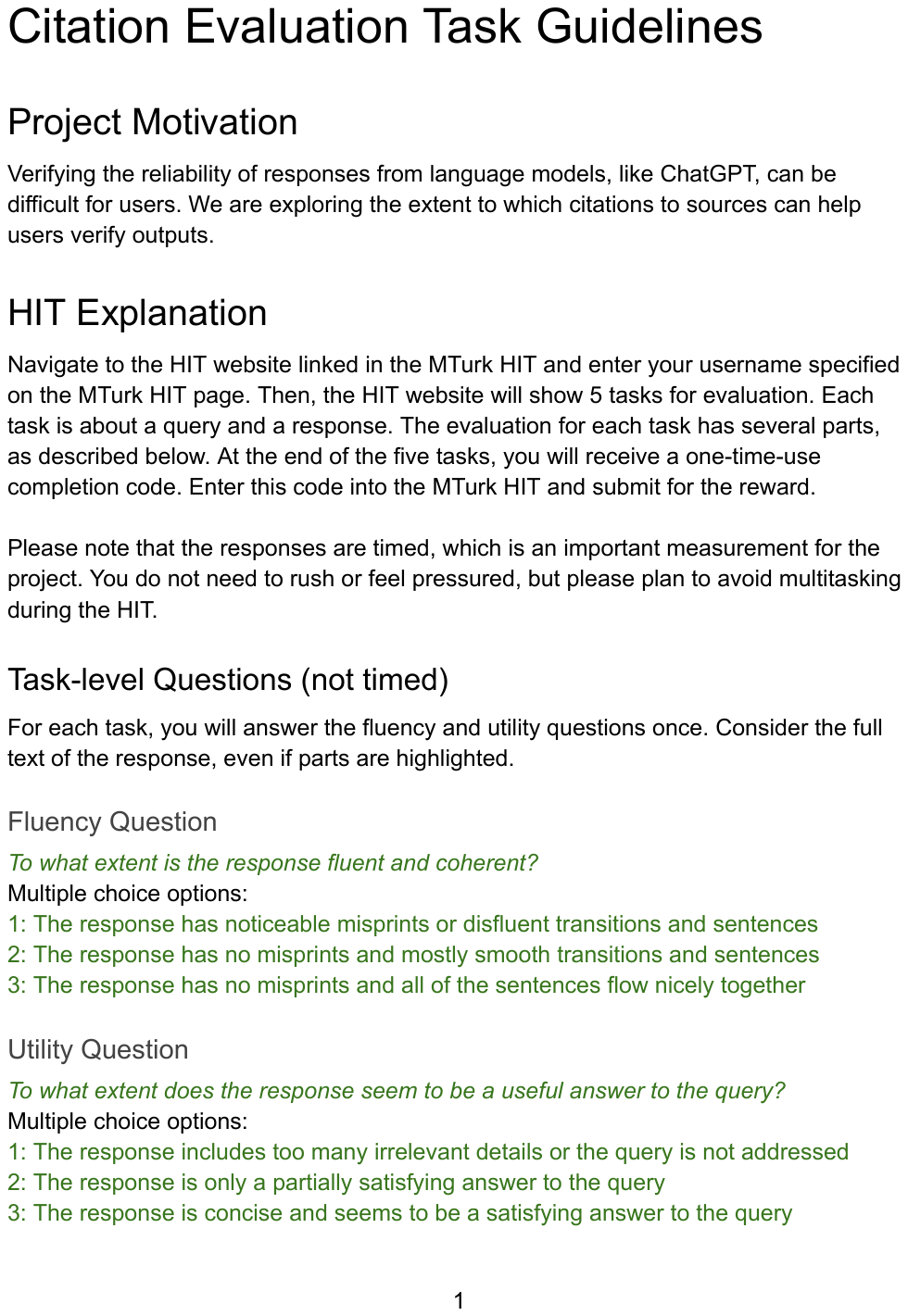}
    \label{fig:Annotation_Task_Guidelines1}
\end{figure}
\begin{figure}[ht]
    \centering
    \includegraphics[width=\textwidth]{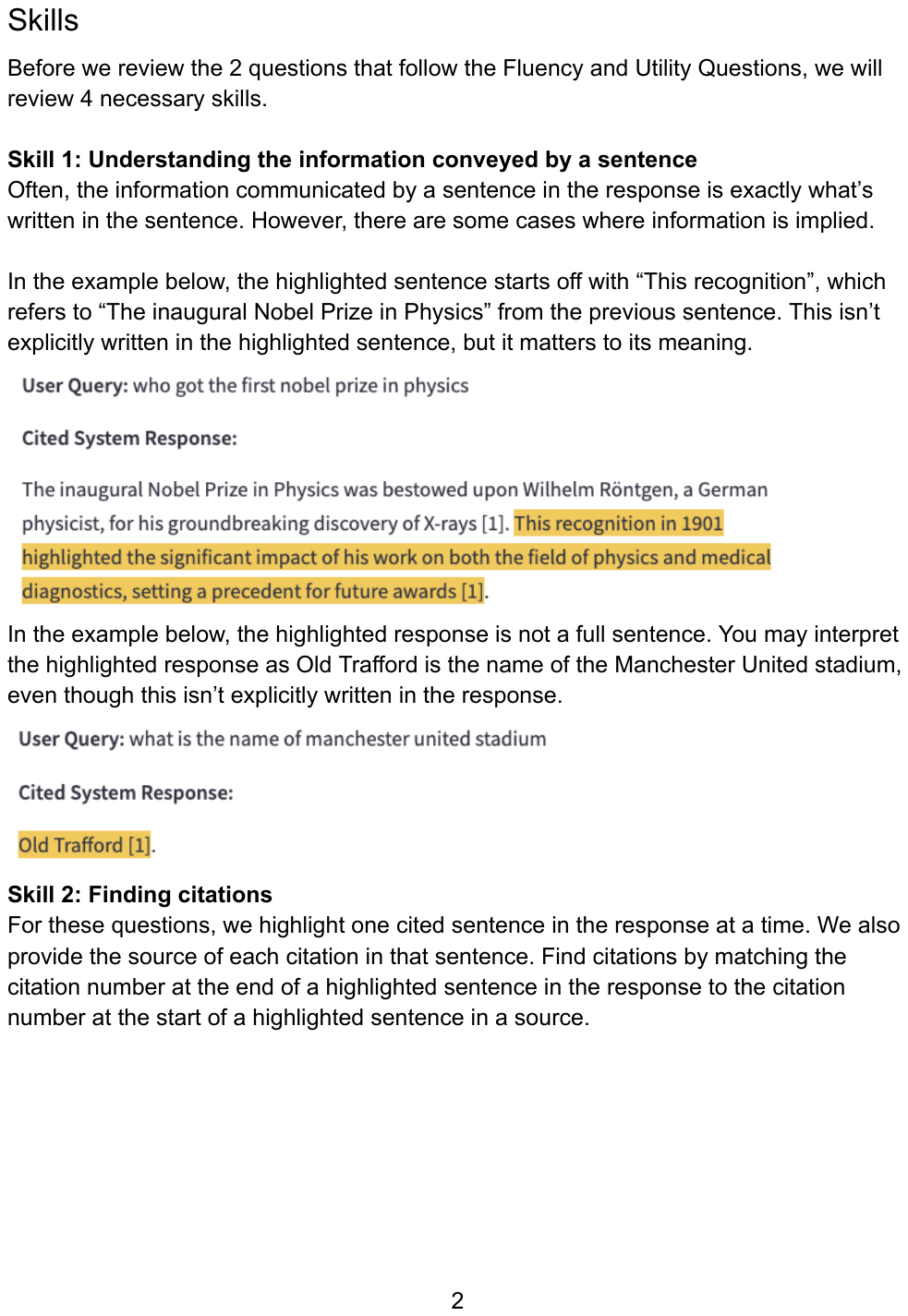}
    \label{fig:Annotation_Task_Guidelines2}
\end{figure}

\begin{figure}[ht]
    \centering
    \includegraphics[width=\textwidth]{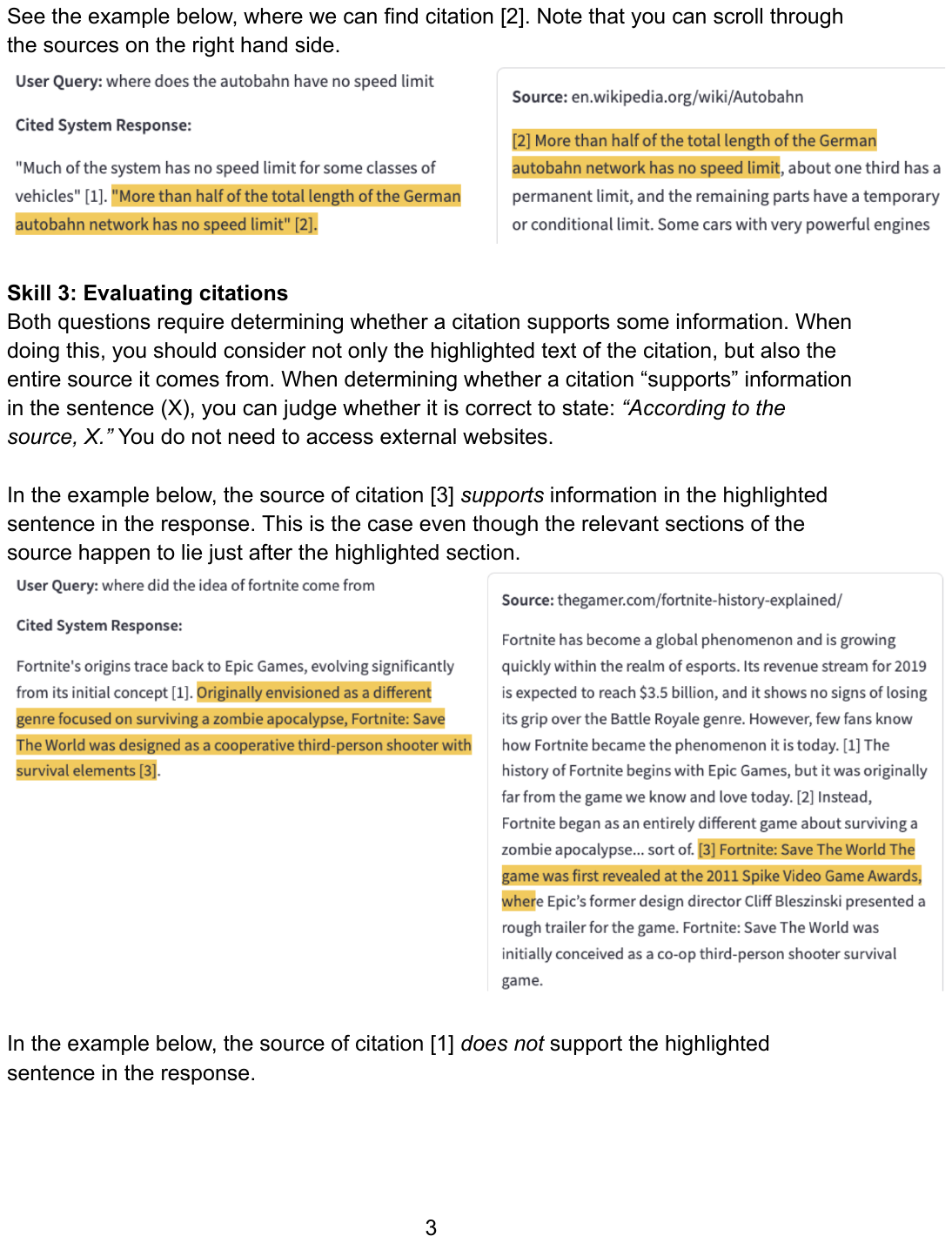}
    \label{fig:Annotation_Task_Guidelines3}
\end{figure}
\begin{figure}[ht]
    \centering
    \includegraphics[width=\textwidth]{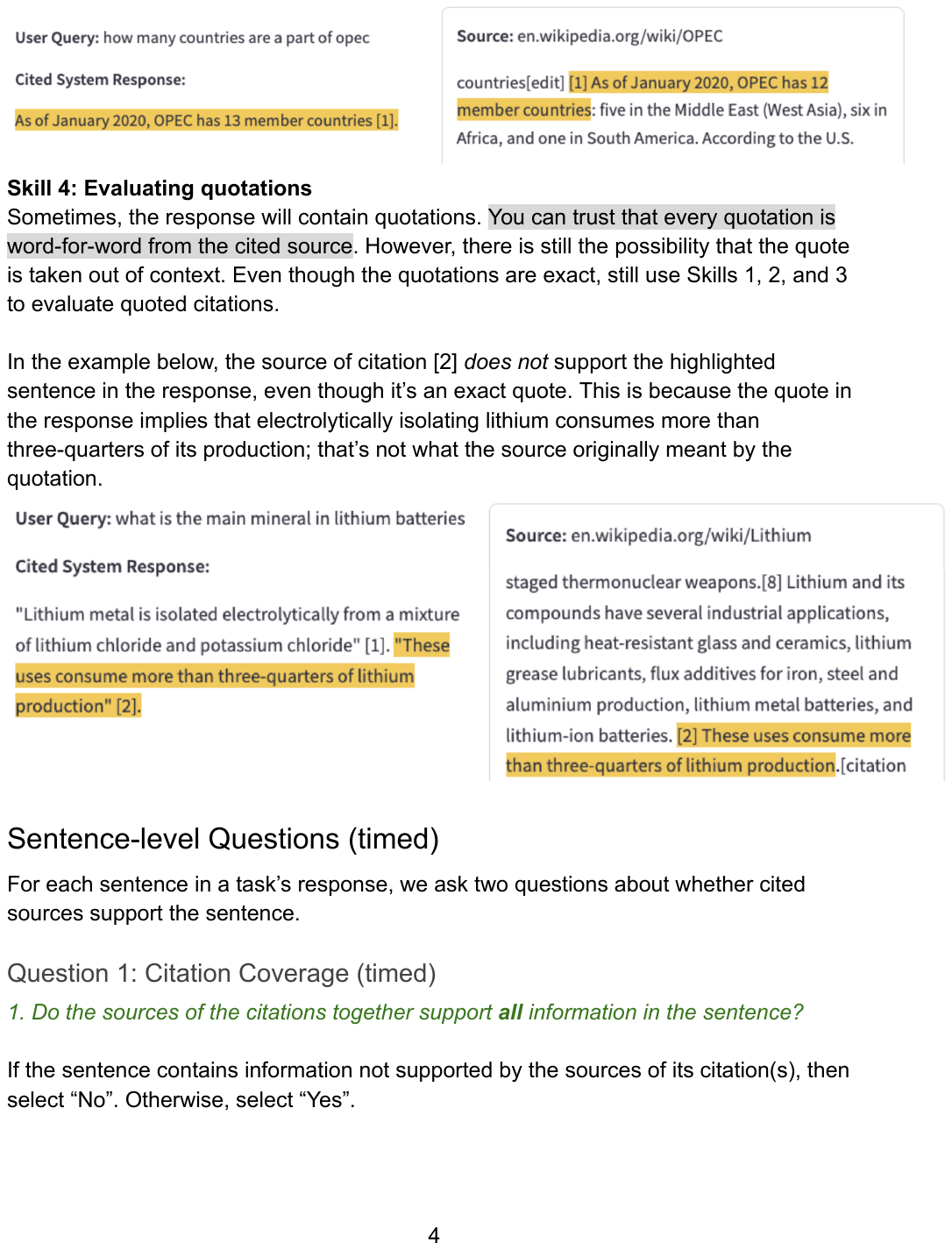}
    \label{fig:Annotation_Task_Guidelines4}
\end{figure}
\begin{figure}[ht]
    \centering
    \includegraphics[width=\textwidth]{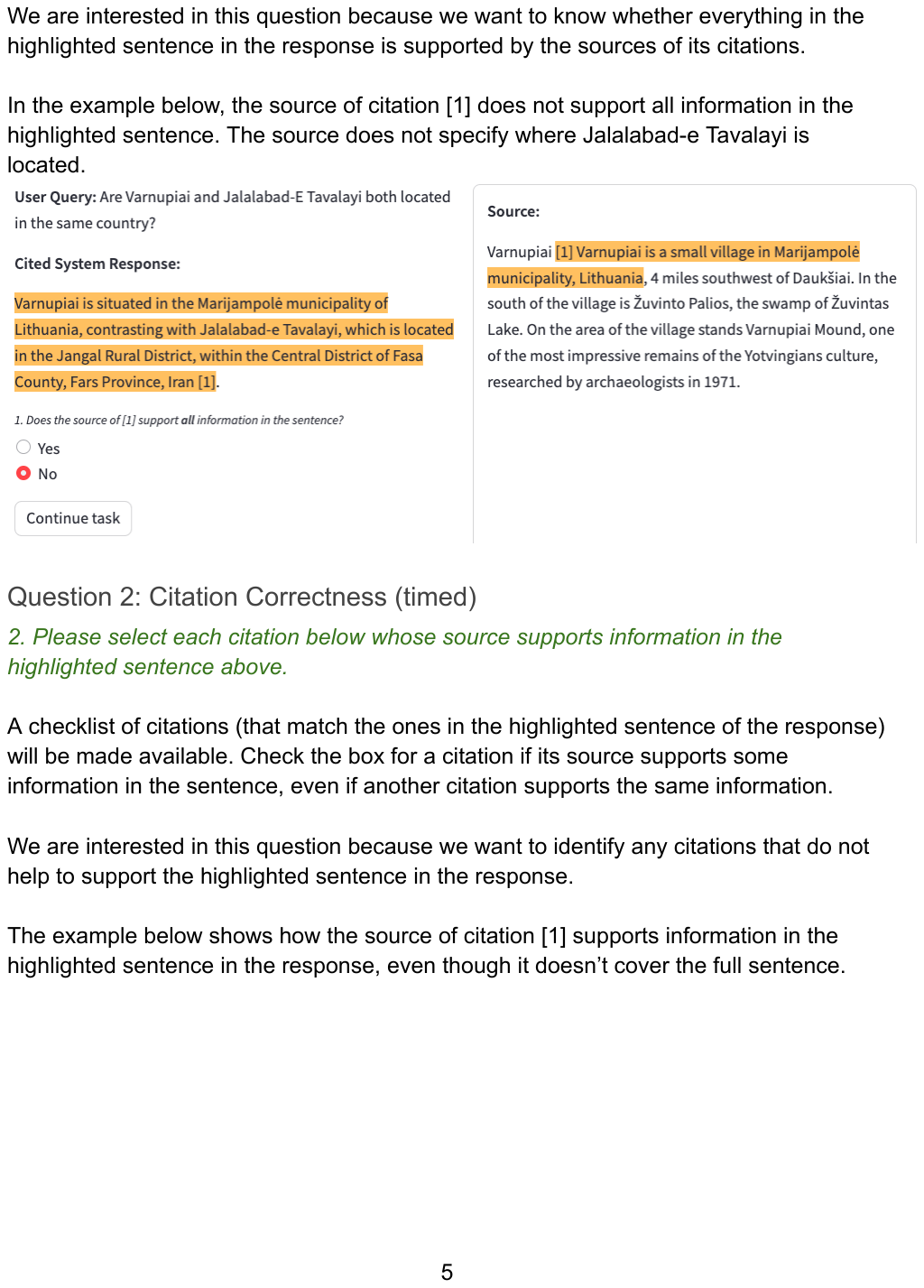}
    \label{fig:Annotation_Task_Guidelines5}
\end{figure}

\subsubsection{Annotation interface}
\label{subsec:annotation_interface}
\begin{figure*}[ht]
    \centering
    \includegraphics[width=\textwidth]{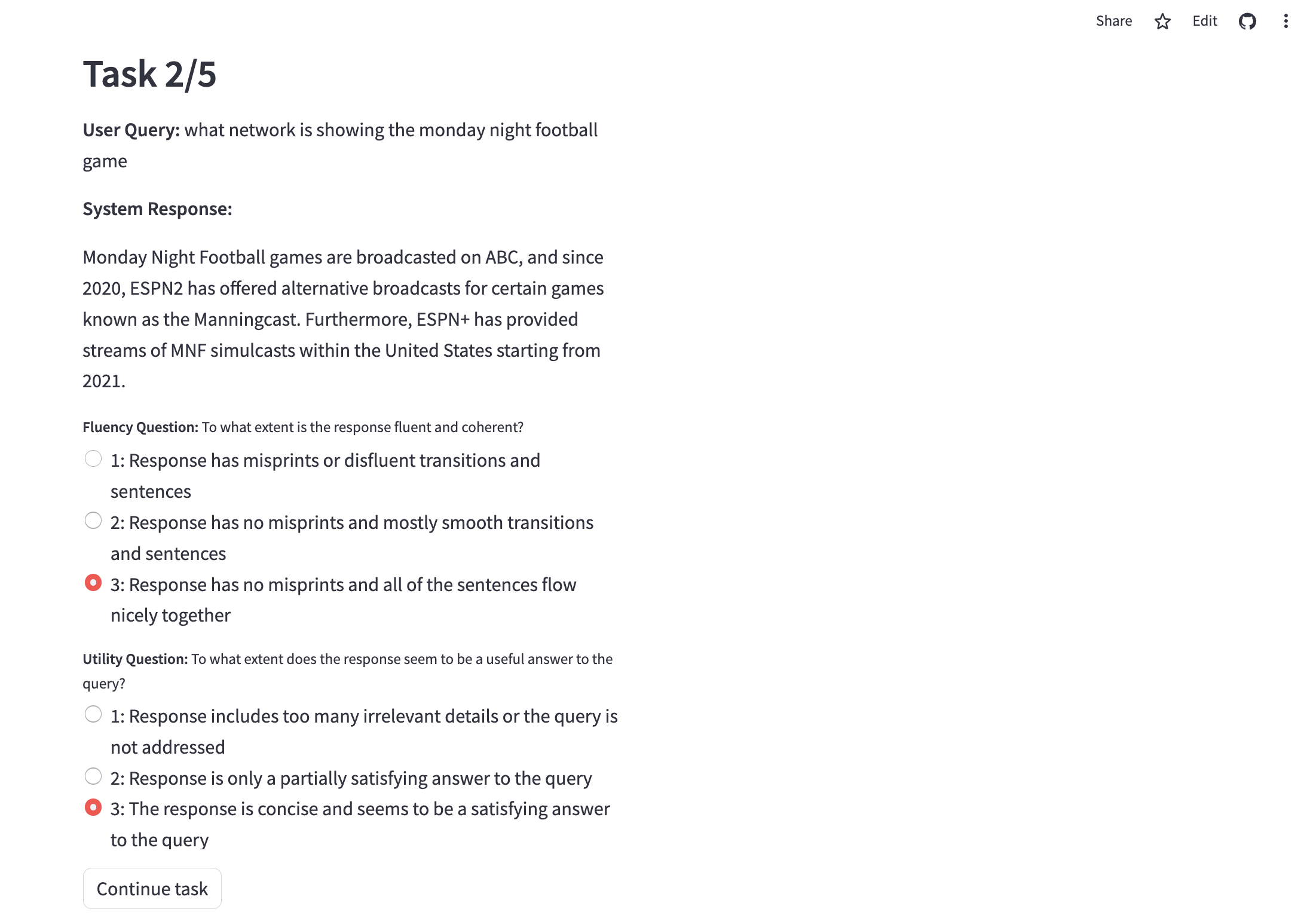}
    \caption{The annotation interface for rating the utility and fluency of a response.}
    \label{fig:annotation_interface_utility_fluency}
\end{figure*}

\begin{figure*}[ht]
    \centering
    \includegraphics[width=\textwidth]{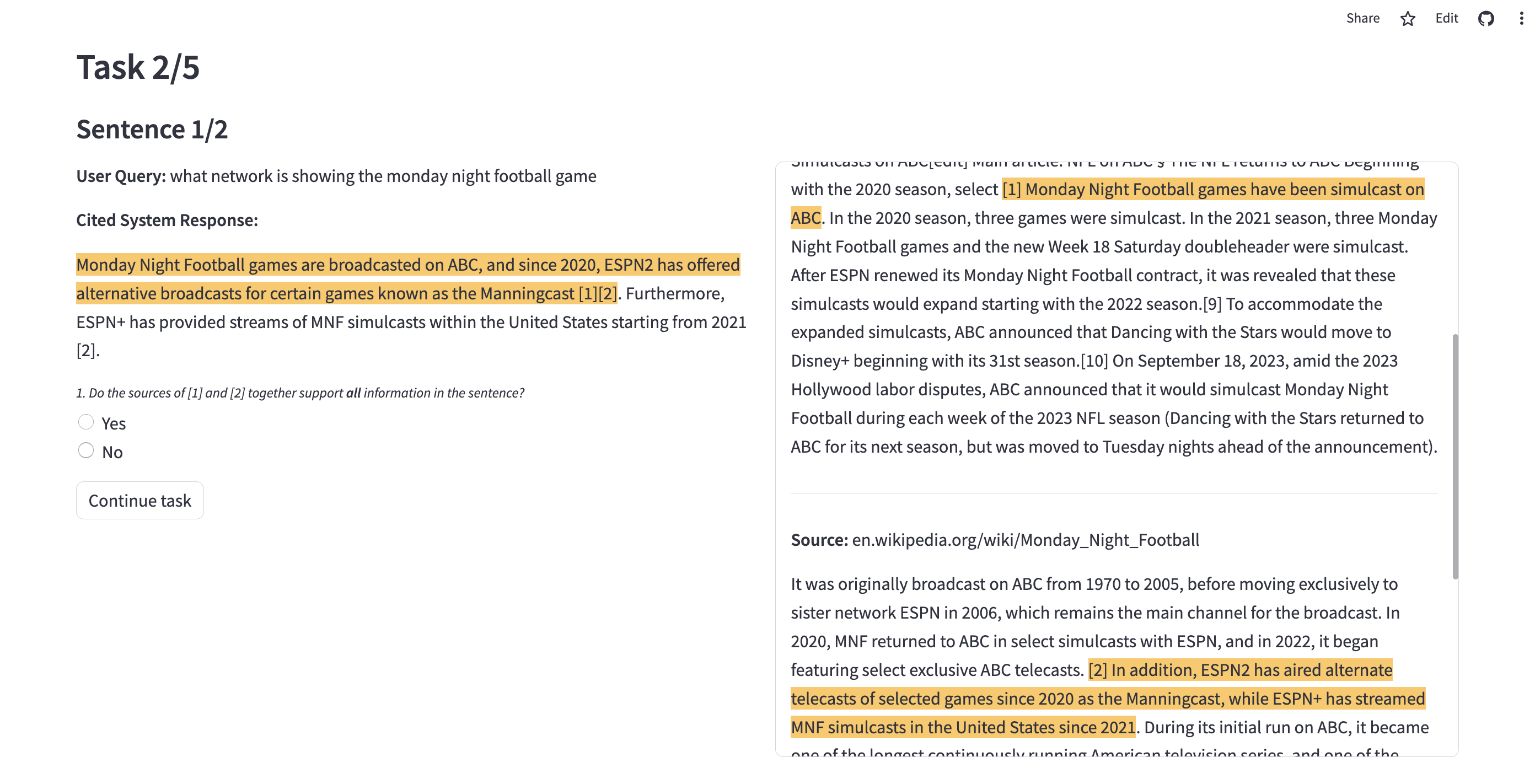}
    \caption{The annotation interface for evaluating the coverage of a sentence.}
    \label{fig:annotation_interface_coverage}
\end{figure*}

\begin{figure*}[ht]
    \centering
    \includegraphics[width=\textwidth]{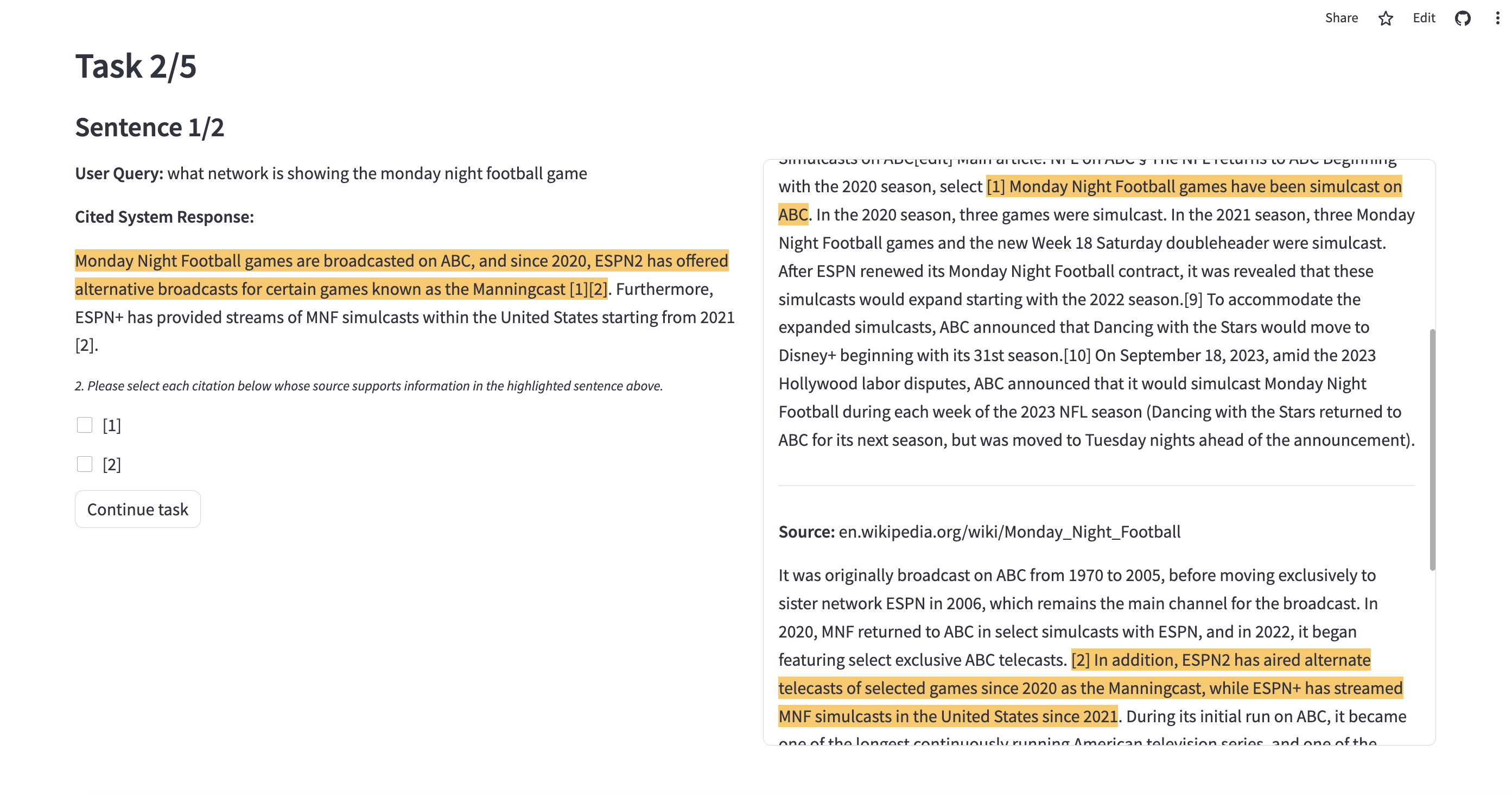}
    \caption{The annotation interface for evaluating the precision of a sentence.}
    \label{fig:annotation_interface_precision}
\end{figure*}

Annotators were directed from MTurk to complete their annotation session through a custom interface. After logging in with their user ID, an annotator is assigned a query-generation pair (a "Task") to evaluate from each operating point implementation being evaluated in the batch. Figure \ref{fig:annotation_interface_utility_fluency} shows the first screen displayed to an annotator for each query-generation pair, used to rate fluency and utility. After this first screen, the coverage and precision annotations are collected for each sentence in the generation. Figure \ref{fig:annotation_interface_coverage} shows the first screen displayed to an annotator for each sentence of the query-generation pair, used to annotate the coverage. Figure \ref{fig:annotation_interface_precision} shows the second screen displayed to an annotator for each sentence of the query-generation pair, used to annotate the precision. For both the coverage and precision evaluation screens, sources with highlighted quotes cited by the generation sentence in question are provided on the right.

The time to verify (T2V) a sentence is collected from a timer that starts when the user clicks the "Continue task" button in Figure \ref{fig:annotation_interface_utility_fluency} and ends when the user clicks the "Continue task" button in Figure \ref{fig:annotation_interface_coverage}.

\clearpage

\subsection{Failure Analysis}
\label{subsec: appendix_failure_analysis}

\subsubsection{Low Utility: Failure to interpret sources in-context}
\refstepcounter{egbox}
\begin{tcolorbox}[adjusted title=Example \theegbox, colback=white, arc=4mm, fontupper=\sffamily]
\raggedright
\noindent
\textbf{NQ Query: }who plays jack's mom in 30 rock 2\vspace{.2cm}\\\textbf{Quoted Generation: }"From 2007 to 2012, she had a recurring role as Colleen Donaghy on the NBC sitcom 30 Rock, a role that won her a third Emmy in 2007" [1].\vspace{.2cm}\\\textbf{Abstractive Generation: }Elaine Stritch played Jack Donaghy's mother, Colleen Donaghy, on the NBC sitcom "30 Rock" from 2007 to 2012, a role that earned her an Emmy in 2007 [1].\vspace{.2cm}\\\textbf{Sources: }

\href{https://en.wikipedia.org/wiki/Elaine_Stritch}{https://en.wikipedia.org/wiki/Elaine\_Stritch}\\
Stritch relocated to London in the 1970s and starred in several West End productions, including Tennessee Williams' Small Craft Warnings (1973) and Neil Simon's The Gingerbread Lady (1974). She also starred with Donald Sinden in the ITV sitcom Two's Company (1975–79), which earned her a 1979 BAFTA TV Award nomination. She won an Emmy Award in 1993 for her guest role on Law \& Order and another for the 2004 television documentary of her one-woman show Elaine Stritch at Liberty. From 2007 to 2012, she had a recurring role as Colleen Donaghy on the NBC sitcom 30 Rock, a role that won her a third Emmy in 2007. Early life 

\href{https://en.wikipedia.org/wiki/Elaine_Stritch}{https://en.wikipedia.org/wiki/Elaine\_Stritch}\\
After returning to the United States, she appeared on The Edge of Night as vinegary nanny Mrs. DeGroot, then was cast as a regular on the short-lived The Ellen Burstyn Show in 1986. She appeared as the stern schoolteacher Mrs. McGee on three episodes of The Cosby Show (1989–90). She had a recurring role in Law \& Order (1992, 1997) as Lanie Stieglitz. Other roles included Judge Grace Lema on Oz (1998); and Martha Albright (mother of Jane Curtin's character) on two episodes of 3rd Rock From the Sun (1997, 2001), alongside her Broadway co-star George Grizzard, who played George Albright. On April 26, 2007, she began guest appearances on the NBC sitcom 30 Rock as Colleen, the fearsome mother of Alec Baldwin's lead character, Jack Donaghy.

\href{https://www.vulture.com/2012/12/30-rock-recap-season-7-jenna-wedding-jacks-mother-dies.html}{https://www.vulture.com/2012/12/30-rock-recap-season-7-jenna-wedding-jacks-mother-dies.html}\\
Tonight’s 30 Rock contained a few things we all probably expected: the fallout from Ms. Elizabeth Lemon getting married without inviting alleged best friend and self-proclaimed moon of the universe Jenna; some transitional stuff to move us into the rest of the season; and a good bit of letdown owing to the fact that it’s not the Liz-Criss wedding episode anymore.   It also had some lovely surprises, including two super-solid Lost references; plenty of guest stars crowding onto the final-season lifeboat (Will Ferrell as Bitch Hunter!); and the revelation of what Jenna looks like under grocery-store-milk-aisle lighting. The latter was a truly Gollum-esque sight, far worse than Liz in HD, and proof of something I’ve believed since high school: She who controls the light board controls the world.   But the one thing about tonight that none of us (who don’t read spoilers) saw coming was the sad passing of Jack’s harpy mom, Colleen, played for seven seasons by the great Elaine Stritch in perhaps the most lovingly drawn New York City in-joke this show ever generated — and yes, I’m counting tonight’s Chuck Scarborough/Sue Simmons bonanza. Colleen was the kind of woman who could provoke a Buddhist monk into whipping a battery at her, who would conceive a child simply because Dragnet got preempted, and who always insisted upon traveling on Pearl Harbor Day to, and I quote, 'show the Emperor we’re not afraid.'
\label{utility_failure_misinterpretation_full}
\end{tcolorbox}

\subsubsection{Low Utility: Information retrieval failure}
In Example \ref{utility_failure_info_retrieval_full}, the retrieval system does not surface sources that answer the query. In contrast, the GPT-4 + Vertex API provides an answer from parametric knowledge. Gemini also provides an answer, perhaps due to a stronger retrieval system or dependence on parametric knowledge in the generation process. 

\refstepcounter{egbox}
\begin{tcolorbox}[adjusted title=Example \theegbox, colback=white, arc=4mm, fontupper=\sffamily]
\raggedright
\noindent
\textbf{NQ Query: }who did cora marry in once upon a time\vspace{.2cm}\\\textbf{Quoted Generation: }Cora "seizes her moment at a masquerade ball, slyly sidling up to the eligible Prince Henry and charming him into asking her to dance" [1].\vspace{.2cm}\\\textbf{Sources: }

\href{https://ew.com/recap/once-upon-a-time-season-2-cora-dies/}{https://ew.com/recap/once-upon-a-time-season-2-cora-dies/}\\
Luckily, young Cora also happens to look like Rose McGowan — still foxy even after the devastating car accident that caused her to need plastic surgery — which means she’s got a chance at marrying up. [1] She seizes her moment at a masquerade ball, slyly sidling up to the eligible Prince Henry and charming him into asking her to dance. The social climber’s ascent is stopped, though, when the king cuts in and sneers that she’ll never be good enough for his son. Kudos to Once for being the only TV show or movie ever to admit that covering just the area around your eyes is not an effective disguise. NEXT: Enter the Dragon — er, the Imp 

\href{https://en.wikipedia.org/wiki/The_Miller%27s_Daughter_(Once_Upon_a_Time)}{https://en.wikipedia.org/wiki/The\_Miller\%27s\_Daughter\_(Once\_Upon\_a\_Time)}\\
Cora and Regina arrive at the shop and overcome the protection spell. While David, Neal, and Emma stand against them, Mary Margaret sneaks away to Regina's mausoleum and uses the candle to curse Cora's heart. Regina follows after Cora senses that someone is there. Emma and Neal retreat to the back room, where she casts a new protection spell. Believing he will die, Gold asks to call Belle (Emilie de Ravin). Although Belle still doesn't remember Gold, she is moved when he tells her he loves her, and that she is a hero for loving a monster like him. He says that she inspires him to be his best self. Neal is surprised to hear such heartfelt words from his father, who then also apologizes to him. Neal affirms that he is still angry, but he tearfully embraces Gold. Regina discovers Mary Margaret at the mausoleum. 

\href{https://once-upon-the-once-upon-the-time.fandom.com/wiki/Cora/}{https://once-upon-the-once-upon-the-time.fandom.com/wiki/Cora}\\
Cora, a miller's daughter, earns extra money from working in a tavern. After chastely declining the flirtatious advances of patrons, she accidentally falls into Jonathan's lap. He, who has been coming to the pub for a week, offers her a seat since she has been working past the end of her shift. Jonathan compliments her for behaving with the grace of a lady, to which Cora talks about doing so due to her wish to be better than what she is. He mentions having to depart soon, and drops a handkerchief, which is embroidered with the royal crest. Citing that he had hoped for her to know him as a normal man rather than a prince, Jonathan proposes to her with a straw ring and promises, in two weeks' time, to return with a gold ring. Prior to his leaving, they consummate the marriage. After two weeks, Cora doesn't hear from Jonathan and finds him, two months later, at the royal estate, where she learns he is not a prince, but a gardener. When Cora admits being pregnant, she asks him to provide her with money for herself and the child to get by, and if not, she will tell the actual prince what he did to her. Jonathan doesn't care, so she calls for help, but he knocks her away and flees. She is found by Prince Leopold and later shows him how to build a campfire. As they bond, he tells her about his engagement to a princess, Eva, but decides to marry Cora instead. Two days before the wedding, Jonathan threatens to reveal their past and her pregnancy unless she provides him with riches. Before she can fulfill the deal, Eva, having overheard the conversation, exposes the truth to Leopold. With her lie exposed, Cora is hauled away by guards, and she gives birth to her child. Unable to give up on her dream to be more than a miller's daughter, Cora abandons her daughter to a passing cyclone, which whisks the infant to another world. ('Bleeding Through', 'It's Not Easy Being Green') 
\label{utility_failure_info_retrieval_full}
\end{tcolorbox}

\subsubsection{Low Fluency and Utility: quotes of dialogue are not identified as such in the quoted OP implementation}
An edge case not gracefully handled by our instantiation of the quoted operating point is the handling of quotations within quotes.

The quotations in the quoted generation below are not credited as an interview response given by LaTavia Roberson\footnote{https://people.com/music/latavia-roberson-getting-dumped-destinys-child-led-to-depression/}. Instead, the system generates this dialogue in the first-person.

\refstepcounter{egbox}
\begin{tcolorbox}[adjusted title=Example \theegbox, colback=white, arc=4mm, fontupper=\sffamily]
\raggedright
\noindent
\textbf{NQ Query: }what happened to the fourth member of destinys child\vspace{.2cm}\\\textbf{Quoted Generation: }\textcolor{orange}{"I never left Destiny’s Child" [1]}. \textcolor{orange}{"That is something that people say" [2]}. \textcolor{orange}{"But who would leave Destiny’s Child?" [3]} \textcolor{orange}{"That’s crazy!" [4]} \textcolor{orange}{"I was dismissed from the group" [5]}, and \textcolor{orange}{"It was very difficult because of the way that I found out about it" [6]}. \textcolor{orange}{"I hate even talking about it, and it’s been 20 years — but it is what it is" [7]}. \textcolor{orange}{"We saw the ‘Say My Name’ video on TV, and that’s how I found out I was no longer in the group" [8]}.
\end{tcolorbox}

\subsubsection{Low Utility: Logical inconsistency in entailed and abstractive OPs}
The director of The Navigator: A Medieval Odyssey was not born earlier, according to the provided dates.

\refstepcounter{egbox}
\begin{tcolorbox}[adjusted title=Example \theegbox, colback=white, arc=4mm, fontupper=\sffamily]
\raggedright
\noindent
\textbf{2WikiMH Query: }Which film has the director born earlier, The Navigator: A Medieval Odyssey or Ulsavapittennu?\vspace{.2cm}\\\textbf{Gemini Generation: }The Navigator: A Medieval Odyssey has the director born earlier. Its director, Vincent Ward, was born in 1956, while the director of Ulsavapittennu, Bharath Gopi, was born in 1937.
\end{tcolorbox}

The generation should say "No, Vavan and Kukuyevka are not located in the same country..." to be consistent with the query.

\refstepcounter{egbox}
\begin{tcolorbox}[adjusted title=Example \theegbox, colback=white, arc=4mm, fontupper=\sffamily]
\raggedright
\noindent
\textbf{2WikiMH Query: }Are Vavan and Kukuyevka both located in the same country?\vspace{.2cm}\\\textbf{Entailed Generation: }Yes, Vavan and Kukuyevka are not located in the same country; Vavan is in Iran, while Kukuyevka is in Russia [1][2].
\end{tcolorbox}

\subsubsection{Subjectivity in citation precision annotation}
\label{subsubsec:citation_subjectivity}
The citation precision failure analysis revealed annotations for precision errors with which we disagreed. We categorize the different false positive imprecise citation annotations we encountered in the failure analysis in Table \ref{tab:precision_prevalence_appendix}. Despite observing this noise in the precision annotations, we still expect the ordering of the OP instantiations by citation precision to hold because generations from each instantiation are subject to the same errors in annotator judgment. We provide examples of subjective citation precision annotation decisions in Examples \ref{specificity_preference_1} and \ref{specificity_preference_2}.

\begin{table}[h]
\centering
\begin{tabular}{lccccccc}
\toprule
               & \begin{tabular}[c]{@{}c@{}}Overall \\ imprecise\\ citation\\ annotation\\ prevalence\end{tabular} 
               & \begin{tabular}[c]{@{}c@{}}Claim \\ taken out \\ of context\end{tabular} 
               & \begin{tabular}[c]{@{}c@{}}Irrelevant \\ source quote:\\ Keywords \\ as proxies\end{tabular} 
               & \begin{tabular}[c]{@{}c@{}}Relevant \\ source \\ quote:\\ Claim too \\ specific\end{tabular} 
               & \begin{tabular}[c]{@{}c@{}}False\\ positive:\\ Ignores \\ context\end{tabular} 
               & \begin{tabular}[c]{@{}c@{}}False \\ positive:\\ User error\end{tabular} 
               & \begin{tabular}[c]{@{}c@{}}False \\ positive:\\ Evaluates\\ coverage\end{tabular} \\ 
\midrule
Quoted   & 16/1466 & 3/16 & 0/16 & 0/16 & 7/16 & 6/16 & 0/16 \\ 
Entailed & 40/998  & 1/40 & 8/40 & 14/40 & 1/40 & 8/40 & 8/40 \\ 
GPT-4 + Vertex & 40/191  & 5/40 & 23/40 & 6/40 & 5/40 & 0/40 & 1/40 \\ 
Gemini         & 40/229  & 1/40 & 21/40 & 6/40 & 3/40 & 2/40 & 7/40 \\ 
\bottomrule
\end{tabular}
\caption{Prevalence of different true positive and false positive precision failure categories in quoted, entailed, GPT-4 + Vertex, and Gemini generations. The left column provides the overall prevalence of imprecise citation annotations, while the remaining columns show occurrences of specific failure categories.}
\label{tab:precision_prevalence_appendix}
\end{table}

The generation calls the scrolls that inspired manga "emaki", but citation 1 (Source 1) refers to them as "emakimono". Further research reveals that "emaki" is a valid abbreviation, but some users may prefer the inclusion of a citation that specifies this.\footnote{https://en.wikipedia.org/wiki/Emakimono}

\refstepcounter{egbox}
\begin{tcolorbox}[adjusted title=Example \theegbox, colback=white, arc=4mm, fontupper=\sffamily]
\raggedright
\noindent
\textbf{ETa3G Query: }Explain to a third-grader: what was the first form of manga in japan\vspace{.2cm}\\\textbf{Gemini Generation: }Imagine drawing stories on long scrolls. Long ago, in Japan, people told stories with pictures instead of just words. These pictures were like comic books we have today, but they were much longer and were called "emaki." They were like a giant picture book that you could unroll to see the whole story. These scrolls were the first form of manga \textcolor{orange}{[1]}!\vspace{.2cm}\\\textbf{Source 1: }\href{https://en.wikipedia.org/wiki/History\_of\_manga\#:\~:text=6\%20External\%20links-,Before\%20World\%20War\%20II,the\%2012th\%20and\%2013th\%20centuries.}{https://en.wikipedia.org/wiki/History\_of\_manga\#:\~:text=6\%20External\%20links-,Before\%20World\%}\\\textcolor{orange}{[1] Manga is said to originate from emakimono (scrolls), Chōjū-jinbutsu-giga, dating back to the 12th and 13th centuries.} During the Edo period (1603–1867), another book of drawings, Toba Ehon, embedded the concept of manga. The word first came into common usage in the late 18th and early 19th centuries, with the publication of such works as Santō Kyōden's picturebook Shiji no yukikai (1798), and Aikawa Minwa's Manga hyakujo (1814); this also includes the celebrated Hokusai Manga books (1814–1834) which contain assorted drawings from the sketchbooks of the famous ukiyo-e artist Hokusai (1760–1849). Kitazawa Rakuten (1876–1955) was the first artist to use the word manga in the modern sense. Another example in the first half of the 19th century is speculated to be Dehōdai mucharon (1822) with prints from the artist Hiroshige, who illustrated several books of this kind between 1820 and 1837.
\label{specificity_preference_1}
\end{tcolorbox}

Citation 1 (Source 1) specifies that the presiding officer in question is referred to as the Chairman, rather than the Chairperson as the response suggests. Chairman and Chairperson are used interchangeably in this context and a citation to a source indicating this\footnote{https://timesofindia.indiatimes.com/city/nagpur/vacancy-of-chairman-in-maharashtra-legislative-council/articleshow/105486799.cms} may be desirable.

\refstepcounter{egbox}
\begin{tcolorbox}[adjusted title=Example \theegbox, colback=white, arc=4mm, fontupper=\sffamily]
\raggedright
\noindent
\textbf{NQ Query: }who is the presiding officer of legislative council in india\vspace{.2cm}\\\textbf{Gemini Generation: }The presiding officer of a legislative council in India is called a Chairperson \textcolor{orange}{[1]}. \textcolor{gray}{This position is analogous to the Speaker in the lower house (Legislative Assembly). The Chairperson is responsible for overseeing the proceedings of the council, maintaining order, and ensuring that the rules of procedure are followed.}\vspace{.2cm}\\\textbf{Source 1: }\href{https://www.business-standard.com/about/what-is-legislative-council\#:\~:text=The\%20legislative\%20Council\%20elects\%20its,Chairman\%20from\%20amongst\%20its\%20members.}{https://www.business-standard.com/about/what-is-legislative-council\#:\~:text=The\%20legislative\%20}\\. How is the Legislative Council created? The members of the Council are either nominated by the Governor of the state or are indirectly elected. Here's the math: One-third of the members of this House are elected by the Legislative Assembly. One-third are elected by the local bodies like a municipality or other local authorities. One-twelfth of the members are elected by graduates. One-twelfth of the members are elected by teachers. About one-sixth of the members are nominated by the Governor. \textcolor{orange}{[1] The legislative Council elects its Chairman, who plays the role of presiding officer and Deputy Chairman from amongst its members.} Eligibility criteria: Indian citizen who is at least 30 years of age; a person cannot simultaneously be a Member of Parliament and State legislature. What is the role of the Legislative Council? The Constitution of India gives limited power to the Legislative Council. It cannot make or break a government. It cannot say ‘no' to Finance Bills. However, there are some p
\label{specificity_preference_2}
\end{tcolorbox}

\subsection{Further Implementation Details}
\label{subsec:more_implementation_details}
\subsubsection{Further Implementation Details of the OPs}
To obtain retrieved snippets given a query, we scrape text from the top 50 Google Search results for the query, chunk the text into snippets about 1000 characters long, and select the ten most relevant snippets to the query using query-document dense passage retrieval \citep{karpukhin-etal-2020-dense}.

The \textbf{quoted generation} is generated by conditioning GPT-4 generation on few-shot examples, the quoting instruction, the sub-queries derived from the query, and either the gold or retrieved snippets. We found that breaking up the query (e.g. "Where was the director of film Out All Night (1933 Film) born?") into sub-queries (e.g. "Who directed the film Out All Night (1933 Film)? Where were they born?") was especially helpful for queries with multiple components, such as those from 2WikiMH.

The prompt instructions used to generate the reference generations remain the same for each OP across query distributions, although the few-shot examples are specific to each query distribution. The citations are all generated using the same prompt instruction, but with different few-shot examples for different query distributions. All of the few-shot examples for the quoted, paraphrased, entailed, and abstractive generations were carefully selected in equal numbers for each OP for each dataset. 

\begin{figure*}[ht]
\centering
    \resizebox{.75\textwidth}{!}{
    \begin{subfigure}
        \centering
        \includegraphics{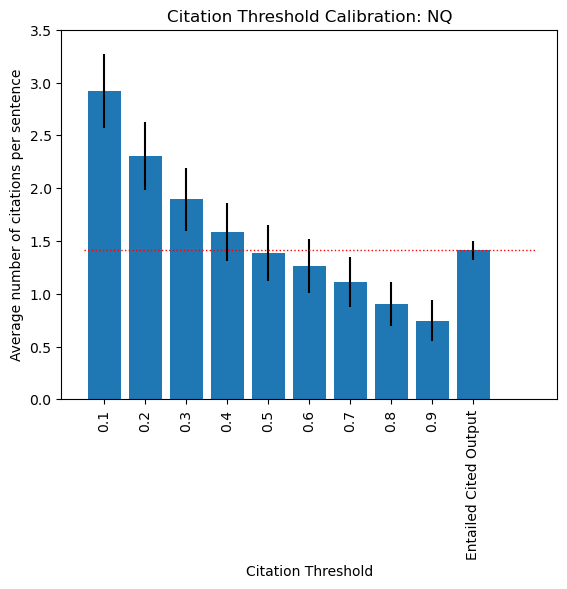}
    \end{subfigure}
    \begin{subfigure}
        \centering
        \includegraphics{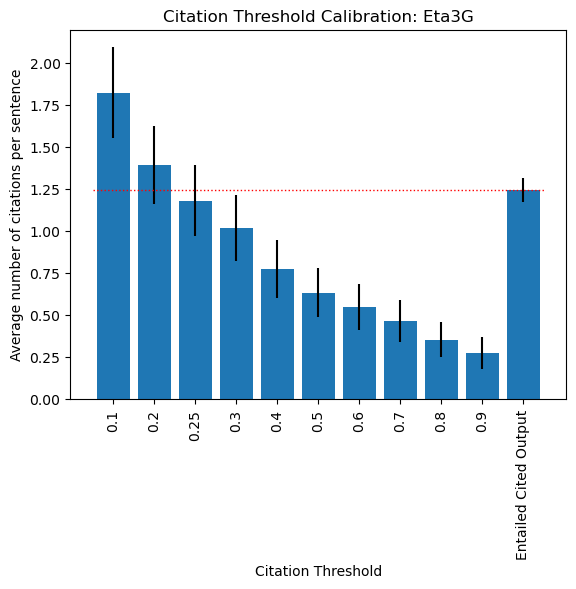}
    \end{subfigure}
    }
    \resizebox{.75\textwidth}{!}{
    \begin{subfigure}
        \centering
        \includegraphics{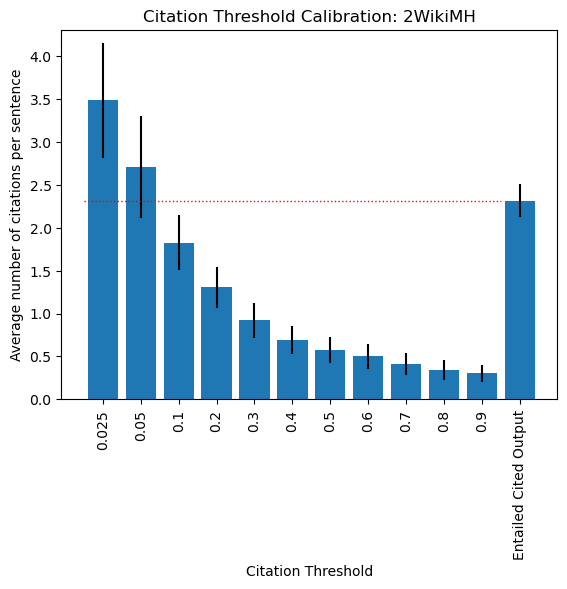}
    \end{subfigure}
    \begin{subfigure}
        \centering
        \includegraphics{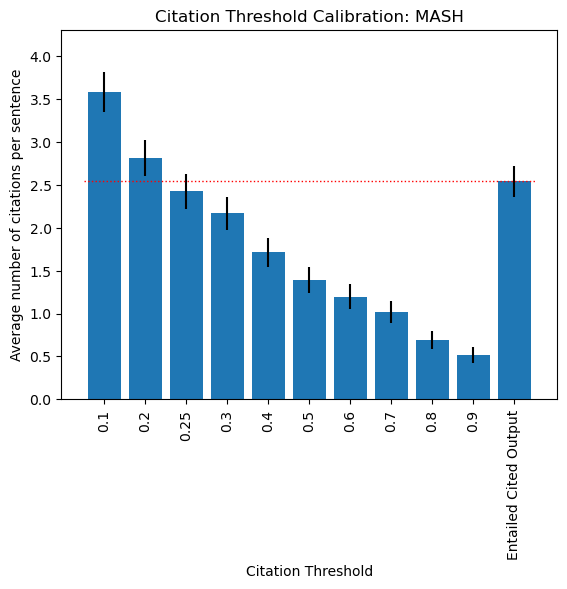}
    \end{subfigure}
    }
\caption{Average number of citations from the Vertex API per sentence by Citation Threshold for GPT-4 generations, compared to the average number of citations per sentence from the entailed generation (red dotted line). Only sentences that require citation, according to the Vertex API, are considered. These results led us to set the citation threshold for the NQ, Eta3G, 2WikiMH, and MASH datasets to 0.5, 0.25, 0.05, and 0.25 respectively.}
\label{fig:post_hoc_calibration}
\end{figure*}

\subsubsection{GPT-4 + Vertex Generation Implementation Details}
\label{gpt-4 output details}

The response is generated by conditioning GPT-4 on an instruction and query, without retrieved sources. In the instruction, we specify the style of the response as paragraph form and the target length of the generation to be the same as the average length of the entailed generations for each dataset. After this response is generated, supporting sources snippets are then retrieved using Google search to identify the top 50 webpage results and a semantic similarity model to identify snippets about 1000 characters long within those webpages that are relevant to the response.\footnote{\href{https://huggingface.co/sentence-transformers/all-MiniLM-L6-v2}{https://huggingface.co/sentence-transformers/all-MiniLM-L6-v2}} These sources are concatenated with the sources used for the same query in generating the quoted generation. Finally, for each sentence in the generation, citations to individual sentences in the retrieved sources are identified using Google's Vertex API.\footnote{\href{https://cloud.google.com/generative-ai-app-builder/docs/check-grounding}{https://cloud.google.com/generative-ai-app-builder/docs/check-grounding}}

Google's Vertex API requires a \textit{citation threshold}, $\alpha \in [0,1]$, that controls the confidence for provided citations. Accordingly, higher levels of $\alpha$ lead to fewer citations per sentence and lower levels of $\alpha$ lead to more citations per sentence. We choose the citation threshold by calibrating the average number of citations per sentence from post hoc cited generations to that of entailed cited generations for each dataset (Figure \ref{fig:post_hoc_calibration}). The citation thresholds for the NQ, Eta3G, 2WikiMH, and MASH datasets are set to 0.5, 0.25, 0.05, and 0.25, respectively.

\begin{table}[ht]
\centering
\begin{tabular}{lllll}
                      & NQ             & ELI3G & 2WikiMH & MASH \\ \cline{2-5} 
Entailed Generation & 30.1 +/- 2.8  & 30.2 +/- 2.9 & 
16.7 +/ 1.2 & 55.2 +/- 3.7\\ \cline{2-5} 
GPT-4 + Vertex Generation & 29.5 +/- 0.5 & 31.2 +/- 0.5 & 
14.9 +/- 0.3 & 61.8 +/- 0.8 \\ \cline{2-5} 
\end{tabular}
\caption{Average number of words per generation with 95\% confidence intervals. The average GPT-4 + Vertex generation length matches that of the entailed generations.}
\label{tab:baseline_output_length_calibration}
\end{table}

We also observe that longer responses often have higher citation counts, likely because additional information in the response calls for more citations. Because of this, we also control the word count of GPT-4 generations through the prompt instruction to be similar to the word count of entailed cited generations (Table \ref{tab:baseline_output_length_calibration}).

\subsubsection{Gemini Generation Implementation Details}
\label{gemini output details}

In our work, a Google Gemini generation was obtained by passing in an instruction (the same used for the GPT-4 + Vertex generation) and the query. To obtain citations, we used the "Double-check response" feature on \href{https://gemini.google.com/app}{https://gemini.google.com/app}. Gemini citations include citations to supporting sources, highlighted in green, and to conflicting sources, highlighted in orange. We only consider Gemini's citations to supporting sources. Gemini provides at most one citation per sentence in the response. Each citation corresponds to a sentence in a website; we scrape each website and use a 1000 character span centered on the cited sentence as the citation's source. The generations were generated and scraped throughout August and September 2024.

\subsection{Efficiency Analysis}

\begin{figure*}[ht]
\centering
    \resizebox{\textwidth}{!}{
    \begin{subfigure}
        \centering
        \includegraphics{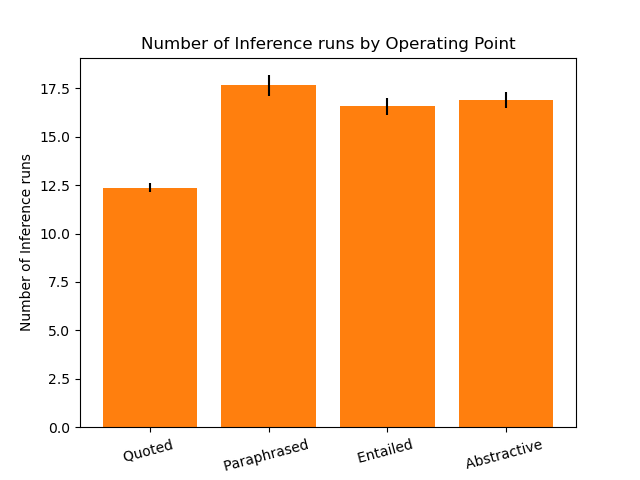}
    \end{subfigure}
    \begin{subfigure}
        \centering
        \includegraphics{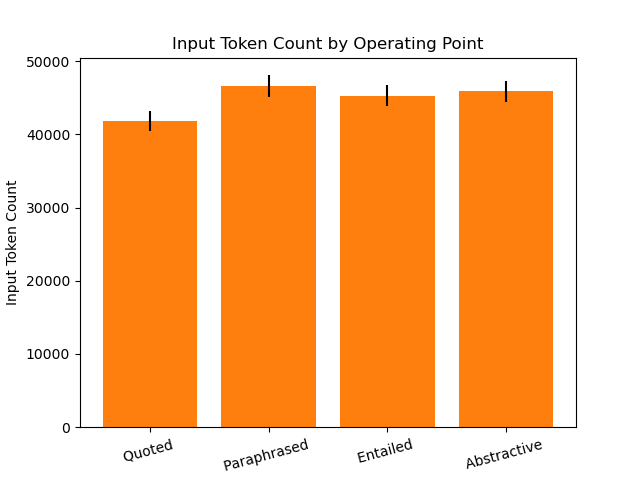}
    \end{subfigure}
    \begin{subfigure}
        \centering
        \includegraphics{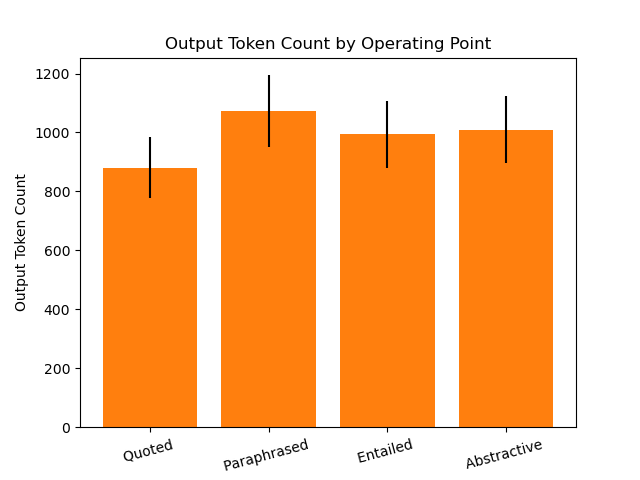}
    \end{subfigure}
    }
\caption{LLM generation efficiency of the generations quantified by the number of inference runs, input tokens, and generation tokens per query, averaged over the four query distributions.}
\label{fig:inference_efficiency}
\end{figure*}

The largest difference in resource usage across the reference OPs is between the quoted generation and the others---citations for the quoted generation are identifiable via exact string matching, rather than calls to the LLM (Figure \ref{fig:inference_efficiency}). Differences in efficiency across the paraphrased, entailed, and abstractive generations are due to differences in the number of sentences and citations.

\clearpage
\subsection{Prompts with Few-Shot Examples}
\label{subsec:prompts}
\subsubsection{Prompts for the Quoted Generation}
\label{Prompts for the Quoted Output}
\vspace{.2cm}
\begin{promptbox}
\raggedright
\textcolor{instructioncolor}{Instruction: }Respond to the following query using word-for-word quotes from the sources provided below. The following sub-questions will help you answer the query; be sure that your response to the query answers each sub-question using a quotation from a source. Clearly indicate the quotes to avoid plagiarizing! Be concise in your response and focus on information that responds to the query. Do not refer to the sources in your response.The provided sources should contain enough information to write a response to the query. However, in the rare case that the sources are insufficient, then respond with "Insufficient information to generate a grounded response."

\textcolor{querycolor}{Query:} What's the life cycle of a frog?

\textcolor{subquestioncolor}{Sub-questions:} What's the life cycle of a frog?

\textcolor{sourcescolor}{Sources:} 

".S., Applied Ecology, Indiana University BloomingtonB.S., Biology and Chemistry, University of Illinois at Urbana-ChampaignLaura Klappenbach, M.S., is a science writer specializing in ecology, biology, and wildlife.Learn about ourEditorial ProcessUpdated on August 11, 2019The life cycle of a frog consists of three stages: egg, larva, and adult. As the frog grows, it moves through these stages in a process known as metamorphosis."

"But in a few species, parents remain with the eggs to look after them as they develop. As the fertilized eggs mature, the yolk in each egg splits into more and more cells and begins to take the form of a tadpole, the larva of a frog. Within one to three weeks, the egg is ready to hatch, and a tiny tadpole breaks free.03of 04Stage 2: Tadpole (Larva)Johner Images / Getty ImagesTadpoles, frogs' larvae, have rudimentary gills, a mouth, and a long tail. For the first week or two after the tadpole hatches, it moves very little. During this time, the tadpole absorbs the remaining yolk left over from the egg, which provides much-needed nourishment. After absorbing the yolk, the tadpole is strong enough to swim on its own.Most tadpoles feed on algae and other vegetation, so they are considered herbivores. They filter material from the water as they swim or tear away bits of plant material. As the tadpole continues to grow, it begins to develop hind limbs."

"The legless, water-bound tadpoles slowly metamorphose into frogs over the next 14 weeks. First, they grow back legs, then front legs too! Soon after, their body starts to change shape, and they’re able to start eating insects. Next, the tadpoles’ tails shrink away, and skin grows over their gills, as they develop lungs and eardrums! These are super important steps, as they prepare the tadpole for life on land. Once their gills and tails are gone forever, tadpoles undergo one last 24-hour push, where the metamorphosis completes. Once this stage is finished, the baby frogs emerge from the water as tiny adults!"

\textcolor{responsecolor}{Response: }"The life cycle of a frog consists of three stages: egg, larva, and adult. As the frog grows, it moves through these stages in a process known as metamorphosis." During the egg stage, "the yolk in each egg splits into more and more cells and begins to take the form of a tadpole, the larva of a frog. Within one to three weeks, the egg is ready to hatch, and a tiny tadpole breaks free." "Tadpoles, frogs' larvae, have rudimentary gills, a mouth, and a long tail." As they grow, "tadpoles feed on algae and other vegetation," and eventually, "they grow back legs, then front legs too! Soon after, their body starts to change shape, and they’re able to start eating insects. Next, the tadpoles’ tails shrink away, and skin grows over their gills, as they develop lungs and eardrums!"

\textcolor{instructioncolor}{Instruction: }Respond to the following query using word-for-word quotes from the sources provided below. The following sub-questions will help you answer the query; be sure that your response to the query answers each sub-question using a quotation from a source. Clearly indicate the quotes to avoid plagiarizing! Be concise in your response and focus on information that responds to the query. Do not refer to the sources in your response.The provided sources should contain enough information to write a response to the query. However, in the rare case that the sources are insufficient, then respond with "Insufficient information to generate a grounded response."

\textcolor{querycolor}{Query:} What are the benefits of hyaluronic acid for skin care?

\textcolor{subquestioncolor}{Sub-questions:} What are the benefits of hyaluronic acid for skin care?

\textcolor{sourcescolor}{Sources:} 

"Hyaluronic acid is a remarkable substance because of all the benefits and uses it has in your body. Here are just a few of the benefits of hyaluronic acid: It helps things move smoothly. Hyaluronic acid helps your joints work like a well-oiled machine. It prevents pain and injury from bones grinding against each other. It helps keep things hydrated."

"Hyaluronic acid is very good at retaining water. A quarter-teaspoon of hyaluronic acid holds about one and a half gallons of water. That’s why hyaluronic acid is often used for treating dry eyes. It’s also used in moisturizing creams, lotions, ointments and serums. It makes your skin flexible. Hyaluronic acid helps skin stretch and flex and reduces skin wrinkles and lines. Hyaluronic acid is also proven to help wounds heal faster and can reduce scarring."

\textcolor{responsecolor}{Response: }Hyaluronic acid "helps keep things hydrated". It "helps skin stretch and flex and reduces skin wrinkles and lines." "Hyaluronic acid is also proven to help wounds heal faster and can reduce scarring."

\textcolor{instructioncolor}{Instruction: }Respond to the following query using word-for-word quotes from the sources provided below. The following sub-questions will help you answer the query; be sure that your response to the query answers each sub-question using a quotation from a source. Clearly indicate the quotes to avoid plagiarizing! Be concise in your response and focus on information that responds to the query. Do not refer to the sources in your response.The provided sources should contain enough information to write a response to the query. However, in the rare case that the sources are insufficient, then respond with "Insufficient information to generate a grounded response."

\textcolor{querycolor}{Query:} What is a heirloom tomato?

\textcolor{subquestioncolor}{Sub-questions:} What is a heirloom tomato?

\textcolor{sourcescolor}{Sources:} 

"They are classified as family heirlooms, commercial heirlooms, mystery heirlooms, or created heirlooms. They usually have a shorter shelf life and are less disease resistant than hybrids. They are grown for various reasons: for food, historical interest, access to wider varieties, and by people who wish to save seeds from year to year, as well as for their taste.[1]Taste[edit]Many heirloom tomatoes are sweeter and lack a genetic mutation that gives tomatoes a uniform red color at the cost of the fruit's taste.[2] Varieties bearing that mutation which have been favored by industry since the 1940s – that is, tomatoes which are not heirlooms – feature fruits with lower levels of carotenoids and a decreased ability to make sugar within the fruit."

"True heirlooms must be 'old' because they have been a variety for anywhere between 25 and 50 years. The seeds, like furniture or jewelry, must also be passed down from generation to generation. There is no exact blueprint for the color and shape of an heirloom tomato, so long as it meets the three characteristics. There are over 100 different types of heirloom tomatoes in the united states and an estimated 3,000 globally, which are suitable for very different climates and growing regions."

\textcolor{responsecolor}{Response: }"Many heirloom tomatoes are sweeter and lack a genetic mutation that gives tomatoes a uniform red color at the cost of the fruit's taste." "True heirlooms must be 'old' because they have been a variety for anywhere between 25 and 50 years."

\textcolor{instructioncolor}{Instruction: }Respond to the following query using word-for-word quotes from the sources provided below. The following sub-questions will help you answer the query; be sure that your response to the query answers each sub-question using a quotation from a source. Clearly indicate the quotes to avoid plagiarizing! Be concise in your response and focus on information that responds to the query. Do not refer to the sources in your response.The provided sources should contain enough information to write a response to the query. However, in the rare case that the sources are insufficient, then respond with "Insufficient information to generate a grounded response."

\textcolor{querycolor}{Query:} What's the flavor of juicy fruit gum?

\textcolor{subquestioncolor}{Sub-questions:} What's the flavor of juicy fruit gum?

\textcolor{sourcescolor}{Sources:} 

 "A Juicy Fruit wrapper from 1946 reportedly advertised that the gum had 'a fascinating artificial flavor,' which doesn't really provide much clarity. In 2002, one curious fan took their question to the company itself and received the following statement in response, 'I'm afraid we can't be very specific because, for competitive reasons, we consider our Juicy Fruit flavoring formula to be a trade secret. I can, however, tell you that the mixture of fruit flavors in Juicy Fruit is comprised of predominately lemon, orange, pineapple, and banana notes.'"

\textcolor{responsecolor}{Response: }"the mixture of fruit flavors in Juicy Fruit is comprised of predominately lemon, orange, pineapple, and banana notes."

\textcolor{instructioncolor}{Instruction: }Respond to the following query using word-for-word quotes from the sources provided below. The following sub-questions will help you answer the query; be sure that your response to the query answers each sub-question using a quotation from a source. Clearly indicate the quotes to avoid plagiarizing! Be concise in your response and focus on information that responds to the query. Do not refer to the sources in your response.The provided sources should contain enough information to write a response to the query. However, in the rare case that the sources are insufficient, then respond with "Insufficient information to generate a grounded response."

\textcolor{querycolor}{Query:} who invented the cornell method of note taking

\textcolor{subquestioncolor}{Sub-questions:} Who invented the Cornell method of note taking?

\textcolor{sourcescolor}{Sources:}

"It gained national popularity when he promoted it in his 1962 book, How to Study in College—which remains in print, in its 11th edition. (The father of two Cornellians, Pauk retired from the University in 1978 and passed away just three years ago—at the age of 105.­) His system divides each notebook page into two unequal columns with a third, 'summary' section across the bottom. Students, whether attending a lecture or reading, record their notes and abbreviated facts in the wider right-hand column. Questions prompted by the material go on the left side—usually written afterward, but when the information is still fresh. 'As you’re taking notes, keep the cue column empty,' Pauk wrote. 'But when you review and recite what you’ve jotted down, draw questions from the ideas in your notes and write them in the cue column. Writing questions helps clarify meanings, reveal relationships, establish continuity, and strengthen memory.' "

"But Cornell Notes remains his most prominent legacy—and for decades, Pauk’s system has been popular across subject areas. It’s used by students in high school, college, and beyond, with free templates available online. The technique remains flexible and adaptable, Bokaer-Smith says, and is just as useful today for online study and with e-textbooks. In a 2021 teaching guide offering insights 'by grad students, for grad students,' two Cornellians studying psychology (doctoral candidate Julia Nolte, MA ’20, and Hamid Turker, PhD ’22, now a postdoc on the Hill) contributed a section about Cornell Notes. Pauk’s system is used by students in high school, college, and beyond. Nolte—who plans to use the system in her fall 2022 class on aging and the media—says it has several key advantages, including its templates and organizational approach. "

"The Cornell method provides a systematic format for condensing and organizing notes. This system of taking notes is designed for use by a high school or college level student. There are several ways of taking notes, but one of the most common is the 'two-column' notes style. The student divides the paper into two columns: the note-taking column (usually on the right) is twice the size of the questions/keyword column, which is on the left. The student leaves five to seven lines open, or about two inches (5 cm), at the bottom of the page. Notes from a lecture or text are written in the note-taking column; notes usually "

"To use Cornell, start by leaving a two-inch column on the left-hand side of the page.  You should also be sure to leave a space of 3-4 lines blank at the bottom of the page.  Once these areas have been designated you can take notes on the rest of the page as you normally would.  Don’t be afraid to use blank space to separate and organize new concepts. After you’ve finished taking notes for the class you’ll want to go back (immediately, if you can) and use that blank column on the left to write questions that can be answered by the corresponding notes.  Remember to put the question as close as possible to the note that answers it.   Later on, you can quiz yourself by covering the main note section of the page and answering the questions in your own words. Finally, use the blank bottom section to briefly summarize the page in one or two sentences. "

"The Cornell Method was invented about sixty years ago (see Walter Pauk’s 1962 classic How to Study at College, now in its tenth edition), though I only found out about it in 2012.  It incorporates a lot of what I was doing already—providing spaces for notes, and margins for reactions, connections and comments.  But it takes it further, and adds some very cool functions. Firstly, the template gives you less space to write notes.  You aren’t supposed to record everything you see, or even everything that is interesting.  Having fewer lines to write notes encourages you to be selective—just to chose the quotes or paraphrases or details you expect to include in your thesis. It’s so liberating to realise there’s no way you will be able to only read an article once for all for all time. Of course you need to reread it each time you want to use it for a new purpose, but you can be focused on what you’re pretty sure you need for this article. (Which, face it, is usually 1 quote.) "

"There are many ways to take notes, and everyone seems to have their preference when it comes to note taking. Many methods involve some or all of the 5 R’s of note taking: record, reduce, recite, reflect, and review. Too many students, however, are still using the same basic note taking method they first learned in middle school. Like any skill, note taking develops with practice. Practicing proven note taking techniques can dramatically improve a student’s academic performance. One popular method of note taking is the Cornell System. It was developed by Walter Pauk, a college professor at CornellUniversity and author of academic success books. The Cornell System is a method for taking effective notes in a systematic manner. HOW TO FORMAT YOUR PAGES "

"Some people think the Cornell Method can only be implemented for handwriting. But remote workers use this method to take digital notes. Although there is a journal that highlighted the advantage of students who utilised study skills by taking notes by hand. The student managed to answer conceptual questions better than students who took notes with laptops. So, this article will cover the Cornell Method and its iconic 5 Rs. What is the Cornell Method? The Cornell Method is often referred to as the Cornell Notes System or the 5 Rs of note-taking. This method was invented in the 1950s by Walter Pauk, a professor of education at Cornell University. Professor Pauk writes about the implementation of the Cornell Method in his book: How to Study in College. The Cornell Method can be applied to both traditional and digital, making this method popular nowadays. Students and professionals can use this method to record lessons or during meetings. What makes the Cornell Method easy to use? The Cornell Method offers efficiency. The five steps which include record, reduce, recite, reflect and review, are easy to remember. This system encourages us to reflect on notes and summarise them in our own words. Despite its simplicity, the method makes us think critically about the key concepts in each lesson. We have to explain the main points in detail, so we are more likely to remember them. Also, it has a logical and easy-to-read layout. The left side contains the main points and the right side for details. This will make it easier to find information when referencing the record. Cornell University suggests closing the right side, which contains detailed definitions. Then see if we can redefine the bullet points. This note-taking tip will help you quickly identify topics that you want to review. Source: Chloe The 5 Rs of Note-Taking "

"Cornell Notes: A Step-by-Step Guide Use CasesFoundersManagersInvestorsSalesProduct ManagersPricingAboutBlogBook demoBook demoTry nowTry nowProductivity HacksCornell Notes: A Step-by-Step GuidebyLeo-07.10.2024Table of ContentsExample H2Example H3Ready to love your workday again?Use Jamie for free.Try nowTry now Your browser does not support the video tag. Reading time:5minThe Cornell note taking method is a structured system for recording, summarizing, and reviewing notes. Developed by Professor Walter Pauk of Cornell University, it divides the note page into three sections: a main note-taking area, a left-hand column for cues and questions, and a summary area at the bottom, facilitating an organized review and retention of information. Enhance your note-taking experience Jamie: Automated summaries for efficient learning Try Jamie for Free "

"How should you take notes in class? Like so many stu­dents who came before me and would come after, I had lit­tle idea in col­lege and even less in high school. The inher­ent­ly ambigu­ous nature of the note-tak­ing task has inspired a vari­ety of meth­ods and sys­tems, few of them as respect­ed as Cor­nell Notes. Invent­ed in the 1940s by Cor­nell Uni­ver­si­ty edu­ca­tion pro­fes­sor Wal­ter Pauk, author of How to Study in Col­lege, Cor­nell Notes involves divid­ing each page up into three sec­tions: one to para­phrase the lec­ture’s main ideas, one to sum­ma­rize those ideas, and one to write ques­tions. After writ­ing down those main ideas dur­ing class, imme­di­ate­ly sum­ma­rize and add ques­tions about the con­tent. Then, while study­ing lat­er, try to answer those ques­tions with­out look­ing at the main body of notes. "
\textcolor{responsecolor}{Response: }\end{promptbox}
\refstepcounter{promptbox}
\textbf{Box \thepromptbox: } Example of the prompt with few-shot examples used to generate the quoted output for NQ queries.
\label{box:nq_quoted_output_prompt}
\vspace{.2cm}

\vspace{.2cm}
\begin{promptbox}
\raggedright
\textcolor{instructioncolor}{Instruction: }Respond to the following query using word-for-word quotes from the sources provided below. The following sub-questions will help you answer the query; be sure that your response to the query answers each sub-question using a quotation from a source. Clearly indicate the quotes to avoid plagiarizing! Be concise in your response and focus on information that responds to the query. Do not refer to the sources in your response.The provided sources should contain enough information to write a response to the query. However, in the rare case that the sources are insufficient, then respond with "Insufficient information to generate a grounded response."

\textcolor{querycolor}{Query:} Explain to a third-grader: what's the life cycle of a frog?

\textcolor{subquestioncolor}{Sub-questions:} Explain to a third-grader: what's the life cycle of a frog?

\textcolor{sourcescolor}{Sources:} 

".S., Applied Ecology, Indiana University BloomingtonB.S., Biology and Chemistry, University of Illinois at Urbana-ChampaignLaura Klappenbach, M.S., is a science writer specializing in ecology, biology, and wildlife.Learn about ourEditorial ProcessUpdated on August 11, 2019The life cycle of a frog consists of three stages: egg, larva, and adult. As the frog grows, it moves through these stages in a process known as metamorphosis."

"But in a few species, parents remain with the eggs to look after them as they develop. As the fertilized eggs mature, the yolk in each egg splits into more and more cells and begins to take the form of a tadpole, the larva of a frog. Within one to three weeks, the egg is ready to hatch, and a tiny tadpole breaks free.03of 04Stage 2: Tadpole (Larva)Johner Images / Getty ImagesTadpoles, frogs' larvae, have rudimentary gills, a mouth, and a long tail. For the first week or two after the tadpole hatches, it moves very little. During this time, the tadpole absorbs the remaining yolk left over from the egg, which provides much-needed nourishment. After absorbing the yolk, the tadpole is strong enough to swim on its own.Most tadpoles feed on algae and other vegetation, so they are considered herbivores. They filter material from the water as they swim or tear away bits of plant material. As the tadpole continues to grow, it begins to develop hind limbs."

"The legless, water-bound tadpoles slowly metamorphose into frogs over the next 14 weeks. First, they grow back legs, then front legs too! Soon after, their body starts to change shape, and they’re able to start eating insects. Next, the tadpoles’ tails shrink away, and skin grows over their gills, as they develop lungs and eardrums! These are super important steps, as they prepare the tadpole for life on land. Once their gills and tails are gone forever, tadpoles undergo one last 24-hour push, where the metamorphosis completes. Once this stage is finished, the baby frogs emerge from the water as tiny adults!"

\textcolor{responsecolor}{Response: }"The life cycle of a frog consists of three stages: egg, larva, and adult. As the frog grows, it moves through these stages in a process known as metamorphosis." During the egg stage, "the yolk in each egg splits into more and more cells and begins to take the form of a tadpole, the larva of a frog. Within one to three weeks, the egg is ready to hatch, and a tiny tadpole breaks free." "Tadpoles, frogs' larvae, have rudimentary gills, a mouth, and a long tail." As they grow, "tadpoles feed on algae and other vegetation," and eventually, "they grow back legs, then front legs too! Soon after, their body starts to change shape, and they’re able to start eating insects. Next, the tadpoles’ tails shrink away, and skin grows over their gills, as they develop lungs and eardrums!"

\textcolor{instructioncolor}{Instruction: }Respond to the following query using word-for-word quotes from the sources provided below. The following sub-questions will help you answer the query; be sure that your response to the query answers each sub-question using a quotation from a source. Clearly indicate the quotes to avoid plagiarizing! Be concise in your response and focus on information that responds to the query. Do not refer to the sources in your response.The provided sources should contain enough information to write a response to the query. However, in the rare case that the sources are insufficient, then respond with "Insufficient information to generate a grounded response."

\textcolor{querycolor}{Query:} Explain to a third-grader: what are the benefits of hyaluronic acid for skin care?

\textcolor{subquestioncolor}{Sub-questions:} Explain to a third-grader: what are the benefits of hyaluronic acid for skin care?

\textcolor{sourcescolor}{Sources:} 

"Hyaluronic acid is a remarkable substance because of all the benefits and uses it has in your body. Here are just a few of the benefits of hyaluronic acid: It helps things move smoothly. Hyaluronic acid helps your joints work like a well-oiled machine. It prevents pain and injury from bones grinding against each other. It helps keep things hydrated."

"Hyaluronic acid is very good at retaining water. A quarter-teaspoon of hyaluronic acid holds about one and a half gallons of water. That’s why hyaluronic acid is often used for treating dry eyes. It’s also used in moisturizing creams, lotions, ointments and serums. It makes your skin flexible. Hyaluronic acid helps skin stretch and flex and reduces skin wrinkles and lines. Hyaluronic acid is also proven to help wounds heal faster and can reduce scarring."

\textcolor{responsecolor}{Response: }Hyaluronic acid "helps keep things hydrated". It "helps skin stretch and flex and reduces skin wrinkles and lines." "Hyaluronic acid is also proven to help wounds heal faster and can reduce scarring."

\textcolor{instructioncolor}{Instruction: }Respond to the following query using word-for-word quotes from the sources provided below. The following sub-questions will help you answer the query; be sure that your response to the query answers each sub-question using a quotation from a source. Clearly indicate the quotes to avoid plagiarizing! Be concise in your response and focus on information that responds to the query. Do not refer to the sources in your response.The provided sources should contain enough information to write a response to the query. However, in the rare case that the sources are insufficient, then respond with "Insufficient information to generate a grounded response."

\textcolor{querycolor}{Query:} Explain to a third-grader: what is a heirloom tomato?

\textcolor{subquestioncolor}{Sub-questions:} Explain to a third-grader: what is a heirloom tomato?

\textcolor{sourcescolor}{Sources:} 

"They are classified as family heirlooms, commercial heirlooms, mystery heirlooms, or created heirlooms. They usually have a shorter shelf life and are less disease resistant than hybrids. They are grown for various reasons: for food, historical interest, access to wider varieties, and by people who wish to save seeds from year to year, as well as for their taste.[1]Taste[edit]Many heirloom tomatoes are sweeter and lack a genetic mutation that gives tomatoes a uniform red color at the cost of the fruit's taste.[2] Varieties bearing that mutation which have been favored by industry since the 1940s – that is, tomatoes which are not heirlooms – feature fruits with lower levels of carotenoids and a decreased ability to make sugar within the fruit."

"True heirlooms must be 'old' because they have been a variety for anywhere between 25 and 50 years. The seeds, like furniture or jewelry, must also be passed down from generation to generation. There is no exact blueprint for the color and shape of an heirloom tomato, so long as it meets the three characteristics. There are over 100 different types of heirloom tomatoes in the united states and an estimated 3,000 globally, which are suitable for very different climates and growing regions."

\textcolor{responsecolor}{Response: }"Many heirloom tomatoes are sweeter and lack a genetic mutation that gives tomatoes a uniform red color at the cost of the fruit's taste." "True heirlooms must be 'old' because they have been a variety for anywhere between 25 and 50 years."

\textcolor{instructioncolor}{Instruction: }Respond to the following query using word-for-word quotes from the sources provided below. The following sub-questions will help you answer the query; be sure that your response to the query answers each sub-question using a quotation from a source. Clearly indicate the quotes to avoid plagiarizing! Be concise in your response and focus on information that responds to the query. Do not refer to the sources in your response.The provided sources should contain enough information to write a response to the query. However, in the rare case that the sources are insufficient, then respond with "Insufficient information to generate a grounded response."

\textcolor{querycolor}{Query:} Explain to a third-grader: what's the flavor of juicy fruit gum?

\textcolor{subquestioncolor}{Sub-questions:} Explain to a third-grader: what's the flavor of juicy fruit gum?

\textcolor{sourcescolor}{Sources:} 

 "A Juicy Fruit wrapper from 1946 reportedly advertised that the gum had 'a fascinating artificial flavor,' which doesn't really provide much clarity. In 2002, one curious fan took their question to the company itself and received the following statement in response, 'I'm afraid we can't be very specific because, for competitive reasons, we consider our Juicy Fruit flavoring formula to be a trade secret. I can, however, tell you that the mixture of fruit flavors in Juicy Fruit is comprised of predominately lemon, orange, pineapple, and banana notes.'"

\textcolor{responsecolor}{Response: }"the mixture of fruit flavors in Juicy Fruit is comprised of predominately lemon, orange, pineapple, and banana notes."

\textcolor{instructioncolor}{Instruction: }Respond to the following query using word-for-word quotes from the sources provided below. The following sub-questions will help you answer the query; be sure that your response to the query answers each sub-question using a quotation from a source. Clearly indicate the quotes to avoid plagiarizing! Be concise in your response and focus on information that responds to the query. Do not refer to the sources in your response.The provided sources should contain enough information to write a response to the query. However, in the rare case that the sources are insufficient, then respond with "Insufficient information to generate a grounded response."

\textcolor{querycolor}{Query:} Explain to a third-grader: who stole the mona lisa from the louvre in 1911

\textcolor{subquestioncolor}{Sub-questions:} Who stole the Mona Lisa from the Louvre in 1911?

\textcolor{sourcescolor}{Sources:}

"After keeping the painting hidden in a trunk in his apartment for two years, Peruggia returned to Italy with it. He kept it in his apartment in Florence, Italy, for some time. However, Peruggia eventually grew impatient and was finally caught when he contacted Mario Fratelli, the owner of an art gallery in Florence. Fratelli's story conflicts with Peruggia's, but it was clear that Peruggia expected a reward for returning the painting to what he regarded as its 'homeland'. Fratelli called in Giovanni Poggi, director of the Uffizi Gallery, who authenticated the painting. Poggi and Fratelli, after taking the painting for 'safekeeping', informed the police, who arrested Peruggia at his hotel. After its recovery, the painting was exhibited all over Italy with banner headlines rejoicing its return. The Mona Lisa was then returned to the Louvre in 1913. While the painting was famous before the theft, the notoriety it received from the newspaper headlines and the large scale police investigation helped the artwork become one of the best known in the world, gaining considerable public interest. "

"La Gioconda — better known as the Mona Lisa — wasn’t really that famous before she was stolen. Advertisement The man who stole her, Vincenzo Peruggia, had once worked at the Louvre and had, ironically, helped build a protective case for the 16th century masterpiece.On Sunday, August 20, 1911, Peruggia and two accomplices took advantage of lax security and hid overnight in a storage closet at the Louvre.The next morning, while the museum was still closed, they calmly took down Leonardo da Vinci’s priceless work, wrapped it in a cloth and waltzed out a side door.It took more than 24 hours to discover the 'art heist of the century', when a visiting artist, intending to paint a copy of the Mona Lisa, found only empty wall space.Police blitzed the Louvre, questioning people, while the media ran wild with the story. When the Louvre reopened, crowds came to stare at the bare wall hooks. Advertisement Because Peruggia was a former Louvre employee, police interviewed him — not once, but twice — but found nothing to incriminate him.Peruggia would wait over two years before writing to Alfredo Geri, an Italian art dealer, ostensibly out of a desire to repatriate the painting to Italy. Geri was stunned when he saw that smile.Geri called the police, who on December 11, 1913, arrested Peruggia. The man who planned the heist would serve only seven months in jail. "

"Part of the reason the Mona Lisa is known worldwide is because of its theft in 1911 by the Italian handyman Vincenzo Peruggia. That year, Peruggia and two others stowed themselves away in a closet of the Louvre, hung around until the museum closed, and then took the painting, which at the time was considered a minor work by Leonardo, with them, hopping on a train out of Paris. As the theft gained more and more press, both within France and outside it, Peruggia held on to the work, at one point stashing it beneath the floorboards of his Paris apartment. A little over two years after the theft, Peruggia made an attempt to sell the work to a dealer in Florence, in an attempt, he believed, to bring back to Italy a treasure that it had lost. The proposed sale backfired when the dealer called the director of the Uffizi Galleries, who obtained the work and called the police. Peruggia went on to spend six months in prison, and the painting was returned to the Louvre. "

"Leonardo da Vinci’s Mona Lisa, also known as La Gioconda, is the most famous painting in the world. Quantities of effort and ink have been spent over the years on identifying who she was and deciding what her enigmatic smile signifies, what she says about femininity, if anything, and why she has no eyebrows. Leonardo took the painting with him when he was invited to France by Francis I in 1516. The king bought it and at the French Revolution it was placed in the Louvre. Napoleon took it away to hang in his bedroom, but it was returned to the Louvre afterwards. The theft of this fabulous object in 1911 created a media sensation. The police were as baffled as everyone else. It was thought that modernist enemies of traditional art must be involved and the avant-garde poet and playwright Guillaume Apollinaire was arrested in September and questioned for a week before being released. Pablo Picasso was the next prominent suspect, but there was no evidence against him either. It was not until November 1913, calling himself Leonardo Vincenzo, that Perugia wrote to an art dealer in Florence named Alfredo Geri offering to bring the painting to Italy for a reward of 500,000 lire. He travelled to Florence by train the following month, taking the Mona Lisa in a trunk, hidden beneath a false bottom. After booking into a hotel, which subsequently shrewdly changed its name to the Hotel La Gioconda, he took the painting to Geri’s gallery. Geri persuaded him to leave it for expert examination and the police arrested Perugia later that day. "

"And on that morning, with the Louvre still closed, they slipped out of the closet and lifted 200 pounds of painting, frame and protective glass case off the wall. Stripped of its frame and case, the wooden canvas was covered with a blanket and hustled off to the Quai d'Orsay station, where the trio boarded a 7:47 a.m. express train out of the city. They'd stolen the 'Mona Lisa.' Famous, Overnight "

"After the Louvre announced the theft, newspapers all over the world ran headlines about the missing masterpiece. '60 Detectives Seek Stolen 'Mona Lisa,' French Public Indignant,' the New York Times declared. The heist had become something of a national scandal. 'In France, there was a great deal of concern that American millionaires were buying up the legacy of France — the best paintings,' Dorothy Hoobler says. At one point, American tycoon and art lover J.P. Morgan was suspected of commissioning the theft. Pablo Picasso was also considered a suspect, and was questioned. And as tensions were escalating between France and Germany ahead of World War I, 'there were people who thought the Kaiser was behind it,' Hoobler says. After a weeklong shutdown, the Louvre re-opened to mobs of people, Franz Kafka among them, all rushing to see the empty spot that had become a 'mark of shame' for Parisians. Meanwhile, the thieves had made a clean getaway. They were three Italians: two brothers, Vincenzo and Michele Lancelotti, and the ringleader, Vincenzo Perugia. He was a handyman who had worked for the Louvre to install the very same protective glass cases he had ripped from the 'Mona Lisa.' Perugia hoped to sell the painting. But the heist had received so much attention that the 'Mona Lisa' became too hot to hock, Zug says. "

"On the 21st of August 1911, a man named Vincenzo Peruggia committed an infamous art crime that made history. He stole what is now the most famous painting in the world, Leonardo da Vinci’s Mona Lisa, 1503, from the Louvre in Paris. The art theft caused a media sensation, and it took police a full two years before they caught up with the real culprit. At one point, they even arrested the writer Guillaume Apollinaire and his friend Pablo Picasso, believing they had masterminded the art heist. But who was Vincenzo Peruggia, how did he steal this famous masterpiece, and did he get away with it? 1. Vincenzo Peruggia Worked at the Louvre Photographic Portrait of Vincenzo Peruggia, the man who stole the Mona Lisa in 1911, via Vintage Everyday "

"Peruggia and his henchmen shamelessly carried the painting out of the Louvre and back to his Parisian apartment. Initially he had hoped he could sell the artwork to a wealthy collector. But when news of the painting’s theft hit the worldwide media, and police began offering a large reward for its return, he realized this might prove more difficult than expected. So, he kept the painting stored in a false-bottom trunk while he worked out what to do next. 5. Vincenzo Peruggia Tried to Sell the Mona Lisa to a Dealer in Florence Newspaper article covering the theft of the Mona Lisa in 1911, via Open Culture Two years later, Peruggia decided to have a go at selling the painting to an art dealer in Florence called Alfred Geri, under the false name of Leonardo Vincenzo. Understandably, Geri was immediately suspicious, so he asked the head of an Italian art gallery to join him in the meeting with Peruggia. After confirming the painting’s authenticity, the two Italian art experts pretended to go along with Perugia’s sale, while secretly calling the police. From there, the painting made its way back to the Louvre, where it still hangs today. Meanwhile, the police caught up with Peruggia, who claimed he only stole the painting so he could return it to its birthplace in Italy. Amazingly, Peruggia’s shady criminal activities helped make the Mona Lisa famous around the world. Cite this ArticleREAD NEXT:Leonardo da Vinci: Bio, Works, and TriviaShareTweetShareShareShareCopyEmail By "

\textcolor{responsecolor}{Response: }\end{promptbox}
\refstepcounter{promptbox}
\textbf{Box \thepromptbox: } Example of the prompt with few-shot examples used to generate the quoted output for ETa3G queries.
\label{box:eta3g_quoted_output_prompt}
\vspace{.2cm}
\vspace{.2cm}
\begin{promptbox}
\raggedright
\textcolor{instructioncolor}{Instruction: }Respond to the following query using word-for-word quotes from the sources provided below. The following sub-questions will help you answer the query; be sure that your response to the query answers each sub-question using a quotation from a source. Clearly indicate the quotes to avoid plagiarizing! Be concise in your response and focus on information that responds to the query. Do not refer to the sources in your response.The information to answer the query is in the provided sources.

\textcolor{querycolor}{Query:} Are the director of film Move (1970 Film) and director of film Méditerranée (1963 Film) from the same country?

\textcolor{subquestioncolor}{Sub-questions:} Who is the director of the film Move (1970 film)? What country are they from? Who is the director of the film Méditerranée (1963 Film)? What country are they from?

\textcolor{sourcescolor}{Sources:}

"Stuart Rosenberg (August 11, 1927 – March 15, 2007) was an American film and television director whose motion pictures include 'Cool Hand Luke' (1967), 'Voyage of the Damned' (1976), 'The Amityville Horror' (1979), and 'The Pope of Greenwich Village' (1984). He was noted for his work with actor Paul Newman."

"'Méditerranée' is a 1963 French experimental film directed by Jean-Daniel Pollet with assistance from Volker Schlöndorff. It was written by Philippe Sollers and produced by Barbet Schroeder, with music by Antione Duhamel. The 45 minute film is cited as one of Pollet's most influential films, which according to Jonathan Rosenbaum directly influenced Jean-Luc Goddard's 'Contempt', released later the same year. Footage for the film was shot around the Mediterranean, including at a Greek temple, a Sicilian garden, the sea, and also features a fisherman, a bullfighter, and a girl on an operating table."

"Move is a 1970 American comedy film starring Elliott Gould, Paula Prentiss and Geneviève Waïte, and directed by Stuart Rosenberg. The screenplay was written by Joel Lieber and Stanley Hart, adapted from a novel by Lieber."

"Jean-Daniel Pollet (1936–2004) was a French film director and screenwriter who was most active in the 1960s and 1970s. He was associated with two approaches to filmmaking: comedies which blended burlesque and melancholic elements, and poetic films based on texts by writers such as the French poet Francis Ponge."

\textcolor{responsecolor}{Response: }"Move is a 1970 American comedy film starring Elliott Gould, Paula Prentiss and Geneviève Waïte, and directed by Stuart Rosenberg." "Stuart Rosenberg (August 11, 1927 – March 15, 2007) was an American film and television director." However, "'Méditerranée' is a 1963 French experimental film directed by Jean-Daniel Pollet" and "Jean-Daniel Pollet (1936–2004) was a French film director and screenwriter."

\textcolor{instructioncolor}{Instruction: }Respond to the following query using word-for-word quotes from the sources provided below. The following sub-questions will help you answer the query; be sure that your response to the query answers each sub-question using a quotation from a source. Clearly indicate the quotes to avoid plagiarizing! Be concise in your response and focus on information that responds to the query. Do not refer to the sources in your response.The information to answer the query is in the provided sources.

\textcolor{querycolor}{Query:} What nationality is the director of film Wedding Night In Paradise (1950 Film)?

\textcolor{subquestioncolor}{Sub-questions:}Who is the director of Wedding Night In Paradise (1950 Film)? What is their nationality?

\textcolor{sourcescolor}{Sources:}

"Wedding Night in Paradise (1950 film)Wedding Night in Paradise  is a 1950 West German musical comedy film directed by Géza von Bolváry and starring Johannes Heesters, Claude Farell and Gretl Schörg. It is an operetta film, based on the 1942 stage work of the same title. The film's sets were designed by Paul Markwitz and Fritz Maurischat."

"Géza von Bolváry Géza von Bolváry (full name Géza Maria von Bolváry-Zahn; 26 December 1897 – 10 August 1961) was a Hungarian actor, screenwriter and film director, who worked principally in Germany and Austria."

\textcolor{responsecolor}{Response: }"Wedding Night in Paradise  is a 1950 West German musical comedy film directed by Géza von Bolváry." "Géza von Bolváry (full name Géza Maria von Bolváry-Zahn; 26 December 1897 – 10 August 1961) was a Hungarian actor, screenwriter and film director."

\textcolor{instructioncolor}{Instruction: }Respond to the following query using word-for-word quotes from the sources provided below. The following sub-questions will help you answer the query; be sure that your response to the query answers each sub-question using a quotation from a source. Clearly indicate the quotes to avoid plagiarizing! Be concise in your response and focus on information that responds to the query. Do not refer to the sources in your response.The information to answer the query is in the provided sources.

\textcolor{querycolor}{Query:} When did the performer of song A Pretty Girl Milking Her Cow die?

\textcolor{subquestioncolor}{Sub-questions:} Who performed the song A Pretty Girl Milking Her Cow? When did they die?

\textcolor{sourcescolor}{Sources:}

"Judy Garland (born Frances Ethel Gumm; June 10, 1922 – June 22, 1969) was an American actress, singer and dancer. During a career that spanned 45 years, she attained international stardom as an actress in both musical and dramatic roles, as a recording artist, and on the concert stage."

"A Pretty Girl Milking Her Cow (Cailin Deas Crúite na mBó in the Irish language) is a traditional 18th-century Irish ballad. The English version is attributed to Thomas Moore (1779–1852). Originally sung in Irish Gaelic, the song was popular through the early 20th century. It enjoyed a revival when an updated swing version sung by Irish-American singer/actress Judy Garland was featured in the 1940 film Little Nellie Kelly. The updated version is true to the original musical air, and incorporated original lyrics by MGM Musical Director Roger Edens, and featured Garland singing the song to George Murphy using some of the original Gaelic lyrics in the first chorus."

\textcolor{responsecolor}{Response: }"Judy Garland (born Frances Ethel Gumm; June 10, 1922 – June 22, 1969) was an American actress, singer and dancer." A Pretty Girl Milking Her Cow "enjoyed a revival when an updated swing version sung by Irish-American singer/actress Judy Garland was featured in the 1940 film Little Nellie Kelly."

\textcolor{instructioncolor}{Instruction: }Respond to the following query using word-for-word quotes from the sources provided below. The following sub-questions will help you answer the query; be sure that your response to the query answers each sub-question using a quotation from a source. Clearly indicate the quotes to avoid plagiarizing! Be concise in your response and focus on information that responds to the query. Do not refer to the sources in your response.The information to answer the query is in the provided sources.

\textcolor{querycolor}{Query:} Do both films The Falcon (Film) and Valentin The Good have the directors from the same country?

\textcolor{subquestioncolor}{Sub-questions:} Who is the director of the film Falcon? What country are they from? Who is the director of Valentin The Good? What country are they from?

\textcolor{sourcescolor}{Sources:}

"Vatroslav Mimica( born 25 June 1923) is a Croatian film director and screenwriter. Born in Omiš, Mimica had enrolled at the University of Zagreb School of Medicine before the outbreak of World War II. In 1942 he joined Young Communist League of Yugoslavia( SKOJ) and in 1943 he went on to join the Yugoslav Partisans, becoming a member of their medical units. After the war Mimica wrote literary and film reviews, and his career in filmmaking began in 1950 when he became the director of the Jadran Film production studio."

"Martin Frič( 29 March 1902 – 26 August 1968) was a Czech film director, screenwriter and actor. He had more than 100 directing credits between 1929 and 1968, including feature films, shorts and documentary films. Throughout his life, Frič struggled with alcoholism. On the day of the Warsaw Pact invasion of Czechoslovakia in 1968, he attempted suicide, after battling cancer. He died in the hospital five days later."

"Valentin the Good is a 1942 Czech comedy film directed by Martin Frič."

"Banović Strahinja(released internationally as The Falcon) is a 1981 Yugoslavian- German adventure film written and directed by Vatroslav Mimica based on Strahinja Banović, a hero of Serbian epic poetry. It entered the section' Officina Veneziana' at the 38th Venice International Film Festival."

\textcolor{responsecolor}{Response: }"Banović Strahinja(released internationally as The Falcon) is a 1981 Yugoslavian- German adventure film written and directed by Vatroslav Mimica." "Vatroslav Mimica( born 25 June 1923) is a Croatian film director and screenwriter." "Valentin the Good is a 1942 Czech comedy film directed by Martin Frič" and "Martin Frič( 29 March 1902 – 26 August 1968) was a Czech film director, screenwriter and actor."

\textcolor{instructioncolor}{Instruction: }Respond to the following query using word-for-word quotes from the sources provided below. The following sub-questions will help you answer the query; be sure that your response to the query answers each sub-question using a quotation from a source. Clearly indicate the quotes to avoid plagiarizing! Be concise in your response and focus on information that responds to the query. Do not refer to the sources in your response.The information to answer the query is in the provided sources.

\textcolor{querycolor}{Query:} What is the date of birth of Mina Gerhardsen's father?

\textcolor{subquestioncolor}{Sub-questions:} Who is Mina Gerhardsen's father? What date was he born?

\textcolor{sourcescolor}{Sources:}

"Mina Gerhardsen (born 14 September 1975) is a Norwegian politician for the Labour Party. She is the daughter of Rune Gerhardsen and Tove Strand, and granddaughter of Einar Gerhardsen. She is married to Eirik Øwre Thorshaug. She led the Oslo branch of Natur og Ungdom from 1993 to 1995, and was deputy leader of the Workers' Youth League in Oslo in 1997. She took the cand.mag. degree at the University of Oslo in 1998, and also has master's degrees in pedagogy from 2000 and human geography from 2003. From 1999 to 2002 she worked part-time as a journalist in 'Dagsavisen' and 'Dagbladet'. She then worked in the Norwegian Red Cross from 2002 to 2004, except for a period from 2003 to 2004 as a journalist in 'Mandag Morgen'. She was hired as a political advisor in the Norwegian Office of the Prime Minister in 2005, when Stoltenberg's Second Cabinet assumed office. In 2009 she was promoted to State Secretary. In 2011 she changed to the Ministry of Culture."

"Rune Gerhardsen (born 13 June 1946) is a Norwegian politician, representing the Norwegian Labour Party. He is a son of Werna and Einar Gerhardsen, and attended Oslo Cathedral School. He chaired the Workers' Youth League from 1973 to 1975 and chaired the City Government of Oslo from 1992 to 1997. He chaired the Norwegian Skating Association from 1986 to 1990 and 2001 to 2003 and also 2013 to 2017."

\textcolor{responsecolor}{Response: }"Mina Gerhardsen (born 14 September 1975) is a Norwegian politician for the Labour Party. She is the daughter of Rune Gerhardsen and Tove Strand." "Rune Gerhardsen (born 13 June 1946) is a Norwegian politician."

\textcolor{instructioncolor}{Instruction: }Respond to the following query using word-for-word quotes from the sources provided below. The following sub-questions will help you answer the query; be sure that your response to the query answers each sub-question using a quotation from a source. Clearly indicate the quotes to avoid plagiarizing! Be concise in your response and focus on information that responds to the query. Do not refer to the sources in your response.The information to answer the query is in the provided sources.

\textcolor{querycolor}{Query:} When is the composer of film Sruthilayalu 's birthday?

\textcolor{subquestioncolor}{Sub-questions:} Who was the composer for film Sruthilayalu? When is their birthday?

\textcolor{sourcescolor}{Sources:}

"K. V. Mahadevan Krishnankoil Venkadachalam Mahadevan (14 March 1918 – 21 June 2001) was an Indian composer, singer-songwriter, music producer, and musician known for his works in Tamil cinema, Telugu cinema, Kannada cinema, and Malayalam cinema. He is best known for his contributions in works such as 'Manchi Manasulu' (1962), 'Lava Kusa' (1963), 'Thiruvilaiyadal' (1965), 'Saraswathi Sabatham' (1966), 'Kandan Karunai' (1967), 'Thillana Mohanambal' (1968), 'Adimai Penn' (1969), 'Balaraju Katha' (1970), 'Athiparasakthi' (1971), 'Sankarabharanam' (1979), 'Saptapadi' (1981), 'Sirivennela' (1986), 'Sruthilayalu' (1987), 'Pelli Pustakam' (1991), and 'Swathi Kiranam' (1992)."

"Sruthilayalu is a 1987 Indian Telugu-language musical drama film, written and directed by K. Viswanath. The film stars Rajasekhar and Sumalata with soundtrack composed by K. V. Mahadevan. The film garnered Nandi Awards for Best feature film; Best direction, and a Filmfare Award for Best Director – Telugu. The film was premiered at the International Film Festival of India, and AISFM Film Festival. The film was dubbed in Tamil as 'Isaikku Oru Koil'."

\textcolor{responsecolor}{Response: }"Sruthilayalu is a 1987 Indian Telugu-language musical drama film, written and directed by K. Viswanath. The film stars Rajasekhar and Sumalata with soundtrack composed by K. V. Mahadevan." "Krishnankoil Venkadachalam Mahadevan (14 March 1918 – 21 June 2001) was an Indian composer, singer-songwriter, music producer, and musician known for his works in Tamil cinema, Telugu cinema, Kannada cinema, and Malayalam cinema."

\textcolor{instructioncolor}{Instruction: }Respond to the following query using word-for-word quotes from the sources provided below. The following sub-questions will help you answer the query; be sure that your response to the query answers each sub-question using a quotation from a source. Clearly indicate the quotes to avoid plagiarizing! Be concise in your response and focus on information that responds to the query. Do not refer to the sources in your response.The information to answer the query is in the provided sources.

\textcolor{querycolor}{Query:} When did Fatima Bint Mubarak Al Ketbi's husband die?

\textcolor{subquestioncolor}{Sub-questions:} Who is Fatima Bint Mubarak Al Ketbi's husband? When did they die?

\textcolor{sourcescolor}{Sources:}

"Sheikh Zayed bin Sultan Al Nahyan ; 6 May 1918 – 2 November 2004) was the ruler of Abu Dhabi for more than 30 years (6 August 1966 – 2 November 2004). He was the founding father and the principal driving force behind the formation of the United Arab Emirates."

"Fatima bint Mubarak Al Ketbi is the third wife of Sheikh Zayed bin Sultan Al Nahyan, the founder and inaugural president of United Arab Emirates, and late emir (ruler) of Abu Dhabi. She is referred to as the mother of sheikhs and as the Mother of the UAE."

\textcolor{responsecolor}{Response: }"Fatima bint Mubarak Al Ketbi is the third wife of Sheikh Zayed bin Sultan Al Nahyan." "Sheikh Zayed bin Sultan Al Nahyan ; 6 May 1918 – 2 November 2004) was the ruler of Abu Dhabi for more than 30 years (6 August 1966 – 2 November 2004)."

\textcolor{instructioncolor}{Instruction: }Respond to the following query using word-for-word quotes from the sources provided below. The following sub-questions will help you answer the query; be sure that your response to the query answers each sub-question using a quotation from a source. Clearly indicate the quotes to avoid plagiarizing! Be concise in your response and focus on information that responds to the query. Do not refer to the sources in your response.The information to answer the query is in the provided sources.

\textcolor{querycolor}{Query:} Who is the mother-in-law of Andrew Murray (Scottish Soldier)?

\textcolor{subquestioncolor}{Sub-questions:} Who is Andrew Murray's spouse? Who is their mother?

\textcolor{sourcescolor}{Sources:}

"Christina Bruce (c. 1278 – 1356/1357), also known as Christina de Brus, was a daughter of Marjorie, Countess of Carrick, and her husband, Robert de Brus, 'jure uxoris' Earl of Carrick, as well as a sister of Robert the Bruce, King of Scots. It is presumed that she and her siblings were born at Turnberry Castle in Carrick."

"Sir Andrew Murray (1298–1338), also known as Sir Andrew Moray, or Sir Andrew de Moray, was a Scottish military and political leader who supported David II of Scotland against Edward Balliol and King Edward III of England during the so-called Second War of Scottish Independence. He held the lordships of Avoch and Petty in north Scotland, and Bothwell in west-central Scotland. In 1326 he married Christina Bruce, a sister of King Robert I of Scotland."

\textcolor{responsecolor}{Response: }Sir Andrew Murray "married Christina Bruce, a sister of King Robert I of Scotland." "Christina Bruce (c. 1278 – 1356/1357), also known as Christina de Brus, was a daughter of Marjorie, Countess of Carrick, and her husband, Robert de Brus, 'jure uxoris' Earl of Carrick, as well as a sister of Robert the Bruce, King of Scots."

\textcolor{instructioncolor}{Instruction: }Respond to the following query using word-for-word quotes from the sources provided below. The following sub-questions will help you answer the query; be sure that your response to the query answers each sub-question using a quotation from a source. Clearly indicate the quotes to avoid plagiarizing! Be concise in your response and focus on information that responds to the query. Do not refer to the sources in your response.The information to answer the query is in the provided sources.

\textcolor{querycolor}{Query:} When did the performer of song A Pretty Girl Milking Her Cow die?

\textcolor{subquestioncolor}{Sub-questions:} Who performed the song A Pretty Girl Milking Her Cow? When did they die?

\textcolor{sourcescolor}{Sources:}

"Judy Garland (born Frances Ethel Gumm; June 10, 1922 – June 22, 1969) was an American actress, singer and dancer. During a career that spanned 45 years, she attained international stardom as an actress in both musical and dramatic roles, as a recording artist, and on the concert stage."

"A Pretty Girl Milking Her Cow (Cailin Deas Crúite na mBó in the Irish language) is a traditional 18th-century Irish ballad. The English version is attributed to Thomas Moore (1779–1852). Originally sung in Irish Gaelic, the song was popular through the early 20th century. It enjoyed a revival when an updated swing version sung by Irish-American singer/actress Judy Garland was featured in the 1940 film Little Nellie Kelly. The updated version is true to the original musical air, and incorporated original lyrics by MGM Musical Director Roger Edens, and featured Garland singing the song to George Murphy using some of the original Gaelic lyrics in the first chorus."

\textcolor{responsecolor}{Response: }"Judy Garland (born Frances Ethel Gumm; June 10, 1922 – June 22, 1969) was an American actress, singer and dancer." A Pretty Girl Milking Her Cow "enjoyed a revival when an updated swing version sung by Irish-American singer/actress Judy Garland was featured in the 1940 film Little Nellie Kelly."

\textcolor{instructioncolor}{Instruction: }Respond to the following query using word-for-word quotes from the sources provided below. The following sub-questions will help you answer the query; be sure that your response to the query answers each sub-question using a quotation from a source. Clearly indicate the quotes to avoid plagiarizing! Be concise in your response and focus on information that responds to the query. Do not refer to the sources in your response.The information to answer the query is in the provided sources.

\textcolor{querycolor}{Query:} What is the date of birth of the director of film Runway (2010 Film)?

\textcolor{subquestioncolor}{Sub-questions:} Who is the director of the film Runway (2010 Film)? What is their date of birth?

\textcolor{sourcescolor}{Sources:}

"Runway is a 2010 Bangladeshi drama film directed by Tareque Masud with his wife Catherine Masud. It was his last feature film. "

"Tareque Masud( 6 December 1956 – 13 August 2011) was a Bangladeshi independent film director, film producer, screenwriter and lyricist. He first found success with the films' Muktir Gaan'( 1995) and' Matir Moina'( 2002), for which he won three international awards, including the International Critics' FIPRESCI Prize, in the Directors' Fortnight section outside competition at the 2002 Cannes Film Festival. The film became Bangladesh's first film to compete for the Academy Award for Best Foreign Language Film. Masud died in a road accident on 13 August 2011 while returning to Dhaka from Manikganj on the Dhaka- Aricha highway after visiting a filming location. Masud was working on' Kagojer Phool'( The Paper Flower). In 2012, he posthumously received Ekushey Padak, the highest civilian award of Bangladesh. In 2013, New York University Asian/ Pacific/ American Institute, and South Asia Solidarity Initiative, hosted the first North American retrospective of his films. "

\textcolor{responsecolor}{Response: }\end{promptbox}
\refstepcounter{promptbox}
\textbf{Box \thepromptbox: } Example of the prompt with few-shot examples used to generate the quoted output for 2WikiMH queries.
\label{box:2WikiMH_quoted_output_prompt}
\vspace{.2cm}

\vspace{.2cm}
\begin{promptbox}
\raggedright
\textcolor{instructioncolor}{Instruction: }Respond to the following query using word-for-word quotes from the sources provided below. The following sub-questions will help you answer the query; be sure that your response to the query answers each sub-question using a quotation from a source. Clearly indicate the quotes to avoid plagiarizing! Be concise in your response and focus on information that responds to the query. Do not refer to the sources in your response.The information to answer the query is in the provided sources.

\textcolor{querycolor}{Query:} What can you do to prevent migraine with brainstem aura?

\textcolor{subquestioncolor}{Sub-questions:} What can you do to prevent migraine with brainstem aura?

\textcolor{sourcescolor}{Sources:} 

"Symptoms of MBA can also seem like the signs of other more serious conditions, like seizure disorders, stroke, meningitis, or brain tumors. To rule those out, you'll need to see a brain doctor, called a neurologist. He'll give you a thorough exam and ask you questions about your symptoms. He'll may also use tests like MRI, CT scans, and nerve tests to see what's causing your symptoms. Treatments for MBA generally aim to relieve symptoms of pain and nausea. You might take pain relievers such as acetaminophen, ibuprofen, and naproxen, as well as nausea medicines such as chlorpromazine, metoclopramide, and prochlorperazine. Your doctor may prescribe a medication that treats regular migraines, such as triptans. To keep from getting a MBA,  it helps to avoid the things that usually cause one."

"Keep a journal of your attacks so you can figure out the things that trigger them. It also helps to live a healthy lifestyle. That means you need to:  Get enough sleep.  Limit your stress.  Exercise daily. Diet can also affect migraines. Do these things:  Eat a balanced diet.  Avoid drugs and alcohol.  Don't skip meals.  Limit caffeine.  Avoid any foods that have been triggers. Some common food triggers include: Dairy Wheat Chocolate Eggs Rye Tomatoes Oranges"

"Along with these changes,  If you don't respond to other treatments and you have 4 or more migraine days a month, your doctor may suggest preventive medicines. You can take these regularly to reduce the severity or frequency of the headaches. These include seizure medicines, blood pressure medicines (like beta blockers and calcium channel blockers), and some antidepressants. CGRP inhibitors are a new class of preventive medicine that your doctor may recommend if other medicines don't help."

\textcolor{responsecolor}{Response: }To prevent migraine with brainstem aura, "it helps to avoid the things that usually cause one." This includes being sure to "Get enough sleep," "Limit your stress," and "Exercise daily." Additionally, "Eat a balanced diet," "Avoid drugs and alcohol," "Don't skip meals,"  and "Limit caffeine." "Avoid any foods that have been triggers." Furthermore, "If you don't respond to other treatments and you have 4 or more migraine days a month, your doctor may suggest preventive medicines," including "seizure medicines, blood pressure medicines (like beta blockers and calcium channel blockers), and some antidepressants." Also, "CGRP inhibitors are a new class of preventive medicine."

\textcolor{instructioncolor}{Instruction: }Respond to the following query using word-for-word quotes from the sources provided below. The following sub-questions will help you answer the query; be sure that your response to the query answers each sub-question using a quotation from a source. Clearly indicate the quotes to avoid plagiarizing! Be concise in your response and focus on information that responds to the query. Do not refer to the sources in your response.The information to answer the query is in the provided sources.

\textcolor{querycolor}{Query:} What should I do if my child has a biting injury?

\textcolor{subquestioncolor}{Sub-questions:} What should I do if my child has a biting injury?

\textcolor{sourcescolor}{Sources:} 

"If your child is prone to biting, keep an eye on any playmates and step in when an altercation appears to be brewing. Even with your best prevention efforts, biting incidents might still occur. When your child bites, firmly let your child know that this behavior is not acceptable by saying, 'No. We don't bite!' Explain that biting hurts the other person. Then remove your child from the situation and give the child time to calm down. You might have heard from other parents that if your child bites you, bite your child back. This isn't good advice. Children learn by imitation. If you bite your child, the child is going to get the impression that this behavior is acceptable and he or she will be more likely to do it again. The same goes for hitting a child for biting. If you are unable to get your child to stop biting, the behavior could begin to have an impact on school and relationships. You or another adult might have to closely supervise interactions between your child and other kids. When biting becomes a habit or continues past age 4 or 5, it might stem from a more serious emotional problem. Talk to your child's health care provider, or enlist the help of a child psychologist or therapist."

"[1] The first thing to do for any biting injury is to wash the area with soap and water. [2] Even little teeth can break the skin. [3] If the bite is bleeding and the wound appears to be deep, call your child's doctor. [4] The bite may need medical treatment, which could include antibiotics or a tetanus shot, or both."

\textcolor{responsecolor}{Response: }"The first thing to do for any biting injury is to wash the area with soap and water." "Even little teeth can break the skin." "If the bite is bleeding and the wound appears to be deep, call your child's doctor." "The bite may need medical treatment, which could include antibiotics or a tetanus shot, or both."

\textcolor{instructioncolor}{Instruction: }Respond to the following query using word-for-word quotes from the sources provided below. The following sub-questions will help you answer the query; be sure that your response to the query answers each sub-question using a quotation from a source. Clearly indicate the quotes to avoid plagiarizing! Be concise in your response and focus on information that responds to the query. Do not refer to the sources in your response.The information to answer the query is in the provided sources.

\textcolor{querycolor}{Query:} Can I give cough and cold medicines to children under 4?

\textcolor{subquestioncolor}{Sub-questions:} Can I give cough and cold medicines to children under 4?

\textcolor{sourcescolor}{Sources:} 

"There are over 200 cold viruses ready to lay your family low with nasal congestion, sore throat, and cough. Combat cold symptoms with these home remedies. Get rest. Rest helps the body focus its energy on getting well, so keep kids home from school to keep them warm, and well-rested."

"Drink lots of fluids. Replenish liquids lost from fever, vomiting and diarrhea. Fluids also help loosen mucus. Use a humidifier. A humidifier in your child's room can keep the air moist and break up nasal and chest congestion. Talk to your pediatrician before giving OTC cold and cough medicines. These medicines should not be given to children under 4 years of age, according to the FDA and the drug makers. Also, evidence indicates medicines such as antihistamines, decongestants, and cough medicines don't really help, but they could pose a small risk of serious side effects."

\textcolor{responsecolor}{Response: }"Talk to your pediatrician before giving OTC cold and cough medicines." "These medicines should not be given to children under 4 years of age, according to the FDA and the drug makers."

\textcolor{instructioncolor}{Instruction: }Respond to the following query using word-for-word quotes from the sources provided below. The following sub-questions will help you answer the query; be sure that your response to the query answers each sub-question using a quotation from a source. Clearly indicate the quotes to avoid plagiarizing! Be concise in your response and focus on information that responds to the query. Do not refer to the sources in your response.The information to answer the query is in the provided sources.

\textcolor{querycolor}{Query:} How do I treat a blister?

\textcolor{subquestioncolor}{Sub-questions:} How do I treat a blister?

\textcolor{sourcescolor}{Sources:} 

"Most blisters caused by friction or minor burns do not require a doctor's care.  New skin will form underneath the affected area and the fluid is simply absorbed."

"Do not puncture a blister unless it is large, painful, or likely to be further irritated.  The fluid-filled blister keeps the underlying skin clean, which prevents infection and promotes healing.  But if you need to pop a blister or it pops by itself: Use a sterilized needle (to sterilize it, put the point or edge in a flame until it is red hot, or rinse it in alcohol).  Wash your hands and the area thoroughly, then make a small hole; the fluid will drain on its own.  If the fluid is white or yellow, the blister may be infected and needs medical attention.  Do not remove the skin over a broken blister.  The new skin underneath needs this protective cover.  Apply an antibiotic ointment or cream.  Look for signs of infection to develop, including pus drainage, red or warm skin surrounding the blister, or red streaks leading away from the blister."

"Wear work gloves. Jobs you do only occasionally, such as shoveling snow or raking leaves, are great for raising a blister or two. Wear shoes that fit and are comfortable. New shoes should be broken in gradually. Wear padded socks or put some adhesive padding where your foot is rubbing."

\textcolor{responsecolor}{Response: }"Most blisters caused by friction or minor burns do not require a doctor's care." "New skin will form underneath the affected area and the fluid is simply absorbed." "Do not puncture a blister unless it is large, painful, or likely to be further irritated." "The fluid-filled blister keeps the underlying skin clean, which prevents infection and promotes healing" "But if you need to pop a blister or it pops by itself: Use a sterilized needle (to sterilize it, put the point or edge in a flame until it is red hot, or rinse it in alcohol)." "Wash your hands and the area thoroughly, then make a small hole; the fluid will drain on its own." "If the fluid is white or yellow, the blister may be infected and needs medical attention." "Do not remove the skin over a broken blister." "The new skin underneath needs this protective cover." "Apply an antibiotic ointment or cream." "Look for signs of infection to develop, including pus drainage, red or warm skin surrounding the blister, or red streaks leading away from the blister."

\textcolor{instructioncolor}{Instruction: }Respond to the following query using word-for-word quotes from the sources provided below. The following sub-questions will help you answer the query; be sure that your response to the query answers each sub-question using a quotation from a source. Clearly indicate the quotes to avoid plagiarizing! Be concise in your response and focus on information that responds to the query. Do not refer to the sources in your response.The information to answer the query is in the provided sources.

\textcolor{querycolor}{Query:} What medications can I use to lessen my child's teething pain?

\textcolor{subquestioncolor}{Sub-questions:} What medications can I use to lessen my child's teething pain?

\textcolor{sourcescolor}{Sources:}

"Teething pain can be hard for babies and parents alike. Treatment at home can help. Your child has a fever of 101 ° F or higher, has diarrhea, is dehydrated, or seems especially cranky; these are not typical symptoms of teething. Use a firm rubber teether. Avoid teethers filled with liquid, which can leak, or freezable teethers, which may be too cold or hard. Never tie a teether around a baby's neck -- it's a choking hazard. Gently massage the gums with a cool wet washcloth or a clean finger. If your child is eating solid foods, offer cold applesauce or yogurt. Use baby acetaminophen ( Tylenol) to dull pain. Do not use ibuprofen unless the baby is at least 6 months old. Do not use topical pain medicine that contains benzocaine. They could cause dangerous side effects. Benzocaine can be found in over-the-counter products such as Baby Orajel. Keep a cloth with you to wipe away drool from baby's chin."

\textcolor{responsecolor}{Response: }\end{promptbox}
\refstepcounter{promptbox}
\textbf{Box \thepromptbox: } Example of the prompt with few-shot examples used to generate the quoted output for MASH queries.
\label{box:mash_quoted_output_prompt}
\vspace{.2cm}

\subsubsection{Prompts for the Paraphrased Generation}
\label{Prompts for the Paraphrased Output}
\vspace{.2cm}
\begin{promptbox}
\raggedright
\textcolor{instructioncolor}{Instruction: }Respond to the following query by building off of the response below. Specifically, rephrase each sentence in the response using a more fluent and useful wording to convey the same information. In other words, paraphrase each sentence of the response as an improved new sentence, with respect to the query. Do not refer to the response in your revised response.

\textcolor{querycolor}{Query:} What's the life cycle of a frog?

\textcolor{sourcescolor}{Response:} The life cycle of a frog consists of three stages: egg, larva, and adult. As the frog grows, it moves through these stages in a process known as metamorphosis. During the egg stage, the yolk in each egg splits into more and more cells and begins to take the form of a tadpole, the larva of a frog. Within one to three weeks, the egg is ready to hatch, and a tiny tadpole breaks free. In the larval stage, [t]adpoles, frogs' larvae, have rudimentary gills, a mouth, and a long tail. As they grow, tadpoles feed on algae and other vegetation, and eventually, they grow back legs, then front legs too! Soon after, their body starts to change shape, and they’re able to start eating insects. Next, the tadpoles’ tails shrink away, and skin grows over their gills, as they develop lungs and eardrums!

\textcolor{responsecolor}{Paraphrased Response: }The stages of a frog’s life cycle are egg, larva, and adult. Growing through these stages is a process called metamorphosis. In a frog egg, the yolk splits into many cells which are the beginnings of a tadpole, or a frog larva. In one to three weeks, a small tadpole hatches from the egg. During the larval stage, tadpoles, or frogs’ larvae, are equipped with simple gills, a mouth, and a lengthy tail. As they develop, tadpoles consume plants and algae and then grow back legs, followed by front legs. Shortly afterward, their body begins to shift in shape and they gain the ability to consume insects. Then, the tails of the tadpoles grow smaller and skin develops over their gills while they grow eardrums and lungs.

\textcolor{instructioncolor}{Instruction: }Respond to the following query by building off of the response below. Specifically, rephrase each sentence in the response using a more fluent and useful wording to convey the same information. In other words, paraphrase each sentence of the response as an improved new sentence, with respect to the query. Do not refer to the response in your revised response.

\textcolor{querycolor}{Query:} What are the benefits of hyaluronic acid for skin care?

\textcolor{sourcescolor}{Response:} Hyaluronic acid helps keep things hydrated. It helps skin stretch and flex and reduces skin wrinkles and lines. Hyaluronic acid is also proven to help wounds heal faster and can reduce scarring.

\textcolor{responsecolor}{Paraphrased Response: }Hyaluronic acid helps to retain moisture. It makes it easier for skin to be flexible and diminishes skin lines and wrinkles. Hyaluronic acid has been shown to decrease the recovery time of wounds and can lessen scarring.

\textcolor{instructioncolor}{Instruction: }Respond to the following query by building off of the response below. Specifically, rephrase each sentence in the response using a more fluent and useful wording to convey the same information. In other words, paraphrase each sentence of the response as an improved new sentence, with respect to the query. Do not refer to the response in your revised response.

\textcolor{querycolor}{Query:} What is a heirloom tomato?

\textcolor{sourcescolor}{Response:} Many heirloom tomatoes are sweeter and lack a genetic mutation that gives tomatoes a uniform red color at the cost of the fruit's taste. True heirlooms must be 'old' because they have been a variety for anywhere between 25 and 50 years.

\textcolor{responsecolor}{Paraphrased Response: }Unlike tomatoes that carry a genetic mutation that makes them look consistently red at the expense of their flavor, many heirloom tomatoes taste sweeter. Genuine heirloom tomatoes have to be old due to being a variety over the last 25 to 50 years.

\textcolor{instructioncolor}{Instruction: }Respond to the following query by building off of the response below. Specifically, rephrase each sentence in the response using a more fluent and useful wording to convey the same information. In other words, paraphrase each sentence of the response as an improved new sentence, with respect to the query. Do not refer to the response in your revised response.

\textcolor{querycolor}{Query:} What's the flavor of juicy fruit gum?

\textcolor{sourcescolor}{Response:} the mixture of fruit flavors in Juicy Fruit is comprised of predominately lemon, orange, pineapple, and banana notes. 

\textcolor{responsecolor}{Paraphrased Response: }Juicy Fruit flavor is a blend of mostly orange, lemon, banana, and pineapple flavors.

\textcolor{instructioncolor}{Instruction: }Respond to the following query by building off of the response below. Specifically, rephrase each sentence in the response using a more fluent and useful wording to convey the same information. In other words, paraphrase each sentence of the response as an improved new sentence, with respect to the query. Do not refer to the response in your revised response.

\textcolor{querycolor}{Query:} Explain to a third-grader: who stole the mona lisa from the louvre in 1911

\textcolor{sourcescolor}{Response:} The man who stole her, Vincenzo Peruggia, had once worked at the Louvre and had, ironically, helped build a protective case for the 16th century masterpiece.

\textcolor{responsecolor}{Paraphrased Response: }\end{promptbox}
\refstepcounter{promptbox}
\textbf{Box \thepromptbox: } Example of the prompt with few-shot examples used to generate the paraphrased output for NQ and ETa3G queries.
\label{box:nq_eta3g_pp_output_prompt}
\vspace{.2cm}
\vspace{.2cm}
\begin{promptbox}
\raggedright
\textcolor{instructioncolor}{Instruction: }Respond to the following query by building off of the response below. Specifically, rephrase each sentence in the response using a more fluent and useful wording to convey the same information. In other words, paraphrase each sentence of the response as an improved new sentence, with respect to the query. Do not refer to the response in your revised response.

\textcolor{querycolor}{Query:} Are the director of film Move (1970 Film) and director of film Méditerranée (1963 Film) from the same country?

\textcolor{sourcescolor}{Response:} Move (1970 Film) was directed by Stuart Rosenberg, who was born on August 11, 1927, and passing on March 15, 2007, and was renowned for his work as a director in both American film and television. On the other hand, Méditerranée was directed by Jean-Daniel Pollet who lived from 1936 to 2004 and was celebrated as a French filmmaker and scriptwriter.

\textcolor{responsecolor}{Paraphrased Response: }The film 'Move' (1970) was directed by Stuart Rosenberg, born on August 11, 1927, and who died on March 15, 2007. He was known for his distinguished career in American cinema and television. Conversely, Jean-Daniel Pollet (1936-2004), a French filmmaker and screenwriter recognized for his notable work, directed 'Méditerranée'.

\textcolor{instructioncolor}{Instruction: }Respond to the following query by building off of the response below. Specifically, rephrase each sentence in the response using a more fluent and useful wording to convey the same information. In other words, paraphrase each sentence of the response as an improved new sentence, with respect to the query. Do not refer to the response in your revised response.

\textcolor{querycolor}{Query:} What nationality is the director of film Wedding Night In Paradise (1950 Film)?

\textcolor{sourcescolor}{Response:} Wedding Night in Paradise  is a 1950 West German musical comedy film directed by Géza von Bolváry. Géza von Bolváry (full name Géza Maria von Bolváry-Zahn; 26 December 1897 – 10 August 1961) was a Hungarian actor, screenwriter and film director.

\textcolor{responsecolor}{Paraphrased Response: }Géza von Bolváry directed the 1950 West German musical comedy 'Wedding Night in Paradise.' Born on December 26, 1897, Géza Maria von Bolváry-Zahn was a Hungarian talent known for his work as an actor, screenwriter, and director, and he passed away on August 10, 1961.

\textcolor{instructioncolor}{Instruction: }Respond to the following query by building off of the response below. Specifically, rephrase each sentence in the response using a more fluent and useful wording to convey the same information. In other words, paraphrase each sentence of the response as an improved new sentence, with respect to the query. Do not refer to the response in your revised response.

\textcolor{querycolor}{Query:} When did the performer of song A Pretty Girl Milking Her Cow die?

\textcolor{sourcescolor}{Response:} Judy Garland (born Frances Ethel Gumm; June 10, 1922 – June 22, 1969) was an American actress, singer and dancer. A Pretty Girl Milking Her Cow enjoyed a revival when an updated swing version sung by Irish-American singer/actress Judy Garland was featured in the 1940 film Little Nellie Kelly.

\textcolor{responsecolor}{Paraphrased Response: }American actress, singer, and dancer Judy Garland, originally named Frances Ethel Gumm, lived from June 10, 1922, to June 22, 1969. Notably, her rendition of 'A Pretty Girl Milking Her Cow' experienced a resurgence in popularity when her updated swing version was showcased in the 1940 film 'Little Nellie Kelly'.

\textcolor{instructioncolor}{Instruction: }Respond to the following query by building off of the response below. Specifically, rephrase each sentence in the response using a more fluent and useful wording to convey the same information. In other words, paraphrase each sentence of the response as an improved new sentence, with respect to the query. Do not refer to the response in your revised response.

\textcolor{querycolor}{Query:} Do both films The Falcon (Film) and Valentin The Good have the directors from the same country?

\textcolor{sourcescolor}{Response:} Banović Strahinja( released internationally as The Falcon) is a 1981 Yugoslavian- German adventure film written and directed by Vatroslav Mimica. Vatroslav Mimica( born 25 June 1923) is a Croatian film director and screenwriter. Valentin the Good is a 1942 Czech comedy film directed by Martin Frič and Martin Frič( 29 March 1902 – 26 August 1968) was a Czech film director, screenwriter and actor.

\textcolor{responsecolor}{Paraphrased Response: }'Banović Strahinja' (internationally known as 'The Falcon') is an adventure film from 1981, produced as a Yugoslavian-German collaboration and crafted by writer and director Vatroslav Mimica. Vatroslav Mimica, born on June 25, 1923, is a screenwriter and director from Croatia. Directed by Martin Frič, 'Valentin the Good' is a Czech comedy film that premiered in 1942. Martin Frič, born on March 29, 1902, and passed away on August 26, 1968, was an actor, screenwriter, and director from Czechia.

\textcolor{instructioncolor}{Instruction: }Respond to the following query by building off of the response below. Specifically, rephrase each sentence in the response using a more fluent and useful wording to convey the same information. In other words, paraphrase each sentence of the response as an improved new sentence, with respect to the query. Do not refer to the response in your revised response.

\textcolor{querycolor}{Query:} What is the date of birth of Mina Gerhardsen's father?

\textcolor{sourcescolor}{Response:} Mina Gerhardsen (born 14 September 1975) is a Norwegian politician for the Labour Party. She is the daughter of Rune Gerhardsen and Tove Strand. Rune Gerhardsen (born 13 June 1946) is a Norwegian politician.

\textcolor{responsecolor}{Paraphrased Response: }Mina Gerhardsen, born on September 14, 1975, is a politician from Norway affiliated with the Labour Party. She is the daughter of Rune Gerhardsen and Tove Strand. Rune Gerhardsen, born on June 13, 1946, is also a Norwegian politician.

\textcolor{instructioncolor}{Instruction: }Respond to the following query by building off of the response below. Specifically, rephrase each sentence in the response using a more fluent and useful wording to convey the same information. In other words, paraphrase each sentence of the response as an improved new sentence, with respect to the query. Do not refer to the response in your revised response.

\textcolor{querycolor}{Query:} When is the composer of film Sruthilayalu 's birthday?

\textcolor{sourcescolor}{Response:} Sruthilayalu is a 1987 Indian Telugu-language musical drama film, written and directed by K. Viswanath. The film stars Rajasekhar and Sumalata with soundtrack composed by K. V. Mahadevan. Krishnankoil Venkadachalam Mahadevan (14 March 1918 – 21 June 2001) was an Indian composer, singer-songwriter, music producer, and musician known for his works in Tamil cinema, Telugu cinema, Kannada cinema, and Malayalam cinema.

\textcolor{responsecolor}{Paraphrased Response: }'Sruthilayalu' is a 1987 musical drama film in the Telugu language from India, crafted by writer and director K. Viswanath. The movie features Rajasekhar and Sumalata in leading roles, with a soundtrack created by K. V. Mahadevan. Krishnankoil Venkadachalam Mahadevan (March 14, 1918 – June 21, 2001) was an Indian singer-songwriter, musician, composer, and music producer renowned for his contributions to Telugu, Tamil, Malayalam, and Kannada cinema.

\textcolor{instructioncolor}{Instruction: }Respond to the following query by building off of the response below. Specifically, rephrase each sentence in the response using a more fluent and useful wording to convey the same information. In other words, paraphrase each sentence of the response as an improved new sentence, with respect to the query. Do not refer to the response in your revised response.

\textcolor{querycolor}{Query:} When did Fatima Bint Mubarak Al Ketbi's husband die?

\textcolor{sourcescolor}{Response:} Fatima bint Mubarak Al Ketbi is the third wife of Sheikh Zayed bin Sultan Al Nahyan. Sheikh Zayed bin Sultan Al Nahyan ; 6 May 1918 – 2 November 2004) was the ruler of Abu Dhabi for more than 30 years (6 August 1966 – 2 November 2004).

\textcolor{responsecolor}{Paraphrased Response: }Fatima bint Mubarak Al Ketbi is the wife of Sheikh Zayed bin Sultan Al Nahyan from his third marriage. Sheikh Zayed bin Sultan Al Nahyan, born on May 6, 1918, and passing on November 2, 2004, ruled Abu Dhabi for over 30 years, from August 6, 1966, to November 2, 2004.

\textcolor{instructioncolor}{Instruction: }Respond to the following query by building off of the response below. Specifically, rephrase each sentence in the response using a more fluent and useful wording to convey the same information. In other words, paraphrase each sentence of the response as an improved new sentence, with respect to the query. Do not refer to the response in your revised response.

\textcolor{querycolor}{Query:} Who is the mother-in-law of Andrew Murray (Scottish Soldier)?

\textcolor{sourcescolor}{Response:} Sir Andrew Murray married Christina Bruce, a sister of King Robert I of Scotland. Christina Bruce (c. 1278 – 1356/1357), also known as Christina de Brus, was a daughter of Marjorie, Countess of Carrick, and her husband, Robert de Brus, 'jure uxoris' Earl of Carrick, as well as a sister of Robert the Bruce, King of Scots.

\textcolor{responsecolor}{Paraphrased Response: }Sir Andrew Murray took Christina Bruce, sister of King Robert I of Scotland, as his wife. Born around 1278 and living until 1356/1357, Christina Bruce, also called Christina de Brus, was the daughter of Marjorie, Countess of Carrick, and Robert de Brus, Earl of Carrick by virtue of his marriage. She was also the sister of Robert the Bruce, the King of Scots.

\textcolor{instructioncolor}{Instruction: }Respond to the following query by building off of the response below. Specifically, rephrase each sentence in the response using a more fluent and useful wording to convey the same information. In other words, paraphrase each sentence of the response as an improved new sentence, with respect to the query. Do not refer to the response in your revised response.

\textcolor{querycolor}{Query:} When did the performer of song A Pretty Girl Milking Her Cow die?

\textcolor{sourcescolor}{Response:} Judy Garland (born Frances Ethel Gumm; June 10, 1922 – June 22, 1969) was an American actress, singer and dancer. A Pretty Girl Milking Her Cow enjoyed a revival when an updated swing version sung by Irish-American singer/actress Judy Garland was featured in the 1940 film Little Nellie Kelly.

\textcolor{responsecolor}{Paraphrased Response: }Frances Ethel Gumm, known professionally as Judy Garland (June 10, 1922 – June 22, 1969), was a multifaceted American performer, excelling in acting, singing, and dancing. The song 'A Pretty Girl Milking Her Cow' experienced a resurgence with a modern swing rendition performed by Irish-American singer and actress Judy Garland, featured in the 1940 movie 'Little Nellie Kelly'.

\textcolor{instructioncolor}{Instruction: }Respond to the following query by building off of the response below. Specifically, rephrase each sentence in the response using a more fluent and useful wording to convey the same information. In other words, paraphrase each sentence of the response as an improved new sentence, with respect to the query. Do not refer to the response in your revised response.

\textcolor{querycolor}{Query:} What is the date of birth of the director of film Runway (2010 Film)?

\textcolor{sourcescolor}{Response:} Runway is a 2010 Bangladeshi drama film directed by Tareque Masud. Tareque Masud( 6 December 1956 – 13 August 2011) was a Bangladeshi independent film director, film producer, screenwriter and lyricist.

\textcolor{responsecolor}{Paraphrased Response: }\end{promptbox}
\refstepcounter{promptbox}
\textbf{Box \thepromptbox: } Example of the prompt with few-shot examples used to generate the paraphrased output for 2WikiMH queries.
\label{box:2WikiMH_pp_output_prompt}
\vspace{.2cm}

\vspace{.2cm}
\begin{promptbox}
\raggedright
\textcolor{instructioncolor}{Instruction: }Respond to the following query by building off of the response below. Specifically, rephrase each sentence in the response using a more fluent and useful wording to convey the same information. In other words, paraphrase each sentence of the response as an improved new sentence, with respect to the query. Do not refer to the response in your revised response.

\textcolor{querycolor}{Query:} What can you do to prevent migraine with brainstem aura?

\textcolor{sourcescolor}{Response:} To prevent migraine with brainstem aura, it helps to avoid the things that usually cause one. This includes being sure to Get enough sleep, Limit your stress, and Exercise daily. Additionally, Eat a balanced diet, Avoid drugs and alcohol, Don't skip meals,  and Limit caffeine. Avoid any foods that have been triggers. Furthermore, If you don't respond to other treatments and you have 4 or more migraine days a month, your doctor may suggest preventive medicines, including seizure medicines, blood pressure medicines (like beta blockers and calcium channel blockers), and some antidepressants. Also, CGRP inhibitors are a new class of preventive medicine.

\textcolor{responsecolor}{Paraphrased Response: }To avert migraines with brainstem aura, steering clear of common triggers is beneficial. This encompasses getting adequate sleep, minimizing stress, and engaging in daily physical activity. Moreover, maintaining a nutritious diet, abstaining from drugs and alcohol, not missing meals, reducing caffeine intake, and steering clear of any known food triggers are good precautions. Additionally, if you experience four or more days of migraines a month and haven't found relief with other treatments, your healthcare provider might recommend preventive medications. Such medications may include seizure medicine, blood pressure medicines (including beta blockers and calcium channel blockers), and certain antidepressants. Additionally, there is a new class of preventative medication called CGRP inhibitors.

\textcolor{instructioncolor}{Instruction: }Respond to the following query by building off of the response below. Specifically, rephrase each sentence in the response using a more fluent and useful wording to convey the same information. In other words, paraphrase each sentence of the response as an improved new sentence, with respect to the query. Do not refer to the response in your revised response.

\textcolor{querycolor}{Query:} What should I do if my child has a biting injury?

\textcolor{sourcescolor}{Response:} The first thing to do for any biting injury is to wash the area with soap and water. Even little teeth can break the skin. If the bite is bleeding and the wound appears to be deep, call your child's doctor. The bite may need medical treatment, which could include antibiotics or a tetanus shot, or both.

\textcolor{responsecolor}{Paraphrased Response: }Initially, cleanse the bitten area thoroughly using soap and water, as even small teeth can penetrate the skin. Should the bite result in bleeding and a deep injury, consult your child's physician. Such injuries might necessitate medical intervention, potentially involving antibiotics, a tetanus injection, or both.

\textcolor{instructioncolor}{Instruction: }Respond to the following query by building off of the response below. Specifically, rephrase each sentence in the response using a more fluent and useful wording to convey the same information. In other words, paraphrase each sentence of the response as an improved new sentence, with respect to the query. Do not refer to the response in your revised response.

\textcolor{querycolor}{Query:} Can I give cough and cold medicines to children under 4?

\textcolor{sourcescolor}{Response:} Talk to your pediatrician before giving OTC cold and cough medicines. These medicines should not be given to children under 4 years of age, according to the FDA and the drug makers.

\textcolor{responsecolor}{Paraphrased Response: }Consult your pediatrician before giving your child over-the-counter cough and cold medicines. According to both the FDA and drug makers, children under 4 years of age should not take OTC cold and cough medicines.

\textcolor{instructioncolor}{Instruction: }Respond to the following query by building off of the response below. Specifically, rephrase each sentence in the response using a more fluent and useful wording to convey the same information. In other words, paraphrase each sentence of the response as an improved new sentence, with respect to the query. Do not refer to the response in your revised response.

\textcolor{querycolor}{Query:} How do I treat a blister?

\textcolor{sourcescolor}{Response:} Most blisters caused by friction or minor burns do not require a doctor's care. New skin will form underneath the affected area and the fluid is simply absorbed. Do not puncture a blister unless it is large, painful, or likely to be further irritated. The fluid-filled blister keeps the underlying skin clean, which prevents infection and promotes healing. But if you need to pop a blister or it pops by itself: Use a sterilized needle (to sterilize it, put the point or edge in a flame until it is red hot, or rinse it in alcohol). Wash your hands and the area thoroughly, then make a small hole; the fluid will drain on its own. If the fluid is white or yellow, the blister may be infected and needs medical attention. Do not remove the skin over a broken blister. The new skin underneath needs this protective cover. Apply an antibiotic ointment or cream. Look for signs of infection to develop, including pus drainage, red or warm skin surrounding the blister, or red streaks leading away from the blister.

\textcolor{responsecolor}{Paraphrased Response: }Blisters resulting from friction or minor burns generally don't necessitate medical intervention. Underneath the blister, new skin will develop, and the fluid will naturally be reabsorbed. It's advisable not to burst a blister unless it's sizable, causing discomfort, or at risk of additional irritation. The blister's fluid acts as a barrier, safeguarding the skin beneath from infection and aiding in the healing process. However, if there's a need to drain a blister, or if it bursts on its own, ensure to use a needle that has been sterilized either by heating it until it glows red or by cleansing it with alcohol. Before and after the procedure, clean your hands and the affected area thoroughly, then puncture the blister gently to let the fluid out without forcing it. Should the fluid appear white or yellow, it's indicative of an infection, and medical advice should be sought. Avoid peeling the skin off an opened blister as the new skin underneath requires this natural bandage for protection. Applying an antibacterial ointment or cream is recommended. Stay vigilant for signs of infection such as pus, reddening or warmth around the blister, or red lines emanating from it.

\textcolor{instructioncolor}{Instruction: }Respond to the following query by building off of the response below. Specifically, rephrase each sentence in the response using a more fluent and useful wording to convey the same information. In other words, paraphrase each sentence of the response as an improved new sentence, with respect to the query. Do not refer to the response in your revised response.

\textcolor{querycolor}{Query:} What medications can I use to lessen my child's teething pain?

\textcolor{sourcescolor}{Response:} Use baby acetaminophen (Tylenol) to dull pain. Do not use ibuprofen unless the baby is at least 6 months old.

\textcolor{responsecolor}{Paraphrased Response: }\end{promptbox}
\refstepcounter{promptbox}
\textbf{Box \thepromptbox: } Example of the prompt with few-shot examples used to generate the paraphrased output for MASH queries.
\label{box:mash_pp_output_prompt}
\vspace{.2cm}

\subsubsection{Prompts for the Entailed Generation}
\label{Prompts for the Entailed Output}
\vspace{.2cm}
\begin{promptbox}
\raggedright
\textcolor{instructioncolor}{Instruction: }Respond to the following query by building off of the response below. Specifically, rephrase and combine the sentences in the response by paraphrasing them, cutting out extraneous or redundant information, and simplifying details that are too fine-grained with respect to the question. Also, simplify premises to their logical conclusions to more directly answer the query. Do not refer to the response in your revised response.

\textcolor{querycolor}{Query:} What's the life cycle of a frog?

\textcolor{sourcescolor}{Response:} The stages of a frog’s life cycle are egg, larva, and adult. Growing through these stages is a process called metamorphosis. In a frog egg, the yolk splits into many cells which are the beginnings of a tadpole, or a frog larva. In one to three weeks, a small tadpole hatches from the egg. During the larval stage, tadpoles, or frogs’ larvae, are equipped with simple gills, a mouth, and a lengthy tail. As they develop, tadpoles consume plants and algae and then grow back legs, followed by front legs. Shortly afterward, their body begins to shift in shape and they gain the ability to consume insects. Then, the tails of the tadpoles grow smaller and skin develops over their gills while they grow eardrums and lungs.

\textcolor{responsecolor}{Revised Response: }The stages of a frog’s life cycle are egg, larva, and adult. Growing through these stages is a process called metamorphosis. Cells in the yolk of a frog egg form a tadpole, or a frog larva. In one to three weeks, a small tadpole hatches from the egg. During the larval stage, tadpoles are equipped with simple gills, a mouth, and a lengthy tail. As they develop, tadpoles consume plants and algae and then grow back legs, followed by front legs. Shortly afterward, they gain the ability to consume insects. Then, the tails of the tadpoles shrink, they lose their gills, and they grow eardrums and lungs.

\textcolor{instructioncolor}{Instruction: }Respond to the following query by building off of the response below. Specifically, rephrase and combine the sentences in the response by paraphrasing them, cutting out extraneous or redundant information, and simplifying details that are too fine-grained with respect to the question. Also, simplify premises to their logical conclusions to more directly answer the query. Do not refer to the response in your revised response.

\textcolor{querycolor}{Query:} What are the benefits of hyaluronic acid for skin care?

\textcolor{sourcescolor}{Response:} Hyaluronic acid helps to retain moisture. It makes it easier for skin to be flexible and diminishes skin lines and wrinkles. Hyaluronic acid has been shown to decrease the recovery time of wounds and can lessen scarring.

\textcolor{responsecolor}{Revised Response: }Hyaluronic acid helps to retain moisture. It makes it easier for skin to be flexible and diminishes wrinkles. Hyaluronic acid may decrease the recovery time of wounds and can lessen scarring.

\textcolor{instructioncolor}{Instruction: }Respond to the following query by building off of the response below. Specifically, rephrase and combine the sentences in the response by paraphrasing them, cutting out extraneous or redundant information, and simplifying details that are too fine-grained with respect to the question. Also, simplify premises to their logical conclusions to more directly answer the query. Do not refer to the response in your revised response.

\textcolor{querycolor}{Query:} What is a heirloom tomato?

\textcolor{sourcescolor}{Response:} Unlike tomatoes that carry a genetic mutation that makes them look consistently red at the expense of their flavor, many heirloom tomatoes taste sweeter. Genuine heirloom tomatoes have to be old due to being a variety over the last 25 to 50 years.

\textcolor{responsecolor}{Revised Response: }Unlike tomatoes that carry a genetic mutation that makes them look consistently red at the expense of their flavor, many heirloom tomatoes taste sweeter. Genuine heirloom tomatoes have been a variety over the last 25 to 50 years.

\textcolor{instructioncolor}{Instruction: }Respond to the following query by building off of the response below. Specifically, rephrase and combine the sentences in the response by paraphrasing them, cutting out extraneous or redundant information, and simplifying details that are too fine-grained with respect to the question. Also, simplify premises to their logical conclusions to more directly answer the query. Do not refer to the response in your revised response.

\textcolor{querycolor}{Query:} What's the flavor of juicy fruit gum?

\textcolor{sourcescolor}{Response:} the mixture of fruit flavors in Juicy Fruit is comprised of predominately lemon, orange, pineapple, and banana notes.

\textcolor{responsecolor}{Revised Response: }Juicy Fruit flavor is a blend of mostly orange, lemon, banana, and pineapple flavors.

\textcolor{instructioncolor}{Instruction: }Respond to the following query by building off of the response below. Specifically, rephrase and combine the sentences in the response by paraphrasing them, cutting out extraneous or redundant information, and simplifying details that are too fine-grained with respect to the question. Also, simplify premises to their logical conclusions to more directly answer the query. Do not refer to the response in your revised response.

\textcolor{querycolor}{Query:} Explain to a third-grader: who stole the mona lisa from the louvre in 1911

\textcolor{sourcescolor}{Response:} The man who stole her, Vincenzo Peruggia, had once worked at the Louvre and had, ironically, helped build a protective case for the 16th century masterpiece.

\textcolor{responsecolor}{Revised Response: }\end{promptbox}
\refstepcounter{promptbox}
\textbf{Box \thepromptbox: } Example of the prompt with few-shot examples used to generate the entailed output for NQ and ETa3G queries.
\label{box:nq_eta3g_ent_output_prompt}
\vspace{.2cm}
\vspace{.2cm}
\begin{promptbox}
\raggedright
\textcolor{instructioncolor}{Instruction: }Respond to the following query by building off of the response below. Specifically, rephrase and combine the sentences in the response by paraphrasing them, cutting out extraneous or redundant information, and simplifying details that are too fine-grained with respect to the question. Also, simplify premises to their logical conclusions to more directly answer the query. Do not refer to the response in your revised response.

\textcolor{querycolor}{Query:} Are the director of film Move (1970 Film) and director of film Méditerranée (1963 Film) from the same country?

\textcolor{sourcescolor}{Response:} Move (1970 Film) was directed by Stuart Rosenberg, who was born on August 11, 1927, and passing on March 15, 2007, and was renowned for his work as a director in both American film and television. On the other hand, Méditerranée was directed by Jean-Daniel Pollet who lived from 1936 to 2004 and was celebrated as a French filmmaker and scriptwriter.

\textcolor{responsecolor}{Revised Response: }No; Stuart Rosenberg directed 'Move' and was American while Jean-Daniel Pollet directed 'Méditerranée' and was French.

\textcolor{instructioncolor}{Instruction: }Respond to the following query by building off of the response below. Specifically, rephrase and combine the sentences in the response by paraphrasing them, cutting out extraneous or redundant information, and simplifying details that are too fine-grained with respect to the question. Also, simplify premises to their logical conclusions to more directly answer the query. Do not refer to the response in your revised response.

\textcolor{querycolor}{Query:} What nationality is the director of film Wedding Night In Paradise (1950 Film)?

\textcolor{sourcescolor}{Response:} Wedding Night in Paradise is a 1950 West German musical comedy film directed by Géza von Bolváry. Géza von Bolváry (full name Géza Maria von Bolváry-Zahn; 26 December 1897 – 10 August 1961) was a Hungarian actor, screenwriter and film director.

\textcolor{responsecolor}{Revised Response: }Géza von Bolváry directed 'Wedding Night in Paradise' and was Hungarian.

\textcolor{instructioncolor}{Instruction: }Respond to the following query by building off of the response below. Specifically, rephrase and combine the sentences in the response by paraphrasing them, cutting out extraneous or redundant information, and simplifying details that are too fine-grained with respect to the question. Also, simplify premises to their logical conclusions to more directly answer the query. Do not refer to the response in your revised response.

\textcolor{querycolor}{Query:} When did the performer of song A Pretty Girl Milking Her Cow die?

\textcolor{sourcescolor}{Response:} Judy Garland (born Frances Ethel Gumm; June 10, 1922 – June 22, 1969) was an American actress, singer and dancer. A Pretty Girl Milking Her Cow enjoyed a revival when an updated swing version sung by Irish-American singer/actress Judy Garland was featured in the 1940 film Little Nellie Kelly.

\textcolor{responsecolor}{Revised Response: }Judy Garland, the performer of 'A Pretty Girl Milking Her Cow' died on June 22, 1969.

\textcolor{instructioncolor}{Instruction: }Respond to the following query by building off of the response below. Specifically, rephrase and combine the sentences in the response by paraphrasing them, cutting out extraneous or redundant information, and simplifying details that are too fine-grained with respect to the question. Also, simplify premises to their logical conclusions to more directly answer the query. Do not refer to the response in your revised response.

\textcolor{querycolor}{Query:} Do both films The Falcon (Film) and Valentin The Good have the directors from the same country?

\textcolor{sourcescolor}{Response:} Banović Strahinja( released internationally as The Falcon) is a 1981 Yugoslavian- German adventure film written and directed by Vatroslav Mimica. Vatroslav Mimica( born 25 June 1923) is a Croatian film director and screenwriter. Valentin the Good is a 1942 Czech comedy film directed by Martin Frič and Martin Frič( 29 March 1902 – 26 August 1968) was a Czech film director, screenwriter and actor.

\textcolor{responsecolor}{Revised Response: }No; the director of 'The Falcon' is Vatroslav Mimica, who is from Croatia, while 'Valentin the Good' was directed by Martin Frič, from Czechia.

\textcolor{instructioncolor}{Instruction: }Respond to the following query by building off of the response below. Specifically, rephrase and combine the sentences in the response by paraphrasing them, cutting out extraneous or redundant information, and simplifying details that are too fine-grained with respect to the question. Also, simplify premises to their logical conclusions to more directly answer the query. Do not refer to the response in your revised response.

\textcolor{querycolor}{Query:} What is the date of birth of Mina Gerhardsen's father?

\textcolor{sourcescolor}{Response:} Mina Gerhardsen (born 14 September 1975) is a Norwegian politician for the Labour Party. She is the daughter of Rune Gerhardsen and Tove Strand. Rune Gerhardsen (born 13 June 1946) is a Norwegian politician.

\textcolor{responsecolor}{Revised Response: }Mina Gerhardsen's father, Rune Gerhardsen, was born on June 13, 1946.

\textcolor{instructioncolor}{Instruction: }Respond to the following query by building off of the response below. Specifically, rephrase and combine the sentences in the response by paraphrasing them, cutting out extraneous or redundant information, and simplifying details that are too fine-grained with respect to the question. Also, simplify premises to their logical conclusions to more directly answer the query. Do not refer to the response in your revised response.

\textcolor{querycolor}{Query:} When is the composer of film Sruthilayalu 's birthday?

\textcolor{sourcescolor}{Response:} Sruthilayalu is a 1987 Indian Telugu-language musical drama film, written and directed by K. Viswanath. The film stars Rajasekhar and Sumalata with soundtrack composed by K. V. Mahadevan. Krishnankoil Venkadachalam Mahadevan (14 March 1918 – 21 June 2001) was an Indian composer, singer-songwriter, music producer, and musician known for his works in Tamil cinema, Telugu cinema, Kannada cinema, and Malayalam cinema.

\textcolor{responsecolor}{Revised Response: }The composer of 'Sruthilayalu', Krishnankoil Venkadachalam Mahadevan, was born on March 14, 1918.

\textcolor{instructioncolor}{Instruction: }Respond to the following query by building off of the response below. Specifically, rephrase and combine the sentences in the response by paraphrasing them, cutting out extraneous or redundant information, and simplifying details that are too fine-grained with respect to the question. Also, simplify premises to their logical conclusions to more directly answer the query. Do not refer to the response in your revised response.

\textcolor{querycolor}{Query:} When did Fatima Bint Mubarak Al Ketbi's husband die?

\textcolor{sourcescolor}{Response:} Fatima bint Mubarak Al Ketbi is the third wife of Sheikh Zayed bin Sultan Al Nahyan. Sheikh Zayed bin Sultan Al Nahyan ; 6 May 1918 – 2 November 2004) was the ruler of Abu Dhabi for more than 30 years (6 August 1966 – 2 November 2004).

\textcolor{responsecolor}{Revised Response: }Fatima bint Mubarak Al Ketbi's husband, Sheikh Zayed bin Sultan Al Nahyan, passed away on November 2, 2004.

\textcolor{instructioncolor}{Instruction: }Respond to the following query by building off of the response below. Specifically, rephrase and combine the sentences in the response by paraphrasing them, cutting out extraneous or redundant information, and simplifying details that are too fine-grained with respect to the question. Also, simplify premises to their logical conclusions to more directly answer the query. Do not refer to the response in your revised response.

\textcolor{querycolor}{Query:} Who is the mother-in-law of Andrew Murray (Scottish Soldier)?

\textcolor{sourcescolor}{Response:} Sir Andrew Murray married Christina Bruce, a sister of King Robert I of Scotland. Christina Bruce (c. 1278 – 1356/1357), also known as Christina de Brus, was a daughter of Marjorie, Countess of Carrick, and her husband, Robert de Brus, 'jure uxoris' Earl of Carrick, as well as a sister of Robert the Bruce, King of Scots.

\textcolor{responsecolor}{Revised Response: }Sir Andrew Murray married Christina Bruce, making her mother, Marjorie, Countess of Carrick, his mother-in-law.

\textcolor{instructioncolor}{Instruction: }Respond to the following query by building off of the response below. Specifically, rephrase and combine the sentences in the response by paraphrasing them, cutting out extraneous or redundant information, and simplifying details that are too fine-grained with respect to the question. Also, simplify premises to their logical conclusions to more directly answer the query. Do not refer to the response in your revised response.

\textcolor{querycolor}{Query:} When did the performer of song A Pretty Girl Milking Her Cow die?

\textcolor{sourcescolor}{Response:} Judy Garland (born Frances Ethel Gumm; June 10, 1922 – June 22, 1969) was an American actress, singer and dancer. A Pretty Girl Milking Her Cow enjoyed a revival when an updated swing version sung by Irish-American singer/actress Judy Garland was featured in the 1940 film Little Nellie Kelly.

\textcolor{responsecolor}{Revised Response: }The song 'A Pretty Girl Milking Her Cow' became especially popular from its rendition by Judy Garland, who passed away on June 22, 1969.

\textcolor{instructioncolor}{Instruction: }Respond to the following query by building off of the response below. Specifically, rephrase and combine the sentences in the response by paraphrasing them, cutting out extraneous or redundant information, and simplifying details that are too fine-grained with respect to the question. Also, simplify premises to their logical conclusions to more directly answer the query. Do not refer to the response in your revised response.

\textcolor{querycolor}{Query:} What is the date of birth of the director of film Runway (2010 Film)?

\textcolor{sourcescolor}{Response:} Runway is a 2010 Bangladeshi drama film directed by Tareque Masud. Tareque Masud( 6 December 1956 – 13 August 2011) was a Bangladeshi independent film director, film producer, screenwriter and lyricist.

\textcolor{responsecolor}{Revised Response: }\end{promptbox}
\refstepcounter{promptbox}
\textbf{Box \thepromptbox: } Example of the prompt with few-shot examples used to generate the entailed output for 2WikiMH queries.
\label{box:2WikiMH_ent_output_prompt}
\vspace{.2cm}
\vspace{.2cm}
\begin{promptbox}
\raggedright
\textcolor{instructioncolor}{Instruction: }Respond to the following query by building off of the response below. Specifically, rephrase and combine the sentences in the response by paraphrasing them, cutting out extraneous or redundant information, and simplifying details that are too fine-grained with respect to the question. Also, simplify premises to their logical conclusions to more directly answer the query. Do not refer to the response in your revised response.

\textcolor{querycolor}{Query:} What can you do to prevent migraine with brainstem aura?

\textcolor{sourcescolor}{Response:} To prevent migraine with brainstem aura, it helps to avoid the things that usually cause one. This includes being sure to Get enough sleep, Limit your stress, and Exercise daily. Additionally, Eat a balanced diet, Avoid drugs and alcohol, Don't skip meals,  and Limit caffeine. Avoid any foods that have been triggers. Furthermore, If you don't respond to other treatments and you have 4 or more migraine days a month, your doctor may suggest preventive medicines, including seizure medicines, blood pressure medicines (like beta blockers and calcium channel blockers), and some antidepressants. Also, CGRP inhibitors are a new class of preventive medicine.

\textcolor{responsecolor}{Revised Response: }To prevent migraine with brainstem aura, avoid common triggers by getting sufficient sleep, reducing stress, exercising regularly, eating a balanced diet, avoiding drugs and alcohol, and limiting caffeine. If experiencing four or more migraine days a month and unresponsive to other treatments, your doctor may recommend preventive medications.

\textcolor{instructioncolor}{Instruction: }Respond to the following query by building off of the response below. Specifically, rephrase and combine the sentences in the response by paraphrasing them, cutting out extraneous or redundant information, and simplifying details that are too fine-grained with respect to the question. Also, simplify premises to their logical conclusions to more directly answer the query. Do not refer to the response in your revised response.

\textcolor{querycolor}{Query:} What should I do if my child has a biting injury?

\textcolor{sourcescolor}{Response:} The first thing to do for any biting injury is to wash the area with soap and water. Even little teeth can break the skin. If the bite is bleeding and the wound appears to be deep, call your child's doctor. The bite may need medical treatment, which could include antibiotics or a tetanus shot, or both.

\textcolor{responsecolor}{Revised Response: }For a biting injury, first wash the area with soap and water. If the bite is bleeding and deep, consult your child's doctor for possible medical treatments, including antibiotics and a tetanus shot.

\textcolor{instructioncolor}{Instruction: }Respond to the following query by building off of the response below. Specifically, rephrase and combine the sentences in the response by paraphrasing them, cutting out extraneous or redundant information, and simplifying details that are too fine-grained with respect to the question. Also, simplify premises to their logical conclusions to more directly answer the query. Do not refer to the response in your revised response.

\textcolor{querycolor}{Query:} Can I give cough and cold medicines to children under 4?

\textcolor{sourcescolor}{Response:} Talk to your pediatrician before giving OTC cold and cough medicines. These medicines should not be given to children under 4 years of age, according to the FDA and the drug makers.

\textcolor{responsecolor}{Revised Response: }The FDA and drug makers advise against giving cough and cold medicines to children under 4.

\textcolor{instructioncolor}{Instruction: }Respond to the following query by building off of the response below. Specifically, rephrase and combine the sentences in the response by paraphrasing them, cutting out extraneous or redundant information, and simplifying details that are too fine-grained with respect to the question. Also, simplify premises to their logical conclusions to more directly answer the query. Do not refer to the response in your revised response.

\textcolor{querycolor}{Query:} How do I treat a blister?

\textcolor{sourcescolor}{Response:} Most blisters caused by friction or minor burns do not require a doctor's care. New skin will form underneath the affected area and the fluid is simply absorbed. Do not puncture a blister unless it is large, painful, or likely to be further irritated. The fluid-filled blister keeps the underlying skin clean, which prevents infection and promotes healing. But if you need to pop a blister or it pops by itself: Use a sterilized needle (to sterilize it, put the point or edge in a flame until it is red hot, or rinse it in alcohol). Wash your hands and the area thoroughly, then make a small hole; the fluid will drain on its own. If the fluid is white or yellow, the blister may be infected and needs medical attention. Do not remove the skin over a broken blister. The new skin underneath needs this protective cover. Apply an antibiotic ointment or cream. Look for signs of infection to develop, including pus drainage, red or warm skin surrounding the blister, or red streaks leading away from the blister.

\textcolor{responsecolor}{Revised Response: }Most blisters from friction or minor burns can heal without medical intervention, as new skin forms underneath and the fluid is naturally absorbed. Avoid puncturing a blister unless it's painful or likely to get irritated. In this case, sterilize a needle with flame or alcohol, clean your hands and the blister area, and then gently puncture the blister to let the fluid drain. Keep the protective skin layer over the blister and apply an antibiotic ointment. Watch for signs of infection like pus, redness, or warmth around the blister.

\textcolor{instructioncolor}{Instruction: }Respond to the following query by building off of the response below. Specifically, rephrase and combine the sentences in the response by paraphrasing them, cutting out extraneous or redundant information, and simplifying details that are too fine-grained with respect to the question. Also, simplify premises to their logical conclusions to more directly answer the query. Do not refer to the response in your revised response.

\textcolor{querycolor}{Query:} What medications can I use to lessen my child's teething pain?

\textcolor{sourcescolor}{Response:} Use baby acetaminophen (Tylenol) to dull pain. Do not use ibuprofen unless the baby is at least 6 months old.

\textcolor{responsecolor}{Revised Response: }
\end{promptbox}
\refstepcounter{promptbox}
\textbf{Box \thepromptbox: } Example of the prompt with few-shot examples used to generate the entailed output for MASH queries.
\label{box:mash_ent_output_prompt}
\vspace{.2cm}

\subsubsection{Prompts for the Abstractive Generation}
\label{Prompts for the Abstractive Output}
\vspace{.2cm}
\begin{promptbox}
\raggedright
\textcolor{instructioncolor}{Instruction: }Respond to the following query by building off of the response below. Specifically, rephrase and combine the sentences in the response by paraphrasing them, cutting out extraneous or redundant information, and simplifying details that are too fine-grained with respect to the question. Also, simplify premises to their logical conclusions to more directly answer the query. Most importantly, use accurate outside information to make the revised response a more useful answer to the query. If the provided response is not an accurate or useful answer to the query, then extensively revise it with accurate outside information. Respond in no more than about 100 words. Do not refer to the response in your revised response.

\textcolor{querycolor}{Query:} What's the life cycle of a frog?

\textcolor{sourcescolor}{Response:} The stages of a frog’s life cycle are egg, larva, and adult. In a frog egg, the yolk splits into many cells which are the beginnings of a tadpole, or a frog larva. In one to three weeks, a small tadpole hatches from the egg. During the larval stage, tadpoles, or frogs’ larvae, are equipped with simple gills, a mouth, and a lengthy tail. As they develop, tadpoles consume plants and algae and then grow back legs, followed by front legs. Shortly afterward, their body begins to shift in shape and they gain the ability to consume insects. Then, the tails of the tadpoles grow smaller and skin develops over their gills while they grow eardrums and lungs.

\textcolor{responsecolor}{Revised Response: }A frog matures from egg, to tadpole, to adult in a process called metamorphosis. After hatching from an egg into a tadpole, it will grow back legs and then front legs, while storing nutrients in its long tail. As the tadpole loses its gills, it gains lungs. At this point, the tadpole is considered to be a froglet, until its tail shrinks away, at which point it is a frog.

\textcolor{instructioncolor}{Instruction: }Respond to the following query by building off of the response below. Specifically, rephrase and combine the sentences in the response by paraphrasing them, cutting out extraneous or redundant information, and simplifying details that are too fine-grained with respect to the question. Also, simplify premises to their logical conclusions to more directly answer the query. Most importantly, use accurate outside information to make the revised response a more useful answer to the query. If the provided response is not an accurate or useful answer to the query, then extensively revise it with accurate outside information. Respond in no more than about 100 words. Do not refer to the response in your revised response.

\textcolor{querycolor}{Query:} What are the benefits of hyaluronic acid for skin care?

\textcolor{sourcescolor}{Response:} Hyaluronic acid helps to retain moisture. It makes it easier for skin to be flexible and diminishes skin lines and wrinkles.

\textcolor{responsecolor}{Revised Response: }Hyaluronic acid helps keep skin moist and wrinkle-free. It can also reduce scarring, which may be especially beneficial for people who have had severe acne in the past.

\textcolor{instructioncolor}{Instruction: }Respond to the following query by building off of the response below. Specifically, rephrase and combine the sentences in the response by paraphrasing them, cutting out extraneous or redundant information, and simplifying details that are too fine-grained with respect to the question. Also, simplify premises to their logical conclusions to more directly answer the query. Most importantly, use accurate outside information to make the revised response a more useful answer to the query. If the provided response is not an accurate or useful answer to the query, then extensively revise it with accurate outside information. Respond in no more than about 100 words. Do not refer to the response in your revised response.

\textcolor{querycolor}{Query:} What is a heirloom tomato?

\textcolor{sourcescolor}{Response:} Unlike tomatoes that carry a genetic mutation that makes them look consistently red at the expense of their flavor, many heirloom tomatoes taste sweeter. Genuine heirloom tomatoes have to be old due to being a variety over the last 25 to 50 years.

\textcolor{responsecolor}{Revised Response: }Heirloom tomatoes are tomatoes that have a sweet flavor profile and are often other colors than red, including yellow, orange, and purple. A tomato variety is considered heirloom once it’s been a variety for 25-50 years, although some insist that at least 100 years is necessary.

\textcolor{instructioncolor}{Instruction: }Respond to the following query by building off of the response below. Specifically, rephrase and combine the sentences in the response by paraphrasing them, cutting out extraneous or redundant information, and simplifying details that are too fine-grained with respect to the question. Also, simplify premises to their logical conclusions to more directly answer the query. Most importantly, use accurate outside information to make the revised response a more useful answer to the query. If the provided response is not an accurate or useful answer to the query, then extensively revise it with accurate outside information. Respond in no more than about 100 words. Do not refer to the response in your revised response.

\textcolor{querycolor}{Query:} What's the flavor of juicy fruit gum?

\textcolor{sourcescolor}{Response:} the mixture of fruit flavors in Juicy Fruit is comprised of predominately lemon, orange, pineapple, and banana notes.

\textcolor{responsecolor}{Revised Response: }Juicy Fruit flavor is a blend of mostly orange, lemon, banana, and pineapple flavors. Some say it tastes like coconut and jackfruit as well.

\textcolor{instructioncolor}{Instruction: }Respond to the following query by building off of the response below. Specifically, rephrase and combine the sentences in the response by paraphrasing them, cutting out extraneous or redundant information, and simplifying details that are too fine-grained with respect to the question. Also, simplify premises to their logical conclusions to more directly answer the query. Most importantly, use accurate outside information to make the revised response a more useful answer to the query. If the provided response is not an accurate or useful answer to the query, then extensively revise it with accurate outside information. Respond in no more than about 100 words. Do not refer to the response in your revised response.

\textcolor{querycolor}{Query:} Explain to a third-grader: who stole the mona lisa from the louvre in 1911

\textcolor{sourcescolor}{Response:} The man who stole her, Vincenzo Peruggia, had once worked at the Louvre and had, ironically, helped build a protective case for the 16th century masterpiece.

\textcolor{responsecolor}{Revised Response: }\end{promptbox}
\refstepcounter{promptbox}
\textbf{Box \thepromptbox: } Example of the prompt with few-shot examples used to generate the abstractive output for NQ and ETa3G queries.
\label{box:nq_eta3g_abs_output_prompt}
\vspace{.2cm}
\vspace{.2cm}
\begin{promptbox}
\raggedright
\textcolor{instructioncolor}{Instruction: }Respond to the following query by building off of the response below. Specifically, rephrase and combine the sentences in the response by paraphrasing them, cutting out extraneous or redundant information, and simplifying details that are too fine-grained with respect to the question. Also, simplify premises to their logical conclusions to more directly answer the query. Most importantly, use accurate outside information to make the revised response a more useful answer to the query. If the provided response is not an accurate or useful answer to the query, then extensively revise it with accurate outside information. Respond in no more than about 100 words. Do not refer to the response in your revised response.

\textcolor{querycolor}{Query:} Are the director of film Move (1970 Film) and director of film Méditerranée (1963 Film) from the same country?

\textcolor{sourcescolor}{Response:} Move (1970 Film) was directed by Stuart Rosenberg, who was born on August 11, 1927, and passing on March 15, 2007, and was renowned for his work as a director in both American film and television. On the other hand, Méditerranée was directed by Jean-Daniel Pollet who lived from 1936 to 2004 and was celebrated as a French filmmaker and scriptwriter.

\textcolor{responsecolor}{Revised Response: }No; the director of 'Move' was born in New York and was American while the director of 'Méditerranée' was born in Nord and was French.

\textcolor{instructioncolor}{Instruction: }Respond to the following query by building off of the response below. Specifically, rephrase and combine the sentences in the response by paraphrasing them, cutting out extraneous or redundant information, and simplifying details that are too fine-grained with respect to the question. Also, simplify premises to their logical conclusions to more directly answer the query. Most importantly, use accurate outside information to make the revised response a more useful answer to the query. If the provided response is not an accurate or useful answer to the query, then extensively revise it with accurate outside information. Respond in no more than about 100 words. Do not refer to the response in your revised response.

\textcolor{querycolor}{Query:} What nationality is the director of film Wedding Night In Paradise (1950 Film)?

\textcolor{sourcescolor}{Response:} Wedding Night in Paradise  is a 1950 West German musical comedy film directed by Géza von Bolváry. Géza von Bolváry (full name Géza Maria von Bolváry-Zahn; 26 December 1897 – 10 August 1961) was a Hungarian actor, screenwriter and film director.

\textcolor{responsecolor}{Revised Response: }The director of 'Wedding Night in Paradise' was Hungarian, but also worked in Germany and Austria.

\textcolor{instructioncolor}{Instruction: }Respond to the following query by building off of the response below. Specifically, rephrase and combine the sentences in the response by paraphrasing them, cutting out extraneous or redundant information, and simplifying details that are too fine-grained with respect to the question. Also, simplify premises to their logical conclusions to more directly answer the query. Most importantly, use accurate outside information to make the revised response a more useful answer to the query. If the provided response is not an accurate or useful answer to the query, then extensively revise it with accurate outside information. Respond in no more than about 100 words. Do not refer to the response in your revised response.

\textcolor{querycolor}{Query:} When did the performer of song A Pretty Girl Milking Her Cow die?

\textcolor{sourcescolor}{Response:} Judy Garland (born Frances Ethel Gumm; June 10, 1922 – June 22, 1969) was an American actress, singer and dancer. A Pretty Girl Milking Her Cow enjoyed a revival when an updated swing version sung by Irish-American singer/actress Judy Garland was featured in the 1940 film Little Nellie Kelly.

\textcolor{responsecolor}{Revised Response: }Judy Garland, the performer of 'A Pretty Girl Milking Her Cow' died on June 22, 1969 at age 47.

\textcolor{instructioncolor}{Instruction: }Respond to the following query by building off of the response below. Specifically, rephrase and combine the sentences in the response by paraphrasing them, cutting out extraneous or redundant information, and simplifying details that are too fine-grained with respect to the question. Also, simplify premises to their logical conclusions to more directly answer the query. Most importantly, use accurate outside information to make the revised response a more useful answer to the query. If the provided response is not an accurate or useful answer to the query, then extensively revise it with accurate outside information. Respond in no more than about 100 words. Do not refer to the response in your revised response.

\textcolor{querycolor}{Query:} Do both films The Falcon (Film) and Valentin The Good have the directors from the same country?

\textcolor{sourcescolor}{Response:} Banović Strahinja( released internationally as The Falcon) is a 1981 Yugoslavian- German adventure film written and directed by Vatroslav Mimica. Vatroslav Mimica( born 25 June 1923) is a Croatian film director and screenwriter. Valentin the Good is a 1942 Czech comedy film directed by Martin Frič and Martin Frič( 29 March 1902 – 26 August 1968) was a Czech film director, screenwriter and actor.

\textcolor{responsecolor}{Revised Response: }No; the director of 'The Falcon' is Vatroslav Mimica, who is from Croatia, while 'Valentin the Good' was directed by Martin Frič, from the Czech Republic. However, during Frič's life, the Czech Republic did not exist; modern-day Czech Republic was first part of the Austro-Hungarian Empire and then, after World War I, was part of Czechoslovakia.

\textcolor{instructioncolor}{Instruction: }Respond to the following query by building off of the response below. Specifically, rephrase and combine the sentences in the response by paraphrasing them, cutting out extraneous or redundant information, and simplifying details that are too fine-grained with respect to the question. Also, simplify premises to their logical conclusions to more directly answer the query. Most importantly, use accurate outside information to make the revised response a more useful answer to the query. If the provided response is not an accurate or useful answer to the query, then extensively revise it with accurate outside information. Respond in no more than about 100 words. Do not refer to the response in your revised response.

\textcolor{querycolor}{Query:} What is the date of birth of Mina Gerhardsen's father?

\textcolor{sourcescolor}{Response:} Mina Gerhardsen (born 14 September 1975) is a Norwegian politician for the Labour Party. She is the daughter of Rune Gerhardsen and Tove Strand. Rune Gerhardsen (born 13 June 1946) is a Norwegian politician.

\textcolor{responsecolor}{Revised Response: }Mina Gerhardsen's father, Rune Gerhardsen, was born on June 13, 1946.

\textcolor{instructioncolor}{Instruction: }Respond to the following query by building off of the response below. Specifically, rephrase and combine the sentences in the response by paraphrasing them, cutting out extraneous or redundant information, and simplifying details that are too fine-grained with respect to the question. Also, simplify premises to their logical conclusions to more directly answer the query. Most importantly, use accurate outside information to make the revised response a more useful answer to the query. If the provided response is not an accurate or useful answer to the query, then extensively revise it with accurate outside information. Respond in no more than about 100 words. Do not refer to the response in your revised response.

\textcolor{querycolor}{Query:} When is the composer of film Sruthilayalu 's birthday?

\textcolor{sourcescolor}{Response:} Sruthilayalu is a 1987 Indian Telugu-language musical drama film, written and directed by K. Viswanath. The film stars Rajasekhar and Sumalata with soundtrack composed by K. V. Mahadevan. Krishnankoil Venkadachalam Mahadevan (14 March 1918 – 21 June 2001) was an Indian composer, singer-songwriter, music producer, and musician known for his works in Tamil cinema, Telugu cinema, Kannada cinema, and Malayalam cinema.

\textcolor{responsecolor}{Revised Response: }The composer of 'Sruthilayalu', Krishnankoil Venkadachalam Mahadevan, was born on March 14, 1918.

\textcolor{instructioncolor}{Instruction: }Respond to the following query by building off of the response below. Specifically, rephrase and combine the sentences in the response by paraphrasing them, cutting out extraneous or redundant information, and simplifying details that are too fine-grained with respect to the question. Also, simplify premises to their logical conclusions to more directly answer the query. Most importantly, use accurate outside information to make the revised response a more useful answer to the query. If the provided response is not an accurate or useful answer to the query, then extensively revise it with accurate outside information. Respond in no more than about 100 words. Do not refer to the response in your revised response.

\textcolor{querycolor}{Query:} When did Fatima Bint Mubarak Al Ketbi's husband die?

\textcolor{sourcescolor}{Response:} Fatima bint Mubarak Al Ketbi is the third wife of Sheikh Zayed bin Sultan Al Nahyan. Sheikh Zayed bin Sultan Al Nahyan ; 6 May 1918 – 2 November 2004) was the ruler of Abu Dhabi for more than 30 years (6 August 1966 – 2 November 2004).

\textcolor{responsecolor}{Revised Response: }Fatima bint Mubarak Al Ketbi's husband was Sheikh Zayed bin Sultan Al Nahyan, who passed away when he was 86, on November 2, 2004.

\textcolor{instructioncolor}{Instruction: }Respond to the following query by building off of the response below. Specifically, rephrase and combine the sentences in the response by paraphrasing them, cutting out extraneous or redundant information, and simplifying details that are too fine-grained with respect to the question. Also, simplify premises to their logical conclusions to more directly answer the query. Most importantly, use accurate outside information to make the revised response a more useful answer to the query. If the provided response is not an accurate or useful answer to the query, then extensively revise it with accurate outside information. Respond in no more than about 100 words. Do not refer to the response in your revised response.

\textcolor{querycolor}{Query:} Who is the mother-in-law of Andrew Murray (Scottish Soldier)?

\textcolor{sourcescolor}{Response:} Sir Andrew Murray married Christina Bruce, a sister of King Robert I of Scotland. Christina Bruce (c. 1278 – 1356/1357), also known as Christina de Brus, was a daughter of Marjorie, Countess of Carrick, and her husband, Robert de Brus, 'jure uxoris' Earl of Carrick, as well as a sister of Robert the Bruce, King of Scots.

\textcolor{responsecolor}{Revised Response: }Sir Andrew Murray's married Christina Bruce, making her mother, Marjorie, Countess of Carrick, his mother-in-law. Marjorie is notable for her son, Robert the Bruce, King of Scots, who led the first War of Scottish Independence against England.

\textcolor{instructioncolor}{Instruction: }Respond to the following query by building off of the response below. Specifically, rephrase and combine the sentences in the response by paraphrasing them, cutting out extraneous or redundant information, and simplifying details that are too fine-grained with respect to the question. Also, simplify premises to their logical conclusions to more directly answer the query. Most importantly, use accurate outside information to make the revised response a more useful answer to the query. If the provided response is not an accurate or useful answer to the query, then extensively revise it with accurate outside information. Respond in no more than about 100 words. Do not refer to the response in your revised response.

\textcolor{querycolor}{Query:} When did the performer of song A Pretty Girl Milking Her Cow die?

\textcolor{sourcescolor}{Response:} Judy Garland (born Frances Ethel Gumm; June 10, 1922 – June 22, 1969) was an American actress, singer and dancer. A Pretty Girl Milking Her Cow enjoyed a revival when an updated swing version sung by Irish-American singer/actress Judy Garland was featured in the 1940 film Little Nellie Kelly.

\textcolor{responsecolor}{Revised Response: }While the song 'A Pretty Girl Milking Her Cow' has been sung by many people, it became especially popular from its rendition by Judy Garland, who passed away on June 22, 1969.

\textcolor{instructioncolor}{Instruction: }Respond to the following query by building off of the response below. Specifically, rephrase and combine the sentences in the response by paraphrasing them, cutting out extraneous or redundant information, and simplifying details that are too fine-grained with respect to the question. Also, simplify premises to their logical conclusions to more directly answer the query. Most importantly, use accurate outside information to make the revised response a more useful answer to the query. If the provided response is not an accurate or useful answer to the query, then extensively revise it with accurate outside information. Respond in no more than about 100 words. Do not refer to the response in your revised response.

\textcolor{querycolor}{Query:} What is the date of birth of the director of film Runway (2010 Film)?

\textcolor{sourcescolor}{Response:} Runway is a 2010 Bangladeshi drama film directed by Tareque Masud. Tareque Masud( 6 December 1956 – 13 August 2011) was a Bangladeshi independent film director, film producer, screenwriter and lyricist.

\textcolor{responsecolor}{Revised Response: }\end{promptbox}
\refstepcounter{promptbox}
\textbf{Box \thepromptbox: } Example of the prompt with few-shot examples used to generate the abstractive output for 2WikiMH queries.
\label{box:2WikiMH_abs_output_prompt}
\vspace{.2cm}
\vspace{.2cm}
\begin{promptbox}
\raggedright
\textcolor{instructioncolor}{Instruction: }Respond to the following query by building off of the response below. Specifically, rephrase and combine the sentences in the response by paraphrasing them, cutting out extraneous or redundant information, and simplifying details that are too fine-grained with respect to the question. Also, simplify premises to their logical conclusions to more directly answer the query. Most importantly, use accurate outside information to make the revised response a more useful answer to the query. If the provided response is not an accurate or useful answer to the query, then extensively revise it with accurate outside information. Respond in no more than about 100 words. Do not refer to the response in your revised response.

\textcolor{querycolor}{Query:} What can you do to prevent migraine with brainstem aura?

\textcolor{sourcescolor}{Response:} To prevent migraine with brainstem aura, it helps to avoid the things that usually cause one. This includes being sure to Get enough sleep, Limit your stress, and Exercise daily. Additionally, Eat a balanced diet, Avoid drugs and alcohol, Don't skip meals,  and Limit caffeine. Avoid any foods that have been triggers.

\textcolor{responsecolor}{Revised Response: }Preventing migraine with brainstem aura involves lifestyle adjustments and possibly medication for those with frequent episodes. Key strategies include maintaining regular sleep patterns, managing stress, exercising regularly, eating a balanced diet, avoiding known food triggers, limiting caffeine and alcohol intake, and not skipping meals. Furthermore, If you don't respond to other treatments and you have 4 or more migraine days a month, your doctor may suggest preventive medicines, including seizure medicines, blood pressure medicines (like beta blockers and calcium channel blockers), and some antidepressants. Also, CGRP inhibitors are a new class of preventive medicine. There is anecdotal evidence that triptans can worsen migraine symptoms, even though it is often prescribed by medical professionals.

\textcolor{instructioncolor}{Instruction: }Respond to the following query by building off of the response below. Specifically, rephrase and combine the sentences in the response by paraphrasing them, cutting out extraneous or redundant information, and simplifying details that are too fine-grained with respect to the question. Also, simplify premises to their logical conclusions to more directly answer the query. Most importantly, use accurate outside information to make the revised response a more useful answer to the query. If the provided response is not an accurate or useful answer to the query, then extensively revise it with accurate outside information. Respond in no more than about 100 words. Do not refer to the response in your revised response.

\textcolor{querycolor}{Query:} What should I do if my child has a biting injury?

\textcolor{sourcescolor}{Response:} The first thing to do for any biting injury is to wash the area with soap and water. Even little teeth can break the skin. If the bite is bleeding and the wound appears to be deep, call your child's doctor. The bite may need medical treatment, which could include antibiotics or a tetanus shot, or both.

\textcolor{responsecolor}{Revised Response: }For a biting injury, immediately clean the wound with soap and water. If it's bleeding or deep, contact your child's doctor for potential treatment, which may involve antibiotics or a tetanus shot. If the child was bitten by an animal, it can be life-saving to seek medical treatment for a rabies vaccine within 24 hours, or at least within 72 hours. Even if more than 72 hours have elapsed since the bite, it is still not too late to get the shot. If left untreated, rabies can be fatal.

\textcolor{instructioncolor}{Instruction: }Respond to the following query by building off of the response below. Specifically, rephrase and combine the sentences in the response by paraphrasing them, cutting out extraneous or redundant information, and simplifying details that are too fine-grained with respect to the question. Also, simplify premises to their logical conclusions to more directly answer the query. Most importantly, use accurate outside information to make the revised response a more useful answer to the query. If the provided response is not an accurate or useful answer to the query, then extensively revise it with accurate outside information. Respond in no more than about 100 words. Do not refer to the response in your revised response.

\textcolor{querycolor}{Query:} Can I give cough and cold medicines to children under 4?

\textcolor{sourcescolor}{Response:} Talk to your pediatrician before giving OTC cold and cough medicines. These medicines should not be given to children under 4 years of age, according to the FDA and the drug makers.

\textcolor{responsecolor}{Revised Response: }The FDA and drug makers advise against giving cough and cold medicines to children under 4. Also, never give adult medicines to children; only use medicines designed for children. Lastly, do not give aspirin to a child due to the risk of Reye's syndrome, which is a rare but serious condition.

\textcolor{instructioncolor}{Instruction: }Respond to the following query by building off of the response below. Specifically, rephrase and combine the sentences in the response by paraphrasing them, cutting out extraneous or redundant information, and simplifying details that are too fine-grained with respect to the question. Also, simplify premises to their logical conclusions to more directly answer the query. Most importantly, use accurate outside information to make the revised response a more useful answer to the query. If the provided response is not an accurate or useful answer to the query, then extensively revise it with accurate outside information. Respond in no more than about 100 words. Do not refer to the response in your revised response.

\textcolor{querycolor}{Query:} How do I treat a blister?

\textcolor{sourcescolor}{Response:} Most blisters caused by friction or minor burns do not require a doctor's care. New skin will form underneath the affected area and the fluid is simply absorbed. Do not puncture a blister unless it is large, painful, or likely to be further irritated. The fluid-filled blister keeps the underlying skin clean, which prevents infection and promotes healing. But if you need to pop a blister or it pops by itself: Use a sterilized needle (to sterilize it, put the point or edge in a flame until it is red hot, or rinse it in alcohol). Wash your hands and the area thoroughly, then make a small hole; the fluid will drain on its own. If the fluid is white or yellow, the blister may be infected and needs medical attention. Do not remove the skin over a broken blister. The new skin underneath needs this protective cover. Apply an antibiotic ointment or cream. Look for signs of infection to develop, including pus drainage, red or warm skin surrounding the blister, or red streaks leading away from the blister.

\textcolor{responsecolor}{Revised Response: }For most blisters, medical care isn't necessary. Allow new skin to form beneath the blister and the fluid to absorb. To protect the blister from further irritation, it is advisable to cover it with moleskin, a durable cotton fabric with an adhesive backing available at most drug stores. If a blister is large, painful, or might be further irritated, and needs to be popped, sterilize a needle with heat or alcohol, clean your hands and the blister area, then gently puncture it to let the fluid drain, but don't remove the overlying skin as it protects the new skin. If the blister's fluid isn't clear (indicating possible infection), seek medical attention. Apply an antibiotic ointment, cover it with a bandage and monitor for signs of infection like pus, redness, or warmth.

\textcolor{instructioncolor}{Instruction: }Respond to the following query by building off of the response below. Specifically, rephrase and combine the sentences in the response by paraphrasing them, cutting out extraneous or redundant information, and simplifying details that are too fine-grained with respect to the question. Also, simplify premises to their logical conclusions to more directly answer the query. Most importantly, use accurate outside information to make the revised response a more useful answer to the query. If the provided response is not an accurate or useful answer to the query, then extensively revise it with accurate outside information. Respond in no more than about 100 words. Do not refer to the response in your revised response.

\textcolor{querycolor}{Query:} What medications can I use to lessen my child's teething pain?

\textcolor{sourcescolor}{Response:} Use baby acetaminophen (Tylenol) to dull pain. Do not use ibuprofen unless the baby is at least 6 months old.

\textcolor{responsecolor}{Revised Response: }\end{promptbox}
\refstepcounter{promptbox}
\textbf{Box \thepromptbox: } Example of the prompt with few-shot examples used to generate the abstractive output for MASH queries.
\label{box:mash_abs_output_prompt}
\vspace{.2cm}

\subsubsection{Prompts for Citation Identification}
\label{Prompts for Citation Generation}
\vspace{.2cm}
\begin{promptbox}
\raggedright
\textcolor{instructioncolor}{Instruction: }Examine the text and numbered quotes below. The text was likely written from one or a few of the quotes. Which of these quotes, if any, could have been used to write the text? It may be none of the quotes, one quote, or several of the quotes. Do not provide redundant quotes. Respond with the number of the quote(s) in brackets.

\textcolor{querycolor}{Text:} As they develop, tadpoles consume plants and algae and then grow back legs, followed by front legs

\textcolor{sourcescolor}{Quotes:}

[1] "The life cycle of a frog consists of three stages: egg, larva, and adult"

[2] "As the frog grows, it moves through these stages in a process known as metamorphosis"

[3] "the yolk in each egg splits into more and more cells and begins to take the form of a tadpole, the larva of a frog"

[4] "Within one to three weeks, the egg is ready to hatch, and a tiny tadpole breaks free"

[5] "[t]adpoles, frogs' larvae, have rudimentary gills, a mouth, and a long tail"

[6] "tadpoles feed on algae and other vegetation"

[7] "they grow back legs, then front legs too"

[8] "Soon after, their body starts to change shape, and they’re able to start eating insects"

[9] "Next, the tadpoles’ tails shrink away, and skin grows over their gills, as they develop lungs and eardrums"

\textcolor{responsecolor}{Answer: }[6][7]

\textcolor{instructioncolor}{Instruction: }Examine the text and numbered quotes below. The text was likely written from one or a few of the quotes. Which of these quotes, if any, could have been used to write the text? It may be none of the quotes, one quote, or several of the quotes. Do not provide redundant quotes. Respond with the number of the quote(s) in brackets.

\textcolor{querycolor}{Text:} Hyaluronic acid has been shown to decrease the recovery time of wounds and can lessen scarring

\textcolor{sourcescolor}{Quotes:}

[1] "helps keep things hydrated"

[2] "helps skin stretch and flex and reduces skin wrinkles and lines"

[3] "Hyaluronic acid is also proven to help wounds heal faster and can reduce scarring"

\textcolor{responsecolor}{Answer: }[3]

\textcolor{instructioncolor}{Instruction: }Examine the text and numbered quotes below. The text was likely written from one or a few of the quotes. Which of these quotes, if any, could have been used to write the text? It may be none of the quotes, one quote, or several of the quotes. Do not provide redundant quotes. Respond with the number of the quote(s) in brackets.

\textcolor{querycolor}{Text:} Unlike tomatoes that carry a genetic mutation that makes them look consistently red at the expense of their flavor, many heirloom tomatoes taste sweeter

\textcolor{sourcescolor}{Quotes:}

[1] "Many heirloom tomatoes are sweeter and lack a genetic mutation that gives tomatoes a uniform red color at the cost of the fruit's taste"

[2] "True heirlooms must be 'old' because they have been a variety for anywhere between 25 and 50 years"

\textcolor{responsecolor}{Answer: }[1]

\textcolor{instructioncolor}{Instruction: }Examine the text and numbered quotes below. The text was likely written from one or a few of the quotes. Which of these quotes, if any, could have been used to write the text? It may be none of the quotes, one quote, or several of the quotes. Do not provide redundant quotes. Respond with the number of the quote(s) in brackets.

\textcolor{querycolor}{Text:} The person who took the Mona Lisa, Vincenzo Peruggia, used to work at the Louvre museum and even helped make a special glass case to keep this famous painting safe before he stole it.

\textcolor{sourcescolor}{Quotes:}

[1] "The man who stole her, Vincenzo Peruggia, had once worked at the Louvre and had, ironically, helped build a protective case for the 16th century masterpiece"

\textcolor{responsecolor}{Answer: }\end{promptbox}
\refstepcounter{promptbox}
\textbf{Box \thepromptbox: } Example of the prompt with few-shot examples used to identify the citations for the paraphrased, entailed, and abstractive responses to NQ and Eta3G queries.
\label{box:nq_eta3g_citation_prompt}
\vspace{.2cm}
\vspace{.2cm}
\begin{promptbox}
\raggedright
\textcolor{instructioncolor}{Instruction: }Examine the text and numbered quotes below. The text was likely written from one or a few of the quotes. Which of these quotes, if any, could have been used to write the text? It may be none of the quotes, one quote, or several of the quotes. Do not provide redundant quotes. Respond with the number of the quote(s) in brackets.

\textcolor{querycolor}{Text:} Theodosius II, an Eastern Roman (Byzantine) emperor, founded the Imperial University of Constantinople, also known as the University of the Palace Hall of Magnaura, in 425 AD when he founded the Pandidakterion.

\textcolor{sourcescolor}{Quotes:}

[1] "The Imperial University of Constantinople, sometimes known as the University of the Palace Hall of Magnaura, can trace its corporate origins to 425 AD, when the Eastern Roman (Byzantine) emperor Theodosius II founded the Pandidakterion"

[2] "Theodosius II (10 April 401 – 28 July 450), commonly surnamed Theodosius the Younger, or Theodosius the Calligrapher, was the Eastern Roman Emperor for most of his life, taking the throne as an infant in 402 and ruling as the Eastern Empire's sole emperor after the death of his father Arcadius in 408"

\textcolor{responsecolor}{Answer: }[1][2]

\textcolor{instructioncolor}{Instruction: }Examine the text and numbered quotes below. The text was likely written from one or a few of the quotes. Which of these quotes, if any, could have been used to write the text? It may be none of the quotes, one quote, or several of the quotes. Do not provide redundant quotes. Respond with the number of the quote(s) in brackets.

\textcolor{querycolor}{Text:} Born on August 11, 1927, Stuart Rosenberg was an American director of television and film and passed away on March 15, 2007.

\textcolor{sourcescolor}{Quotes:}

[1] "Move is a 1970 American comedy film starring Elliott Gould, Paula Prentiss and Geneviève Waïte, and directed by Stuart Rosenberg."

[2] "Stuart Rosenberg (August 11, 1927 – March 15, 2007) was an American film and television director."

[3] "Méditerranée is a 1963 French experimental film directed by Jean-Daniel Pollet"

[4] "Jean-Daniel Pollet (1936–2004) was a French film director and screenwriter."

\textcolor{responsecolor}{Answer: }[1][2][3][4]

\textcolor{instructioncolor}{Instruction: }Examine the text and numbered quotes below. The text was likely written from one or a few of the quotes. Which of these quotes, if any, could have been used to write the text? It may be none of the quotes, one quote, or several of the quotes. Do not provide redundant quotes. Respond with the number of the quote(s) in brackets.

\textcolor{querycolor}{Text:} The third wife of Sheikh Zayed bin Sultan Al Nahyan is Fatima bint Mubarak Al Ketbi.

\textcolor{sourcescolor}{Quotes:}

[1] "Fatima bint Mubarak Al Ketbi is the third wife of Sheikh Zayed bin Sultan Al Nahyan."

[2] "Sheikh Zayed bin Sultan Al Nahyan 6 May 1918 – 2 November 2004) was the ruler of Abu Dhabi for more than 30 years (6 August 1966 – 2 November 2004)."

\textcolor{responsecolor}{Answer: }[1][2]

\textcolor{instructioncolor}{Instruction: }Examine the text and numbered quotes below. The text was likely written from one or a few of the quotes. Which of these quotes, if any, could have been used to write the text? It may be none of the quotes, one quote, or several of the quotes. Do not provide redundant quotes. Respond with the number of the quote(s) in brackets.

\textcolor{querycolor}{Text:} Martin Frič directed the Czech comedy film, 'Valentine the Good', which came out in 1942.

\textcolor{sourcescolor}{Quotes:}

[1] "Banović Strahinja(released internationally as The Falcon) is a 1981 Yugoslavian- German adventure film written and directed by Vatroslav Mimica."

[2] "Vatroslav Mimica( born 25 June 1923) is a Croatian film director and screenwriter."

[3] "Valentin the Good is a 1942 Czech comedy film directed by Martin Frič."

[4] "Martin Frič( 29 March 1902 – 26 August 1968) was a Czech film director, screenwriter and actor."

\textcolor{responsecolor}{Answer: }[1][2][3][4]

\textcolor{instructioncolor}{Instruction: }Examine the text and numbered quotes below. The text was likely written from one or a few of the quotes. Which of these quotes, if any, could have been used to write the text? It may be none of the quotes, one quote, or several of the quotes. Do not provide redundant quotes. Respond with the number of the quote(s) in brackets.

\textcolor{querycolor}{Text:} The 2010 Bangladeshi drama 'Runway' was helmed by Tareque Masud, who was born on December 6, 1956.

\textcolor{sourcescolor}{Quotes:}

[1] "Runway is a 2010 Bangladeshi drama film directed by Tareque Masud"

[2] "Tareque Masud( 6 December 1956 – 13 August 2011) was a Bangladeshi independent film director, film producer, screenwriter and lyricist"

\textcolor{responsecolor}{Answer: }\end{promptbox}
\refstepcounter{promptbox}
\textbf{Box \thepromptbox: } Example of the prompt with few-shot examples used to identify the citations for the paraphrased, entailed, and abstractive responses to 2WikiMH queries.
\label{box:2WikiMH_citation_prompt}
\vspace{.2cm}
\vspace{.2cm}
\begin{promptbox}
\raggedright
\textcolor{instructioncolor}{Instruction: }Examine the text and numbered quotes below. The text was likely written from one or a few of the quotes. Which of these quotes, if any, could have been used to write the text? It may be none of the quotes, one quote, or several of the quotes. Do not provide redundant quotes. Respond with the number of the quote(s) in brackets.

\textcolor{querycolor}{Text:} Moreover, maintaining a nutritious diet, abstaining from drugs and alcohol, not missing meals, reducing caffeine intake, and steering clear of any known food triggers are good precautions.

\textcolor{sourcescolor}{Quotes:}

[1] "it helps to avoid the things that usually cause one"

[2] "Get enough sleep"

[3] "Limit your stress"

[4] "Eat a balanced diet"

[5] "Avoid drugs and alcohol"

[6] "Don't skip meals"

[7] "Limit caffeine"

[8] "Avoid any foods that have been triggers."

\textcolor{responsecolor}{Answer: }[4][5][6][7][8]

\textcolor{instructioncolor}{Instruction: }Examine the text and numbered quotes below. The text was likely written from one or a few of the quotes. Which of these quotes, if any, could have been used to write the text? It may be none of the quotes, one quote, or several of the quotes. Do not provide redundant quotes. Respond with the number of the quote(s) in brackets.

\textcolor{querycolor}{Text:} According to both the FDA and drug makers, children under 4 years of age should not take OTC cold and cough medicines.

\textcolor{sourcescolor}{Quotes:}

[1] "Talk to your pediatrician before giving OTC cold and cough medicines"

[2] "These medicines should not be given to children under 4 years of age, according to the FDA and the drug makers"

\textcolor{responsecolor}{Answer: }[2]

\textcolor{instructioncolor}{Instruction: }Examine the text and numbered quotes below. The text was likely written from one or a few of the quotes. Which of these quotes, if any, could have been used to write the text? It may be none of the quotes, one quote, or several of the quotes. Do not provide redundant quotes. Respond with the number of the quote(s) in brackets.

\textcolor{querycolor}{Text:} However, if there's a need to drain a blister, or if it bursts on its own, ensure to use a needle that has been sterilized either by heating it until it glows red or by cleansing it with alcohol.

\textcolor{sourcescolor}{Quotes:}

[1] "The fluid-filled blister keeps the underlying skin clean, which prevents infection and promotes healing"

[2] "But if you need to pop a blister or it pops by itself: Use a sterilized needle (to sterilize it, put the point or edge in a flame until it is red hot, or rinse it in alcohol)"

[3] "Wash your hands and the area thoroughly, then make a small hole; the fluid will drain on its own"

\textcolor{responsecolor}{Answer: }[2]

\textcolor{instructioncolor}{Instruction: }Examine the text and numbered quotes below. The text was likely written from one or a few of the quotes. Which of these quotes, if any, could have been used to write the text? It may be none of the quotes, one quote, or several of the quotes. Do not provide redundant quotes. Respond with the number of the quote(s) in brackets.

\textcolor{querycolor}{Text:} Administer acetaminophen formulated for infants (such as Tylenol) to alleviate discomfort.

\textcolor{sourcescolor}{Quotes:}

[1] "Use baby acetaminophen (Tylenol) to dull pain"

[2] "Do not use ibuprofen unless the baby is at least 6 months old"

\textcolor{responsecolor}{Answer: }\end{promptbox}
\refstepcounter{promptbox}
\textbf{Box \thepromptbox: } Example of the prompt with few-shot examples used to identify the citations for the paraphrased, entailed, and abstractive responses to MASH queries.
\label{box:mash_citation_prompt}
\vspace{.2cm}

\end{document}